\documentclass{article}
 
\usepackage{microtype}
\usepackage{graphicx}
\usepackage{subfigure}
\usepackage{booktabs}

\usepackage{hyperref}

\usepackage[accepted]{icml2018}
\usepackage{caption}
\usepackage{natbib}
\usepackage{hyperref}
\usepackage{framed}
\usepackage{array}
\usepackage[mathscr]{euscript}
\usepackage{mathtools}
\usepackage{amsmath}
\usepackage{amssymb}
\usepackage{placeins}
\usepackage{etoolbox}
\usepackage{float}
\makeatletter
\patchcmd{\algorithmic}{\addtolength{\ALC@tlm}{\leftmargin} }
{\addtolength{\ALC@tlm}{\leftmargin}}{}{}
\makeatother

\newlength{\oldtextfloatsep}

\newtheorem{theorem}{Theorem}
\newtheorem{proposition}{Proposition}
\newtheorem{corollary}{Corollary}

\newtheorem{remark}{Remark}

\newenvironment{proof}
{\textit{Proof.}}
{$\square$}

\newenvironment{retheo}
{\textbf{Theorem}}
{}

\floatname{algorithm}{\textsc{Algorithm}}
\usepackage{caption}
\def\Rset{\mathbb{R}}

\let\Pr\relax
\DeclareMathOperator*{\Pr}{\mathbb{P}}
\let\P\relax
\DeclareMathOperator*{\P}{\Pr}

\DeclarePairedDelimiter\floor{\lfloor}{\rfloor}
\DeclareMathOperator*{\E}{\mathbb{E}}

\DeclareMathOperator*{\argmin}{argmin}

\newcommand{\cA}{\mathscr A}
\newcommand{\cC}{\mathscr C}
\newcommand{\cD}{\mathscr D}
\newcommand{\cE}{\mathscr E}
\newcommand{\cF}{\mathscr F}
\newcommand{\cH}{\mathscr H}
\newcommand{\cJ}{\mathscr J}

\newcommand{\cO}{\mathscr O}
\newcommand{\cR}{\mathscr R}

\newcommand{\cX}{\mathscr X}
\newcommand{\cY}{\mathscr Y}

\newcommand{\Reg}{R}
\newcommand{\EXP}{\textsc{exp3}}
\newcommand{\EXPABS}{\textsc{exp3-abs}}
\newcommand{\ContEXP}{\textsc{ContextualExp3}}
\newcommand{\UCB}{\textsc{ucb}}
\newcommand{\FTL}{\textsc{fs}}
\newcommand{\UCBN}{\textsc{ucb-n}}
\newcommand{\UCBNT}{\textsc{ucb-nt}}

\newcommand{\UCBGT}{\textsc{ucb-gt}}
\newcommand{\ContEXPABS}{\textsc{Cont\EXPABS}}
\newcommand{\GABS}{G^{\text{\sc abs}}}
\newcommand{\GSUB}{G^{\text{\sc sub}}}

\newcommand{\ov}{\overline}
\newcommand{\h}{\widehat}
\newcommand{\e}{\epsilon}
\newcommand{\set}[2][]{#1 \{ #2 #1 \} }
\newcommand{\comments}[1]{}
\newcommand{\ignore}[1]{}

\hypersetup{
  colorlinks   = true,
  urlcolor     = blue,
  linkcolor    = blue,
  citecolor    = blue
}

\renewcommand{\epsilon}{\varepsilon}
\renewcommand{\leq}{\leqslant}
\renewcommand{\geq}{\geqslant}
\renewcommand{\phi}{\varphi}
\renewcommand{\hat}{\widehat}
\renewcommand{\tilde}{\widetilde}

\newcommand{\R}{\mathbb{R}}

\newcommand{\ve}{\varepsilon}
\newcommand{\scX}{\mathcal{X}}

\newcommand{\loss}{{\mathcal{L}}}
\newcommand{\tL}{{\tilde{L}}}

\icmltitlerunning{Online Learning with Abstention}

\begin{document}

\renewcommand*{\thefootnote}{\fnsymbol{footnote}}

\twocolumn[
\icmltitle{Online Learning with Abstention}

\begin{icmlauthorlist}
\icmlauthor{Corinna Cortes}{google}
\icmlauthor{Giulia DeSalvo}{google}
\icmlauthor{Claudio Gentile}{google,inria}
\icmlauthor{Mehryar Mohri}{courant,google}
\icmlauthor{Scott Yang\footnotemark}{deshawe}
\end{icmlauthorlist}

\icmlaffiliation{google}{Google Research, New York, NY.}
\icmlaffiliation{inria}{INRIA Lille Nord Europe.}
\icmlaffiliation{courant}{Courant Institute of Mathematical Sciences, New York, NY.}
\icmlaffiliation{deshawe}{D. E. Shaw \& Co., New York, NY}

\icmlcorrespondingauthor{Giulia DeSalvo}{giuliad@google.com}

\icmlkeywords{online learning, abstention option, feedback graphs}
\vskip 0.3in
]
\footnotetext{Work done at the Courant Institute of Mathematical Sciences.}

\printAffiliationsAndNotice{}

\renewcommand*{\thefootnote}{\arabic{footnote}}
\setcounter{footnote}{0}

\begin{abstract}
  We present an extensive study of a key problem in online learning
  where the learner can opt to abstain from making a prediction, at a
  certain cost. In the adversarial setting, we show how existing
  online algorithms and guarantees can be adapted to this problem.  In
  the stochastic setting, we first point out a bias problem that
  limits the straightforward extension of algorithms such as \UCBN\ to
  this context.  Next, we give a new algorithm, \UCBGT, that exploits
  historical data and time-varying feedback graphs. We show that this
  algorithm benefits from more favorable regret guarantees than a
  natural extension of \UCBN. We further report the results of a
  series of experiments demonstrating that \UCBGT\ largely outperforms
  that extension of \UCBN, as well as other standard baselines.
\end{abstract}

\section{Introduction}
\label{sec:introduction}
We consider an online learning scenario, prevalent in many
applications, where the learner is granted the option of abstaining
from making a prediction, at a certain cost. For example, in the
classification setting, at each round, the learner can choose to make
a prediction and incur a standard zero-one misclassification cost, or
elect to abstain, in which case she incurs an abstention cost,
typically less than one. Abstention can thus represent an attractive
option to avoid a higher cost of misclassification. Note, however,
that when the learner abstains, she does not receive the true label
(correct class), which results in a loss of information.

This scenario of online learning with abstention is relevant to many
real-life problems. As an example, consider the scenario where a
doctor can choose to make a diagnosis based on the current information
available about a patient, or abstain and request further laboratory
tests, which can represent both a time delay and a financial cost. In
this case, the abstention cost is usually substantially lower than
that of a wrong diagnosis. The online model is appropriate since it
captures the gradual experience a doctor gains by testing, examining
and following new patients.

Another instance of this problem appears in the design of
spoken-dialog applications such as those in modern personal
assistants. Each time the user asks a question, the assistant can
either offer a direct response to the question, at the risk of
providing an inaccurate response, or choose to say ``I am sorry, I do
not understand?'', which results in a longer and thereby more costly
dialog requesting the user to reformulate his question.  Similar
online learning problems arise in the context of self-driving cars
where, at each instant, the assistant must determine whether to
continue steering the car or return the control to the driver. Online
learning with abstention also naturally models many problems arising
in electronic commerce platforms such as an Ad Exchange, an online
platform set up by a publisher where several advertisers bid in order to
compete for an ad slot, the abstention cost being the opportunity loss
of not bidding for a specific ad slot.

In the batch setting, the problem of learning with abstention has been
studied in a number of publications, starting with
\cite{Chow1957,Chow1970}. Its theoretical aspects have been analyzed
by several authors in the last decade.
\citet{YanivWiener2010,YanivWiener2011} studied the trade-off between
the coverage and accuracy of classifiers. \citet{BartlettWegkamp2008}
introduced a loss function including explicitly an abstention cost and
gave a consistency analysis of a surrogate loss that they used to
derive an algorithm. More recently,
\citet{CortesDeSalvoMohri2016a,CortesDeSalvoMohri2016b} presented a
comprehensive study of the problem, including an analysis of the
properties of a corresponding abstention (or rejection) loss with a
series of theoretical guarantees and algorithmic results both for
learning with kernel-based hypotheses and for boosting.

This paper presents an extensive study of the problem
of online learning with abstention, in both the adversarial and the
stochastic settings. We consider the common scenario of prediction
with expert advice \citep{LittlestoneWarmuth1994} and adopt the same
general abstention loss function as in
\citep{CortesDeSalvoMohri2016a}, with each expert formed by a pair
made of a predictor and an abstention function.

A key aspect of the problem we investigate, which makes it distinct
from both batch learning with abstention, where labels are known for
all training points, and standard online learning (in the full
information setting) is the following: if the algorithm abstains from
making a prediction for the input point received at a given round, the
true label of that point is not revealed. As a result, the loss of the
experts that would have instead made a prediction on that point cannot
be determined at that round. Thus, we are dealing with an online
learning scenario with partial feedback.  If the algorithm chooses to
predict, then the true label is revealed and the losses of all
experts, including abstaining ones, are known. But, if the algorithm
elects to abstain, then only the losses of the abstaining experts are
known, all of them being equal to the same abstention cost.

As we shall see, our learning problem can be cast as a specific
instance of online learning with a feedback graph, a framework
introduced by \citet{MannorShamir2011} and later extensively analyzed
by several authors
\citep{CaronKvetonLelargeBhagat2012,AlonCesaBianchiGentileMansour2013,AlonCesaGentileMannorMansourShamir2014,AlonCesaBianchiDekelKoren2015,KocakNeuValkoMunos2014,Neu2015,CohenHazanKoren2016}). In
our context, the feedback graph varies over time, a scenario for which
most of the existing algorithms and analyses (specifically, in the
stochastic setting) do not readily apply. Our setting is distinct from
the KWIK (knows what it knows) framework of \citet{LiLittmanWalsh2008}
and its later extensions, though there are some connections, as
discussed in Appendix~\ref{sec:relatedworks}.

Our contribution can be summarized as follows.  In
Section~\ref{sec:adversarial}, we analyze an adversarial setting both
in the case of a finite family of experts and that of an infinite
family.  We show that the problem of learning with abstention can be
cast as that of online learning with a time-varying feedback graph
tailored to the problem.  In the finite case, we show how ideas from
\cite{AlonCesaGentileMannorMansourShamir2014,AlonCesaBianchiDekelKoren2015}
can be extended and combined with this time-varying feedback graph to
devise an algorithm, \EXPABS, that benefits from favorable
guarantees. In turn, \EXPABS\ is used as a subroutine for the infinite
case where we show how a surrogate loss function can be carefully
designed for the abstention loss, while maintaining the same partial
observability. We use the structure of this loss function to extend
\ContEXP\ \citep{CesaBianchiGaillardGentileGerchinovitz2017} to the
abstention scenario and prove regret guarantees for its performance.

In Section~\ref{sec:stochastic}, we shift our attention to the
stochastic setting. Stochastic bandits with a fixed feedback graph
have been previously studied by \citet{CaronKvetonLelargeBhagat2012}
and \citet{CohenHazanKoren2016}. We first show that an immediate
extension of these algorithms to the time-varying graphs in the
abstention scenario faces a technical \emph{bias problem} in the
estimation of the expert losses. Next, we characterize a set of
feedback graphs that can circumvent this bias problem in the general
setting of online learning with feedback graphs. We further design a
new algorithm, \UCBGT, whose feedback graph is \emph{estimated} based
on past observations. We prove that the algorithm admits more
favorable regret guarantees than the \UCBN\ algorithm
\citep{CaronKvetonLelargeBhagat2012}. Finally, in
Section~\ref{sec:experiments} we report the results of several
experiments with both artificial and real-world datasets demonstrating
that \UCBGT\ in practice significantly outperforms an unbiased, but
limited, extension of \UCBN, as well as a standard bandit baseline,
like UCB \cite{acf02}.

\section{Learning Problem}
\label{sec:problem}
Let $\cX$ denote the input space (e.g., $\cX$ is a bounded subset of
$\R^d$).  We denote by $\cH$ a family of predictors
$h \colon \cX \to \Rset$, and consider the familiar binary
classification problem where the loss $\ell(y,h(x))$ of $h \in \cH$ on
a labeled pair $(x, y) \in \cX \times \set{\pm 1}$ is defined by
either the 0/1-loss $1_{y h(x) \leq 0}$, or some Lipschitz variant
thereof (see Section \ref{sec:adversarial}).  In all cases, we assume
$\ell(\cdot,\cdot) \in [0, 1]$.  We also denote by $\cR$ a family of
abstention functions $r\colon \cX \to \Rset$, with $r(x) \leq 0$
indicating an abstention on $x \in \cX$ (or that $x$ is
\emph{rejected}), and $r(x) > 0$ that $x$ is predicted upon (or that
$x$ is \emph{accepted}).

We consider a specific online learning scenario whose regime lies
between bandit and full information, sometimes referred to as {\em
  bandit with side-information} (e.g.,
\citet{MannorShamir2011,CaronKvetonLelargeBhagat2012,AlonCesaBianchiGentileMansour2013,
  AlonCesaGentileMannorMansourShamir2014,AlonCesaBianchiDekelKoren2015,KocakNeuValkoMunos2014,Neu2015,CohenHazanKoren2016}).
In our case, the arms are pairs made of a predictor function $h$ and
an abstention function $r$ in a given family
$\cE \subseteq \cH \times\cR$.  We will denote by
$\xi_j = (h_j, r_j)$, $j \in [K]$, the elements of $\cE$. In fact,
depending on the setting, $K$ may be finite or (uncountably) infinite.
Given $h_j$, one natural choice for the associated abstention function
$r_j$ is a confidence-based abstention function of the form
$r_j(x) = |h_j(x)| - \theta$, for some threshold $\theta > 0$. Yet,
more general pairs $(h_j, r_j)$ can be considered here. This provides
an important degree of flexibility in the design of algorithms where
abstentions are allowed, as shown in
\cite{CortesDeSalvoMohri2016a,CortesDeSalvoMohri2016b}.
Appendix~\ref{sec:relatedworks} presents a concrete example
illustrating the benefits of learning with these pair of functions.

The online learning protocol is described as follows. The set $\cE$ is
known to the learning algorithm beforehand. At each round $t \in [T]$,
the online algorithm receives an input $x_t \in \cX$ and chooses
(possibly at random) an arm (henceforth also called ``expert" or
``pair") $\xi_{I_t} = (h_{I_t}, r_{I_t}) \in \cE$. If the inequality
$r_{I_t}(x_t) \leq 0$ holds, then the algorithm abstains and incurs as
loss an abstention cost $c(x_t) \in [0,1]$. Otherwise, it predicts
based on the sign of $h_{I_t}(x_t)$, receives the true label
$y_t \in \set{\pm 1}$, and incurs the loss $\ell(y_t,
h_{I_t}(x_t))$. Thus, the overall \emph{abstention loss} $L$ of expert
$\xi = (h, r) \in \cE$ on the labeled pair
$z = (x, y) \in \cX \times \set{\pm 1}$ is defined as follows:
\begin{equation}
\label{e:absloss}
L( \xi, z) =  \ell(y,h(x)) 1_{r(x) > 0} + c(x) 1_{r(x) \leq 0}~.
\end{equation}
For simplicity, we will assume throughout that the abstention cost
$c(x)$ is a (known) constant $c \in [0,1]$, independent of $x$, though
all our results can be straightforwardly extended to the case when $c$
is a (Lipschitz) function of $x$, which is indeed desirable in some
applications.

\begin{figure}[t]
\centering
\includegraphics[scale=0.27]{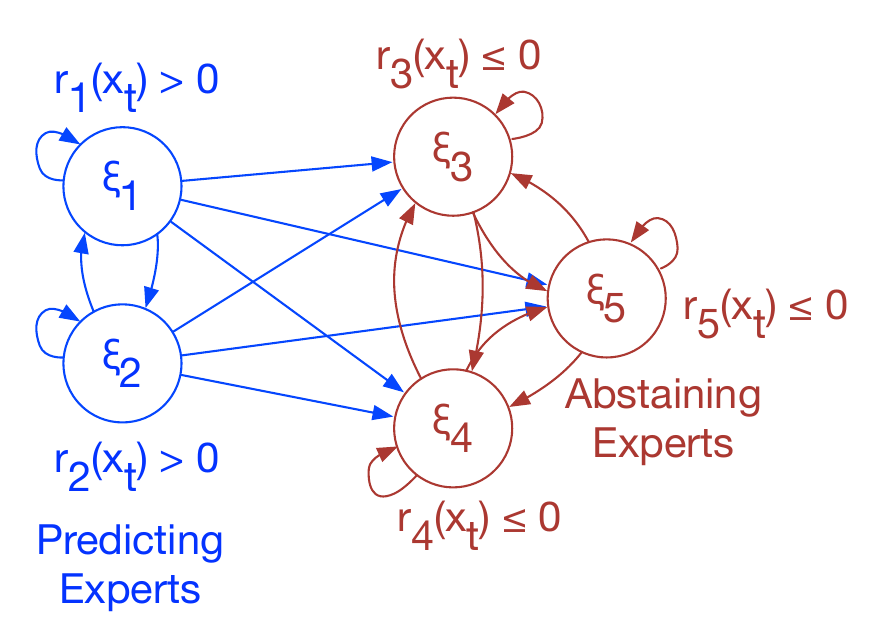}
\vskip -0.1in
\caption{Feedback graph $\GABS_t$ for the scenario of online
  learning with abstention, with $K = 5$.}
\vskip -0.2in
\label{fig:feedback}
\end{figure}

Our problem can be naturally cast as an online learning problem with
side information in the form of a feedback graph.  Online learning
with a feedback graph is a general framework that covers a variety of
problems with partial information, including the full information
scenario, where the graph is fully connected, and the bandit scenario
where all vertices admit only self-loops and are disconnected
\citep{AlonCesaBianchiGentileMansour2013,AlonCesaGentileMannorMansourShamir2014}.
In our case, we have a directed graph $\GABS_t = (V, E_t)$ that
depends on the instance $x_t$ received by the algorithm at round
$t \in [T]$. Here, $V$ denotes the finite set of vertices of this
graph, which, in the case of a finite set of arms, coincides with the
set of experts $\cE$, while $E_t$ denotes the set of directed edges at
round $t$. The directed edge $\xi_i \rightarrow \xi_j$ is in $E_t$
if the loss of expert $\xi_j \in V$ is observed when expert $\xi_i$ is
selected by the algorithm at round $t$.  In our problem, if the
learner chooses to predict at round $t$ (i.e., if $r_{I_t}(x_t) > 0$),
then she observes the loss $L(\xi_j,z_t)$ of all experts $\xi_j$,
since the label $y_t$ is revealed to her. If instead she abstains at
round $t$ (i.e., if $r_{I_t}(x_t) \leq 0$), then she only observes
$L(\xi_j,z_t)$ for those experts $\xi_j$ that are abstaining in that
round, that is, the set of $j$ such that $r_j(x_t) \leq 0$, since for
all such $\xi_j$, we have $L(\xi_j,z_t) = c$.  Notice that in both
cases the learner can observe the loss of her own action. Thus,
the feedback graph we are operating with is a nearly fully connected
directed graph with self-loops, except that it admits only one-way
edges from predicting to abstaining vertices (see
Figure~\ref{fig:feedback} for an example).  Observe also that
the feedback graph $\GABS_t$ is fully determined by $x_t$.

We will consider both an adversarial setting (Section
\ref{sec:adversarial}), where no distributional assumption is made
about the sequence $z_t = (x_t, y_t)$, $t \in [T]$, and a stochastic
setting (Section \ref{sec:stochastic}), where $z_t$ is assumed to be
drawn i.i.d.\ from some unknown distribution $\cD$ over
$\cX\times\{\pm 1\}$. For both settings, we measure the performance of
an algorithm $\cA$ by its (\emph{pseudo}-)\emph{regret} $\Reg_T(\cA)$,
defined as
$\Reg_T(\cA) = \sup_{\xi \in \cE} \E [ \sum_{t = 1}^T L( \xi_{I_t},
z_t) - \sum_{t = 1}^T L(\xi, z_t) ]~,$ where the expectation is taken
both with respect to the algorithm's choice of actions $I_t$s and, in
the stochastic setting, the random draw of the $z_t$s.

In the stochastic setting, we will be mainly concerned with the case
where $\cE$ is a finite set of experts
$\cE = \set{\xi_1, \ldots, \xi_K}$. We then denote by $\mu_j$ the
expected loss of expert $\xi_j \in \cE$,
$\mu_j = \E_{z \sim \cD}[L(\xi_j, z)]$, by $\mu^*$ the expected loss
of the best expert, $\mu^* = \min_{j \in [K]} \mu_j$, and by
$\Delta_j$ the loss gap to the best, $\Delta_j = \mu_j - \mu^*$. In
the adversarial setting, we will analyze both the finite and infinite
expert scenarios. In the infinite case, since $L$ is non-convex in the
relevant parameters (Eq.~\eqref{e:absloss}), further care is needed.

\section{Adversarial setting}
\label{sec:adversarial}

As a warm-up, we start with the adversarial setting with finitely-many
experts.  Following ideas from
\citet{AlonCesaGentileMannorMansourShamir2014,AlonCesaBianchiDekelKoren2015},
we design an online algorithm for the abstention scenario by combining
standard finite-arm bandit algorithms, like \EXP\ \cite{Exp3}, with
the feedback graph $\GABS_t$ of Section \ref{sec:problem}. We call the
resulting algorithm \EXPABS\ (\EXP\ with abstention). The algorithm is
a variant of \EXP\ where the importance weighting scheme to achieve
unbiased loss estimates is based on the probability of the loss of an
expert being {\em observed} as opposed to that of an expert being
selected --- see Appendix~\ref{app:adversarial}
(Algorithm~\ref{alg:exp3rej}). The following guarantee holds for this
algorithm.
\begin{theorem}
\label{th:exp3rej}
Let \EXPABS\ be run with learning rate $\eta$ over a set of $K$
experts $\xi_1,\ldots,\xi_K$.  Then, the algorithm admits the
following regret guarantee after $T$ rounds:
\begin{equation*}
\Reg_T(\EXPABS) \leq (\log K)/\eta + \eta\,T (c^2 + 1)/2.
\end{equation*}
In particular, if \EXPABS\ is run with
$\eta = \sqrt{\frac{2\log K}{(c^2 + 1)T}}$, then
\(
\Reg_T(\EXPABS) \leq \sqrt{2(c^2+1)T\log K}.
\)
\end{theorem}
The proof of this result, as well as all other proofs, is given in the
appendix. The dependency of the bound on the number of experts is
clearly more favorable than the standard bound for \EXP\
($\sqrt{\log K}$ instead of $\sqrt{K}$). Theorem \ref{th:exp3rej} is
in fact reminiscent of what one can achieve using the
contextual-bandit algorithm
EXP4~\cite{AuerCesaBianchiFreundSchapire2002} run on $K$ experts, each
one having two actions.

We now turn our attention to the case of an uncountably infinite
$\cE$. To model this more general framework, one might be tempted to
focus on parametric classes of functions $h$ and $r$, e.g., the family
$\cE$ of linear functions
\[
\set[\big]{(h, r) : h(x) = w^\top x,r(x) = |w^\top x| - \theta,w \in \R^d, \theta > 0},
\]
%
introduce some convex surrogate of the abstention loss
(\ref{e:absloss}), and work in the parametric space of $(w, \theta)$
through some Bandit Convex Optimization technique (e.g.,
\cite{hazan16}). Unfortunately, this approach is not easy to put in
place, since the surrogate loss not only needs to ensure convexity and
some form of calibration, but also the ability for the algorithm to
observe the loss of its own action (the self-loops in the graph of
Figure \ref{fig:feedback}).

\begin{figure}[t]
\centering
 \includegraphics[scale=0.3]{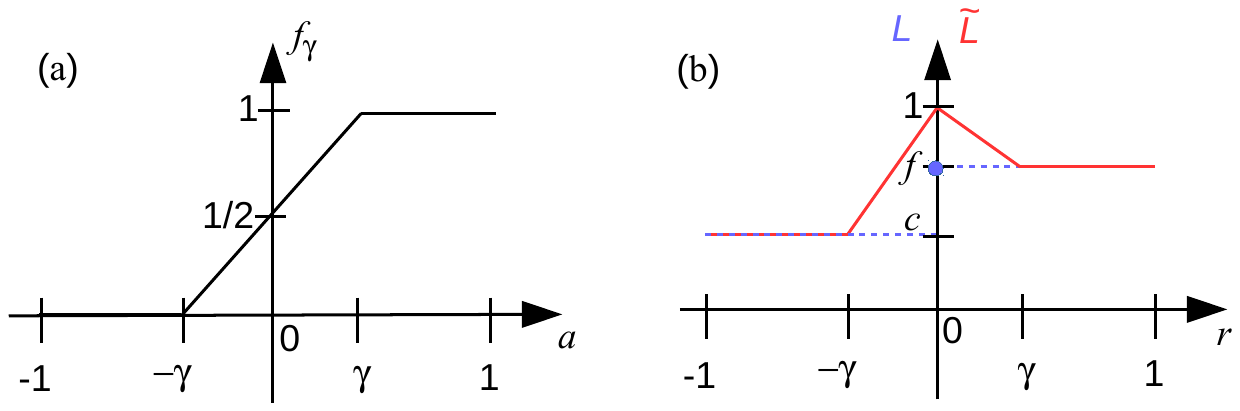}
\vskip -0.1in
 \caption{(a) The 0/1-loss function with slope $1/(2 \gamma)$ at the origin.
   (b) For a given value of $x$ and margin $a = -yh(x)$ (which in
   turn sets the value of $f = f_{\gamma}(a) \in [0, 1]$), plots of the
   abstention loss function $L(a, r)$ (dotted blue curve), and the surrogate
   abstention loss $\tilde{L}(a, r)$ (red curve), both as a
   function of $r = r(x) \in [-1, 1]$.}
\vskip -0.15in
\label{f:1}
\end{figure}

We have been unable to get around this problem by just resorting to
convex surrogate losses (and we strongly suspect that it is not
possible), and in what follows we instead introduce a surrogate
abstention loss which is Lipschitz but not convex. Moreover, we take
the more general viewpoint of competing with pairs $(h,r)$ of
Lipschitz functions with bounded Lipschitz constant. Let us then
consider the version of the abstention loss (\ref{e:absloss}) with
$\ell(y, h(x)) = f_{\gamma}(-yh(x))$, where $f_{\gamma}$ is the 0/1-loss
with slope $1/(2\gamma)$ at the origin,\,
\( f_{\gamma}(a) = \left(\frac{\gamma+a}{2\gamma}\right) 1_{|a| \leq
  \gamma} + 1_{a\geq 0} 1_{|a| > \gamma}\, \)
%
%
(see Figure \ref{f:1} (a)), and the class of experts
${\cE} = \set[\big]{\xi = (h, r)\, |\, h, r\colon \cX \subseteq \R^d
  \to [-1, 1]}$. Here, functions $h$ and $r$ in the definition of
$\cE$ are assumed to be $L_{\cE}$-Lipschitz with respect to an
appropriate distance on $\R^d$, for some constant $L_{\cE}$
which determines the size of the family $\cE$.
\ignore{
, e.g., the standard Euclidean norm $||\cdot||_2$, $|h(x)-h(x')| \leq C_{\cF}||x-x'||_2$ for all $x,x' \in
\R^d$, and similarly for the functions $r$.
}
%


Using ideas from \cite{CesaBianchiGaillardGentileGerchinovitz2017}, we present an
algorithm that approximates the action space by a finite
cover while using the structure of the abstention setting. The crux of
the problem is to define a Lipschitz function $\tL$ that uppers bounds
the abstention loss while maintaining the same feedback assumptions,
namely the feedback graph given in Figure~\ref{fig:feedback}. One
Lipschitz function $\tL$ that precisely solves this problem is the following:
\begin{center}
\(
\tL(\xi,z) =
\begin{cases}
c &{\mbox{if $r(x) \leq -\gamma$}}\\
1+\left(\frac{1-c}{\gamma}\right)r(x) &{\mbox{if $r(x) \in (-\gamma,0)$}}\\
1-\left(\frac{1-f_{\gamma}(-y h(x))}{\gamma}\right)r(x) &{\mbox{if $r(x) \in [0,\gamma)$}}\\
f_{\gamma}(-y h(x)) &{\mbox{if $r(x) \geq \gamma$}}~,
\end{cases}
\)
\end{center}
for $\gamma \in (0,1)$. $\tL(\xi,z)$ is plotted in Figure \ref{f:1}(b).
Notice that this function is consistent with the feedback requirements of Section
\ref{sec:problem}: $r_{I_t}(x_t) \leq 0$ implies that
$\tL((h(x_t),r(x_t)),z_t)$ is known to the algorithm (i.e., is
independent of $y_t$) for all $(h,r) \in \cE$ such that
$r(x_t) \leq 0$, while $r_{I_t}(x_t) > 0$ gives complete knowledge of
$\tL((h(x_t),r(x_t)),z_t)$ for all $(h,r) \in \cE$, since $y_t$ is
observed.

We can then adapt the machinery from
\cite{CesaBianchiGaillardGentileGerchinovitz2017} so as to apply a
contextual version of \EXPABS\ to the sequence of losses
$\tL(\xi,z_t), t \in [T]$.
The algorithm adaptively covers 
$\cX$ with 
balls of a fixed radius $\ve$, each ball hosting
an instance of \EXPABS. 
We call this algorithm \ContEXPABS\ -- see Appendix \ref{app:adversarialinfinite} for details.
\begin{theorem}
\label{th:infarms}
Consider the abstention loss
\begin{equation*}
L(\xi,z) = f_{\gamma}(-yh(x)) 1_{r(x) > 0} + c 1_{r(x) \leq 0}~,
\end{equation*}
and let
$\xi^* = (h^*, r^*) = \argmin_{\xi \in \cE} \sum_{t = 1}^T L(\xi, z_t)
$, with $\cE = \set{(h, r)}$ made of pairs of Lipschitz functions as
described above.  If \ContEXPABS\ is run with parameter
$\ve \simeq T^{-\frac{1}{2+d}}\,\gamma^{\frac{2}{2+d}}$ and an
appropriate learning rate (see Appendix~\ref{app:adversarial}), then,
it admits the following regret guarantee:
\[
\Reg_T(\ContEXPABS)
\leq
{\tilde \cO}\left(T^{\frac{d+1}{d+2}}\,\gamma^{-\frac{d}{d+2}}\right) + M^*_{T}(\gamma),
\]
where  $M^*_T(\gamma)$ is the number of $x_t$ such that $|r^*(x_t)| \leq \gamma$.
\end{theorem}
In the above, $ {\tilde \cO}$ hides constant and 
$\ln(T)$ factors, while $ \simeq $ disregards constants like $L_\cE$, and various log factors. \ContEXPABS\ is also computationally efficient,
thereby providing a compelling solution to the infinite armed case of
online learning with abstention.

\ignore{
Let's close with some quick comments:
\begin{itemize}
\item The alg is efficient (notice that the number of balls and the
  the size of the action space within each ball decreases as we
  increase $d$).
\item The alg gets suboptimal bounds, a chaining on the function space
  would probably get regret $T^{\frac{d}{d+1}}$, instead of
  $T^{\frac{d+1}{d+2}}$ but this comes at the cost of an inefficient
  algorithm (exponential time or more).
\item There is room for improvements:
\begin{itemize}
\item It might be possible to do transfer learning within each ball
  $B$ \ContEXP operates on, and leverage the specific structure
  of the loss function to get faster rates through an efficient
  algorithm. Currently none of the two is done here.
\item The above algorithm needs to know $\gamma$ beforehand, it would
  be nice to have in (\ref{e:finalbound}) a min over $\gamma > 0$
  instead.
\end{itemize}
\end{itemize}
}

\section{Stochastic setting}
\label{sec:stochastic}

We now turn to studying the stochastic setting.  As pointed out in
Section~\ref{sec:problem}, the problem can be cast as an instance of
online learning with time-varying feedback graphs $\GABS_t$. Thus, a
natural method for tackling the problem 
would be to extend existing algorithms designed for the stochastic
setting with feedback graphs to our abstention
scenario~\citep{CohenHazanKoren2016,CaronKvetonLelargeBhagat2012}.  We
cannot benefit from the algorithm of \citet{CohenHazanKoren2016} in
our scenario. This is because at the heart of its design and
theoretical guarantees lies the assumption that the graphs and losses
are \emph{independent}. The dependency of the feedback graphs on the
observations $z_t$, which also define the losses, is precisely a
property that we wish to exploit in our scenario.

An alternative is to extend the \UCBN\ algorithm of
\citet{CaronKvetonLelargeBhagat2012}, for which the authors provide
gap-based regret guarantees. This algorithm is defined for a
stochastic setting with an undirected feedback graph that is fixed
over time. The algorithm can be straightforwardly extended to the case
of directed time-varying feedback graphs (see
Algorithm~\ref{alg:ucbncaron}). We will denote that extension by
\UCBNT\ to explicitly differentiate it from \UCBN. Let $N_t(j)$ denote
the set of out-neighbors of vertex $\xi_j$ in the directed graph at
time $t$, i.e., the set of vertices $\xi_k$ destinations of an edge
from $\xi_j$.  Then, as with \UCBN, the algorithm updates, at each
round $t$, the upper-confidence bound of every expert for which a
feedback is received (those in $N_t(I_t)$), as opposed to updating
only the upper-confidence bound of the expert selected, as in the
standard \UCB\ of \citet{acf02}.

In the context of learning with abstention, the natural feedback
graph $\GABS_t$ at time $t$ depends on the observation $x_t$ and
varies over time.  Can we extend the regret guarantees of
\citet{CaronKvetonLelargeBhagat2012} to \UCBNT\ with such graphs?
We will show in Section~\ref{subsec:bias} that
vanishing regret guarantees do not hold for \UCBNT\ run with
graphs $\GABS_t$. This is because of a fundamental
estimation bias problem that arises when the graph at time $t$
depends on the observation $x_t$.
This issue affects more generally any natural method using
the $\GABS_t$ graphs. Nevertheless, we will show in Section~\ref{subsec:ucbn} that \UCBNT\
does benefit from favorable
guarantees,
provided the feedback graph $\GABS_t$ it uses at round $t$ is replaced by one that
only depends on events up to time $t - 1$.
%

\subsection{Bias problem}
\label{subsec:bias}

\begin{figure}[t]
\centering
\includegraphics[scale=0.21]{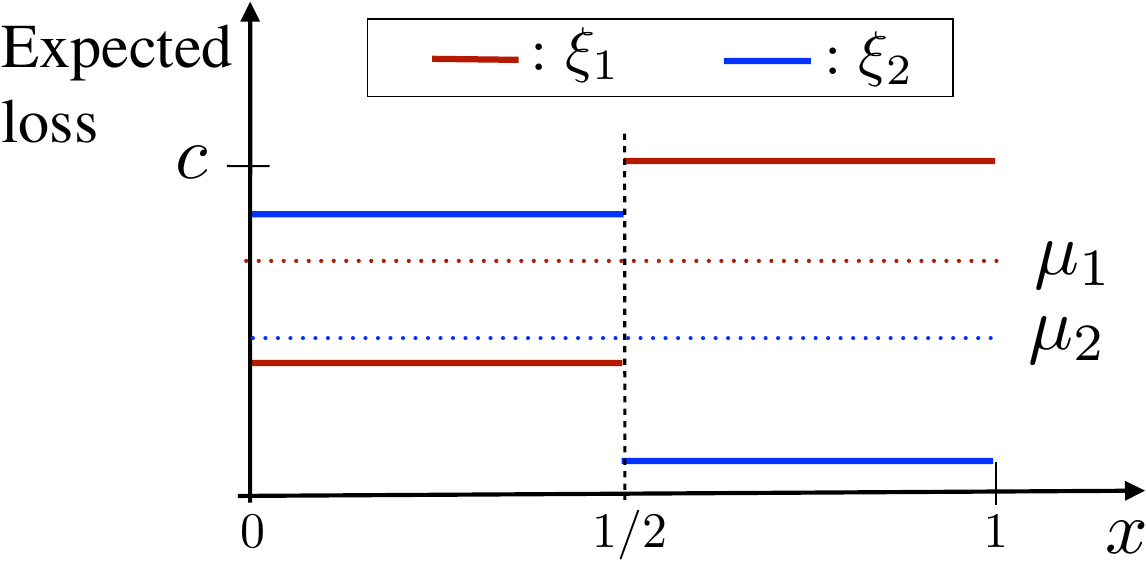}
\vskip -0.1in
\caption{Illustration of the bias problem.}
\vskip -0.2in
\label{fig:biasfeedback}
\end{figure}

Assume there are two experts: $\xi_1$ (red) and $\xi_2$ (blue) with
$\mu_2 < \mu_1$ and $\cX = [0, 1]$ (see
Figure~\ref{fig:biasfeedback}).  For $x > \frac{1}{2}$, the red expert
$\xi_1$ is abstaining and incurring a loss $c$, whereas the blue
expert is never abstaining.  Assume that the probability mass is
quasi-uniform over the interval $[0, 1]$ but with slightly more mass
over the region $x < \frac{1}{2}$. The algorithm may then start out by
observing points in this region. Here, both experts accept and the
algorithm obtains error estimates corresponding to the solid red and
blue lines for $x < \frac{1}{2}$. When the algorithm observes a point
$x > \frac{1}{2}$, it naturally selects the red abstaining expert since
it admits a better current estimated loss. However, for
$x > \frac{1}{2}$, the red expert is worse than the blue expert
$\xi_2$. Furthermore, it is abstaining and thus providing no updates
for expert $\xi_2$ (which is instead predicting). Hence, the algorithm
continues to maintain an estimate of $\xi_2$'s loss at the level of
the blue solid line indicated for $x < \frac{1}{2}$; it then continues to
select the red expert for all $x$s and incurs a high regret.\footnote
{
For the
sake of clarity, we did not introduce specific real values for the
expected loss of each expert on each of the half intervals, but that
can be done straightforwardly. We have also verified experimentally
with such values that the bias problem just pointed out indeed
leads to poor regret for \UCBNT.
}

This simple example shows that, unlike the adversarial scenario
(Section \ref{sec:adversarial}), $\GABS_t$, here, cannot depend on the
input $x_t$, and that, in general, the indiscriminate use of feedback
graphs may result in biased loss observations. On the other hand, we
know that if we were to avoid using feedback graphs at all (which is
always possible using \UCB), we would always be able to define
unbiased loss estimates.  A natural question is then: can we construct
time-varying feedback graphs that lead to unbiased loss observations?
In the next section, we show how to design such a sequence of
auxiliary feedback graphs, which in turn allows us to then extend
\UCBNT\ to the setting of time-varying feedback graphs for general
loss functions.  Under this assumption, we can achieve unbiased
empirical estimates of the average losses $\mu_j$ of the experts,
which will allow us to apply standard concentration bounds in the
proof of this algorithm.

\subsection{Time-varying graphs for \UCBNT}
\label{subsec:ucbn}

\begin{algorithm}[t]
\begin{algorithmic}
 \FOR{$t \geq 1$}
 \STATE \textsc{Receive}($x_t$);
 \STATE $\xi_{I_t} \leftarrow \argmin_{\xi_j \in \cE} \Big \{ \widehat{\mu}_{j,t-1} -  S_{j,t-1} \Big \}$;
 \FOR{$\xi_j\in\mathcal{E}$}
  \STATE  $Q_{j,t} \leftarrow \sum_{s=1}^{t} 1_{j\in N_s(I_s)}~$;
  \STATE $S_{j,t} \leftarrow \sqrt{\frac{5\,\log t}{Q_{j,t}} }$;
  \STATE ${\hat \mu}_{j,t} \leftarrow \frac{1}{Q_{j,t}}\sum_{s=1}^{t} L(\xi_j,z_s) 1_{j\in N_s(I_s)}$.
\ENDFOR
 \ENDFOR
\end{algorithmic}
\caption{\UCBNT}
\label{alg:ucbncaron}
\end{algorithm}
We now show that \UCBNT\ benefits from favorable guarantees, so long
as the feedback graph $\GABS_t$ it uses at time $t$ depends only on
events up to time $t - 1$. This extension works for general bounded
losses and does not only apply to our specific abstention loss $L$.

\ignore{
  A natural extension of \UCBN\ for this purpose consists of updating
  the upper-confidence bounds of each expert according to the feedback
  graph defined at round $t$. That is, t The \UCBN\ algorithm for
  time-varying feedback graphs is similar to the standard \UCBN\ of
  \citet{CaronKvetonLelargeBhagat2012} except that it updates its
  estimate of the losses according to the time-varying
  out-neighborhood $N_t(I_t)$ at each round $t$, which depends on the
  expert $\xi_{I_t}$ chosen at round $t$.
  Algorithm~\ref{alg:ucbncaron} shows the pseudocode where $Q_j(t)$ is
  the number of times expert $j$ has been observed and $S_{j,t-1}$ is
  a slack term. Please see Appendix~\ref{app:stochastic} for a formal
  definition of these quantities.
}
%
So, let us assume that the feedback graph in round $t$ (and the associated
out-neighborhoods $N_t(\cdot)$) in Algorithm \ref{alg:ucbncaron} only
depends on the observed losses $L(\xi_i,z_s)$ and inputs $x_s$, for
$s = 1, \ldots, t - 1$, and $i \in [K]$, and let us denote this
feedback graph by $G_t$, so as not to get confused with $\GABS_t$.
Under this assumption, we can derive strong regret guarantees for
\UCBNT\ with time-varying graphs using a newly introduced notion of admissible coverings. For
the feedback graph at time $G_t$, let $\cC_t$ be
a collection of subsets of $V$ covering $G_t$, such that $\forall C_t \in \cC_t$, $i,j\in C_t$ means that
$i \in N_t(j)$ and $j \in N_t(i)$. We denote such a collection an \emph{admissible covering} of $G_t$.
Let $\cF_t$ denote the set of all admissible coverings of $G_t$, and let $\cF = \cap_{t=1}^T \cF_t$, i.e.
the collection of \emph{shared admissible coverings} that apply across all time steps. Then by construction, for any $\cC \in \cF$ and $C \in \cC$,
$i,j \in C$ means that $i \in N_t(j)$ and $j \in N_t(i)$
for every $t \in [T]$. Note that the definition of $\cF$ is equivalent to considering the
set of edges that are shared across all $G_t$, and then considering admissible coverings
over the graph induced by these shared edges.
Moreover, since $\cup_{i=1}^K \{i\} \in \cF_t$ for every $t \in [T]$, $\cF$ is
always non-empty.

\begin{theorem}
\label{th:ucbn}
Assume that, for all $t \in[T]$, the feedback graph $G_t$ depends only
on information up to time $t - 1$.  Then, the regret
of \UCBNT\
is bounded as follows:
\[
\cO \Bigl( \E \Big[\min_{\cC \in \cF} \sum_{C \in \cC} \frac{\max_{j\in
  C} \Delta_j}{\min_{j \in C}\Delta_j^2} \log(T) + K \Big] \Bigl ) ~.
\]
\end{theorem}
%
The theorem gives a bound on the regret based on any admissible covering that
applies to every feedback graph seen during learning, and the minimum chooses
the admissible covering with the smallest regret. 

Theorem~\ref{th:ucbn} can be interpreted as an extension of Theorem 2
in \citet{CaronKvetonLelargeBhagat2012} to time-varying feedback
graphs. Its proof involves
showing that the use of feedback graphs $G_t$ that
depend only on information up to $t-1$ can result in unbiased
loss estimates, and it considers shared admissible coverings that apply across
the sequence of feedback graphs to derive a time-varying bound that leverages the shared
updates from the graph.

Moreover, the bound illustrates that if the feedback graphs in a problem admit a
shared admissible covering with a small number of elements $|\cC| \ll K$ (e.g. if the feedback graphs
can be decomposed into a small number of components that are fixed across time) for which
$\max_{j \in C} \Delta_j \approx \min_{j \in C} \Delta_j$,
then this bound can be up to a factor $\frac{|\cC|}{K}$ tighter than the
bound guaranteed by the standard UCB algorithm.  Moreover, this regret
guarantee is always more favorable than that of the standard UCB since
the (trivial) admissible covering that splits $V$ into $K$ singletons for
all $t$ is always an admissible covering of every $G_t$.
Furthermore, note that if the feedback graph is fixed throughout all rounds
and we interpret the doubly-directed edges as edges of an undirected
graph $G_U$, it follows that $\cF = \cF_t$ for every $t \in [T]$. Thus, we
straightforwardly obtain the following result, which
is comparable to Theorem 2 in \citep{CaronKvetonLelargeBhagat2012}.
\begin{corollary}
  If the feedback graph $G_t = G$ is fixed over time, then the
  guarantee of Theorem~\ref{th:ucbn} is upper-bounded by:
\[
\cO \Bigl( \min_{\mathcal{C}} \sum_{C \in \mathcal{C}} \frac{\max_{i \in C} \Delta_i }{\min_{
  i \in C}\Delta_i^2} \log(T) + K \Bigr )~,
\]
the outer minimum being over all admissible coverings $\mathcal{C}$ of
$G_U$.
\end{corollary}
\citet{CaronKvetonLelargeBhagat2012} present matching lower bounds for
the case of stochastic bandits with a fixed feedback graph. Since we
can again design abstention scenarios with fixed feedback graphs,
these bounds carry over to our setting.

Now, how can we use the results of this section to design an algorithm
for the abstention scenario? The natural feedback graphs we discussed
in Section \ref{sec:adversarial} are no longer applicable since
$\GABS_t$ depends on $x_t$. Nevertheless, we will present two
solutions to this problem. In Section \ref{subsec:subsetfeedback}, we
present a solution with a fixed graph $G$ that closely captures the
problem of learning with abstention.  Next, in
Section~\ref{subsec:ucbgt}, we will show how to define and leverage a
time-varying graph $G_t$ that is estimated based on past observations.

\subsection{\UCBN\ with the subset feedback graph}
\label{subsec:subsetfeedback}

In this section, we define a \emph{subset feedback graph}, $\GSUB$,
that captures the most informative feedback in the problem of learning
with abstention and yet is safe in the sense that it does not depend
on $x_t$. The definition of the graph is based on the following simple
observation: if the abstention region associated with $\xi_i$ is a
subset of that of $\xi_j$, then, if $\xi_i$ is selected at some round
$t$ and is abstaining, so is $\xi_j$. For an example, see $\xi_i$ and
$\xi_j$ in  Figure~\ref{f:aaa} (top). Crucially, this implication
holds regardless of the particular input point $x_t$ received in the
region of abstention of $\xi_i$. Thus, the set of vertices of $\GSUB$
is $\cE$, and $\GSUB$ admits an edge from $\xi_i$ to $\xi_j$, iff
$\set{x \in \cX \colon r_i(x) \leq 0} \subseteq \set{x \in \cX \colon r_j(x) \leq 0}$.
Since $\GSUB$ does not vary with time, it trivially verifies the
condition of the previous section. Thus, \UCBNT\ run with $\GSUB$
admits the regret guarantees of Theorem~\ref{th:ucbn}, where
we only need to consider the set of admissible coverings of the fixed graph $\GSUB$.

The example of Section~\ref{subsec:bias} illustrated a bias problem in
a special case where the feedback graphs $G_t$ were not subgraphs of
$\GSUB$. 
The following result shows more generally that feedback graphs not
included in $\GSUB$ may result in catastrophic regret behavior.

\begin{proposition}
\label{prop:subset}
Assume that \UCBNT\ is run with feedback graphs $G_t$ that are not
subsets of $\GSUB$. Then, there exists a family of predictors $\cH$, a
Lipschitz loss function $\ell$ in (\ref{e:absloss}), and a distribution $\cD$ over $z_t$s for
which \UCBNT\ incurs linear regret with arbitrarily high probability.
\end{proposition}
The proof of the proposition is given in Appendix~\ref{app:subset}.
In view of this result, no fixed feedback graph for \UCBNT\ can be
more informative than $\GSUB$. But how can we leverage past
observations (up to time $t - 1$) to derive a feedback graph that
would be more informative than the simple subset graph $\GSUB$? The
next section provides a solution based on feedback graphs estimated
based on past observations and a new algorithm.

\subsection{UCB-GT algorithm}
\label{subsec:ucbgt}

\begin{figure}[t]
\centering
\includegraphics[scale=0.39]{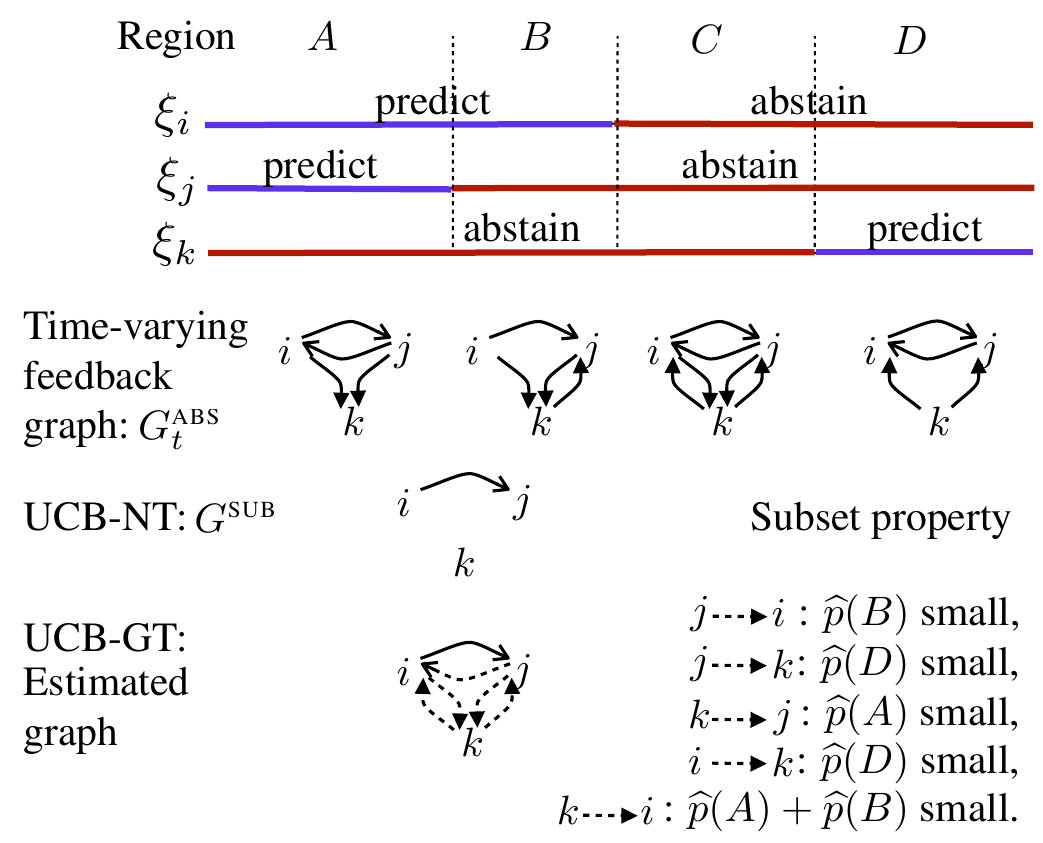}
\vskip -0.1in
\caption{The top row shows three experts $\xi_i$, $\xi_j$, and $\xi_k$
  on a one-dimensional input space marked by their prediction and
  abstention regions. Below each region, the time-varying graph
  $\GABS_t$ is shown.  To avoid the bias problem affecting the graphs
  $\GABS_t$, one option is to use $\GSUB$. Yet, as illustrated,
  $\GSUB$ is minimal and in this example admits only one edge
  (excluding self-loops). Thus, a better option is to use the
  time-varying graphs of \UCBGT\ since they are richer and more
  informative. For these graphs, an edge is added from $\xi_j$ to
  $\xi_i$ if the probability of the region where $\xi_j$ is abstaining
  but $\xi_i$ is predicting is (estimated to be) small.}
\vskip -0.15in
\label{f:aaa}
\end{figure}

We seek graphs $G_t$ that admit $\GSUB$ as a subgraph. We will show
how certain types of edges can be safely added to $\GSUB$ based on
past observations. This leads to a new algorithm, \UCBGT\ (\UCB\ with
estimated time-varying graph), whose pseudocode is given in
Algorithm~\ref{alg:ucbtype1}.
\begin{algorithm}[t]
\begin{algorithmic}
 \FOR{$t \geq 1$}
 \STATE \textsc{Receive}($x_t$);
 \STATE $\xi_{I_t} \leftarrow \argmin_{\xi_i \in\mathcal{E}} \big \{ \widehat{\mu}_{i,t - 1} -  S_{i,t - 1} \big \}$,\\
 where $S_{i,t - 1}$ is as in Algorithm \ref{alg:ucbncaron};
 \FOR {$\xi_i\in \mathcal{E}$}
   \STATE {\bf if}  $ \hat{p}^{t-1}_{I_t, i} \leq \gamma_{i,t-1} $ {\bf then} $Q_{i,t}\leftarrow Q_{i,t - 1}+1$; \\
    \hspace{0.15in}{\bf if} $r_{I_t}(x_t) \leq 0 \wedge r_i(x_t) > 0$ {\bf then}\\
    $\hspace{0.26in}\widehat{\mu}_{i,t} \leftarrow  \left(1-\frac{1}{Q_{i,t}}\right) \widehat{\mu}_{i,t - 1}$;\qquad\qquad\qquad (*) \\
  \hspace{0.15in}{\bf else} $\widehat{\mu}_{i,t} \leftarrow \frac{ L( \xi_i,z_t)}{Q_{i,t}}+ \left(1-\frac{1}{Q_{i,t}}\right) \widehat{\mu}_{i,t - 1}$;\\
    {\bf else} $Q_{i,t}\leftarrow Q_{i,t - 1}$,\,\,  $\widehat{\mu}_{i,t} \leftarrow  \widehat{\mu}_{i,t - 1}$~.
 \ENDFOR
 \ENDFOR
\end{algorithmic}
\caption{\UCBGT}
\label{alg:ucbtype1}
\end{algorithm}
As illustrated by Figure~\ref{f:aaa}, the key idea of \UCBGT\ is to
augment $\GSUB$ with edges from $\xi_j$ to $\xi_i$ where the subset
property
$\set{x \colon r_j(x) \leq 0} \subseteq \set{x \colon r_i(x) \leq 0}$
may not hold, but where the implication
$(r_j(x) \leq 0 \Rightarrow r_i(x) \leq 0)$ holds with high
probability over the choice of $x \in \cX$, that is, the region
$\set{x \colon r_j(x) \leq 0 \wedge r_i(x) > 0}$ admits low
probability.
Of course, adding such an edge $\xi_j \rightarrow \xi_i$ can cause the
estimation bias of Section~\ref{subsec:bias}.  But, if we
restrict ourselves to cases where $p_{j,i} = \Pr[r_j(x) \leq 0 \wedge r_i(x) >
0]$ is upper bounded by some carefully chosen quantity that changes
over rounds, the effect of this bias 
will be limited.
\ignore{
Now, if $\Pr[r_i(x) \leq 0 \wedge r_j(x) > 0] > 0$ and we add edge $\xi_i \rightarrow \xi_j$,
then the empirical estimates of the experts may be affected by
the bias problem described in Section~\ref{subsec:bias}.
However, if we only add this edge when
$\Pr[r_i(x) \leq 0 \wedge r_j(x) > 0]$ is upper bounded by
some carefully chosen quantity that changes over rounds,
the effect of this bias on the incurred regret will be
limited.
}
In reverse, as illustrated in Figure~\ref{f:aaa},
the resulting feedback graph can be substantially more
beneficial since it may have many more edges than $\GSUB$,
hence leading to more frequent updates of the experts'
losses and more favorable regret guarantees. This benefit
is further corroborated by our experimental results (Section
\ref{sec:experiments}).

Since we do not have access to $p_{j,i}$,
we use instead
empirical estimates
$\h p_{j, i}^{t -1} := \tfrac{1}{t-1} \sum_{s = 1}^{t-1} 1_{r_j(x_s) \leq 0, r_i(x_s) > 0}$.
At time $t$, if expert $\xi_j$ is selected, we
update expert $\xi_i$ if the condition
$\h p_{j, i}^{t-1} \leq \gamma_{i, t-1 }$ holds
with $\gamma_{i, t-1} = \sqrt{5 Q_i(t - 1) \log (t )}/ ( (K-1)(t-1) ). $
If the expert $\xi_{I_t}$ chosen abstains while expert $\xi_j$ predicts and
satisfies $\h p_{I _ t, j}^{t -1}\leq \gamma_{j, t-1 }$, then we do
not have access to the true label $y_t$. In that case, we update
optimistically our empirical estimate as if the expert had loss $0$ at
that round (Step (*) in Alg.~\ref{alg:ucbtype1}).

The feedback graph $G_t$ just described can be defined
via the out-neighborhood of vertex $\xi_j$:
$N_t(j) = \set{\xi_i \in \cE \colon\h p_{j, i}^{t-1} \leq \gamma_{i, t -1}}$.
The following regret guarantee holds for \UCBGT.
\begin{theorem}
\label{th:ucbabs}
For any $t \in [T]$, let the feedback graph $G_t$ be defined by the out-neighborhood
$N_t(j) = \set{\xi_i \in \cE \colon \h p_{j, i}^{t -1} \leq  \gamma_{i,t-1 }}$.
Then, the regret of \UCBGT\ is bounded as follows:
\begin{equation*}
 \cO\Bigl( \E \Big[ \min_{\cC \in \cF} \sum_{C \in \cC} \frac{\max_{j\in C }  \Delta_j}{\min_{j \in C }\Delta_j^2} \log(T) + K \Big] \Bigr )~.
\end{equation*}
\end{theorem}
Since the graph $G_t$ of \UCBGT\ has more edges than
$\GSUB$, it admits at least as many admissible coverings as $\GSUB$,
which leads to a more favorable guarantee than that of $\UCBNT$ run
with $\GSUB$. The proof of this result differs from the standard UCB analysis and that of Theorem~\ref{th:ucbn} in that it
involves showing that the \UCBGT\ algorithm can adequately control the amount of
bias introduced by the skewed loss estimates. The experiments in the next section provide an empirical validation of
this theoretical comparison.

\begin{figure*}[t]
\begin{center}
\begin{tabular}{c@{\hspace{1.5cm}}c@{\hspace{1.5cm}}c}
\includegraphics[scale=0.162,trim= 5 10 10 5, clip=true]{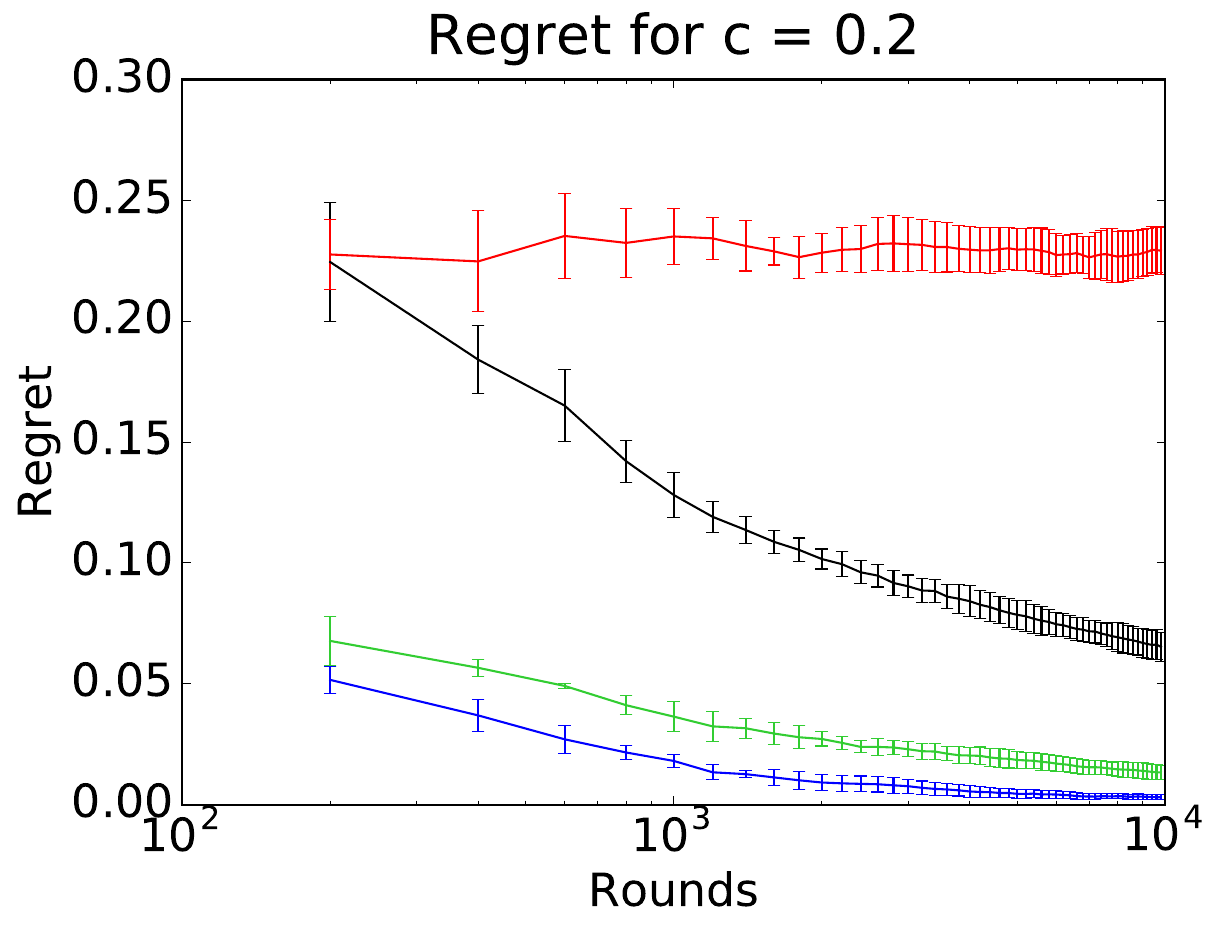} &
\hspace*{-5mm} \includegraphics[scale=0.162,trim= 5 10 10 5, clip=true]{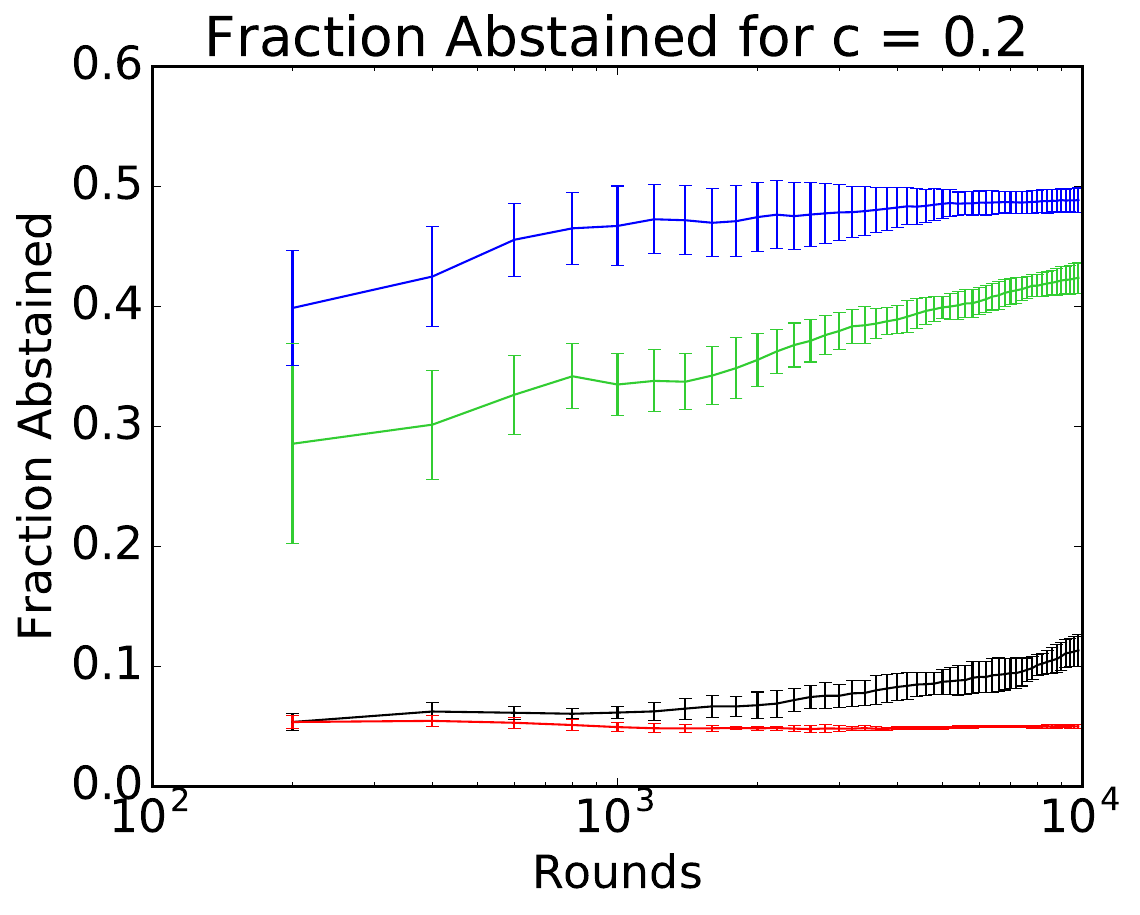} & 
\hspace*{-5mm}\includegraphics[scale=0.162,trim= 5 10 10 5, clip=true]{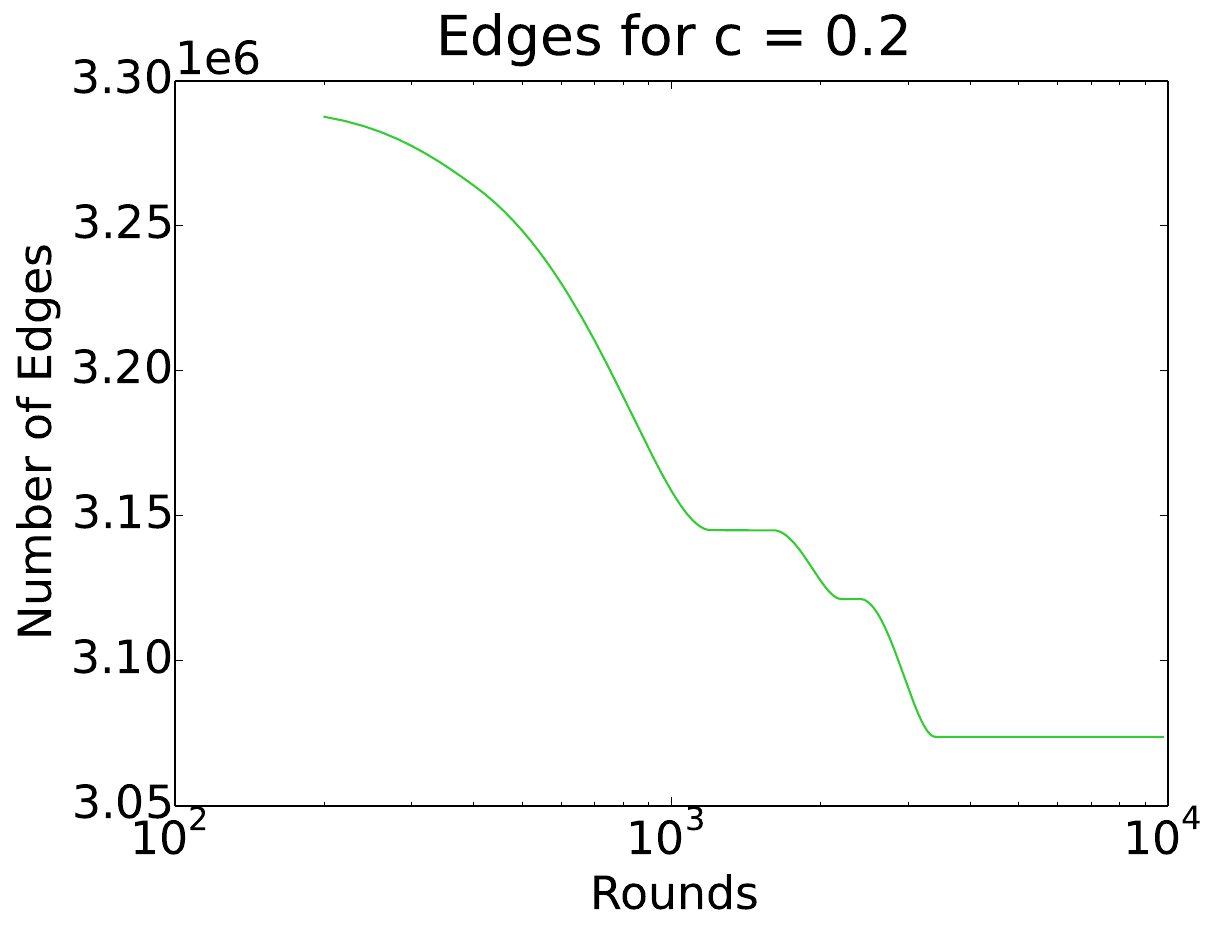} \\
\includegraphics[scale=0.162,trim= 5 10 10 5, clip=true]{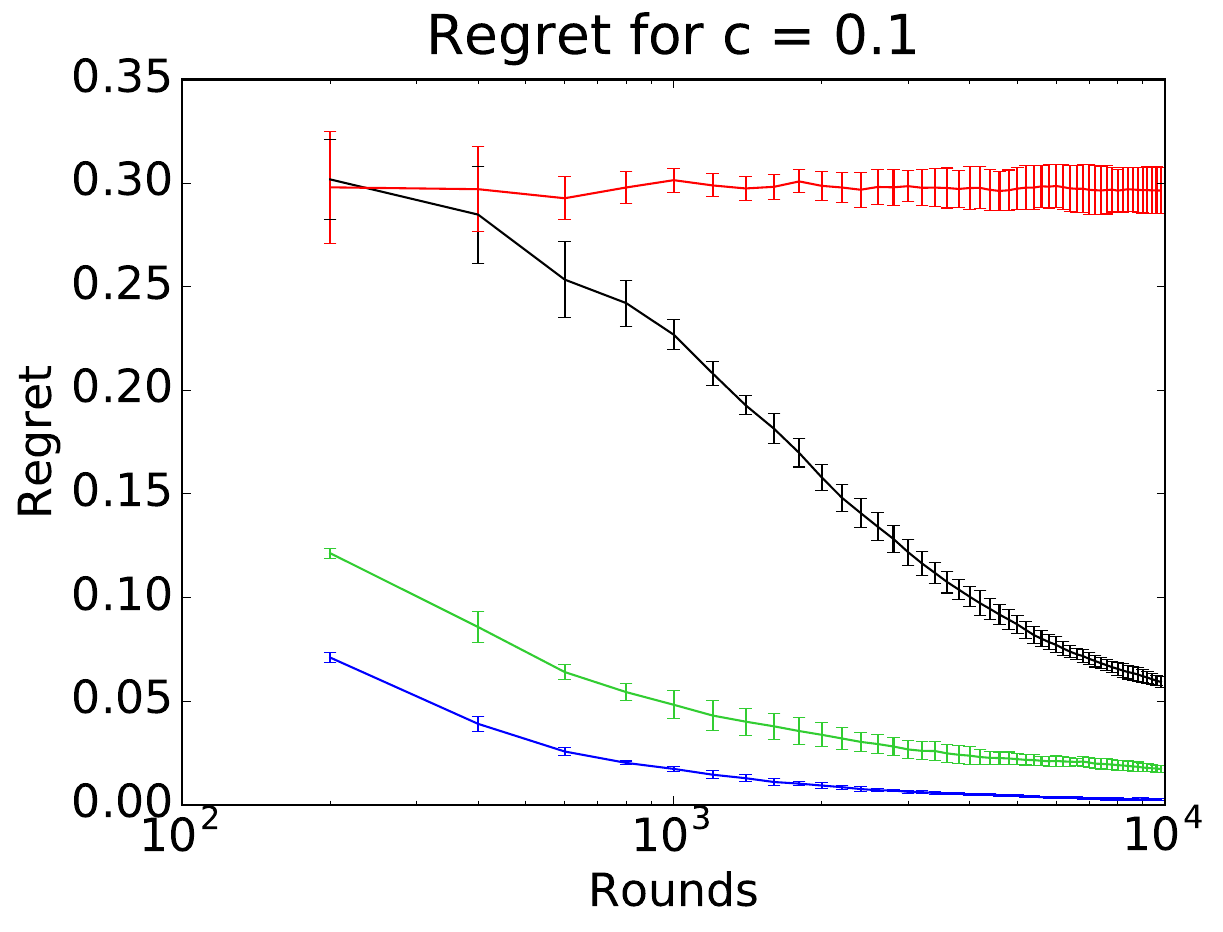} & 
\hspace*{-5mm} \includegraphics[scale=0.162,trim= 5 10 10 5, clip=true]{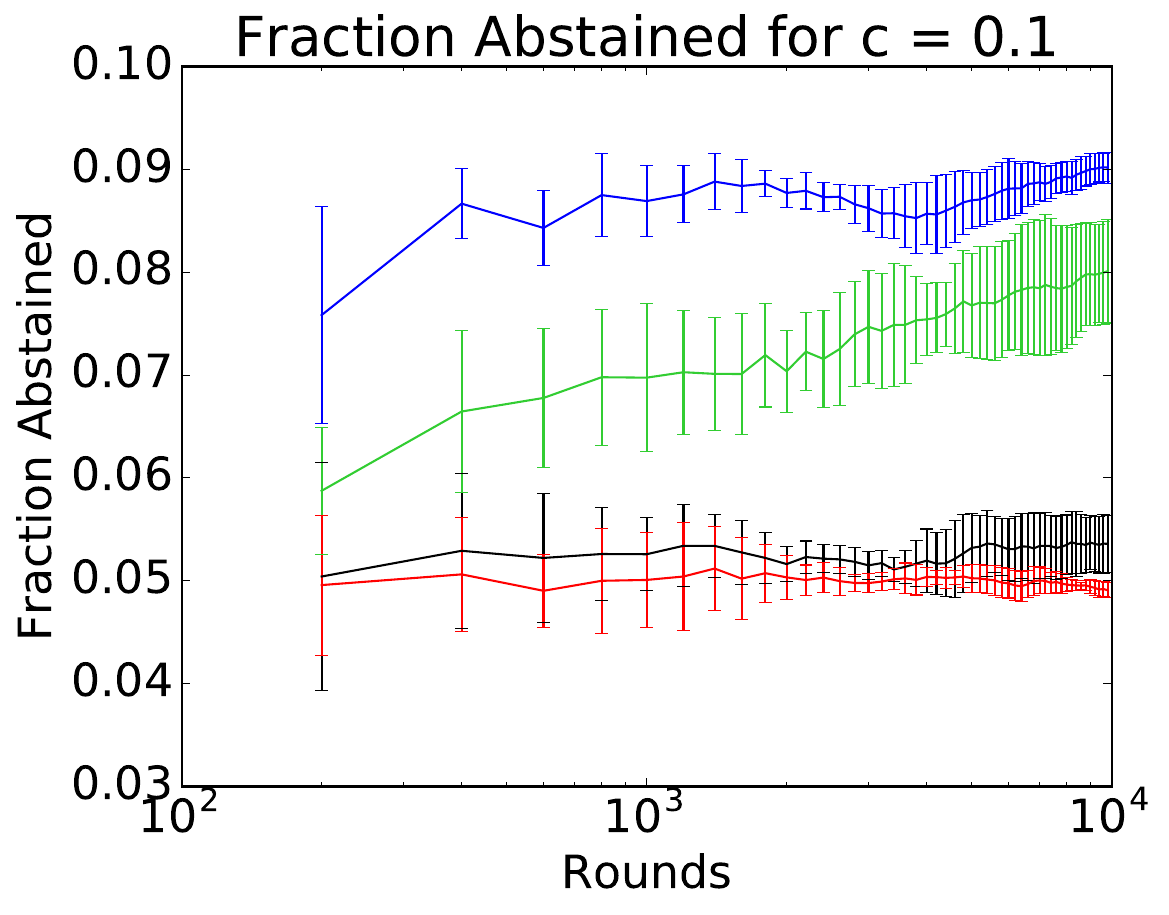} & 
\hspace*{-5mm}\includegraphics[scale=0.162,trim= 5 10 10 5, clip=true]{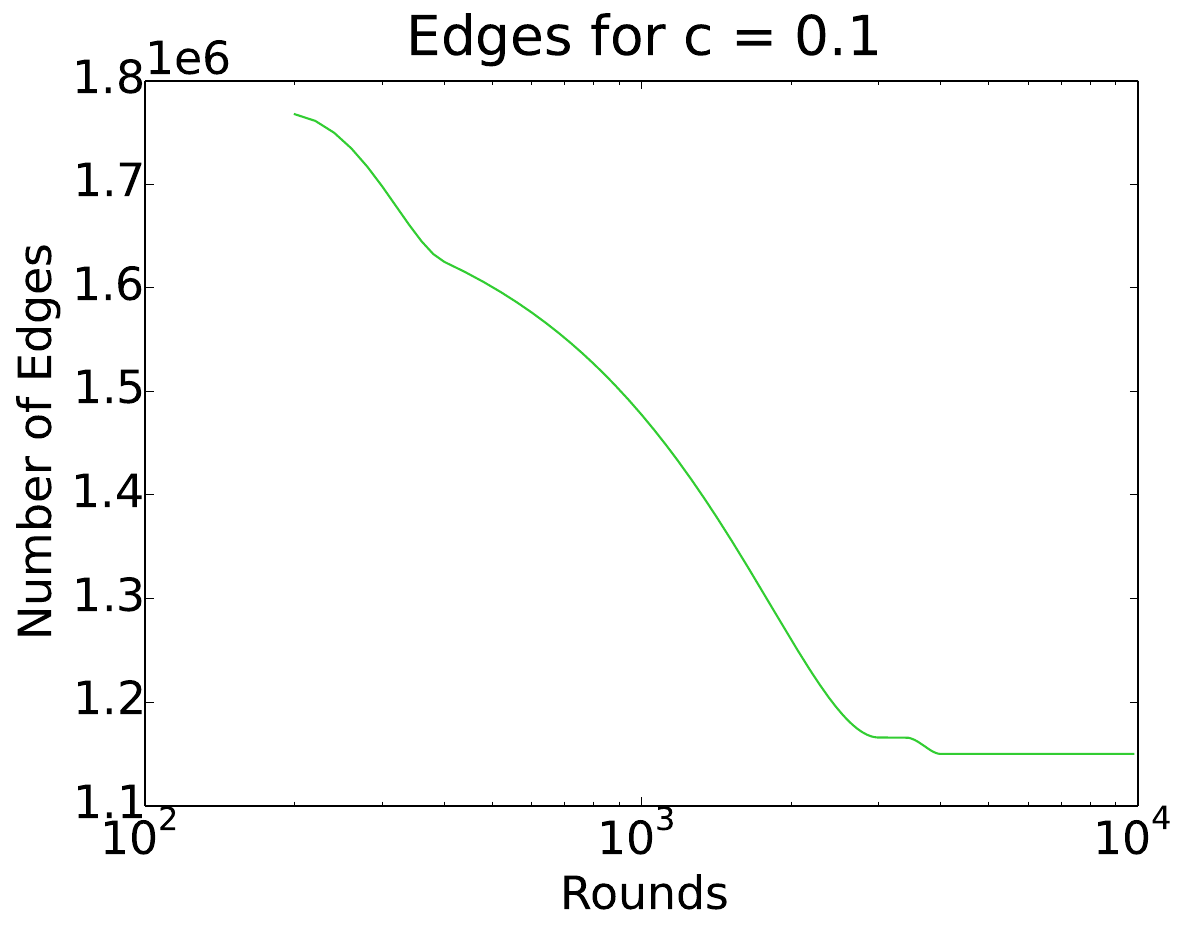} \\
\end{tabular}
\end{center}
\vskip -0.1in
\caption{From the left, we show graphs of the average regret
  $R_t(\cdot)/t$, fraction of points the chosen expert abstained on,
  and the number of edges of the feedback graph as a function of $t$
  (log-scale) for {\color[rgb]{0.16,0.67,0.16}\UCBGT }, \UCBNT ,
  {\color{red} \UCB }, and {\color{blue} \FTL }.  Top row is the
  results for {\tt cod-rna} for cost $c=0.2$ and bottom row is the
  {\tt guide} for cost $c=0.1$. More results are in Appendix
  \ref{app:expresults}.}
\label{fig:expmain}
\vskip -0.2in
\end{figure*}

\section{Experiments}
\label{sec:experiments}

In this section, we report the results of several experiments on ten
datasets comparing \UCBGT, \UCBNT\ with feedback graph $\GSUB$,
vanilla \UCB\ (with no sharing information across experts), as well as Full-Supervision, \FTL. 
\FTL\ is an algorithm that at each round chooses the expert $\xi_j$ with the smallest abstention
loss so far, ${\widehat \mu}_{j,t-1}$, and even if this expert abstains, the algorithm
receives the true label and can update the empirical abstention loss
estimates for all experts. \FTL\ reflects an unrealistic and overly optimistic scenario that
clearly falls outside the abstention setting, but it provides an upper
bound for the best performance we may hope for. 

We used the following eight datasets from the UCI data repository: {\small
{\tt HIGGS}, {\tt phishing}, {\tt ijcnn}, {\tt covtype}, {\tt eye},
{\tt skin}, {\tt cod-rna},} and {\small {\tt guide}}. We also used the {\small \tt
  CIFAR} dataset from \cite{Krizhevsky2009}, where we extracted the
first twenty-five principal components and used their projections
as features, and a synthetic dataset of points drawn according to
the uniform distribution in $[-1,1]^2$.
For each dataset, we generated a total of $K = 2\mathord{,}100$
experts and all the algorithms were tested for a total of
$T = 10\mathord{,}000$ rounds. The experts, $\xi = (h, r)$, were
chosen in the following way. The predictors $h$ are hyperplanes
centered at the origin whose normal vector in $\mathbb{R}^d$ is drawn
randomly from the Gaussian distribution, $\mathcal{N}(0, 1)^d$, where
$d$ is the dimension of the feature space of the dataset. The
abstention functions $r$ are concentric annuli around the origin with
radii in
$(0, \tfrac{\sqrt{d}}{20},\tfrac{2\sqrt{d}}{20} \ldots,
\sqrt{d})$. For each dataset, we generated $100$ predictors and each
predictor $h$ is paired with the 21 abstention functions $r$.  For a
fixed set of experts, we first calculated the regret by averaging over
five random draws of the data, where the best-in-class expert was
determined in hindsight as the one with the minimum average cumulative
abstention loss. We then repeated this experiment five times over
different sets of experts and averaged the results. We report these
results for $c \in \set{0.1, 0.2, 0.3}$.

Figure~\ref{fig:expmain} shows the averaged regret $R_t(\cdot)/t$ with
standard deviations across the five repetitions for the different
algorithms as a function of $t \in [T]$ for two datasets. In
Appendix~\ref{app:expresults}, we present plots of the regret for all
ten datasets. These results show that \UCBGT\ outperforms both \UCBNT\
and \UCB\ on all datasets for all abstention cost values.  Remarkably,
\UCBGT's performance is close to that of \FTL\ for most datasets,
thereby implying that \UCBGT\ attains almost the best regret that we
could hope for. We also find that \UCBNT\ performs better than the
vanilla \UCB.

Figure~\ref{fig:expmain} also illustrates the fraction of points in
which the chosen expert abstains, as well as the number of edges in
the feedback graph as a function of rounds. We only plot the number of
edges of \UCBGT\, since that is the only graph that varies with
time. For both experiments depicted and in general for the rest of the
datasets, the number of edges for \UCBGT\ is between 1 million to 3
million, which is at least a factor of 5 more than for \UCBNT, where
the number of edges we observed are of the order
$200\mathord,000$. \FTL\ enjoys the full information property and the
number of edges is fixed at 4 million (complete graph).  The increased
information sharing of \UCBGT\ is clearly a strong contributing factor
to the algorithm's improvement in regret relative to \UCBNT. In
general, we find that, provided that the estimation bias is
controlled, the higher is the number of edges, the smaller the regret.
Regarding the value of the cost $c$, as expected, we observe that the
fraction of points that the chosen expert abstains on always decreases
as $c$ increases, but also that that fraction depends on the dataset
and the experts used.

\ignore{Finally, Appendix D also provides more experiments analyzing different aspects of the problem. We tested how the choice of experts and the number of experts might impact our results as well as extreme values of the abstention cost, $c\in \{0.001, 0.9\}$.}

Finally, Appendix~\ref{app:expresults} includes more experiments for
different aspects of the problem. In particular, we tested how the
number of experts or a different choice of experts (confidence-based
experts) affected the results. We also experimented with some extreme
abstention costs and, as expected, found the fraction of abstained
points to be large for $c = 0.001$ and small for $c = 0.9$. In all of
these additional experiments, \UCBGT\ outperformed \UCBNT.

\section{Conclusion}
We presented a comprehensive analysis of the novel setting of online
learning with abstention, including algorithms with favorable
guarantees both in the stochastic and adversarial scenarios, and
extensive experiments demonstrating the performance of \UCBGT\ in
practice. Our algorithms and analysis can be straightforwardly
extended to similar problems, including the multi-class and regression
settings, as well as other related scenarios, such as online learning with
budget constraints.
A key idea behind the design of our algorithms in the stochastic
setting is to leverage the stochastic sequence of feedback graphs.
This idea can perhaps be generalized and applied to other problems
where time-varying feedback graphs naturally appear. Furthermore, our regret guarantees can be instead expressed in terms
of the independence number of time-varying graphs by proceeding
as in \cite{LykourisTardosWali2019}.

\ignore{
We presented a comprehensive analysis of the novel setting of online
learning with abstention, showing that efficient learning is
possible in both the adversarial and stochastic scenarios. We drew
connections between this new online setting and existing work
involving time-varying feedback graphs, generalizing and extending
prior work while resolving the issue of biased loss observations. We
presented a novel algorithm, \UCBGT, that carefully uses
biased estimates and which admits favorable regret guarantees. 

Finally, we
presented a thorough experimental comparison showing that
\UCBGT\ significantly outperforms \UCBNT\ with the fixed graph $\GABS$ (an adaptation
to the abstention scenario of a competitor available in the literature),
and achieves performance that is close to an unrealistic and overly optimistic
full-information benchmark. Finally, we remark that the concept of online
learning with abstention is general. This work can
be extended to similar problems, including the multi-class and
regression settings, as well as other scenarios, such as online
learning with budget constraints.
}

\clearpage

\bibliography{olr}

\begin{thebibliography}{29}
\providecommand{\natexlab}[1]{#1}
\providecommand{\url}[1]{\texttt{#1}}
\expandafter\ifx\csname urlstyle\endcsname\relax
  \providecommand{\doi}[1]{doi: #1}\else
  \providecommand{\doi}{doi: \begingroup \urlstyle{rm}\Url}\fi

\bibitem[Alon et~al.(2013)Alon, Cesa-Bianchi, Gentile, and
  Mansour]{AlonCesaBianchiGentileMansour2013}
Alon, N., Cesa-Bianchi, N., Gentile, C., and Mansour, Y.
\newblock From bandits to experts: A tale of domination and independence.
\newblock In \emph{NIPS}, 2013.

\bibitem[Alon et~al.(2014)Alon, Cesa-Bianchi, Gentile, Mannor, Mansour, and
  Shamir]{AlonCesaGentileMannorMansourShamir2014}
Alon, N., Cesa-Bianchi, N., Gentile, C., Mannor, S., Mansour, Y., and Shamir,
  O.
\newblock Nonstochastic multi-armed bandits with graph-structured feedback.
\newblock In \emph{CoRR}, 2014.

\bibitem[Alon et~al.(2015)Alon, Cesa-Bianchi, Dekel, and
  Koren]{AlonCesaBianchiDekelKoren2015}
Alon, N., Cesa-Bianchi, N., Dekel, O., and Koren, T.
\newblock Online learning with feedback graphs: Beyond bandits.
\newblock \emph{JMLR}, 2015.

\bibitem[Auer et~al.(2002{\natexlab{a}})Auer, Cesa-Bianchi, and Fischer]{acf02}
Auer, P., Cesa-Bianchi, N., and Fischer, P.
\newblock Finite-time analysis of the multi-armed bandit problem.
\newblock \emph{Mach. Learn.}, 47\penalty0 (2-3):\penalty0 235--256,
  2002{\natexlab{a}}.

\bibitem[Auer et~al.(2002{\natexlab{b}})Auer, Cesa-Bianchi, Freund, and
  Schapire]{AuerCesaBianchiFreundSchapire2002}
Auer, P., Cesa-Bianchi, N., Freund, Y., and Schapire, R.~E.
\newblock The nonstochastic multi-armed bandit problem.
\newblock \emph{SIAM J. Comput.}, 32\penalty0 (1):\penalty0 48--77,
  2002{\natexlab{b}}.

\bibitem[Auer et~al.(2003)Auer, Cesa-Bianchi, Freund, and Schapire]{Exp3}
Auer, P., Cesa-Bianchi, N., Freund, Y., and Schapire, R.~E.
\newblock The nonstochastic multi-armed bandit problem.
\newblock \emph{SIAM J. Comput.}, 32\penalty0 (1):\penalty0 48--77, 2003.

\bibitem[Bartlett \& Wegkamp(2008)Bartlett and Wegkamp]{BartlettWegkamp2008}
Bartlett, P. and Wegkamp, M.
\newblock Classification with a reject option using a hinge loss.
\newblock \emph{JMLR}, pp.\  291--307, 2008.

\bibitem[Bubeck \& Cesa-Bianchi(2012)Bubeck and Cesa-Bianchi]{Bubeck2012}
Bubeck, S. and Cesa-Bianchi, N.
\newblock Regret analysis of stochastic and nonstochastic multi-armed bandit
  problems.
\newblock \emph{Foundations and Trends in Machine Learning}, 5\penalty0
  (1):\penalty0 1--122, 2012.

\bibitem[Caron et~al.(2012)Caron, Kveton, Lelarge, and
  Bhagat]{CaronKvetonLelargeBhagat2012}
Caron, S., Kveton, B., Lelarge, M., and Bhagat, S.
\newblock Leveraging side observations in stochastic bandits.
\newblock In \emph{UAI}, 2012.

\bibitem[Cesa-Bianchi et~al.(2017)Cesa-Bianchi, Gaillard, Gentile, and
  Gerchinovitz]{CesaBianchiGaillardGentileGerchinovitz2017}
Cesa-Bianchi, N., Gaillard, P., Gentile, C., and Gerchinovitz, S.
\newblock Algorithmic chaining and the role of partial feedback in online
  nonparametric learning.
\newblock In \emph{MLR}, 2017.

\bibitem[Chow(1957)]{Chow1957}
Chow, C.
\newblock An optimum character recognition system using decision function.
\newblock \emph{IEEE T. C.}, 1957.

\bibitem[Chow(1970)]{Chow1970}
Chow, C.
\newblock On optimum recognition error and reject trade-off.
\newblock \emph{IEEE T. C.}, 1970.

\bibitem[Clarkson(2006)]{Clarkson06nearest-neighborsearching}
Clarkson, K.~L.
\newblock Nearest-neighbor searching and metric space dimensions.
\newblock In \emph{Nearest-Neighbor Methods for Learning and Vision: Theory and
  Practice}. MIT Press, 2006.

\bibitem[Cohen et~al.(2016)Cohen, Hazan, and Koren]{CohenHazanKoren2016}
Cohen, A., Hazan, T., and Koren, T.
\newblock Online learning with feedback graphs without the graphs.
\newblock In \emph{ICML}, 2016.

\bibitem[Cortes et~al.(2016{\natexlab{a}})Cortes, DeSalvo, and
  Mohri]{CortesDeSalvoMohri2016a}
Cortes, C., DeSalvo, G., and Mohri, M.
\newblock Learning with rejection.
\newblock In \emph{ALT}, pp.\  67--82. Springer, Heidelberg, Germany,
  2016{\natexlab{a}}.

\bibitem[Cortes et~al.(2016{\natexlab{b}})Cortes, DeSalvo, and
  Mohri]{CortesDeSalvoMohri2016b}
Cortes, C., DeSalvo, G., and Mohri, M.
\newblock Boosting with abstention.
\newblock In \emph{NIPS}. MIT Press, 2016{\natexlab{b}}.

\bibitem[El-Yaniv \& Wiener(2010)El-Yaniv and Wiener]{YanivWiener2010}
El-Yaniv, R. and Wiener, Y.
\newblock On the foundations of noise-free selective classification.
\newblock \emph{JMLR}, 2010.

\bibitem[El-Yaniv \& Wiener(2011)El-Yaniv and Wiener]{YanivWiener2011}
El-Yaniv, R. and Wiener, Y.
\newblock Agnostic selective classification.
\newblock In \emph{NIPS}, 2011.

\bibitem[Hazan(2016)]{hazan16}
Hazan, E.
\newblock \emph{Introduction to Online Convex Optimization}.
\newblock Foundations and Trends in Optimization. Now Publishers Inc., 2016.

\bibitem[Hazan \& Megiddo(2007)Hazan and Megiddo]{hm07}
Hazan, E. and Megiddo, N.
\newblock Online learning with prior knowledge.
\newblock In \emph{COLT}, pp.\  499--513, 2007.

\bibitem[Koc{\'a}k et~al.(2014)Koc{\'a}k, Neu, Valko, and
  Munos]{KocakNeuValkoMunos2014}
Koc{\'a}k, T., Neu, G., Valko, M., and Munos, R.
\newblock Efficient learning by implicit exploration in bandit problems with
  side observations.
\newblock In \emph{NIPS}, pp.\  613--621, 2014.

\bibitem[Krizhevsky et~al.(2009)Krizhevsky, Nair, and Hinton]{Krizhevsky2009}
Krizhevsky, A., Nair, V., and Hinton, G.
\newblock {CIFAR}-10 ({C}anadian {I}nstitute for {A}dvanced {R}esearch), 2009.
\newblock URL \url{http://www.cs.toronto.edu/~kriz/cifar.html}.

\bibitem[Li et~al.(2008)Li, Littman, and Thomas]{LiLittmanWalsh2008}
Li, L., Littman, M., and Thomas, W.
\newblock {K}nows {W}hat {I}t {K}nows: A framework for self-aware learning.
\newblock In \emph{ICML}, 2008.

\bibitem[Littlestone \& Warmuth(1994)Littlestone and
  Warmuth]{LittlestoneWarmuth1994}
Littlestone, N. and Warmuth, M.~K.
\newblock The weighted majority algorithm.
\newblock \emph{Information and computation}, 108\penalty0 (2):\penalty0
  212--261, 1994.

\bibitem[Lykouris et~al.(2019)Lykouris, Tardos, and
  Wali]{LykourisTardosWali2019}
Lykouris, T., Tardos, E., and Wali, D.
\newblock Feedback graph regret bounds for thompson sampling and ucb.
\newblock In \emph{ArXiv}, 2019.

\bibitem[Mannor \& Shamir(2011)Mannor and Shamir]{MannorShamir2011}
Mannor, S. and Shamir, O.
\newblock From bandits to experts: On the value of side-observations.
\newblock In \emph{NIPS}, pp.\  291--307, 2011.

\bibitem[Neu(2015)]{Neu2015}
Neu, G.
\newblock Explore no more: Improved high-probability regret bounds for
  non-stochastic bandits.
\newblock In \emph{NIPS}, pp.\  3168--3176, 2015.

\bibitem[Sayedi et~al.(2010)Sayedi, Zadimoghaddam, and
  Blum]{SayediZadimoghaddamBlum2010}
Sayedi, A., Zadimoghaddam, M., and Blum, A.
\newblock Trading off mistakes and don't-know predictions.
\newblock In \emph{NIPS}, 2010.

\bibitem[Zhang \& Chaudhuri(2016)Zhang and Chaudhuri]{ZhangKamalika2016}
Zhang, C. and Chaudhuri, K.
\newblock The extended {L}ittlestone's dimension for learning with mistakes and
  abstentions.
\newblock In \emph{COLT}, 2016.

\end{thebibliography}
\bibliographystyle{icml2018}

\FloatBarrier
\newpage
\appendix

\onecolumn
\section{Further Related Work}
\label{sec:relatedworks}

\begin{figure}[t]
\centering
\includegraphics[scale=0.3]{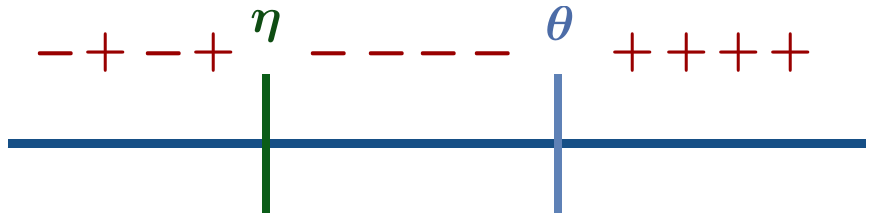}
\vskip -0.1in
\caption{Simple example of the benefits of learning with abstention
  \citep{CortesDeSalvoMohri2016a}.}
\vskip -0.2in
\label{fig:example}
\end{figure}

Learning with abstention is a useful paradigm in applications where
the cost of misclassifying a point is high. More concretely, suppose
the cost of abstention $c$ is less than $1/2$ and consider the set of
points along the real line illustrated in Figure~\ref{fig:example}
where $+$ and $-$ indicate their labels. The best threshold classifier
is the hypothesis given by threshold $\theta$, since it correctly
classifies points to the right of $\eta$, with an expected loss of
$(1/2) \Pr[x \leq \eta]$. On the other hand, the best abstention pair
$(h, r)$ would abstain on the region left of $\eta$ and correctly
classify the rest, with an expected loss of $c\Pr(x \leq \eta)$. Since
$c < 1/2$, the abstention pair always admits a better loss then the best
threshold classifier.

Within the online learning literature, work related to our scenario
includes the KWIK (\emph{knows what it knows}) framework of
\citet{LiLittmanWalsh2008} in which the learning algorithm is required
to make only correct predictions but admits the option of abstaining
from making a prediction. The objective is then to learn a concept
exactly with the fewest number of abstentions.  If in our framework we
received the label at every round, KWIK could be seen as a special
case of our framework for online learning with abstention with an
infinite misclassification cost and some finite abstention cost. A
relaxed version of the KWIK framework was introduced and analyzed by
\citet{SayediZadimoghaddamBlum2010} where a fixed number $k$ of
incorrect predictions are allowed with a learning algorithm related to
the solution of the 'mega-egg game puzzle'.  A theoretical
analysis of learning in this framework was also recently given by
\citet{ZhangKamalika2016}.  Our framework does not strictly cover this
relaxed framework. However, for some choices of the misclassification
cost depending on the horizon, the framework is very close to ours.
The analysis in these frameworks was given in terms of mistake bounds
since the problem is assumed to be realizable. We will not restrict
ourselves to realizable problems and, instead, will provide regret
guarantees.

\section{Additional material for the adversarial setting}
\label{app:adversarial}

We first present the pseudocode and proofs for the finite arm setting
and next analyze the infinite arm setting.

\subsection{Finite arm setting}
Algorithm~\ref{alg:exp3rej} contains the pseudocode for
\EXPABS, an algorithm for online learning with abstention
under an adversarial data model that guarantees small
regret. The algorithm itself is a simple adaptation of the ideas in \cite{AlonCesaGentileMannorMansourShamir2014,AlonCesaBianchiDekelKoren2015}, where we incorporate the side information that the loss of an abstaining arm is always observed, while the loss of a predicting arm is observed only if the algorithm actually plays a predicting arm. In the pseudocode and in the proof that follows, $L_t(\xi_j)$ is a shorthand for $L(\xi_j,(x_t,y_t))$.

\begin{algorithm}[t]
  \caption{\EXPABS}
\label{alg:exp3rej}
\begin{algorithmic}
  \INPUT Set of experts $\cE = \{\xi_1,\ldots, \xi_K\}$; learning rate $\eta > 0$~; \\
  {\bf Init:}  $q_1$ is the uniform distribution over $\cE$~;\\
\FOR{$t\leftarrow 1, 2, \ldots$}
    \STATE \textsc{Receive}($x_t$);
    \STATE $\xi_{I_t}  \gets $ \textsc{Sample}($q_t$);
    \IF {$r_{I_t}(x_t) > 0$}
      \STATE \textsc{Receive}($y_t$);
    \ENDIF
    \STATE For all $\xi_j= (h_j,r_j)$, set :
    \begin{align*}
    &P_t(\xi_j) \leftarrow
    \begin{cases}
    1 &{\mbox{if $r_j(x_t) \leq 0$}}\\
    \sum_{\xi_i\in \mathcal{E}\,:\, r_i(x_t) > 0} q_t(\xi_i) & {\mbox{if $r_j(x_t) > 0$}}~,
    \end{cases}\hspace{3.0in}\\
    &\widehat{L}_t(\xi_j) \leftarrow \frac{L_t(\xi_j)}{P_t(\xi_j)} \left(1_{r_{I_t}(x_t) \leq 0}1_{r_j(x_t) \leq 0} + 1_{r_{I_t}(x_t) > 0}\right)~,\\
    &q_{t+1}(\xi_j) \leftarrow \frac{q_t(\xi_j) \exp(-\eta \widehat{L}_t(\xi_j)) }{\sum_{\xi_i\in \mathcal{E}} q_t(\xi_i ) \exp (-\eta \widehat{L}_t(\xi_i))}~.
    \end{align*}
\ENDFOR
\end{algorithmic}
\end{algorithm}


\noindent{\bf Proof of Theorem \ref{th:exp3rej}.}\\
\begin{proof}
  By applying the standard regret bound of Hedge (e.g., \cite{Bubeck2012}) to distributions
  $q_1,\ldots, q_T$ generated by \EXPABS\ and to the
  non-negative loss estimates $\widehat{L}_t(\xi_j)$, the following holds:
\begin{align}
\label{eq:hedge}
&\E \Bigg [ \sum_{t=1}^T \sum_{\xi_j\in \mathcal{E}} q_t(\xi_j) \E\Big[\widehat{L}_t(\xi_j)\Big]- \sum_{t=1}^T \E\Big[\widehat{L}_t(\xi^\star)\Big] \Bigg ] \leq \frac{\log K}{\eta} + \frac{\eta}{2}\sum_{t=1}^T \E \left [  \sum_{\xi_j\in \mathcal{E}} q_t(\xi_j) \E\left[\widehat{L}_t(\xi_j)^2\right] \right ],
\end{align}
for any fixed $\xi^\star \in \mathcal{E}$. Using the fact that
$\E\Big[\widehat{L}_t(\xi_j)\Big]=L_t( \xi_j)$ and
$ \E\Big[\widehat{L}_t(\xi_j)^2\Big]=\frac{L_t( \xi_j)^2}{P_t(\xi_j)}$, we can write
\begin{align*}
\E \Bigg [ \sum_{t=1}^T \sum_{\xi_j\in \cE} q_t(\xi_j) L_t( \xi_j) - \sum_{t=1}^T L_t( \xi^\star)\Bigg ]
\leq
\frac{\log K}{\eta} + \frac{\eta}{2}\,\sum_{t=1}^T \E \Bigg [  \sum_{\xi_j\in\cE} \frac{q_t(\xi_j)}{P_t(\xi_j)} L_t(\xi_j)^2 \Bigg ].
\end{align*}
%
For each $t$, we can split the nodes $V$ of $\GABS_t$ into the two subsets $V_{abs,t}$ and $V_{acc,t}$ where if a node $\xi_j$
is abstaining at time $t$ then $\xi_j \in V_{abs,t}$, and otherwise $\xi_j \in V_{acc,t}$. Thus, for any round $t$, we can write
\begin{align*}
\sum_{\xi_j\in \cE} \frac{q_t(\xi_j)}{P_t(\xi_j)} L_t( \xi_j)^2
& =     \sum_{\xi_j\in V_{abs,t}} \frac{q_t(\xi_j)}{P_t(\xi_j)} L_t( \xi_j)^2  + \sum_{\xi_j\in V_{acc,t}} \frac{q_t(\xi_j)}{P_t(\xi_j)} L_t(\xi_j)^2 \\
& \leq  \sum_{\xi_j\in V_{abs,t}} q_t(\xi_j)\,c^2+ \sum_{\xi_j\in V_{acc,t}} \frac{q_t(\xi_j)}{P_t(\xi_j)} \\
& \leq c^2 +1~.
\end{align*}
The first inequality holds since if $\xi_j$ is an abstaining expert at time $t$,
we know that $L_t(\xi_j)=c$ and $P_t(\xi_j) =1$, while for the accepting experts we know
that $L_t(\xi_j)\leq 1$ anyway. The second inequality holds because if $\xi_j$ is an
accepting expert, we have $P_t(\xi_j) = \sum_{\xi_j\in V_{acc,t}} q_t(\xi_j)$.
Combining this inequality with \eqref{eq:hedge} concludes the proof.
\end{proof}

\subsection{Infinite arm setting}
\label{app:adversarialinfinite}

Here, the input space $\cX$ is assumed to be totally bounded, so that
there exists a constant $C_{\cX} > 0$ such that, for all
$0 < \ve \leq 1$, $\cX$ can be covered with at most $C_{\cX}\ve^{-d}$
balls of radius $\ve$. Let $\cY$ be a shorthand for $[-1, 1]^2$, the
range space of the pairs $(h, r)$. An $\ve$-covering $\cY_{\epsilon}$
of $\cY$ with respect to the Euclidean distance on $\cY$ has size
$K_{\epsilon} \leq C_{\cY}\ve^{-2}$ for some constant $C_{\cY}$.

\begin{algorithm}[t]
\begin{algorithmic}
\INPUT Ball radius $\ve > 0$, $\ve$-covering $\cY_{\ve}$ of $\cY$ such that $|\cY_{\ve}| \le C_{\cY}\,\ve^{-2}$;
  \FOR{$t=1,2,\dots$}
\STATE \textsc{Receive}($x_t$);
\STATE If $x_t$ does not belong to any existing ball, create new ball of radius $\ve$ centered on $x_t$, and allocate fresh instance of \EXPABS;
\STATE Let ``Active \EXPABS" be the instance allocated to the existing ball whose center $x_s$ is closest to $x_t$;
\STATE Draw action $\xi_{I_t} \in \cY_{\ve}$ using Active \EXPABS;
\STATE Get loss feedback associated with $\xi_{I_t}$ and use it to update state of ``Active \EXPABS".
      \ENDFOR
\end{algorithmic}
\caption{\ContEXPABS.}
\label{ContextualExp3-W}
\end{algorithm}

The online learning scenario for the loss $\tL$ under the abstention setting's feedback graphs is as follows. Given an unknown
sequence $z_1, z_2,\dots$ of pairs $z_t = (x_t,y_t) \in \cX\times\{\pm 1\}$, for every round $t = 1,2,\dots$:
\begin{enumerate}
\item The environment reveals input $x_t \in \cX$;
\item The learner selects an action $\xi_{I_t} \in \cY$ and incurs loss $\tL(\xi_{I_t},z_t)$;
\item The learner obtains feedback from the environment.
\end{enumerate}

Our algorithm is described as Algorithm \ref{ContextualExp3-W}. The algorithm essentially works as follows.
At each round $t$, if a new incoming input $x_t \in \scX$ is not contained in any existing ball generated so far,
then a new ball centered at $x_t$ is created, and a new instance of \EXPABS\ is allocated to handle $x_t$. Otherwise,
the \EXPABS\ instance associated with the closest input so far is used. Each allocated \EXPABS\ instance operates on 
the discretized action space $\cY_{\ve}$.

Consider the function
\[
\tL(a,r) =
\begin{cases}
c &{\mbox{if $r \leq -\gamma$}}\\
1+\left(\frac{1-c}{\gamma}\right)r &{\mbox{if $r \in (-\gamma,0)$}}\\
1-\left(\frac{1-f_{\gamma}(-a)}{\gamma}\right)\,r &{\mbox{if $r \in [0,\gamma)$}}\\
f_{\gamma}(-a) &{\mbox{if $r \geq \gamma$}}~,
\end{cases}
\]
where $f_{\gamma}$ is the Lipschitz variant of the 0/1-loss mentioned
in Section \ref{sec:adversarial} of the main text (Figure \ref{f:1}
(a)).  For any fixed $a$, the function $\tL(a,r)$ is
$1/\gamma$-Lipschitz when viewed as a function of $r$, and is
$1/(2\gamma)$-Lipschitz for any fixed $r$ when viewed as a function of
$a$. Hence
\begin{align*}
|\tL(a,r)-\tL(a',r')|
&\leq |\tL(a,r)-\tL(a,r')| + |\tL(a,r')-\tL(a',r')| \\
&\leq \frac{1}{\gamma}\,|r-r'| + \frac{1}{2\gamma}\,|a-a'|\\
&\leq \sqrt{\frac{1}{\gamma^2}+\frac{1}{4\gamma^2}}\,\sqrt{(a-a')^2 + (r-r')^2} \\
&< \frac{2}{\gamma}\,\sqrt{(a-a')^2 + (r-r')^2}~,
\end{align*}
so that $\tL$ is $\frac{2}{\gamma}$-Lipschitz w.r.t. the Euclidean distance on $\cY$.
Furthermore, a quick comparison to the abstention loss
\[
L(a,r) =  f_{\gamma}(a) 1_{r > 0} + c 1_{r \leq 0}
\]
reveals that (recall Figure \ref{f:1} (b) in the main text) :
\begin{itemize}
\item $\tL$ is an upper bound on $L$, i.e.,
\[
\tL(a,r) \geq L(a,r),\quad \forall\ (a,r) \in \cY~;
\]
\item $\tL$ approximates $L$ in that
\begin{equation}
\label{e:approx}
\tL(a,r) = L(a,r),\quad \forall\ (a,r) \in \cY\,:\, |r| \geq \gamma~.
\end{equation}
\end{itemize}
With the above properties of $\tL$ at hand, we are ready to prove
Theorem~\ref{th:infarms}.

\noindent{\bf Proof of Theorem \ref{th:infarms}.}\\
%
%
\begin{proof}
On each ball $B \subseteq \cX$ that \ContEXPABS\ allocates during its online execution, Theorem \ref{th:exp3rej} supplies the following
regret guarantee for the associated instance of \EXPABS:
%
\[
\frac{\log K_{\ve}}{\eta} + \frac{\eta}{2}\,T_B (c^2+1)\,,
\]
where $T_B$ is the number of points $x_t$ falling into ball $B$. Now, taking into
account that $\tL$ is $\frac{2}{\gamma}$-Lipschitz, and that the functions $h$ and $r$ are assumed to be
$L_{\cE}$-Lipschitz on $\cX$, a direct adaptation of the proof of
Theorem 1 in \cite{CesaBianchiGaillardGentileGerchinovitz2017} gives the bound
\[
\sup_{\xi \in \cE}  \E \Bigg[  \sum_{t = 1}^T  \tL( \xi_{I_t}, z_t) - \sum_{t = 1}^T  \tL(\xi, z_t) \Bigg]
\leq
\frac{N_T\log K_{\ve}}{\eta} + \frac{\eta}{2}\,T (c^2+1) +
  L_{\cE}\,\ve\,\frac{2}{\gamma}\,T~,
\]
being $N_T \leq C_{\cX}\ve^{-d}$ the maximum number of balls created by \ContEXPABS.
Using $c \leq 1$ and setting $\eta = \sqrt{\frac{N_T\,\log K_{\ve}}{T}}$ yields
\[
\sup_{\xi \in \cE}  \E \Bigg[  \sum_{t = 1}^T  \tL( \xi_{I_t}, z_t) - \sum_{t = 1}^T  \tL(\xi, z_t) \Bigg]
\leq
  2\,\sqrt{T\,N_T\,\log K_{\ve}} + L_{\cE}\,\ve\,\frac{2}{\gamma}\,T~.
\]
Next, optimizing for $\ve$ by setting
$\ve \simeq
T^{-\frac{1}{2+d}}\,\left(\frac{1}{\gamma}\right)^{-\frac{2}{2+d}}$
(and disregarding $L_{\cE}$ and log factors) gives
\begin{equation}
\label{e:boundrt}
\sup_{\xi \in \cE}  \E \Bigg[  \sum_{t = 1}^T  \tL( \xi_{I_t}, z_t) - \sum_{t = 1}^T  \tL(\xi, z_t) \Bigg]
=
{\tilde \cO}\left(T^{\frac{d+1}{d+2}}\,\left(\frac{1}{\gamma}\right)^{\frac{d}{d+2}}\right)~.
\end{equation}
Finally, we are left with connecting the above bound on the regret
with a bound on the regret for $L$. Now, observe that
\begin{equation}
\label{e:upperloss}
\E\left[\sum_{t=1}^T \tL(\xi_{I_t},z_t) \right] \geq \E\left[\sum_{t=1}^T L(\xi_{I_t},z_t) \right]~,
\end{equation}
since $\tL(\xi,z_t)$ is an upper bound on $L(\xi,z_t)$ for any $\xi$
and $z_t$. Moreover, if we assume for the sake of brevity that the 
minima are reached (the general case is straightforward to handle 
in a similar way), we can define
\[
\xi^* = (h^*,r^*) = \argmin_{\xi\in \cE} \sum_{t=1}^T
L(\xi,z_t),
\qquad \tilde{\xi}^* = \argmin_{\xi\in \cE} \sum_{t=1}^T \tL(\xi,z_t)~.
\]
We denote by $M^*_T(\gamma)$ the number of $x_t$ such that
$|r^*(x_t)| \leq \gamma$. Then, we can write
\begin{eqnarray*}
\sum_{t=1}^T \tL(\tilde{\xi}^*,z_t)
& \leq & \sum_{t=1}^T \tL(\xi^*,z_t)\\
& \leq & \sum_{t\,:\,|r^*(x_t)| > \gamma}^T \tL(\xi^*,z_t) + M^*_{T}(\gamma)\\
& & \text{(since $\tL \leq 1$))}\\
& = & \sum_{t\,:\,|r^*(x_t)| > \gamma}^T L(\xi^*,z_t) + M^*_{T}(\gamma)\\
&&{\mbox{(using (\ref{e:approx}))}}\\
&\leq&
\sum_{t=1}^T L(\xi^*,z_t) + M^*_{T}(\gamma)~.
\end{eqnarray*}
Combining with (\ref{e:boundrt}) and (\ref{e:upperloss}) gives
the following regret bound
\[
\sup_{\xi \in \cE}  \E \Bigg[  \sum_{t = 1}^T  L( \xi_{I_t}, z_t) - \sum_{t = 1}^T  L(\xi, z_t) \Bigg]
\leq
{\tilde \cO}\left(T^{\frac{d+1}{d+2}}\,\left(\frac{1}{\gamma}\right)^{\frac{d}{d+2}}\right) + M^*_{T}(\gamma)~,
\]
thereby concluding the proof.
\end{proof}

\begin{remark}
  The reader should observe that, since the algorithm is competing
  against an \emph{uncountably infinite} set of experts, the standard
  regret guarantee of $\sqrt{T}$ that one can achieve in the finite
  case cannot be obtained in general (see, e.g., the lower bound on
  regret of $T^{(d-1)/d}$ by \cite{hm07}, which holds in the easier
  full information setting). Notice that, while our algorithm
  \ContEXPABS\ admits a slightly worse bound of the form
  $T^{(d + 1)/(d + 2)}$, it has the advantage of being computationally
  feasible. In particular, the covering of the input space $\cX$ can
  be done adaptively, as the points $x_t$ are observed. In doing so,
  the number of $\epsilon$-balls allocated can never exceed the total
  number of rounds $T$. Given a new $x_t$, the algorithm has to decide
  if a new ball needs to be created or an old ball can be used. Known
  data-structures exist to efficiently implement this decision (e.g.,
  \cite{Clarkson06nearest-neighborsearching}).  The extra additive
  term $M_T^*(\gamma)$ in Theorem~\ref{th:infarms} is due to the fact
  that the loss function $L$ therein is not Lipschitz.
  In fact, one can further improve the term $T^{\frac{d+1}{d+2}}$ to
  $T^{\frac{d}{d+1}}$ by adopting a hierarchical covering technique of
  the function space $\cE$, each layer of the hierarchy being a pool
  of experts for the layer above it, see, e.g.,
  \cite{CesaBianchiGaillardGentileGerchinovitz2017}. However, the
  resulting algorithm would be of theoretical interest only, since it
  would be computationally very costly.
\end{remark}

\section{Additional material for the stochastic setting}
\label{app:stochastic}

In this section, we present the proofs of the theoretical guarantees
for \UCBNT\ and \UCBGT, as well as the proof of
Proposition~\ref{prop:subset}.  The following theorems hold more
generally with
$S_{j,t}= \sqrt{\frac{2\beta \log t}{Q_{j,t}} }$ for $\beta>2$, which
implies slightly better constants in the regret bound. However, for
the sake of the simplicity of the presentation, below we set
$\beta = \frac{5}{2}$. Moreover, we prove Theorem~\ref{th:ucbn} for
the abstention loss $L$, but it holds for any general loss function.
\ignore{
\begin{align*}
 T_j(t)= \sum_{s=1}^t 1_{I_s=j} \hspace{10mm}  Q_{j,t} = \sum_{s=1}^t 1_{j\in N_t(I_s)} \hspace{10mm} S_{j,t}= \sqrt{\frac{5 \log t}{Q_{j,t-1}} }
  \end{align*}

Please see the different framed boxes for definitions needed for
\UCBNT\ and \UCBGT.   Recall that $\ell_s$ is a general
loss function and $L_s$ is the abstention loss function defined in the
Section~\ref{sec:problem}.}

\subsection{Regret of \UCBNT\ }

\noindent{\bf Proof of Theorem \ref{th:ucbn}.}\\
\begin{proof} Consider a sequence of graph realizations $G_1,\ldots,G_t$ denoted by $\mathbf{G}_t$. By conditioning on this quantity, the regret can be decomposed according to each arm $i$:
\begin{align*}
 \sum_{t=1}^T  \E[L(\xi_{I_t},z_t) - L(\xi_{*},z_t)] & =  \sum_{t=1}^T\E [ \E[L(\xi_{I_t},z_t) - L(\xi_{*},z_t)|\mathbf{G}_{t}]]\\
&=  \sum_{t=1}^T \E \Bigg[\E \Bigg[\sum_{i=1}^K 1_{I_t=i}(L(\xi_{i},z_t) - L(\xi_{*},z_t))\bigg |\mathbf{G}_{t}\Bigg]\Bigg]\\ 
& = \sum_{i=1}^K \sum_{t=1}^T \E [\E [L(\xi_i,z_t) -L(\xi_{*},z_t)|\mathbf{G}_{t}] \E[1_{I_t = i}|\mathbf{G}_{t}]]\\
&= \sum_{i=1}^K \sum_{t=1}^T \E [\E [L(\xi_i,z_t) - L(\xi_{*},z_t)] \E[1_{I_t = i}|\mathbf{G}_{t}]] 
= \E \Bigg[ \sum_{i=1}^K \sum_{t=1}^T \Delta_i \E[1_{I_t = i}|\mathbf{G}_{t}] \Bigg ]
\end{align*}
where, in the last step, we used the fact that $L(\cdot, z_t)$s are
independent of $\mathbf{G}_{t}$ since, by assumption, $\mathbf{G}_{t}$
only depends on information up $t - 1$. Next, we focus on bounding
$\sum_{t=1}^T \E[1_{I_t = i}|\mathbf{G}_{t}]$ for each arm $i$.

We split the expectation according to the events $Q_{i,t-1}>s_i$ and
$Q_{i,t-1}\leq s_i$, where $s_i$ is a quantity determined later:
\begin{align*}
\sum_{t=1}^T  \E[1_{I_t = i}|\mathbf{G}_{t}] 
& = \sum_{t=1}^T 
\E[1_{I_t = i} (1_{Q_{i,t-1} \leq s_i} + 1_{Q_{i,t-1} > s_i})|\mathbf{G}_{t}]\\
& \leq  s_i + \sum_{t=1}^T  \E[1_{I_t = i}  1_{Q_{i,t-1} > s_i}|\mathbf{G}_{t}].
\end{align*}
We wish to choose $s_i$ sufficiently large so that the second term is
bounded but so that it admits a mild dependence on $T$.
Now, whenever $I_t=i$, by the design of the algorithm, it must be the case that the upper confidence bound of $i$ is smaller than that of any other expert. Thus,
\begin{align*}
& \E[1_{I_t = i} 1_{Q_{i,t-1} > s_i}|\mathbf{G}_{t}]= \Pr[I_t = i, Q_{i,t-1} > s_i|\mathbf{G}_{t}] \leq \Pr[\widehat{\mu}_{i,t-1} - S_{i,t-1} \leq \widehat{\mu}_{*,t-1} - S_{*,t-1}, Q_{i,t-1} > s_i|\mathbf{G}_{t}],
\end{align*}
where $*$ denotes the best-in-class expert. We now use the terms
$\mu_{*}$, $\mu_i$ and $S_{i,t-1}$ to reorder the first event in the
probability on the right-hand side of the last expression as follows:
\begin{align*}
& 0 \leq \widehat{\mu}_{*,t-1} - S_{*,t-1} - \widehat{\mu}_{i,t-1} +
  S_{i,t-1} \\
\Leftrightarrow \ & 0 \leq \left(\widehat{\mu}_{*,t-1} - S_{*,t-1} - \mu_{*} \right)  + \left(\mu_i - \widehat{\mu}_{i,t-1} + S_{i,t-1} - 2 S_{i,t-1} \right) + \left( \mu_{*} - \mu_i + 2 S_{i,t-1} \right).
\end{align*}
If we can show that the third term is negative, then the first and
second term must be positive. Moreover, we will further show that the
first and second terms can only be positive with an extremely low
probability that is bounded by a constant independent of
$T$. Furthermore, the third term will be negative whenever the slack
term in the upper confidence bound is small enough, which amounts to
choosing $s_i$ large enough.

In particular, by setting $s_i = \frac{20 \log(T)}{\Delta_i^2}$, we ensure that the event $Q_{i,t-1} > s_i$ implies that
\begin{align*}
&Q_{i,t-1} > \frac{20 \log(t)}{\Delta_i^2}
\Leftrightarrow \mu_{*} - \mu_i + 2 S_{i,t-1} < 0.
\end{align*}
As explained above, it then follows that
\begin{align*}
&\Pr[\widehat{\mu}_{i,t-1} - S_{i,t-1} \leq \widehat{\mu}_{*,t-1} - S_{*,t-1}, Q_{i,t-1} > s_i|\mathbf{G}_{t}] \\
  & \leq \Pr[\widehat{\mu}_{*,t-1} - S_{*,t-1} - \mu_{*} \geq 0|\mathbf{G}_{t}]  + \Pr[\mu_i - \widehat{\mu}_{i,t-1} + S_{i,t-1} - 2 S_{i,t-1} \geq 0|\mathbf{G}_{t}].
\end{align*}

We can bound these last probabilities using the union bound and
a concentration inequality such as Hoeffding's Inequality:
\begin{align*}
& \P[ \mu_i - \widehat{\mu}_{i,t-1} + S_{i,t-1} - 2 S_{i,t-1} \geq 0|\mathbf{G}_{t}] \\
&=  \P\Big[ -\tfrac{1}{Q_{i,t-1}}\sum_{s=1}^{t-1}  L(\xi_i,z_s) 1_{i\in N_s(I_s)} + \mu_i   -  \sqrt{\tfrac{5 \log (t)}{Q_{i,t-1}}}  \geq 0\Big |\mathbf{G}_{t}\Big].
\end{align*}

Now, the estimate $\widehat{\mu}_{i,t-1}$ is an average of i.i.d.\
realizations of the random variable $ L(\xi_i,z)$, with $z \sim \cD$,
since the out-neighborhood of the chosen expert only depends on
previous observations.  That is,
\begin{align*}
\frac{  \E[ \sum_{s=1}^{t-1}L(\xi_i,z_s) 1_{i\in N_s(I_s)} ]}{\E[ \sum_{s=1}^{t-1}  1_{i\in N_s(I_s)} ]}
&= \frac{  \E[ \sum_{s=1}^{t-1} \E[L(\xi_i,z_s) 1_{i\in N_s(I_s)}| i\in N_s(I_s) ]]}{\E[ \sum_{s=1}^{t-1}  1_{i\in N_s(I_s)} ]}\\
&= \frac{  \E[ \sum_{s=1}^{t-1}1_{i\in N_s(I_s)}  \E[L(\xi_i,z_s) | i\in N_s(I_s) ]]}{\E[ \sum_{s=1}^{t-1}  1_{i\in N_s(I_s)} ]}\\
&= \frac{  \E[ \sum_{s=1}^{t-1}1_{i\in N_s(I_s)}  \E[L(\xi_i,z_s)]]}{\E[ \sum_{s=1}^{t-1}  1_{i\in N_s(I_s)} ]}\\
&= \E[L(\xi_i,z)].
\end{align*}
Hence,
$\widehat{\mu}_{i,t-1}$ can be turned into an empirical estimate of
$\mu_{i}$ using the union bound as follows:
\begin{align*}
\P\Big [ -\tfrac{1}{Q_{i,t-1}}\sum_{s=1}^{t-1}  L(\xi_i,z_s) 1_{i\in N_s(I_s)} + \mu_i   -  \sqrt{\tfrac{5 \log (t)}{Q_{i,t-1}}}  \geq 0 \Big|\mathbf{G}_{t}\Big]
& \leq \P \Big [ \exists n \in [1,t] :   - \hat{\mu}_{i}^n+\mu_{i}   -  \sqrt{\tfrac{5 \log (t)}{n}}\Big |\mathbf{G}_{t}\Big] \\
& \leq \sum_{n=1}^t \frac{1}{t^{\frac{5}{2}}} = \frac{1}{t^{\frac{3}{2}}}~,
\end{align*}
where $ \hat{\mu}_{i}^n = \frac{1}{n}\sum_{s=1}^n  L(\xi_i,z_s)$. By the same reasoning, we can also bound the probability of the best arm :
\[
\P\Big[ \widehat{\mu}_{*,t-1} - S_{*,t-1} - \mu_{*} \geq 0 \Big |\mathbf{G}_{t} \Big ] \leq \sum_{n=1}^t \frac{1}{t^{\frac{5}{2}}} = \frac{1}{t^{\frac{3}{2}}}~.
\]
Now let $\cC$ be any element of $\cF$ and $C$ any element of $\cC$. Then
for any $i \in C$, it follows that
\begin{align*}
    Q_{i,t}
    &= \sum_{s=1}^t 1_{i \in N_s(I_s)}
    = \sum_{s=1}^t \sum_{j=1}^K 1_{i \in N_s(j)} 1_{I_s = j}
    \geq \sum_{s=1}^t \sum_{j \in C} 1_{i \in N_s(j)} 1_{I_s = j}
    = \sum_{s=1}^t \sum_{j \in C} 1_{I_s=j}.
\end{align*}

This implies that
\begin{align*}
&\sum_{i=1}^K \sum_{t=1}^T \Delta_i \E[1_{I_t = i} 1_{Q_{i,t-1} \leq s_i}|\mathbf{G}_{t}] \\
    &\leq \sum_{C \in \cC} \Bigg[ \max_{i \in C} \Delta_i \Bigg] \sum_{t=1}^T \sum_{i \in C} \E\Bigg[1_{I_t = i} 1_{Q_{i,t-1} \leq s_i}\Bigg|\mathbf{G}_{t}\Bigg] \\
    &\leq \sum_{C \in \cC} \Bigg[ \max_{i \in C} \Delta_i \Bigg] \sum_{t=1}^T \sum_{i \in C} \E\Bigg[1_{I_t = i} 1_{Q_{i,t-1} \leq \max_{j \in C} s_j}\Bigg|\mathbf{G}_{t}\Bigg] \\
    &\leq \sum_{C \in \cC} \Bigg[ \max_{i \in C} \Delta_i \Bigg] \max_{j \in C} s_j. 
\end{align*}

Combining the above calculations, applying our definition for $s_i$,
and using the fact that the above analysis holds for any shared admissible covering shows that
\begin{align*}
& \E\Bigg[
  \min_{\cC \in \cF} \sum_{C \in \cC} \Bigg[\max_{j \in C} \Delta_j
  \Bigg] \Bigg[ \max_{j \in C} \frac{20
  \log(T)}{\Delta_j^2}\Bigg]  + 5K \Bigg],
\end{align*}
which proves the bound of the theorem.
\end{proof}

\subsection{Regret of \UCBGT\ }

Next, we prove the regret bound for \UCBGT, which demonstrates how one can exploit the bias and feedback structure in the problem.

\noindent{\bf Proof of Theorem \ref{th:ucbabs}.}\\
\begin{proof}
  As in the previous proof, we focus on bounding
  $\sum_{t=1}^T \E[1_{I_t = i}|\mathbf{G}_{t}]$ for each arm $i$.  We again split the expectation according to the events based on $Q_{i,t-1}$ as follows:
    $$ \sum_{t=1}^T \E[1_{I_t
    = i}1_{Q_{i,t-1} \leq s_i} |\mathbf{G}_{t}] + \E[1_{I_t=i}  1_{Q_{i,t-1} > s_i}
  |\mathbf{G}_{t}],$$
    where $s_i$ is to be
  determined later. We then bound the second term using the algorithm's choice of arm, $I_t$:
\begin{align*}
& \E[1_{I_t = i} 1_{Q_{i,t-1} > s_i}|\mathbf{G}_{t}]= \Pr[I_t = i, Q_{i,t-1} > s_i|\mathbf{G}_{t}]\leq \Pr[\widehat{\mu}_{i,t-1} - S_{i,t-1} \leq \widehat{\mu}_{*,t-1} - S_{*,t-1}, Q_{i,t-1} > s_i|\mathbf{G}_{t}].
\end{align*} 
$\widehat{\mu}_{i,t-1}$ is a biased estimate of $\mu_i$. This is because
whenever $x_s$ falls in the region
$\{x \colon r_i(x) > 0 \wedge r_{I_s}(x) \leq 0\}$ and the condition
$\hat{p}_{I_s,i}^{s - 1} \leq \gamma_{i, s - 1}$ holds, the label
$y_s$ is not accessible. In this case, the \UCBGT\ algorithm updates the average loss of
expert $i$ optimistically, as if the expert were correct at that time
step.
    
    We can decompose this biased estimate $\widehat{\mu}_{i, t - 1}$
into two terms:
$\widehat{\mu}_{i, t - 1} = \widetilde{\mu}_{i, t - 1} - \e_{i, t -
  1}$.  The first term, $\widetilde{\mu}_{i, t - 1}$, is an unbiased
estimate of arm $i$ and similar to the estimates in Theorem~\ref{th:ucbn}. The second term is the
misclassification rate $\e_{i, t - 1}$ over
$\set{s \in [t - 1] \colon r_i(x_s) > 0 \cap r_{I_s}(x_s) \leq 0}$
whenever the condition $\hat{p}_{I_s,i}^{s-1} \leq \gamma_{i,s-1}$
holds, that is,
$\e_{i,t-1}= \frac{1}{Q_{i,t-1}} \sum_{s=1}^{t-1} 1_{y_sh_i(x_s) \leq
  0}1_{r_i(x_s)>0, r_{I_s}(x_s) \leq 0} 1_{\hat{p}_{I_s,i}^{s-1} \leq
  \gamma_{i,s-1} }$.

Now, by the design of the \UCBGT, if arm $i$ is chosen at time $t$, it
must be the case that
$\widehat{\mu}_{i,t-1} - S_{i,t-1} \leq \widehat{\mu}_{*,t-1} -
S_{*,t-1}$. We can expand and rewrite this expression as follows:
\begin{align*}
& 0 \leq \widehat{\mu}_{*,t-1} + \e_{i^*,t-1} - \e_{i^*,t-1} - S_{*,t-1} - \widehat{\mu}_{i,t-1}-  \e_{i,t-1} + \e_{i,t-1}  + S_{i,t-1} \\
\Leftrightarrow \ & 0 \leq \left(\widetilde{\mu}_{*,t-1} - S_{*,t-1} - \mu_{*} \right)  + \left(\mu_i - \widetilde{\mu}_{i,t-1} + S_{i,t-1} - 2 S_{i,t-1} \right) + \left( \mu_{*} - \mu_i + (2+C) S_{i,t-1}  \right),
\end{align*}
where we used the fact that $ - \e_{i^*,t-1} \leq 0 $, and where
we bounded $\e_{i,t-1}$ as follows:
\begin{align*}
\e_{i,t-1}
& =  \frac{1}{Q_{i,t-1}} \sum_{s=1}^{t-1} 1_{y_sh_i(x) \leq 0} 1_{r_i(x_s)>0, r_{I_s}(x_s) \leq 0} 1_{\hat{p}_{I_s,i}^{s-1} \leq \gamma_{i,s-1} }\\
& \leq \frac{1}{Q_{i,t-1}}   \sum_{s=1}^{t-1} 1_{r_i(x_s)>0, r_{I_s}(x_s) \leq 0} 1_{\hat{p}_{I_s,i}^{s-1} \leq \gamma_{i,s-1} } = \frac{1}{Q_{i,t-1}}  \sum_{s=1}^{t-1} \sum_{\xi_j \in \mathcal{E}-\xi_i} 1_{r_i(x_s)>0, r_{j}(x_s) \leq 0} 1_{\hat{p}_{I_s,i}^{s-1} \leq \gamma_{i,s-1} } 1_{I_s=j}\\
&\leq  \frac{1}{Q_{i,t-1}}   \sum_{\xi_j \in \mathcal{E}-\xi_i} \sum_{s=1}^{t-1} 1_{r_i(x_s)>0, r_{j}(x_s) \leq 0} 1_{\hat{p}_{I_s,i}^{s-1} \leq \gamma_{i,s-1} }.
\end{align*}
The condition $\hat{p}_{j,i}^{s-1} \leq \gamma_{i,s-1}$ is equivalent
to
$\sum_{k = 1}^{s - 1} 1_{r_i(x_k)>0, r_{j}(x_k) \leq 0} \leq (s -
1)\gamma_{i, s - 1}$. Since the sum above is non-zero only when this
condition holds, there exists $s_j\in[1,t-1]$ such that
$\sum_{s=1}^{t-1} 1_{r_i(x_s)>0, r_{j}(x_s) \leq 0} 1_{\hat{p}_{j,i}^s
  \leq \gamma_{i,s}} \leq (s_j-1)\gamma_{i,s_j-1} +1$.  Moreover,
using the fact that
$(s_j-1)\gamma_{i,s_j-1} = \sqrt{5 Q_{i,s_j-1}\log(s_j)} /(K-1) \leq
\sqrt{5 Q_{i,t-1}\log(t-1)} /(K-1) $, we can conclude that
\begin{align*}
  \e_{i,t-1} \leq \frac{1}{Q_{i,t-1}} \sum_{\xi_j \in \mathcal{E}-\xi_i}  \sum_{s=1}^{t-1}  1_{r_i(x_s)>0, r_{j}(x_s) \leq 0} 1_{\hat{p}_{j,i}^{s-1} \leq \gamma_{i,s-1}}  \leq   \frac{K-1}{Q_{i,t-1}} \Bigg[  \frac{\sqrt{5 Q_{i,t-1}\log(t-1)} } {K-1}+1 \Bigg] \leq   C \sqrt{\frac{5\log(t-1)}{Q_{i,t-1}}} 
\end{align*}
for some constant $C > 0$.  The rest of the proof now follows by
similar arguments as in the proof of Theorem~\ref{th:ucbn}.
Specifically, we can choose $s_i$ such that the term
$\mu_{*} - \mu_i + (2 + C) S_{i, t - 1}$ is negative, and since now
$\widetilde{\mu}_{*,t-1}$ and $\widetilde{\mu}_{i,t-1}$ are unbiased
estimates, we can bound the probabilities
$\Pr [\widetilde{\mu}_{*,t-1} - S_{*,t-1} - \mu_{*} \geq
0|\mathbf{G}_{t}]$ and
$\Pr[\mu_i - \widetilde{\mu}_{i,t-1} - S_{i,t-1}\geq 0
|\mathbf{G}_{t}]$ using standard concentration inequalities.
\end{proof}

\subsection{Linear regret without the subset property}
\label{app:subset}

In this section, we prove Proposition~\ref{prop:subset}, which
shows that when the subset property does not hold for a feedback
graph, then it is possible to incur linear regret.

\noindent{\bf Proof of Proposition \ref{prop:subset}.}\\
\begin{proof}
  Let $p^* \in (0,1)$. We design a setting in which with probability
  at least $p^*$, the \UCBNT\ algorithm incurs linear regret.

   Since the family of abstention functions induces a feedback graph
   that violates the subset property, there exist pairs $(h_i, r_i)$
   and $(h_j, r_j)$ and points $x^*$, $\tilde{x}$ for which
   $x^* \in \mathcal{A}_i \setminus \mathcal{A}_j$,
   $\tilde{x} \in \mathcal{A}_i \cap \mathcal{A}_j$, where
   $\mathcal{A}_i$ and $\mathcal{A}_j$ are the acceptance regions
   associated with $r_i$ and $r_j$, respectively, and the feedback
   graph is designed such that the algorithm updates the pair
   $(h_i, r_i)$ when the pair $(h_j, r_j)$ is selected.

   Now, for some $p \in (0, 1)$ to be determined later, consider a
   distribution with probability $p$ on $(\tilde{x}, \tilde{y})$ and
   $(1 - p)$ on $(x^*, y^*)$.

   We choose the set of hypothesis functions $\cH = \{h_i, h_j\}$, the
   loss function $\ell$ in (\ref{e:absloss}), and the labels $y^*$ and
   $\tilde{y}$ in such a way that
   $\ell(\tilde{y}, h_i(\tilde{x})) = c - \beta$,
   $\ell(\tilde{y}, h_j(\tilde{x})) = c - \alpha$, and
   $\ell(y^*, h_i(x^*)) = 0$, where $\alpha, \beta$ are values that
   will be later specified. For instance, we can consider the hinge
   loss $\ell(y,\hat{y}) = (1 - y \hat{y})_+$, and $h_i$, $h_j$ such
   that $h_i(\tilde{x}) = \frac{1-c+\beta}{\tilde{y}}$,
   $h_j(\tilde{x}) = \frac{1-c+\alpha}{\tilde{y}}$, and
   $h_i(x^*) = \frac{1}{y^*}$. Note that, since $r_j(x^*) < 0$,
   $\ell(y^*, h_j(x^*))$ may admit any value.

   Now, by construction, $\mu_i = (c - \beta) p$ and
   $\mu_j = (c-\alpha) p + c (1-p) = c - \alpha p$.  We claim that we
   can choose $\alpha$, $\beta$ and $p$ such that (1)
   $\alpha > \beta$; (2) $\mu_i < \mu_j$; (3)
   $\mu_j < \ell(\tilde{y}, h_i(\tilde{x}))$.

   The first condition is immediate. The second condition is
   equivalent to $cp - \beta p < c - \alpha p$, which is itself
   equivalent to $\alpha - \beta < \frac{c(1-p)}{p}$. By continuity,
   we can choose $\alpha$ and $\beta$ close enough such that this is
   true for any $p \in (0,1)$. The third condition is equivalent to
   $c - \alpha p < c - \beta$, which is itself equivalent to
   $\beta < \alpha p$. This is true for $p$ close enough to $1$.

   Now let $n \in \mathbb{N}$ be large enough such that
   $\mu_j < \ell(\tilde{y}, h_i(\tilde{x})) -
   \sqrt{\frac{5\log(n)}{n}}$. By continuity, we can choose $p$ large
   enough such that $p > (p^*)^{1/n}$, and for this choice of $p$, we
   can choose $\alpha$ and $\beta$ such that $\alpha > \beta$,
   $\alpha, \beta < c$, $\alpha - \beta < \frac{c(1-p)}{p}$, and
   $\beta < \alpha p$. For instance, if we, without loss of
   generality, assume that $p > \frac{1}{2}$, then we can choose,
   $\alpha = \frac{c(1-p)}{2p}$ and $\beta=\frac{c(1-p)}{4}$.

   Then, with probability $p^n > p^*$, the point $\tilde{x}$ will be
   sampled $n$ times at the start of the game, such that the pair
   $(h_j, r_j)$ will have a lower confidence bound than the pair
   $(h_i, r_i)$ at all time steps. Thus, \UCBNT\ will choose the pair
   $(h_j, r_j)$ throughout the entire game, even though
   $\mu_i < \mu_j$. Consequently, the regret of the algorithm will be
   at least $T (\mu_j - \mu_i)$.
\end{proof}

\section{Additional experimental results}
\label{app:expresults}

\begin{table}[t]
\centering
\begin{tabular}{ c | c}
Dataset & Number of features \\
\hline
 {\tt covtype} & 54 \\
    {\tt ijcnn} & 22 \\
      {\tt skin} & 3 \\
      {\tt HIGGS} & 28 \\
              {\tt guide} & 4 \\
        {\tt phishing} & 68 \\
        {\tt cod} & 8 \\
          {\tt eye} & 14 \\
          {\tt CIFAR} & 25 \\
\end{tabular}
\caption{Table shows the number of features of each dataset.}
\label{tb:features}
\end{table}

\begin{figure*}[!ht]
\begin{center}
\begin{tabular}{ c c c }
\includegraphics[scale=0.25,trim= 5 10 10 5, clip=true]{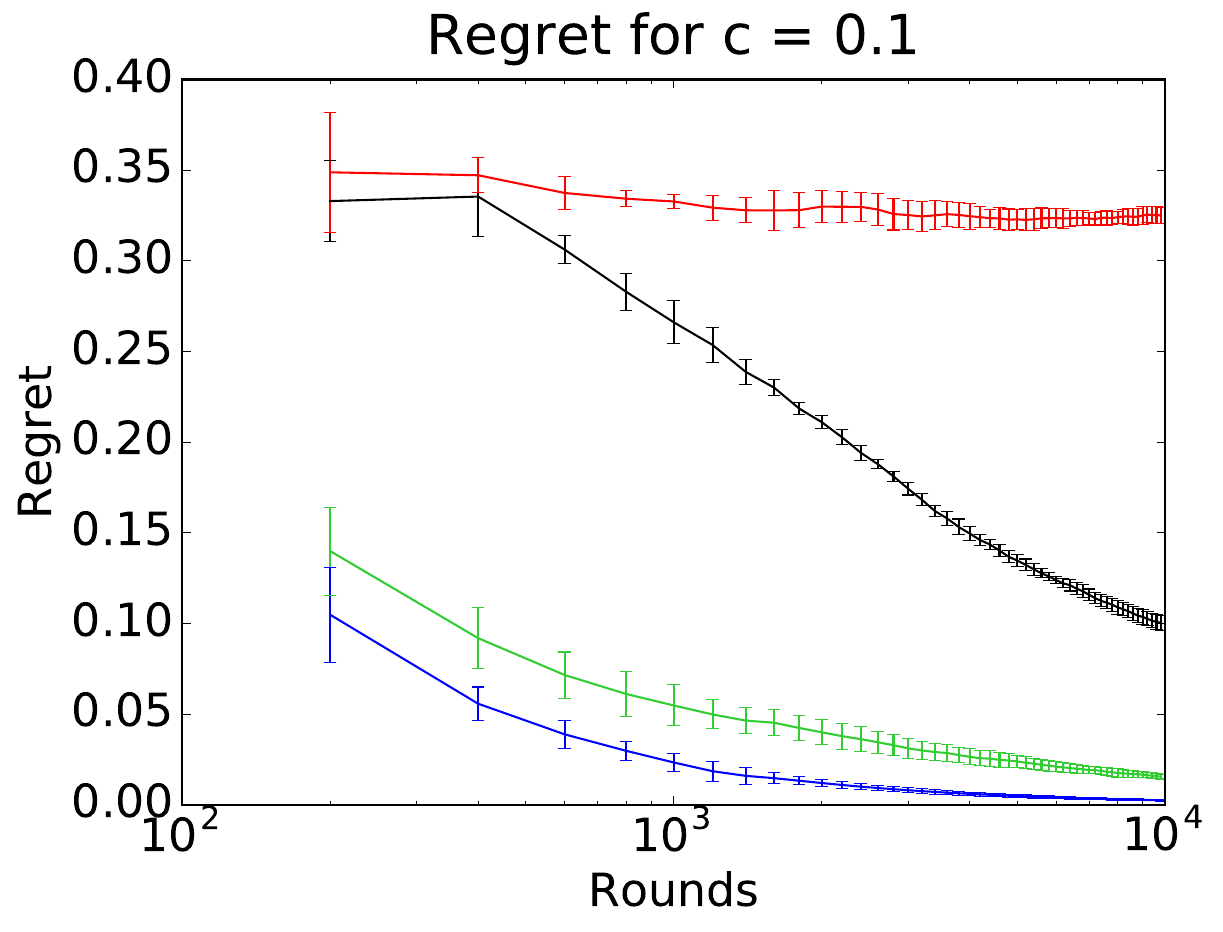} &
\hspace*{-5mm} \includegraphics[scale=0.25,trim= 5 10 10 5, clip=true]{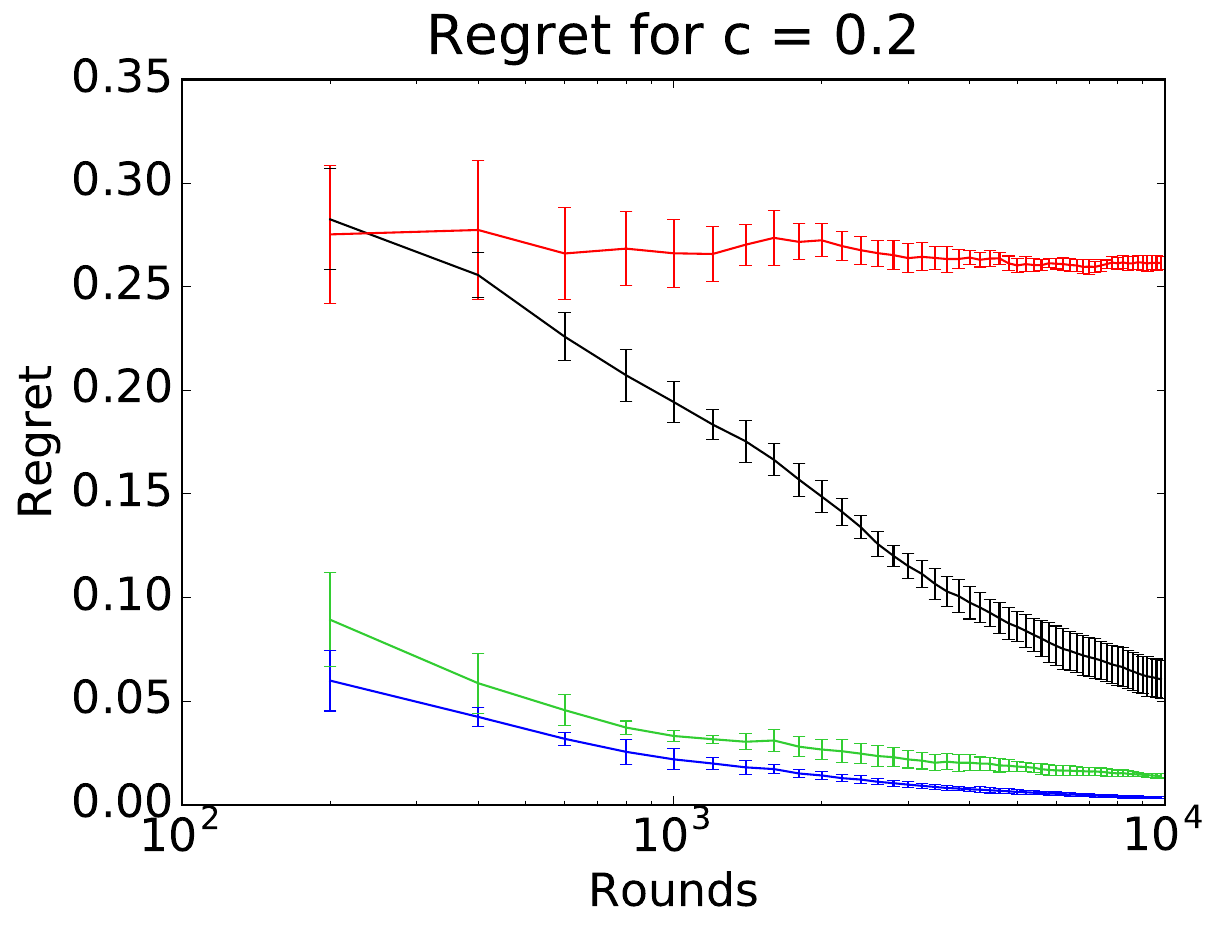} &
\hspace*{-5mm}\includegraphics[scale=0.25,trim= 5 10 10 5, clip=true]{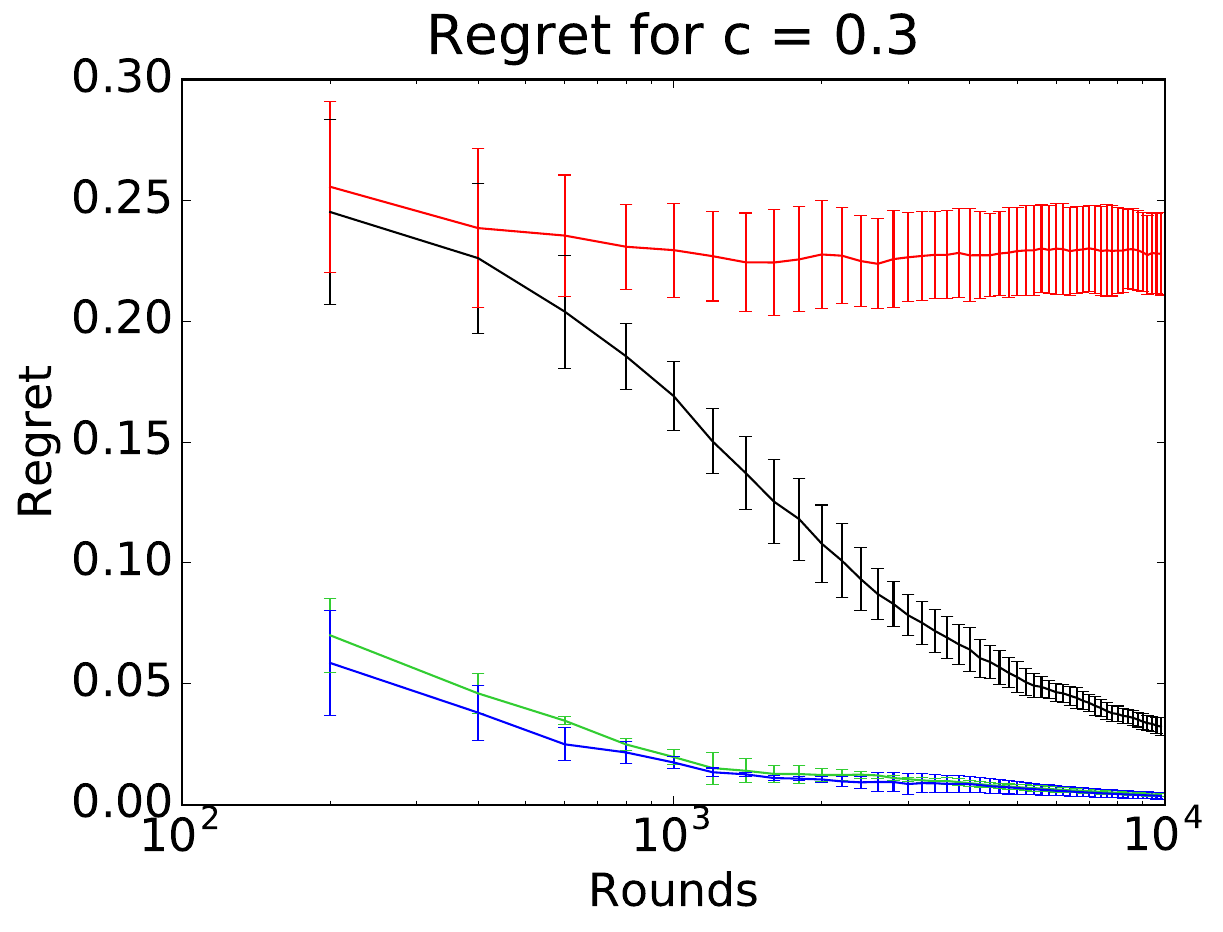} \\
\includegraphics[scale=0.25,trim= 5 10 10 5, clip=true]{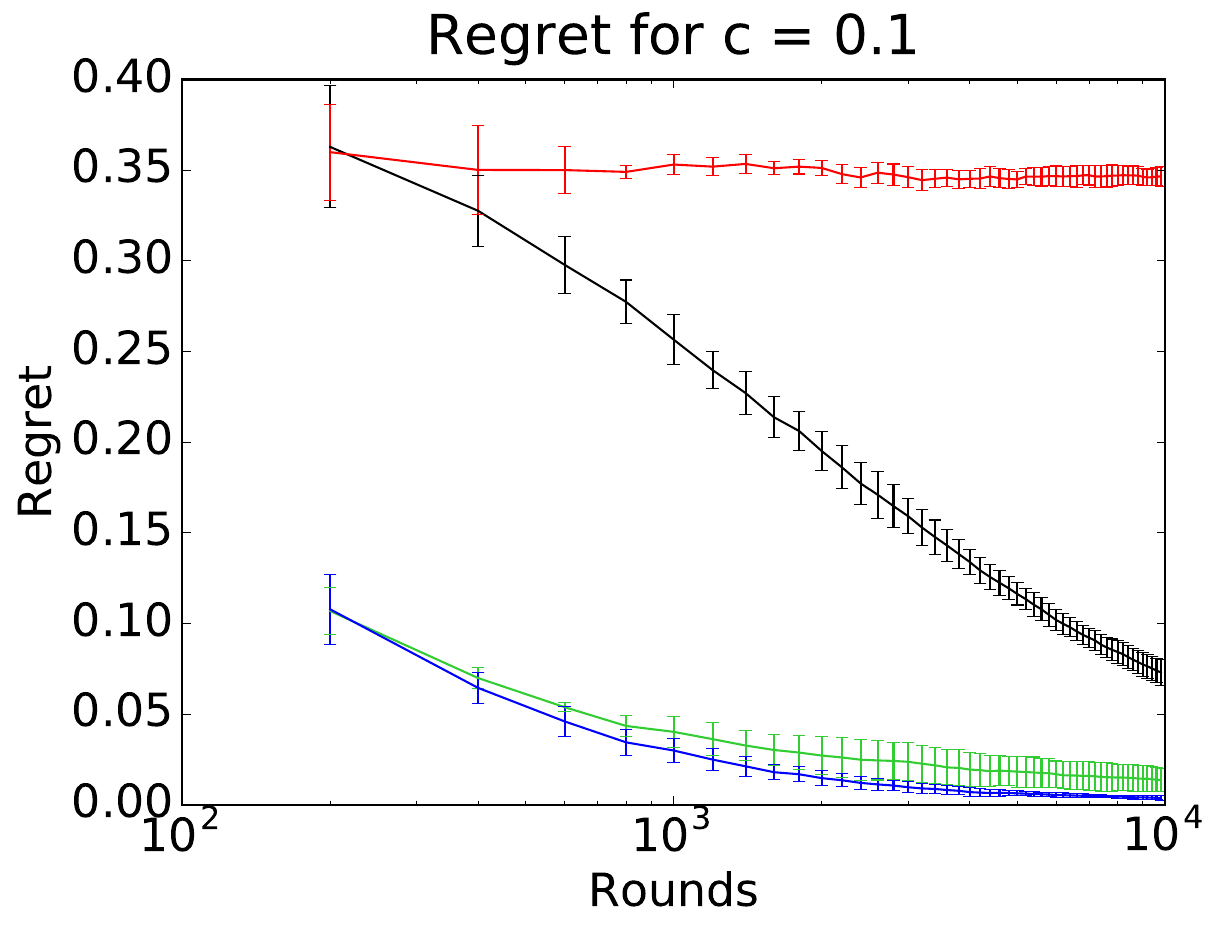} &
\hspace*{-5mm} \includegraphics[scale=0.25,trim= 5 10 10 5, clip=true]{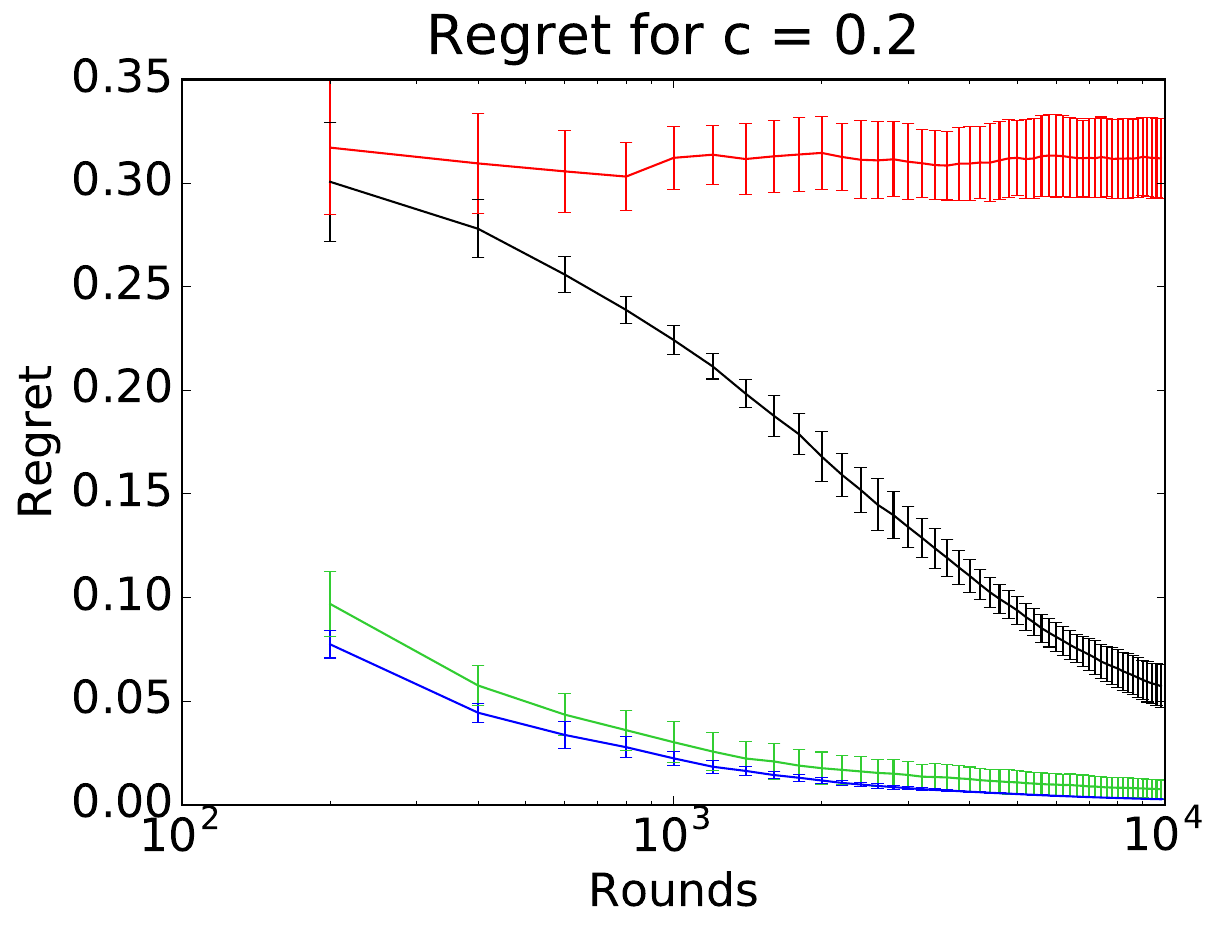} &
\hspace*{-5mm}\includegraphics[scale=0.25,trim= 5 10 10 5, clip=true]{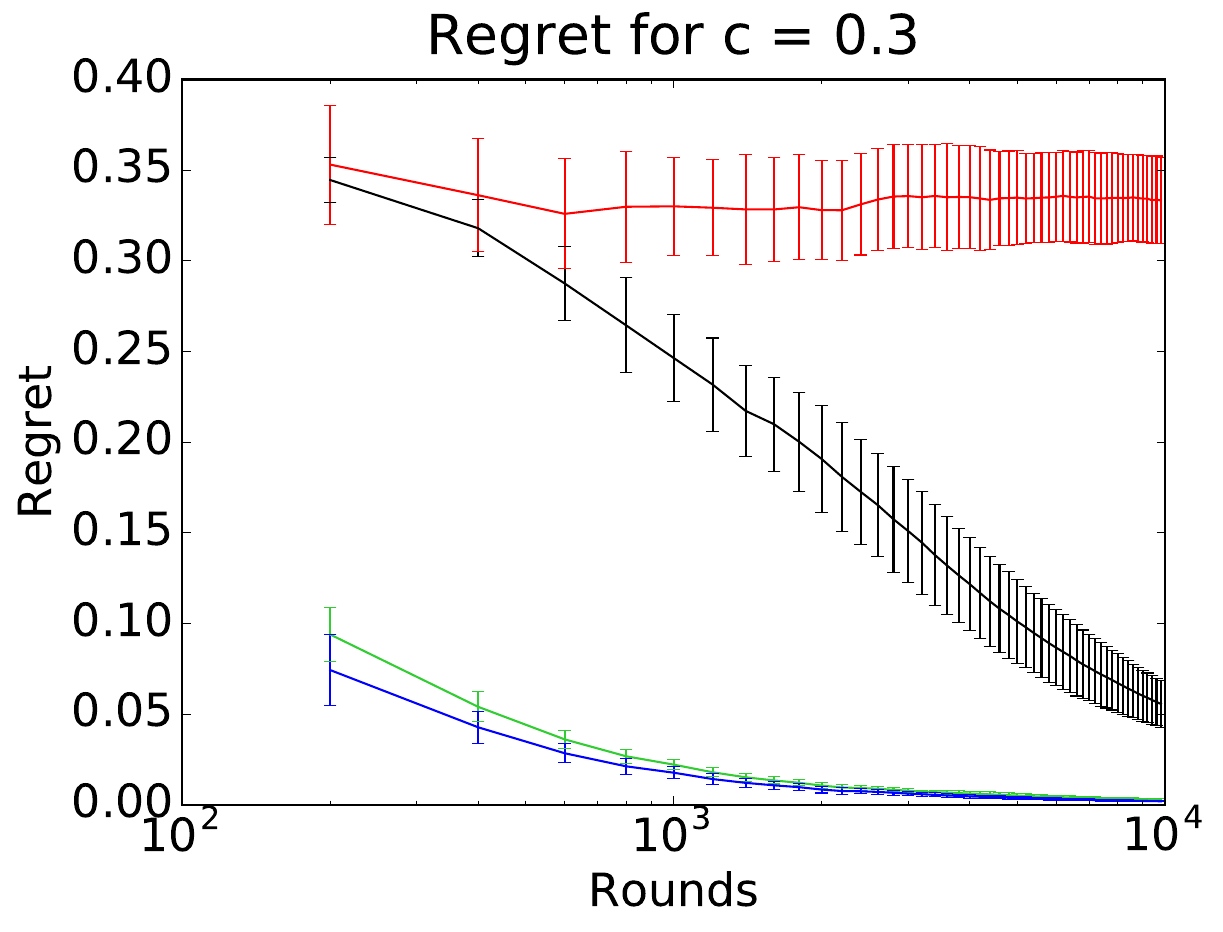} \\
\includegraphics[scale=0.25,trim= 5 10 10 5, clip=true]{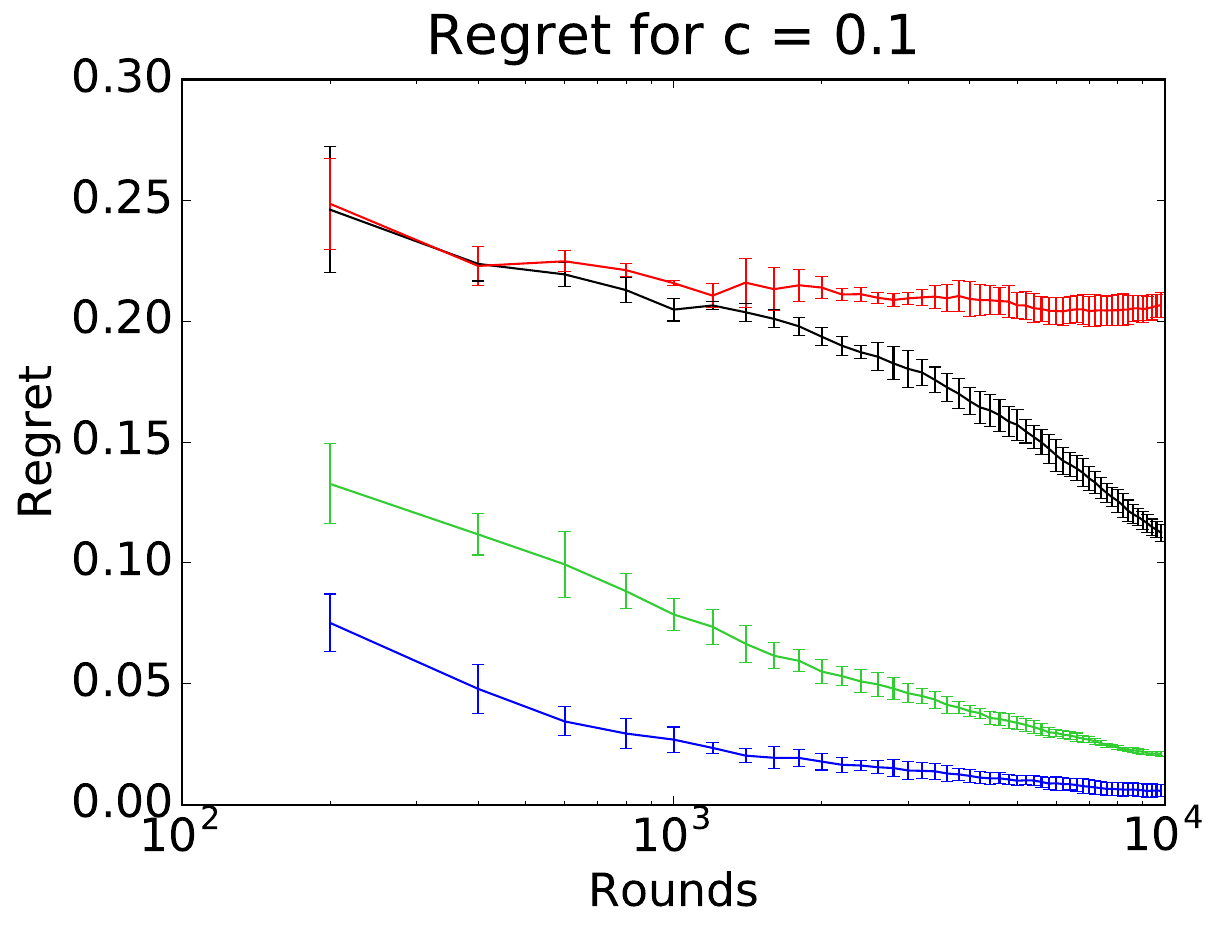} &
\hspace*{-5mm} \includegraphics[scale=0.25,trim= 5 10 10 5, clip=true]{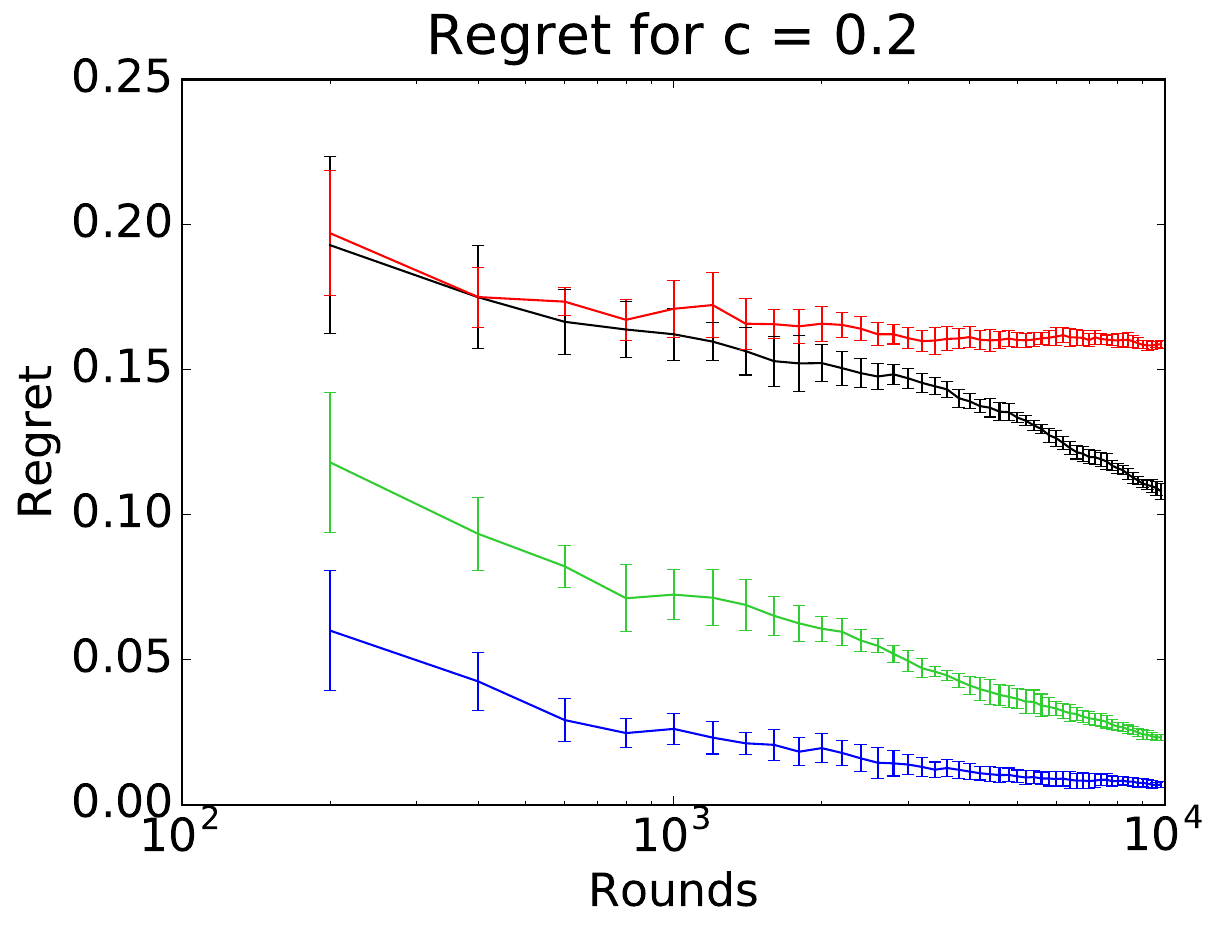} &
\hspace*{-5mm}\includegraphics[scale=0.25,trim= 5 10 10 5, clip=true]{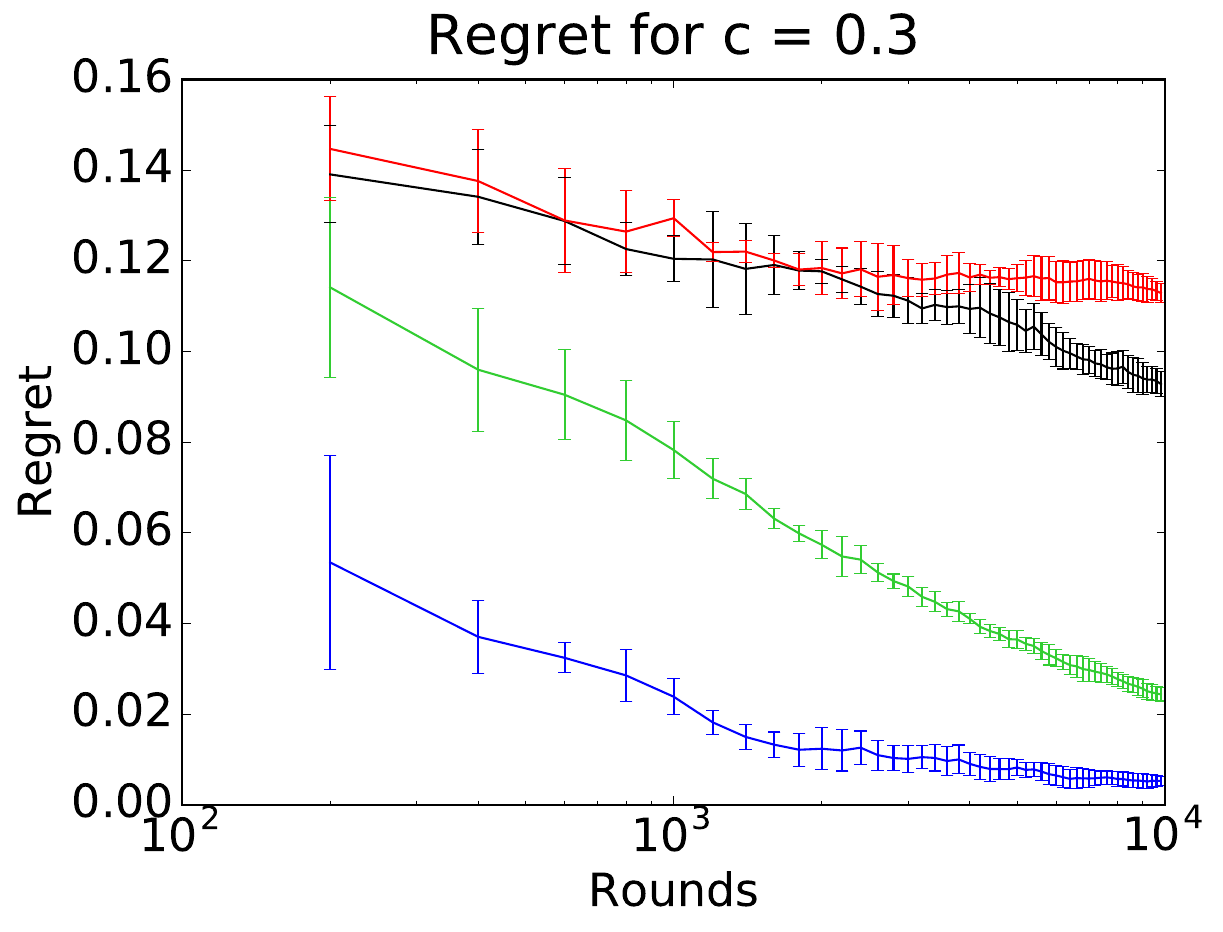} \\
\includegraphics[scale=0.25,trim= 5 10 10 5, clip=true]{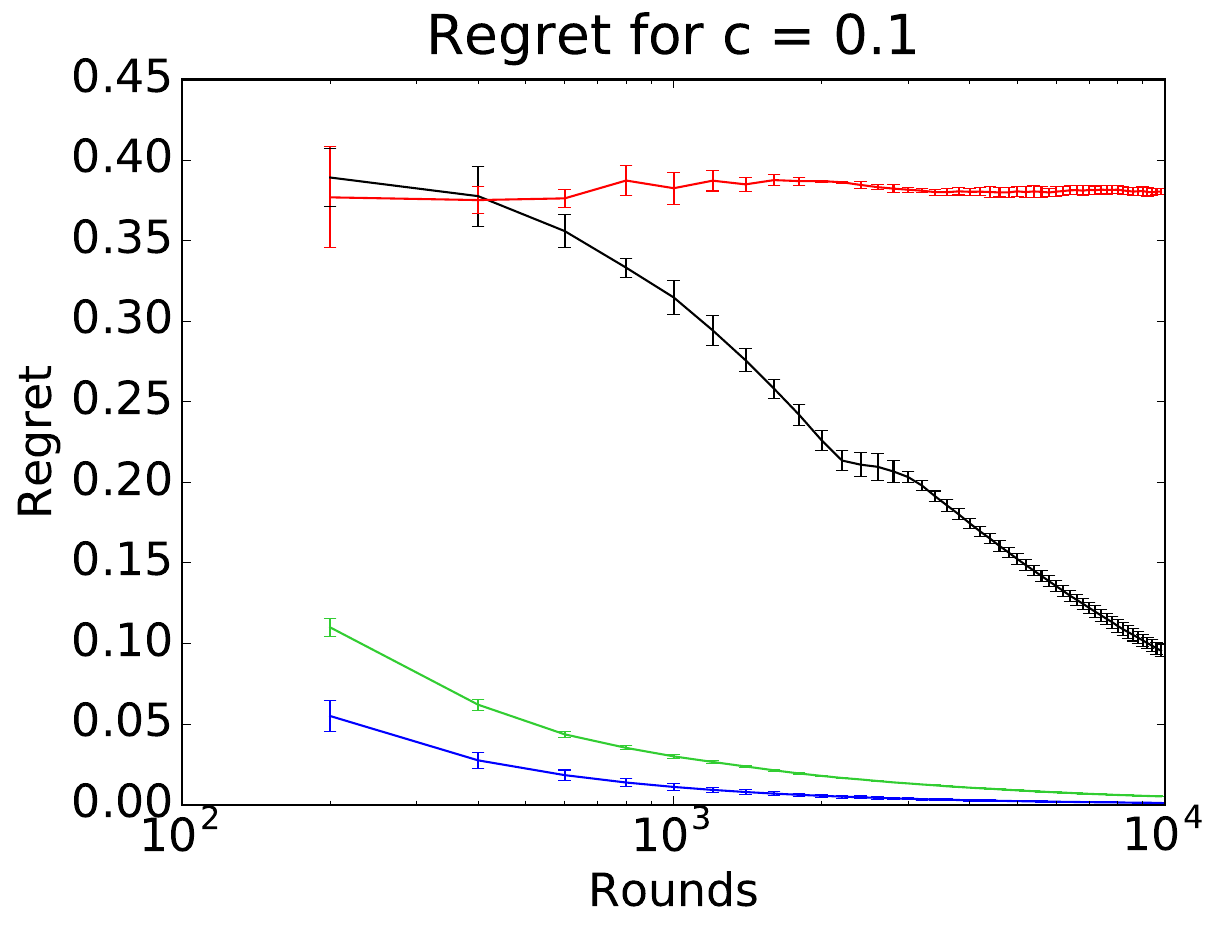} &
\hspace*{-5mm} \includegraphics[scale=0.25,trim= 5 10 10 5, clip=true]{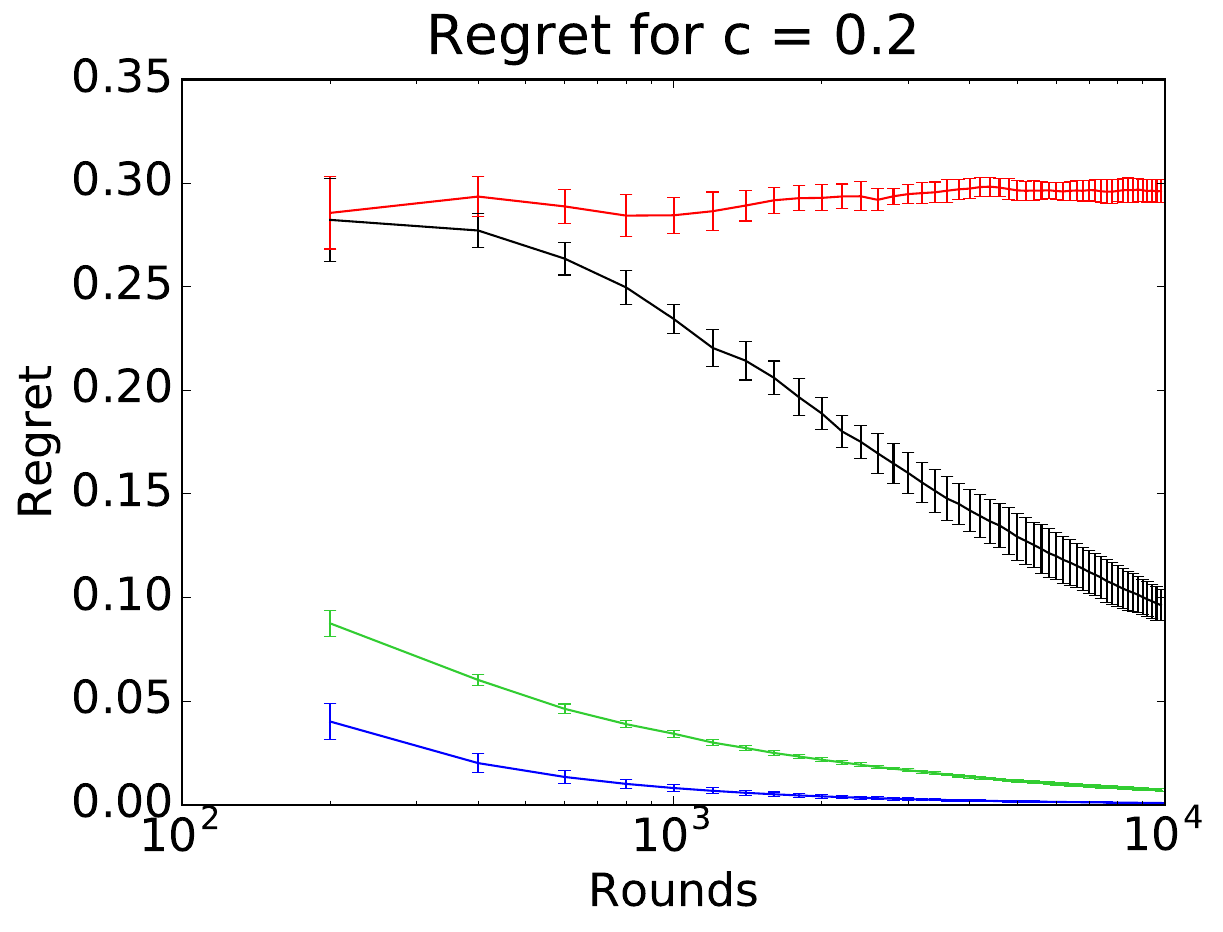} &
\hspace*{-5mm}\includegraphics[scale=0.25,trim= 5 10 10 5, clip=true]{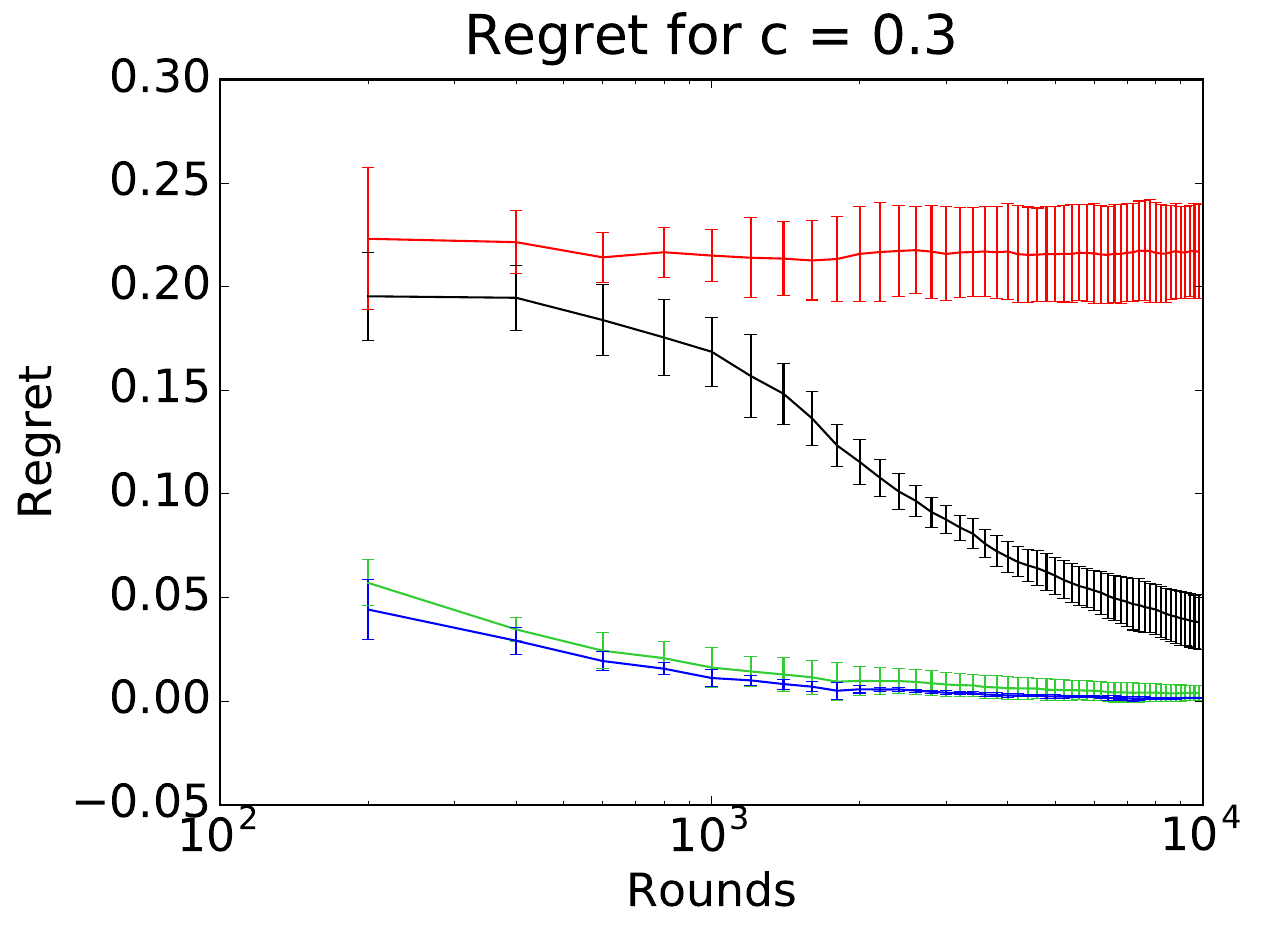} \\
\includegraphics[scale=0.25,trim= 5 10 10 5, clip=true]{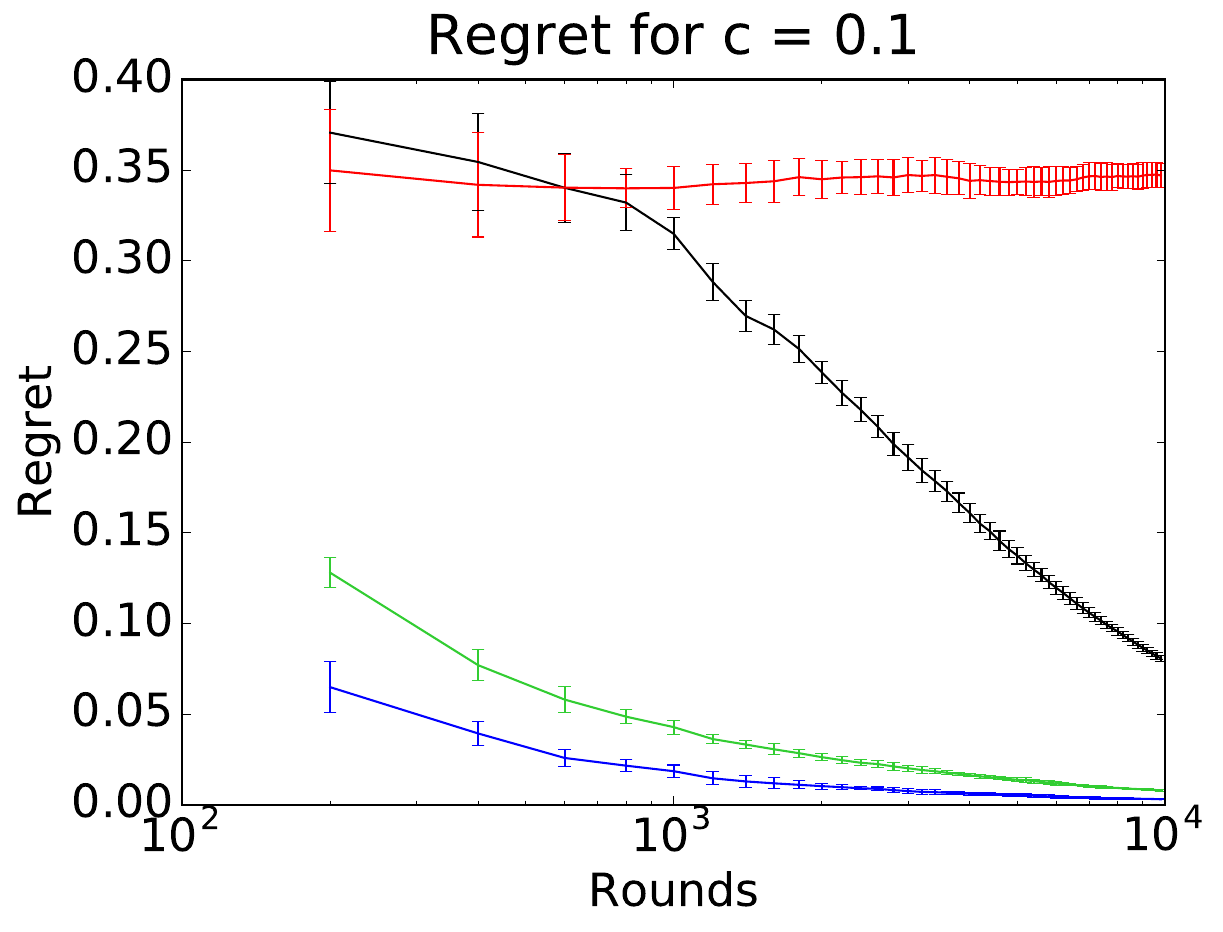} &
\hspace*{-5mm} \includegraphics[scale=0.25,trim= 5 10 10 5, clip=true]{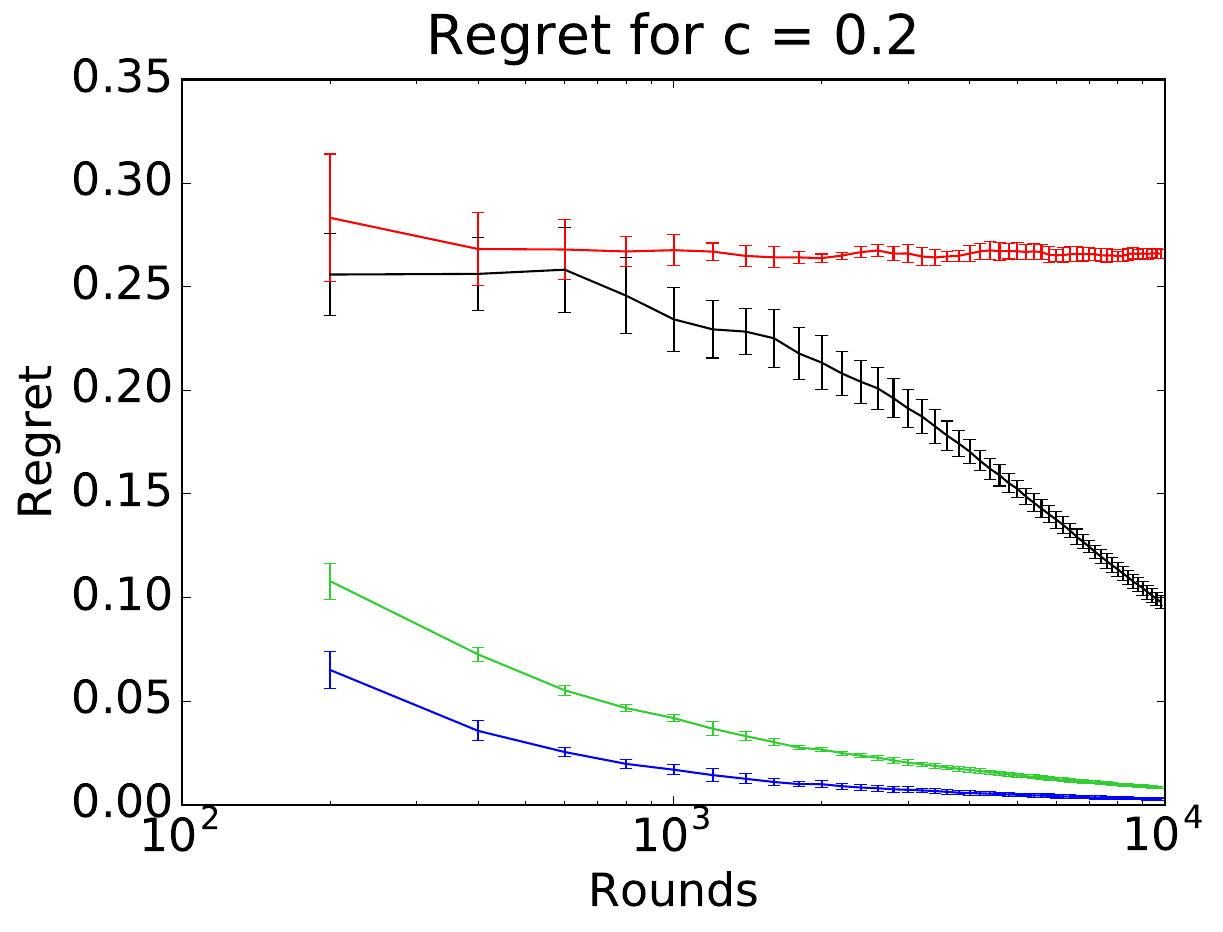} &
\hspace*{-5mm}\includegraphics[scale=0.25,trim= 5 10 10 5, clip=true]{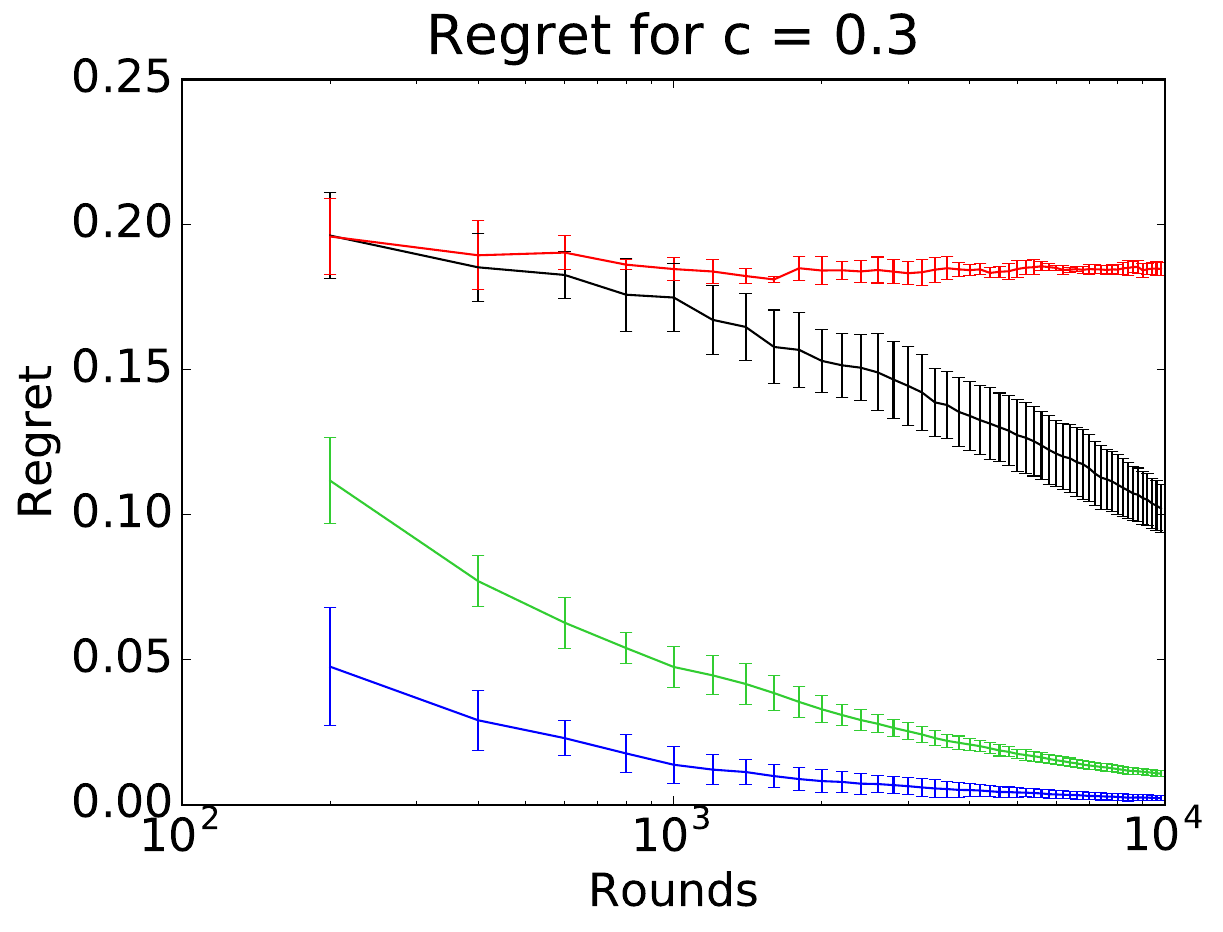} \\
\end{tabular}
\end{center}
\vskip -.15in
\caption{A graph of the averaged regret $R_t(\cdot)/t$ with standard
  deviations as a function of $t$ (log scale) for
  {\color[rgb]{0.16,0.67,0.16}\UCBGT }, \UCBNT , {\color{red} \UCB },
  and {\color{blue} \FTL } for different values of abstention costs.
  Each row is a dataset, starting from the top row we have: {\tt
    CIFAR}, {\tt ijcnn}, {\tt HIGGS}, {\tt phishing}, and {\tt
    covtype}. }
\label{fig:fullres1}
\vskip -.1in
\end{figure*}

\begin{figure*}[!ht]
\begin{center}
\begin{tabular}{ c c c }
\includegraphics[scale=0.25,trim= 5 10 10 5, clip=true]{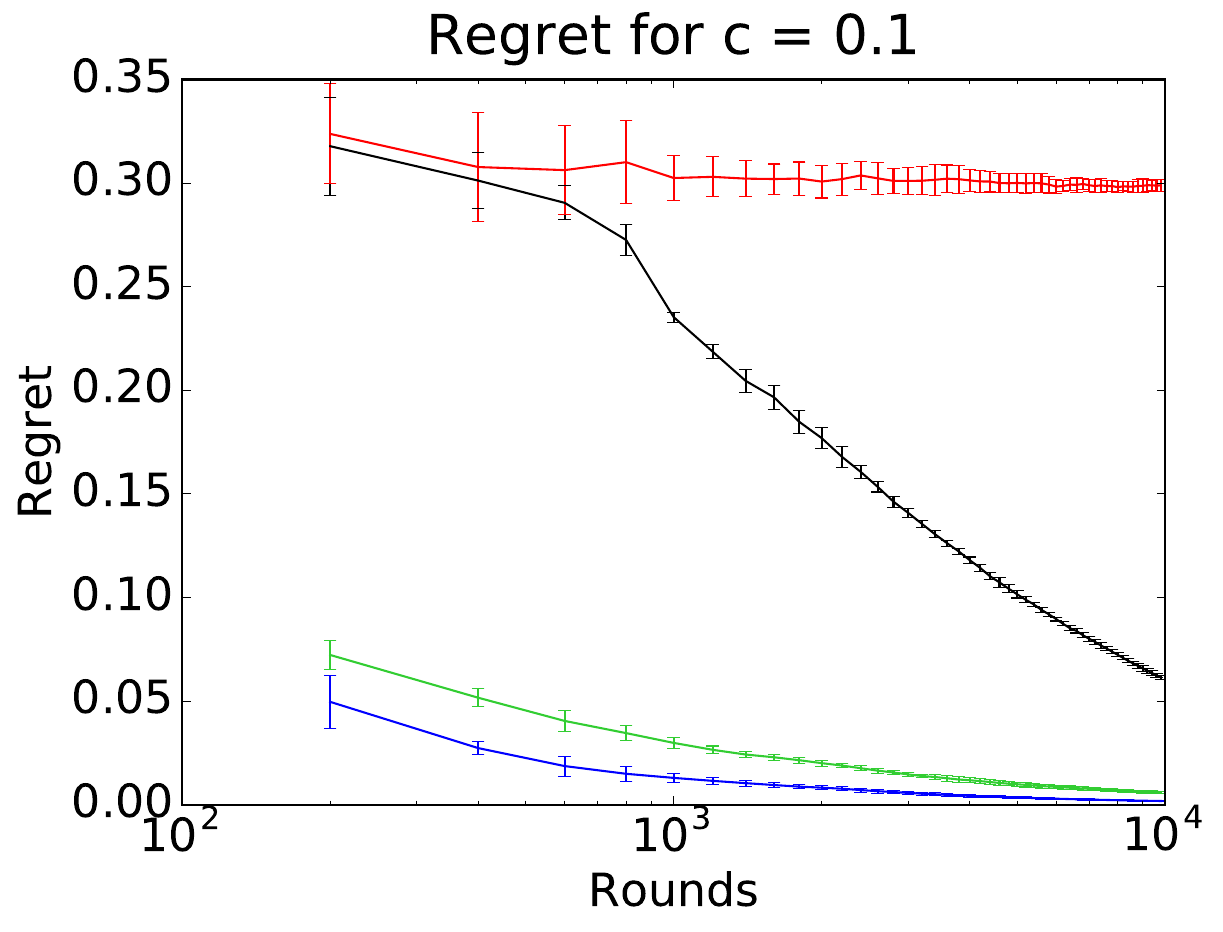} &
\hspace*{-5mm} \includegraphics[scale=0.25,trim= 5 10 10 5, clip=true]{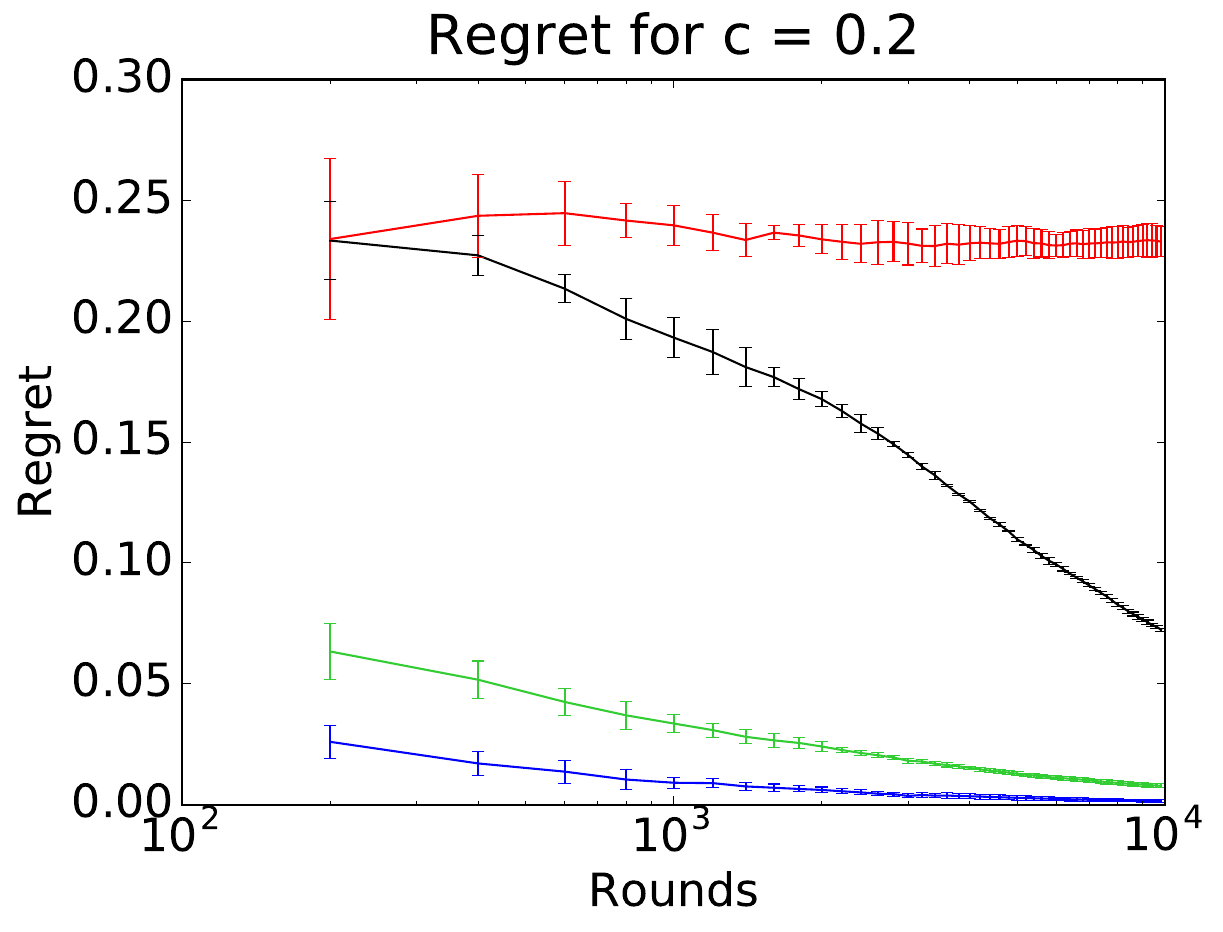} &
\hspace*{-5mm}\includegraphics[scale=0.25,trim= 5 10 10 5, clip=true]{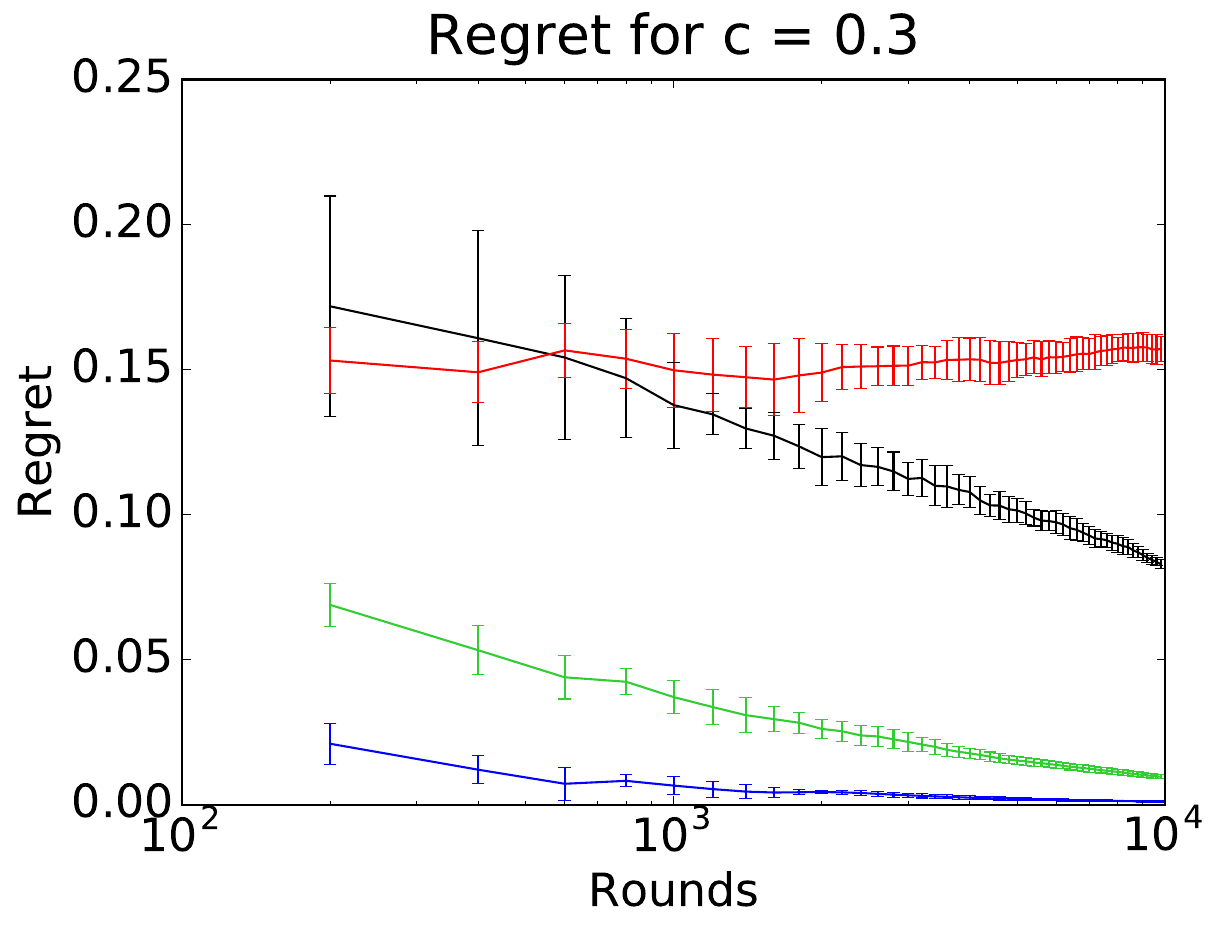} \\
\includegraphics[scale=0.25,trim= 5 10 10 5, clip=true]{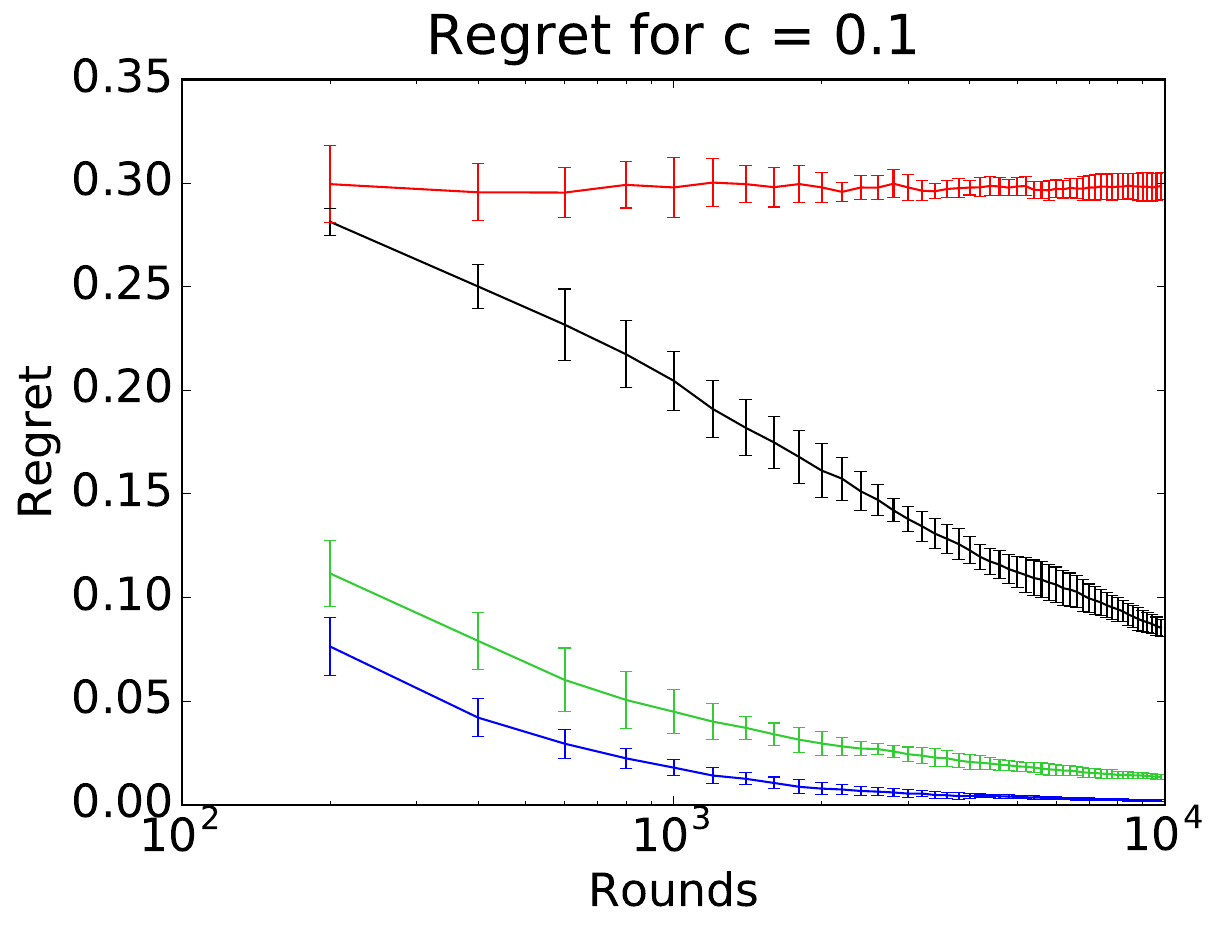} &
\hspace*{-5mm} \includegraphics[scale=0.25,trim= 5 10 10 5, clip=true]{exp_results3/regret_c02_d_8.pdf} &
\hspace*{-5mm}\includegraphics[scale=0.25,trim= 5 10 10 5, clip=true]{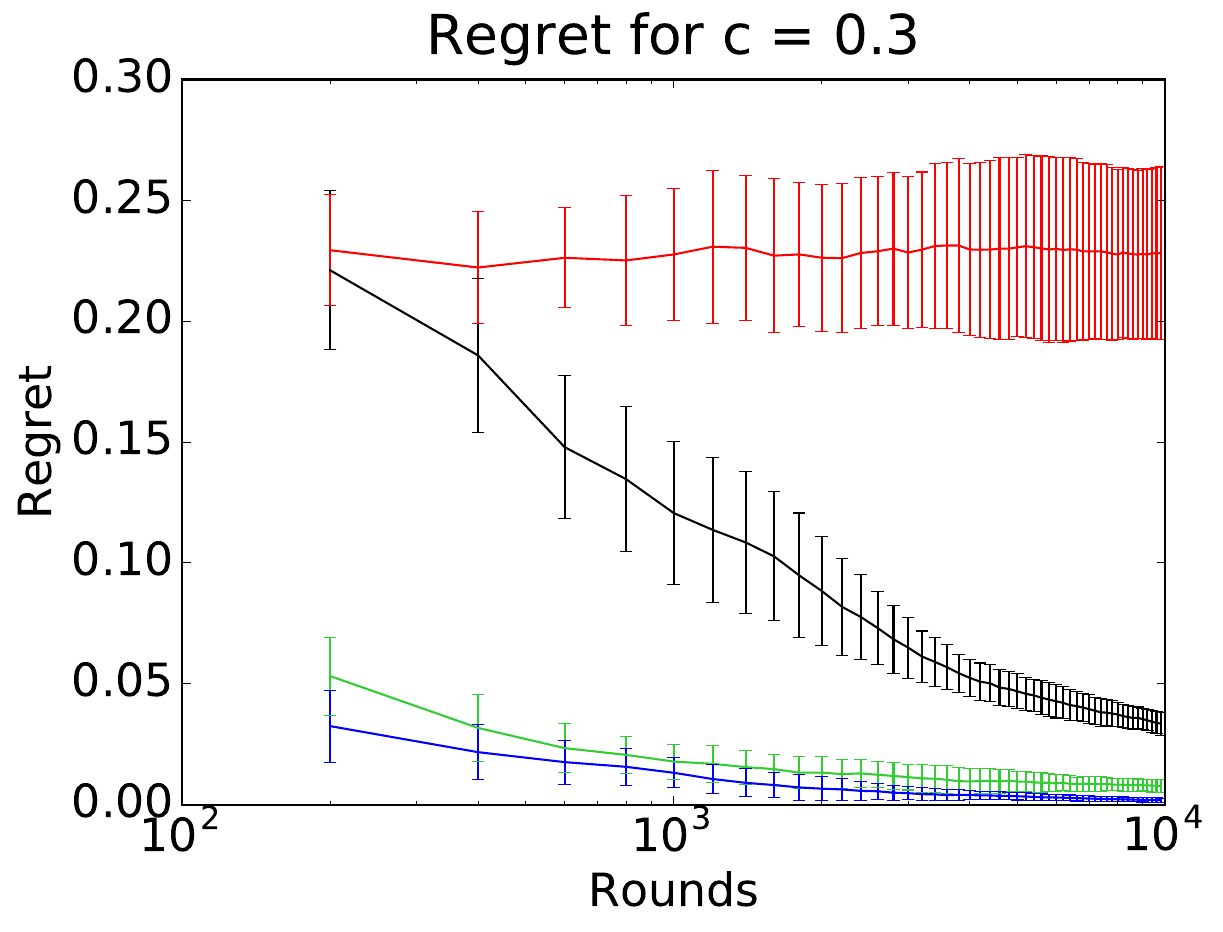} \\
\includegraphics[scale=0.25,trim= 5 10 10 5, clip=true]{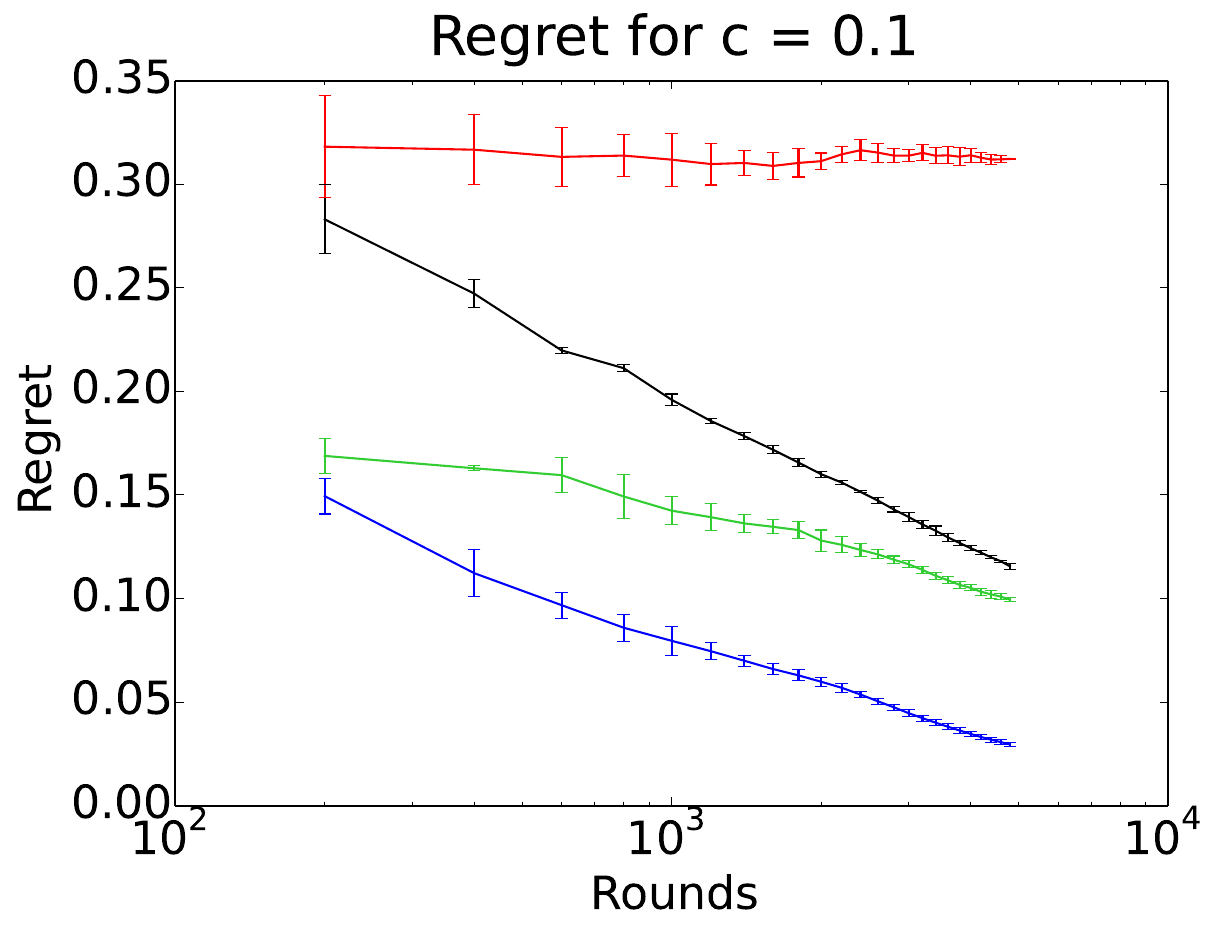} &
\hspace*{-5mm} \includegraphics[scale=0.25,trim= 5 10 10 5, clip=true]{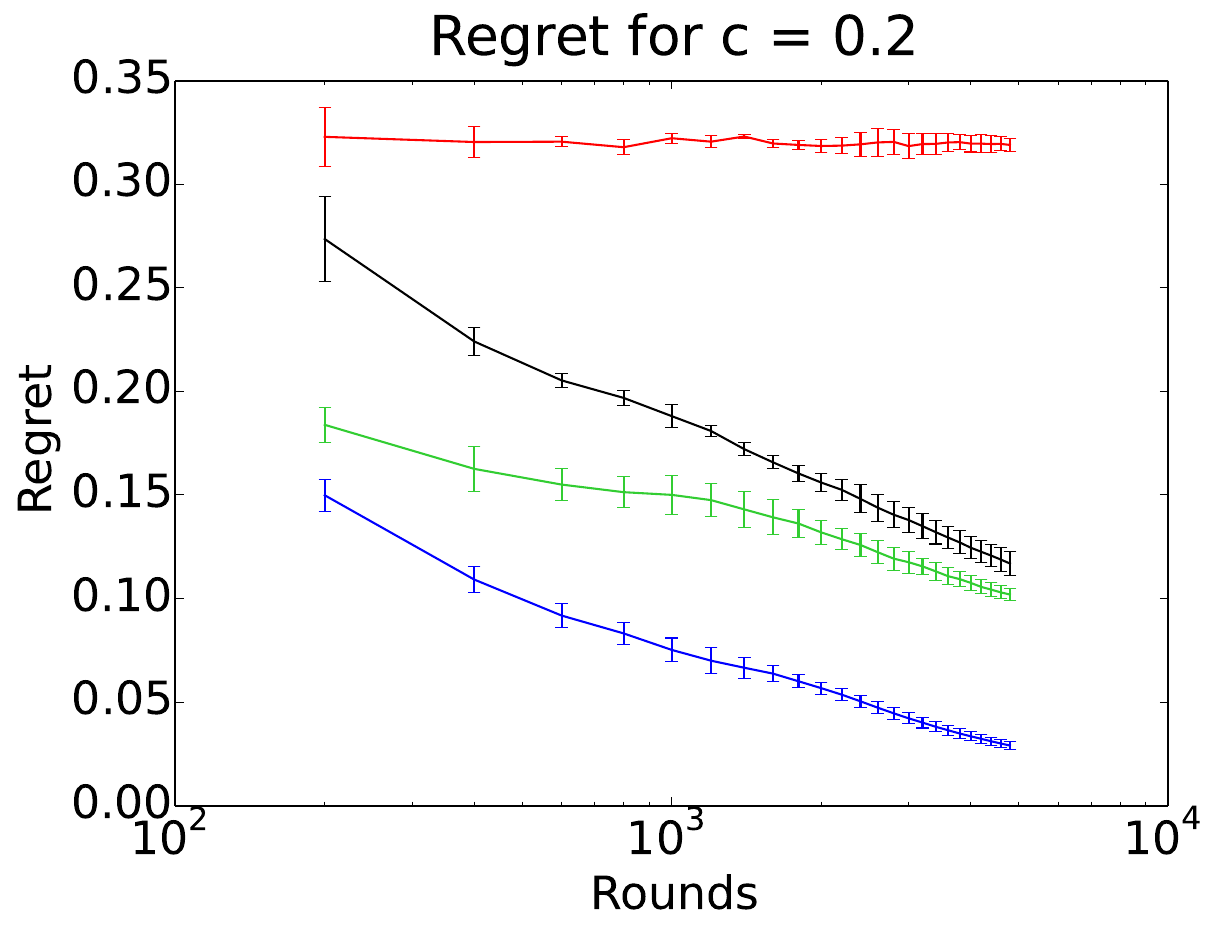} &
\hspace*{-5mm}\includegraphics[scale=0.25,trim= 5 10 10 5, clip=true]{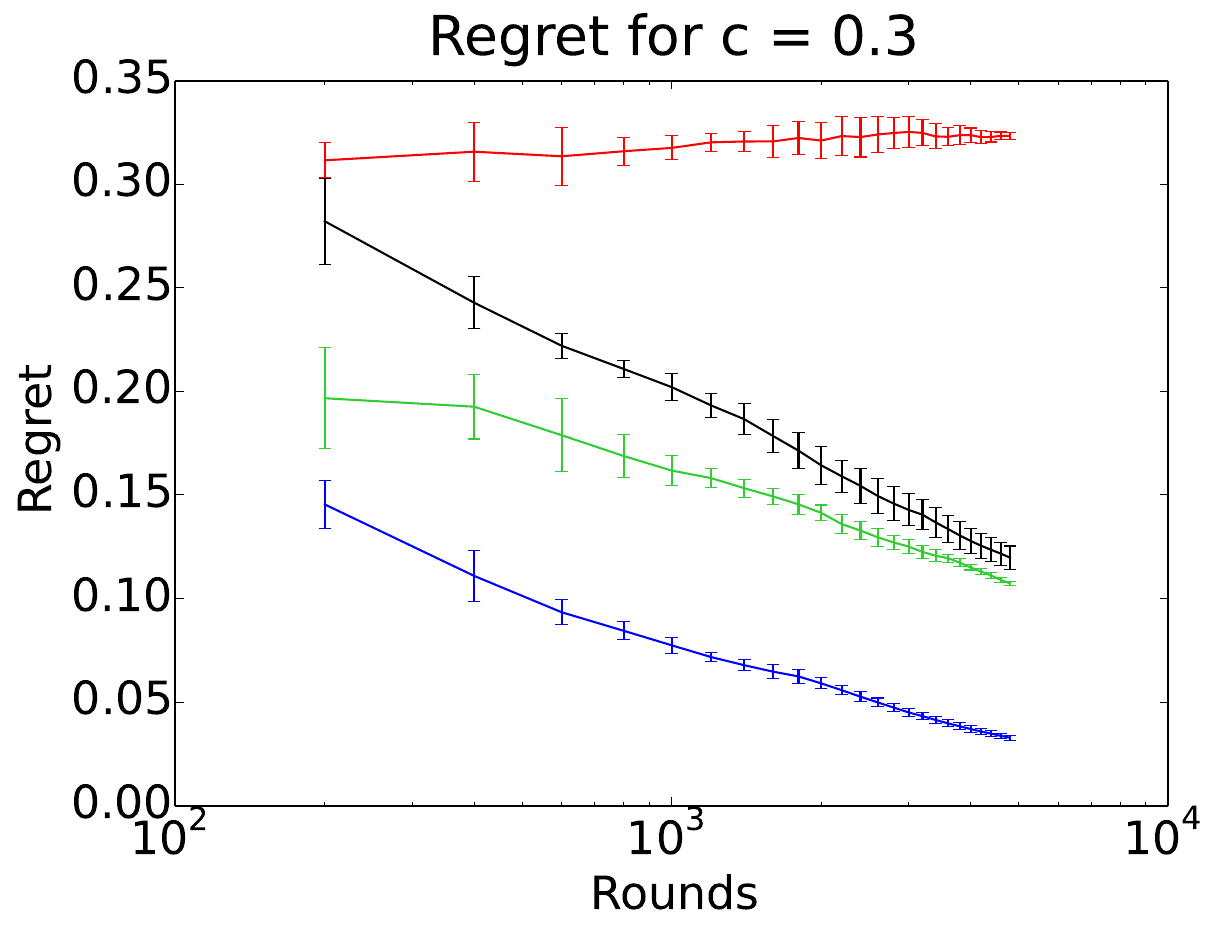} \\
\includegraphics[scale=0.25,trim= 5 10 10 5, clip=true]{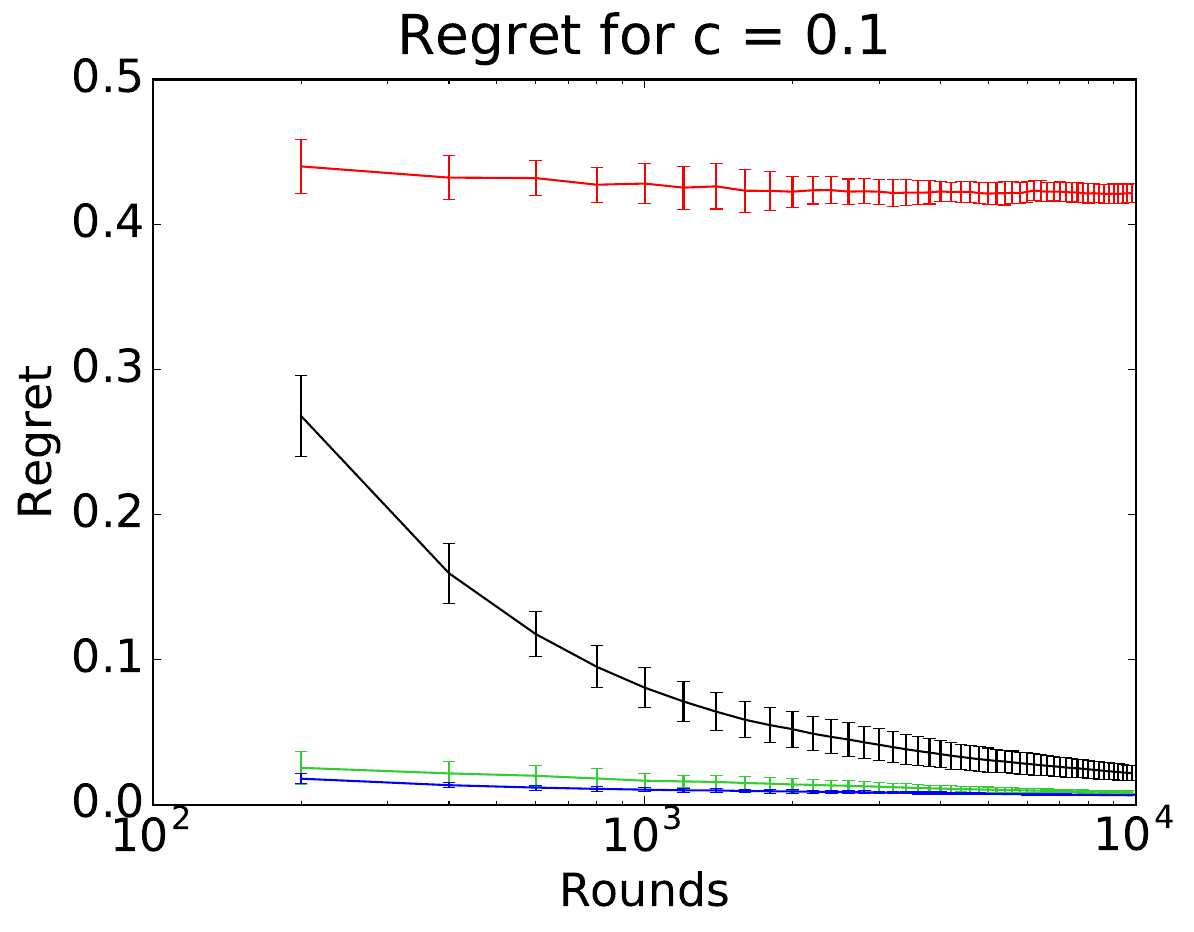} &
\hspace*{-5mm} \includegraphics[scale=0.25,trim= 5 10 10 5, clip=true]{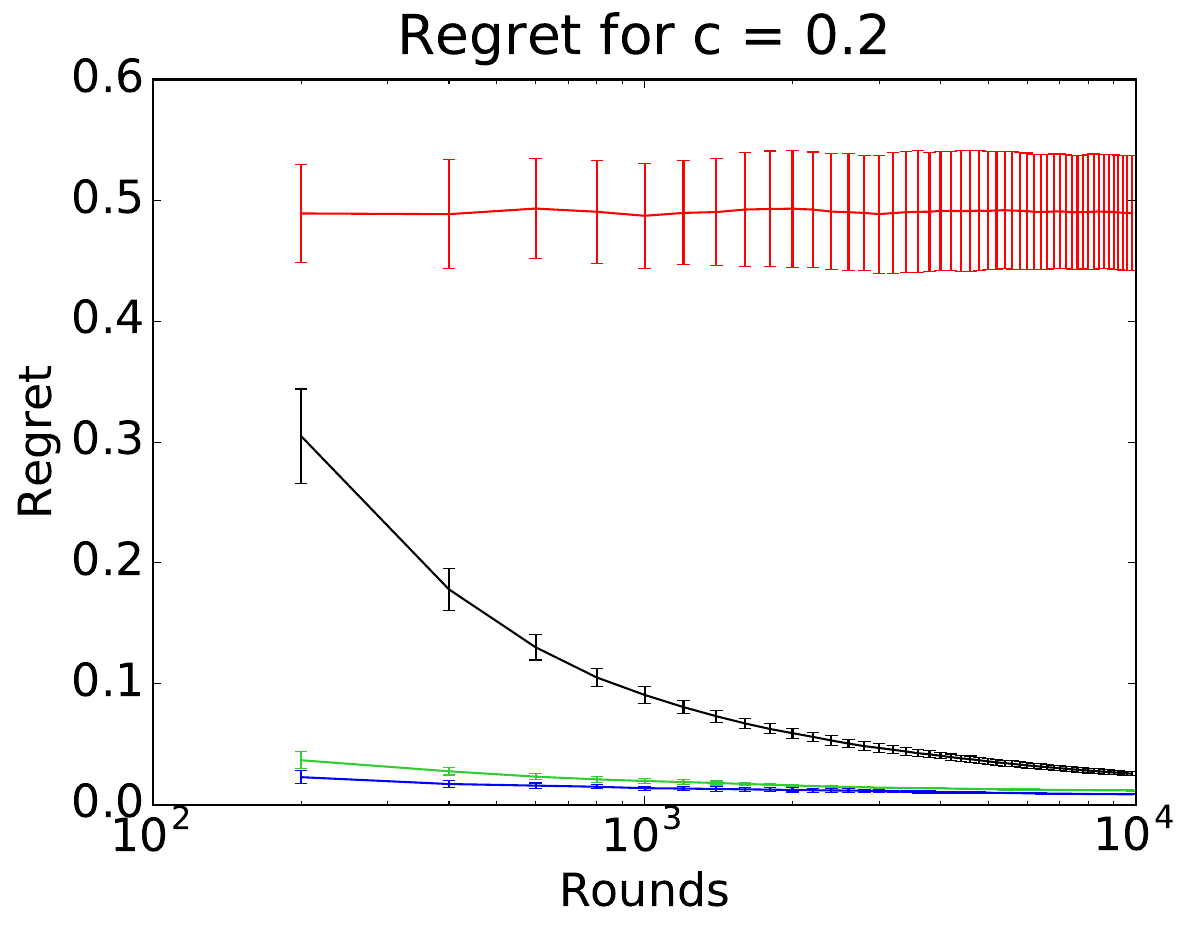} &
\hspace*{-5mm}\includegraphics[scale=0.25,trim= 5 10 10 5, clip=true]{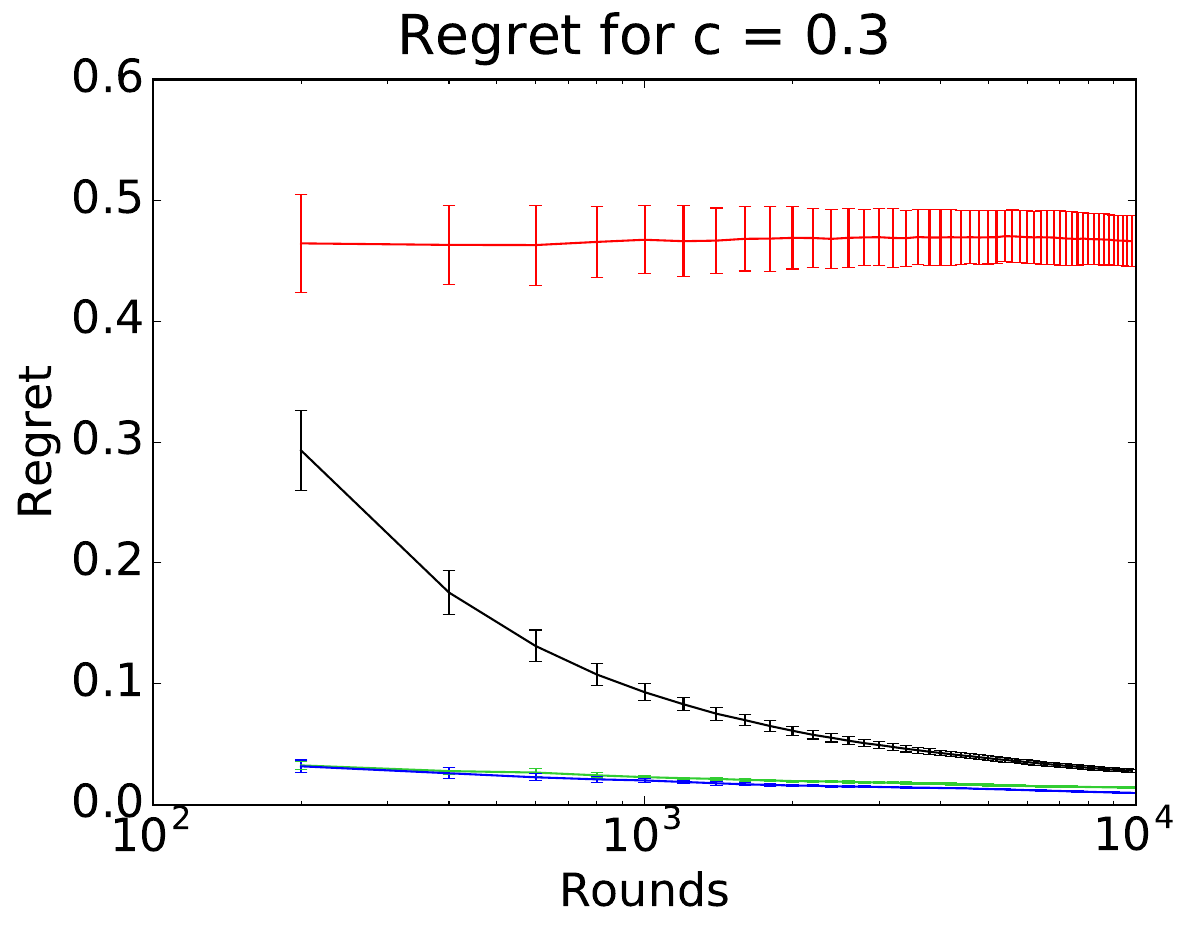} \\
\includegraphics[scale=0.25,trim= 5 10 10 5, clip=true]{exp_results3/regret_c01_d_4.pdf} &
\hspace*{-5mm} \includegraphics[scale=0.25,trim= 5 10 10 5, clip=true]{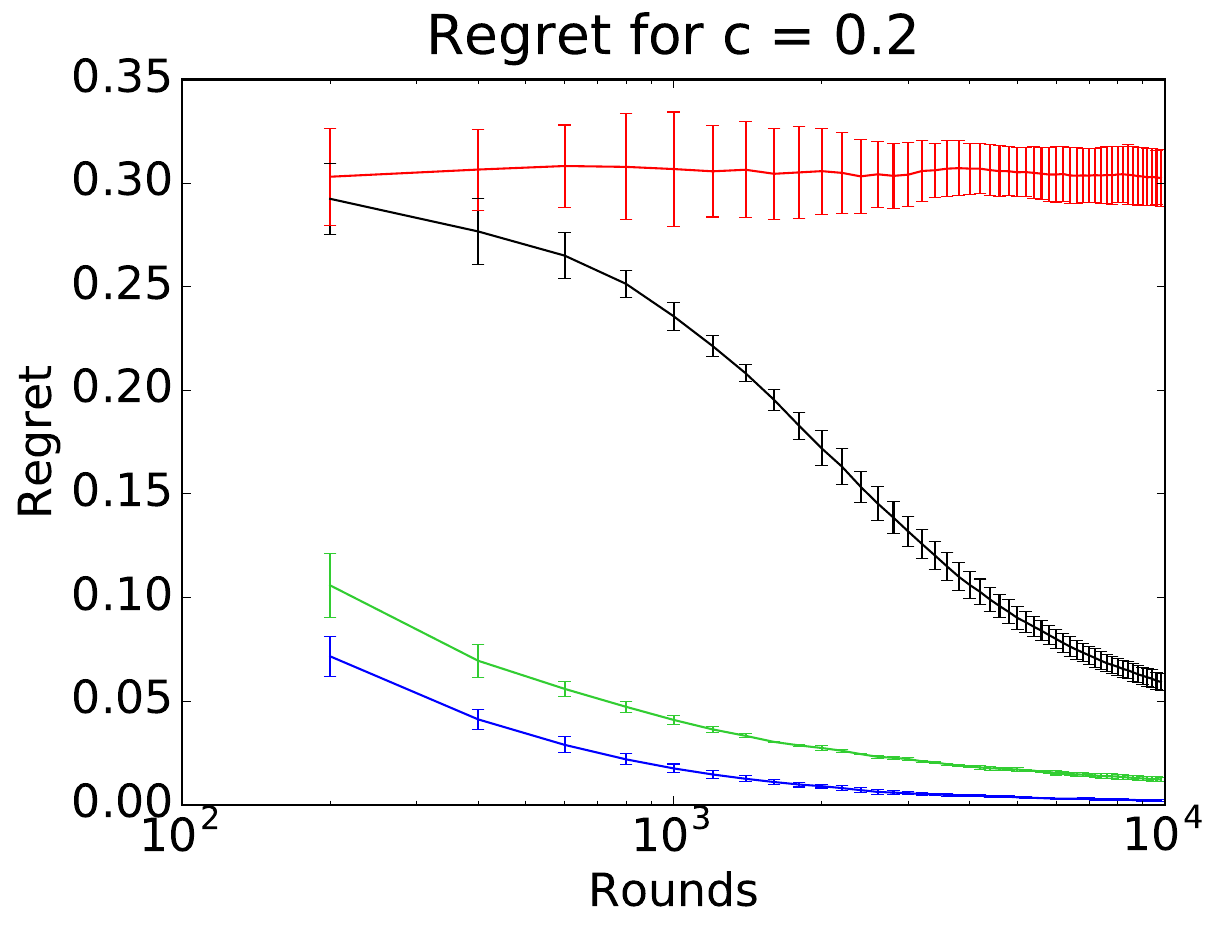}&
\hspace*{-5mm}\includegraphics[scale=0.25,trim= 5 10 10 5, clip=true]{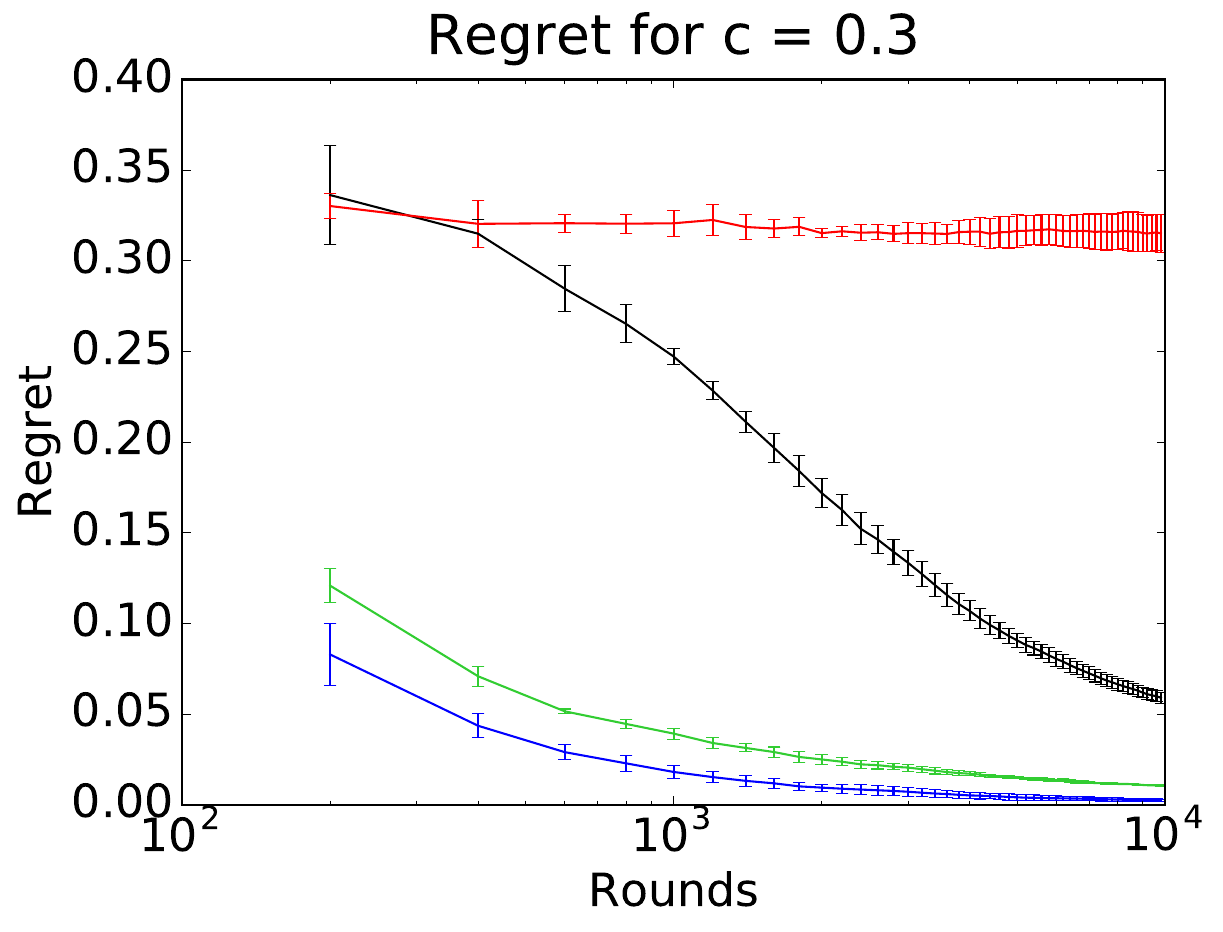} \\
\end{tabular}
\end{center}
\vskip -.15in
\caption{A graph of the averaged regret $R_t(\cdot)/t$ with standard
  deviations as a function of $t$ (log scale) for
  {\color[rgb]{0.16,0.67,0.16}\UCBGT }, \UCBNT , {\color{red} \UCB },
  and {\color{blue} \FTL } for different values of abstention costs.
  Each row is a dataset, starting from the top row we have: {\tt eye},
  {\tt cod-ran}, {\tt synthetic}, {\tt skin}, and {\tt guide}.  }
\label{fig:fullres2}
\vskip -.1in
\end{figure*}

In this section, we present several figures showing our experimental
results. Figure~\ref{fig:fullres1} and Figure~\ref{fig:fullres2} show
the regret for different abstention costs $c\in \{0.1,0.2,0.3\}$ for
all our datasets. We observe that, in general, \UCBGT\ outperforms \UCBNT\
and \UCB\ for all datasets and is even within the standard deviation
of the \FTL 's regret for some of datasets. The figures also indicate
that the regret of \UCB\ decreases slowly. This is expected, since there
are $2\mathord,100$ experts, $10\mathord,000$ time steps, and the algorithm only
updates a single expert per time step. 

\begin{figure*}[!ht]
\begin{center}
\begin{tabular}{ c c c }
\includegraphics[scale=0.25,trim= 5 10 10 5, clip=true]{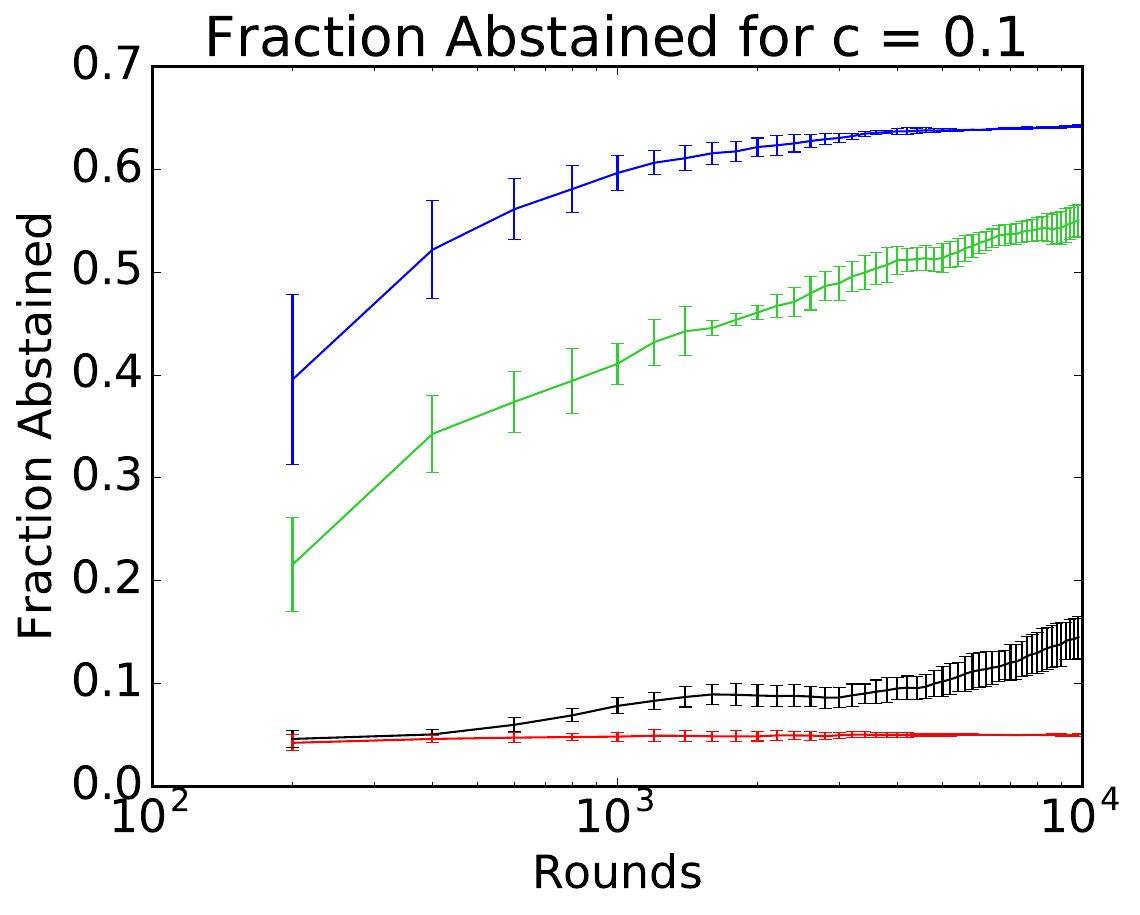} &
\hspace*{-5mm} \includegraphics[scale=0.25,trim= 5 10 10 5, clip=true]{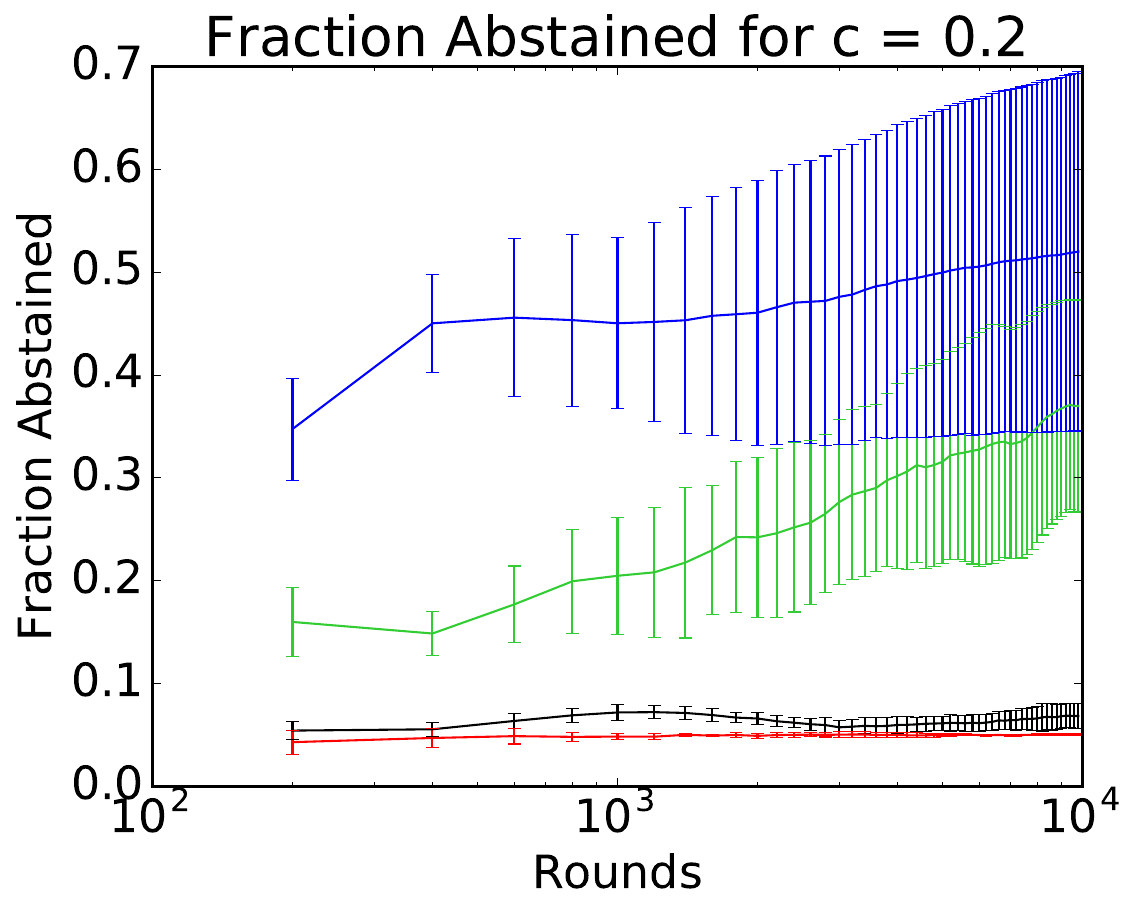} &
\hspace*{-5mm}\includegraphics[scale=0.25,trim= 5 10 10 5, clip=true]{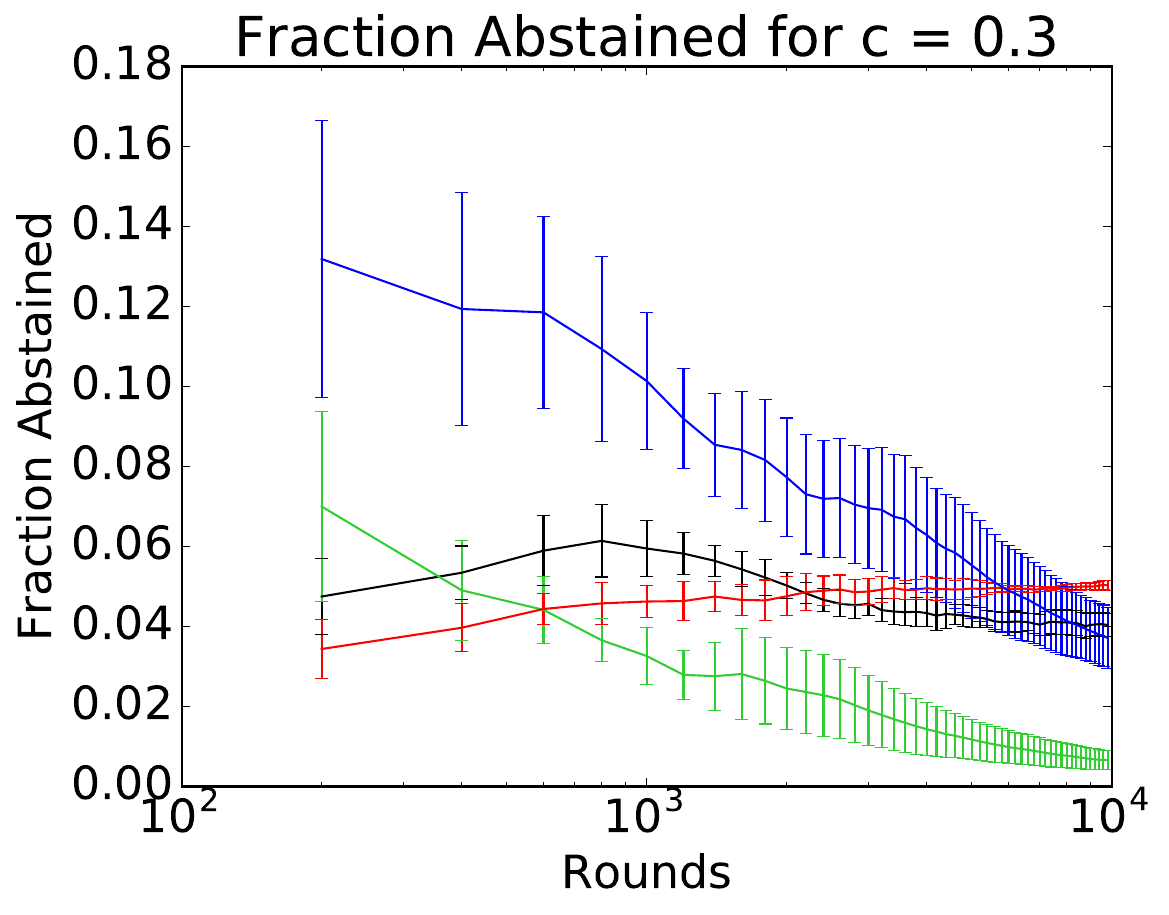} \\
\includegraphics[scale=0.25,trim= 5 10 10 5, clip=true]{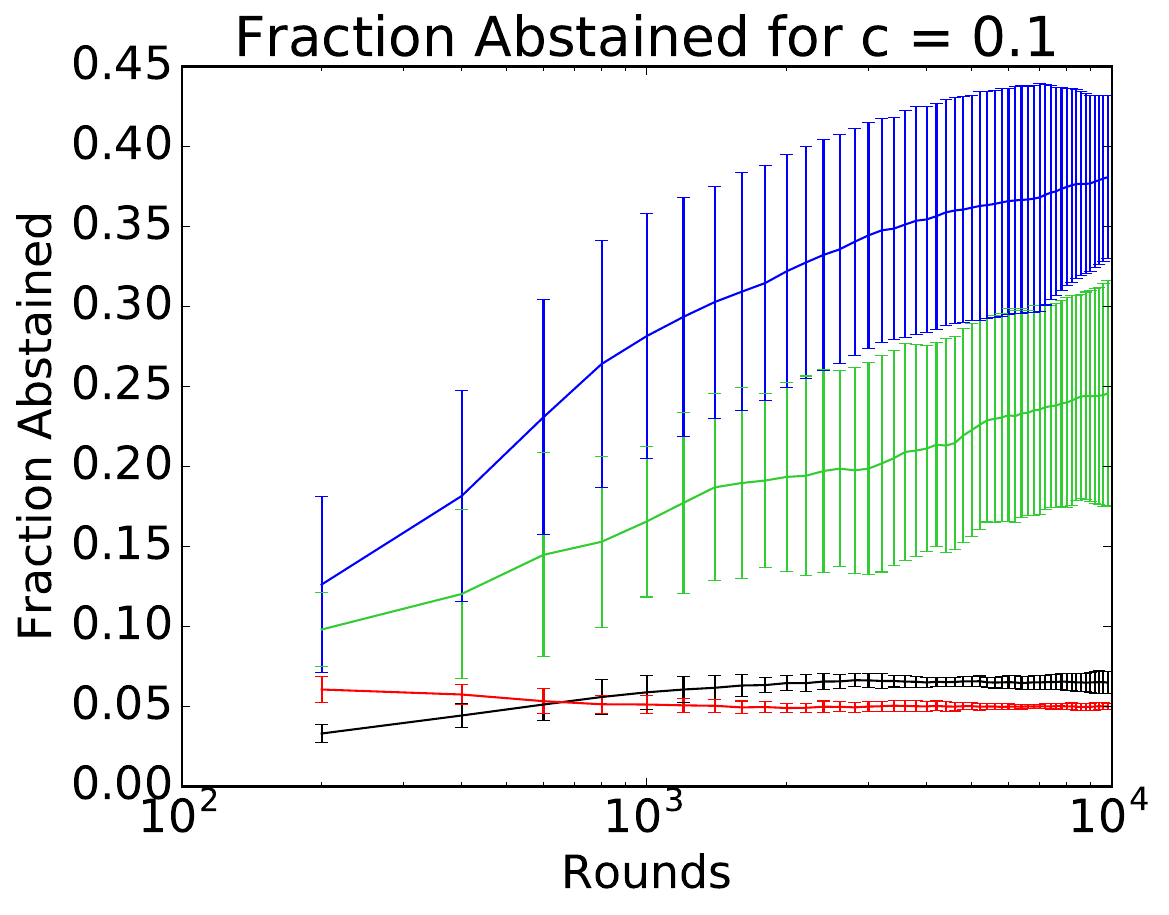} &
\hspace*{-5mm} \includegraphics[scale=0.25,trim= 5 10 10 5, clip=true]{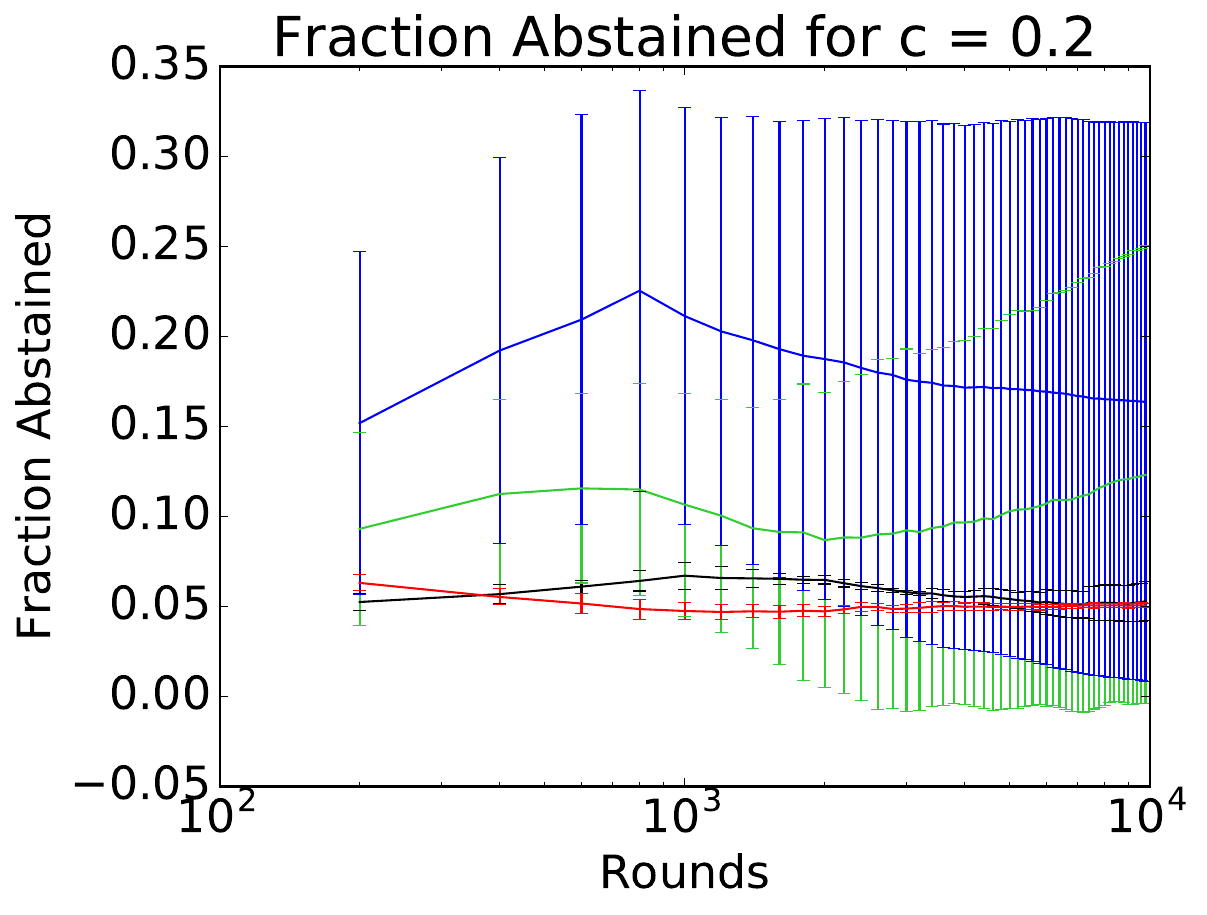} &
\hspace*{-5mm}\includegraphics[scale=0.25,trim= 5 10 10 5, clip=true]{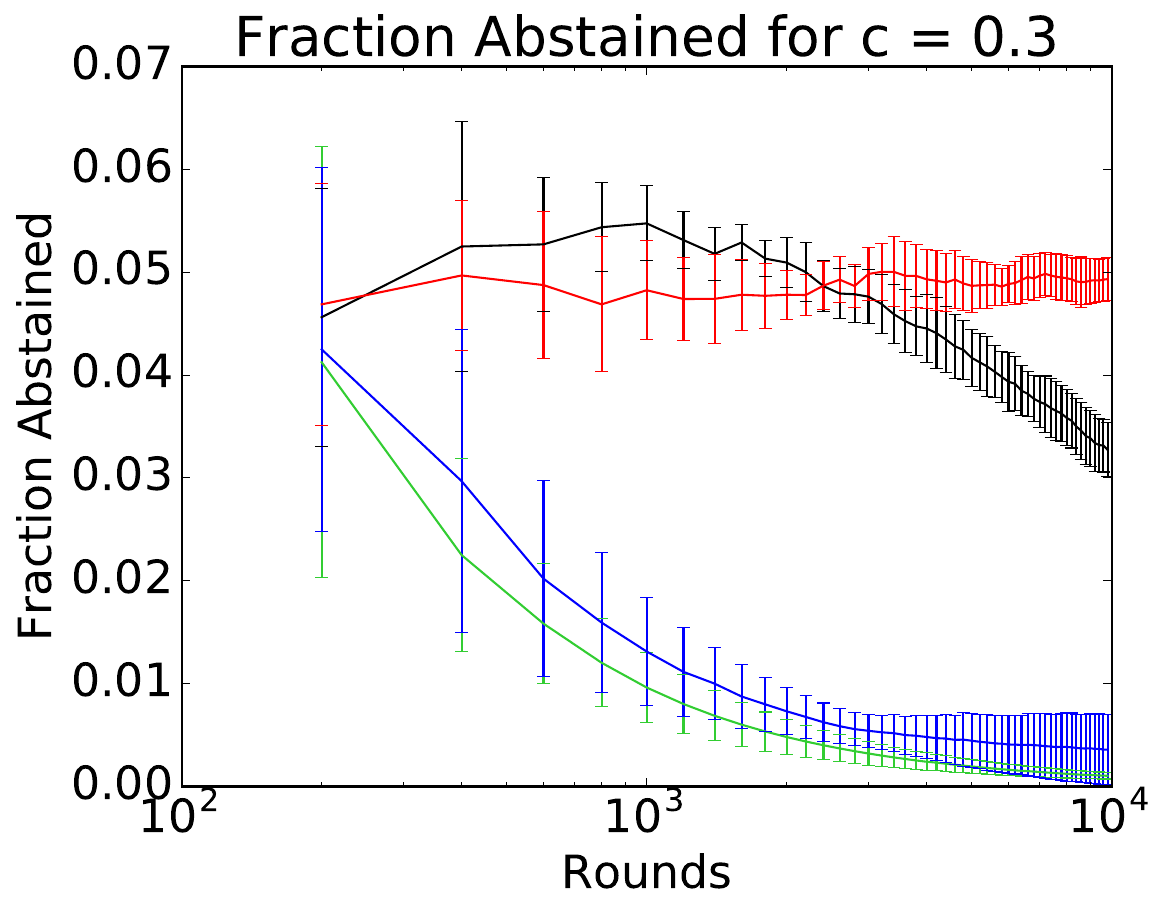} \\
\includegraphics[scale=0.25,trim= 5 10 10 5, clip=true]{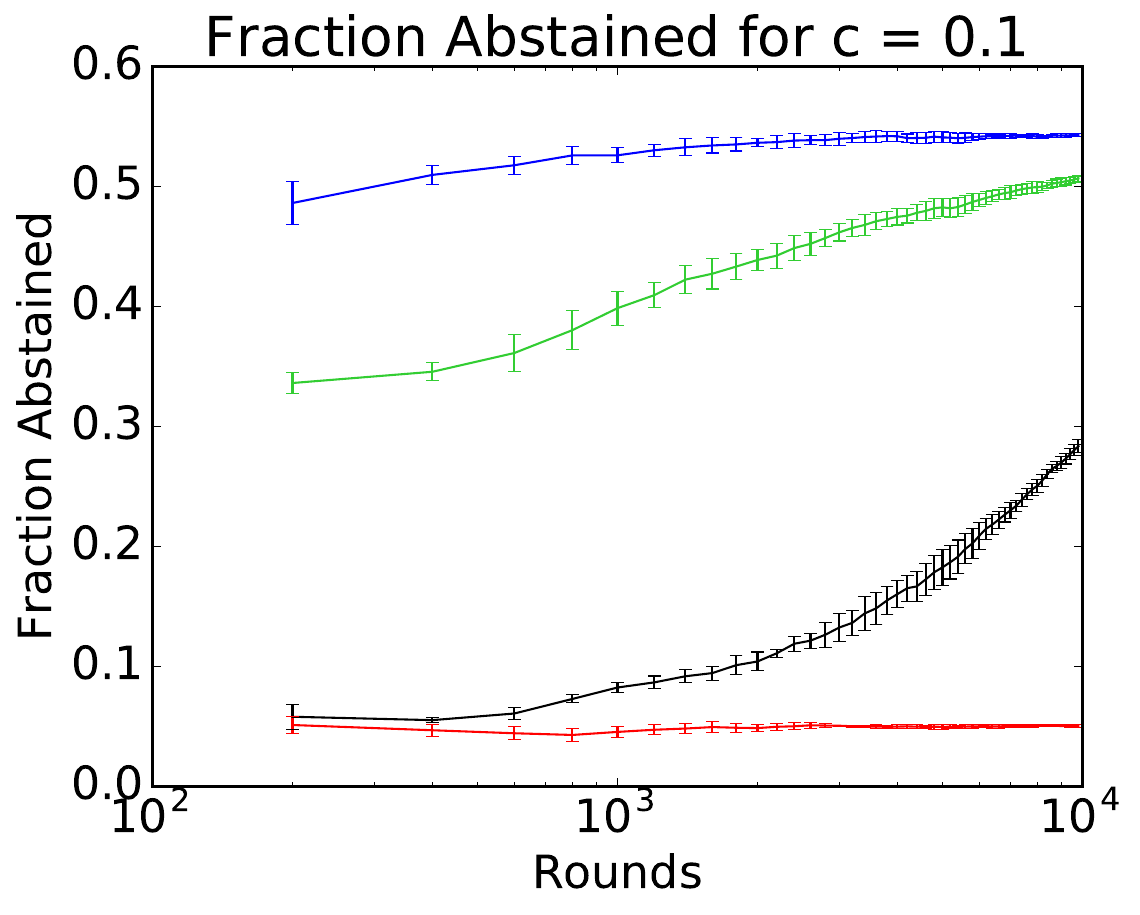} &
\hspace*{-5mm} \includegraphics[scale=0.25,trim= 5 10 10 5, clip=true]{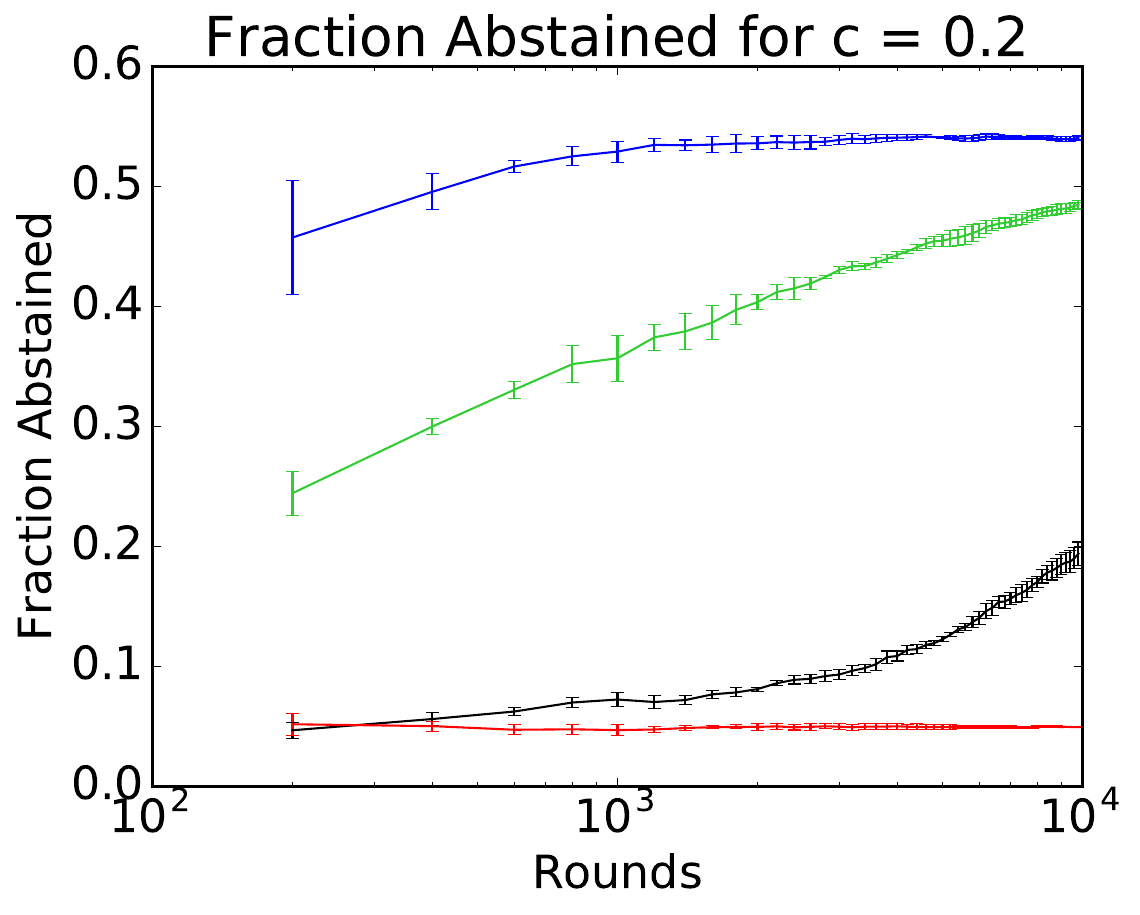} &
\hspace*{-5mm}\includegraphics[scale=0.25,trim= 5 10 10 5, clip=true]{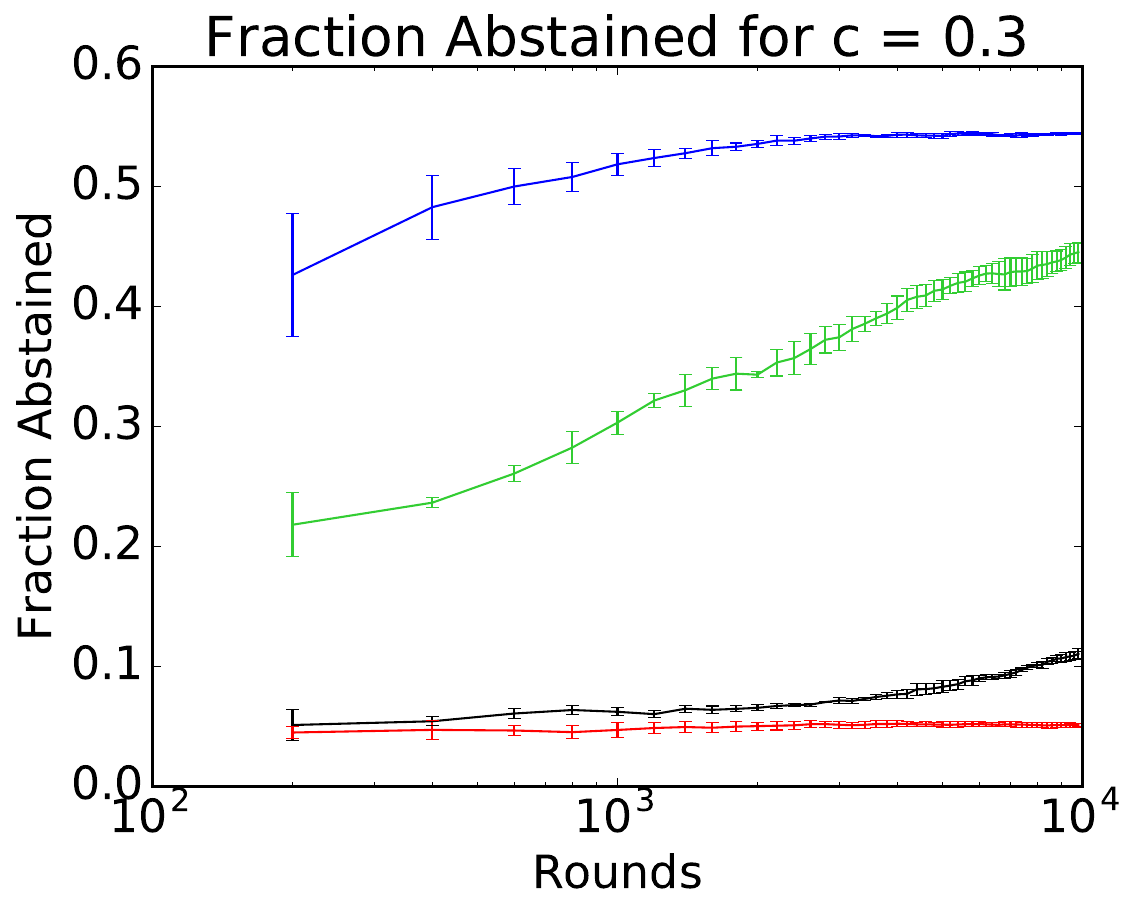} \\
\includegraphics[scale=0.25,trim= 5 10 10 5, clip=true]{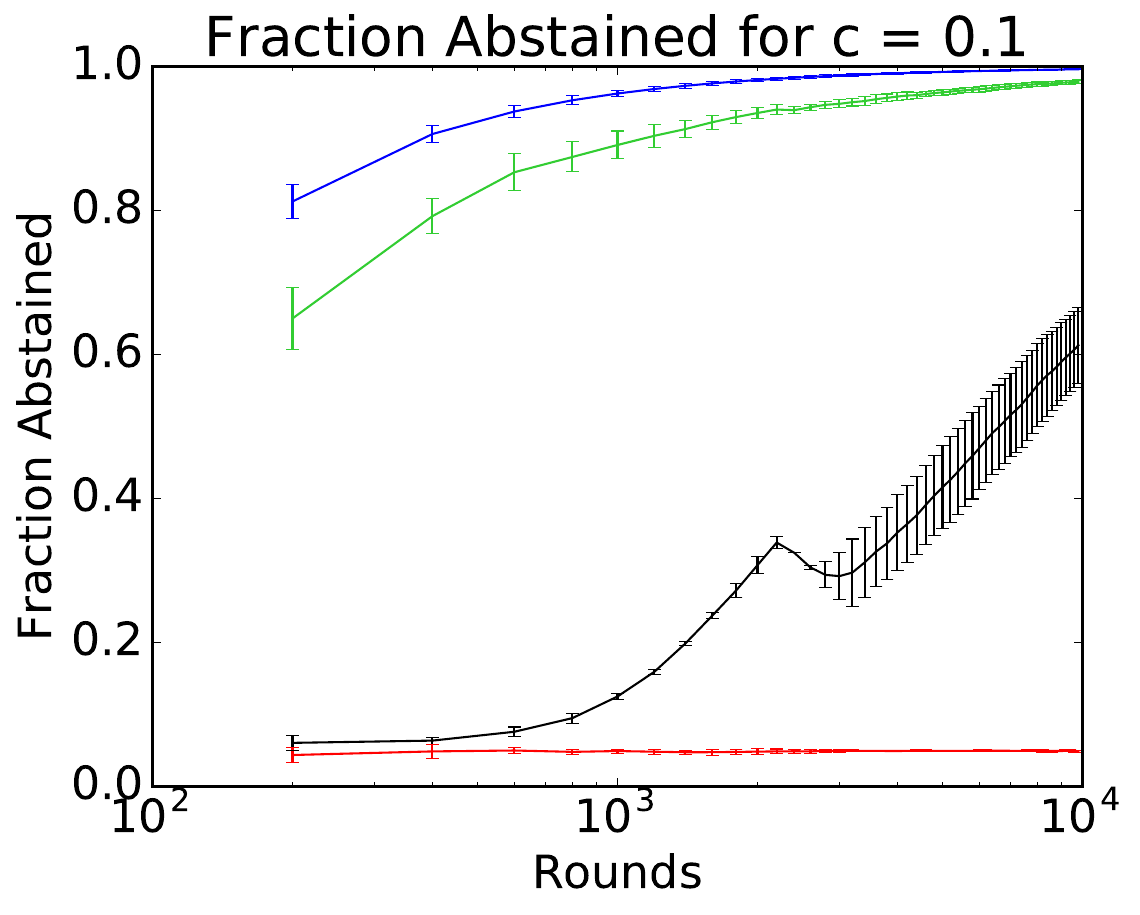} &
\hspace*{-5mm} \includegraphics[scale=0.25,trim= 5 10 10 5, clip=true]{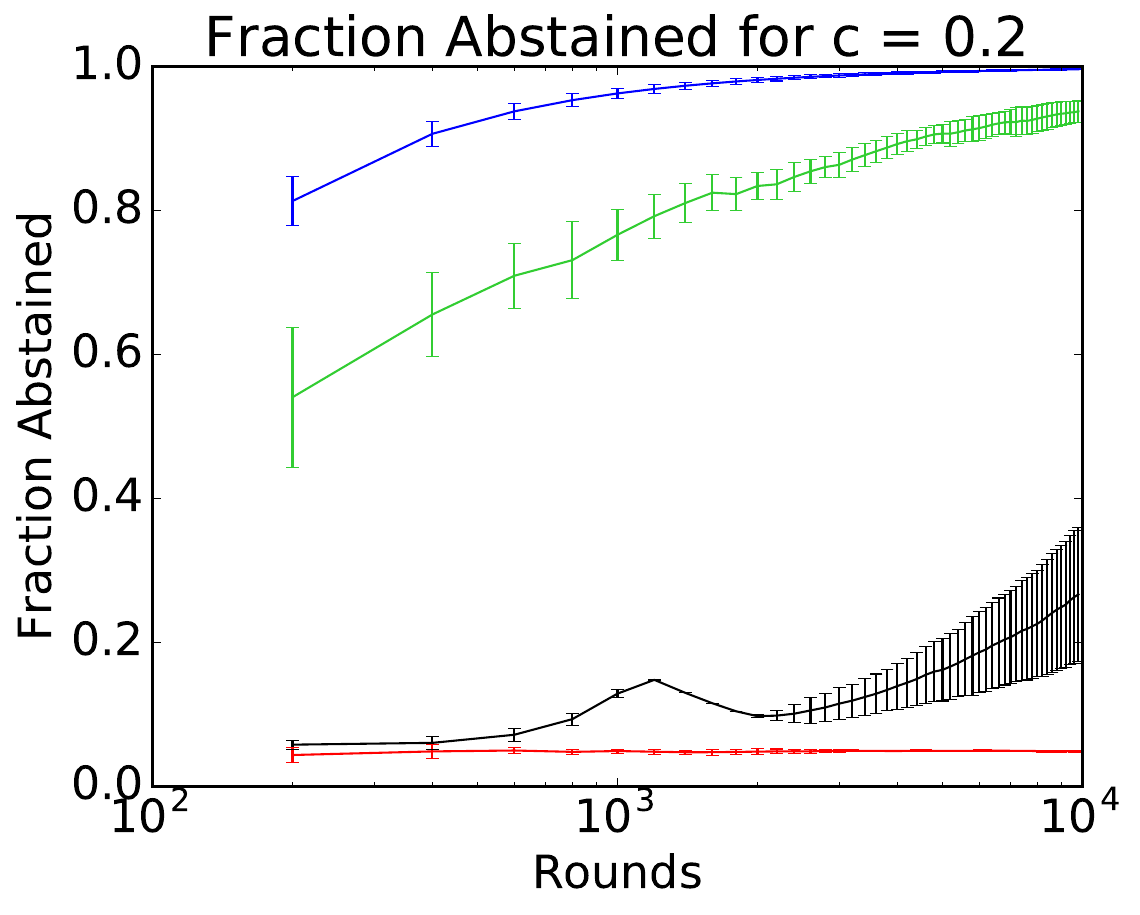} &
\hspace*{-5mm}\includegraphics[scale=0.25,trim= 5 10 10 5, clip=true]{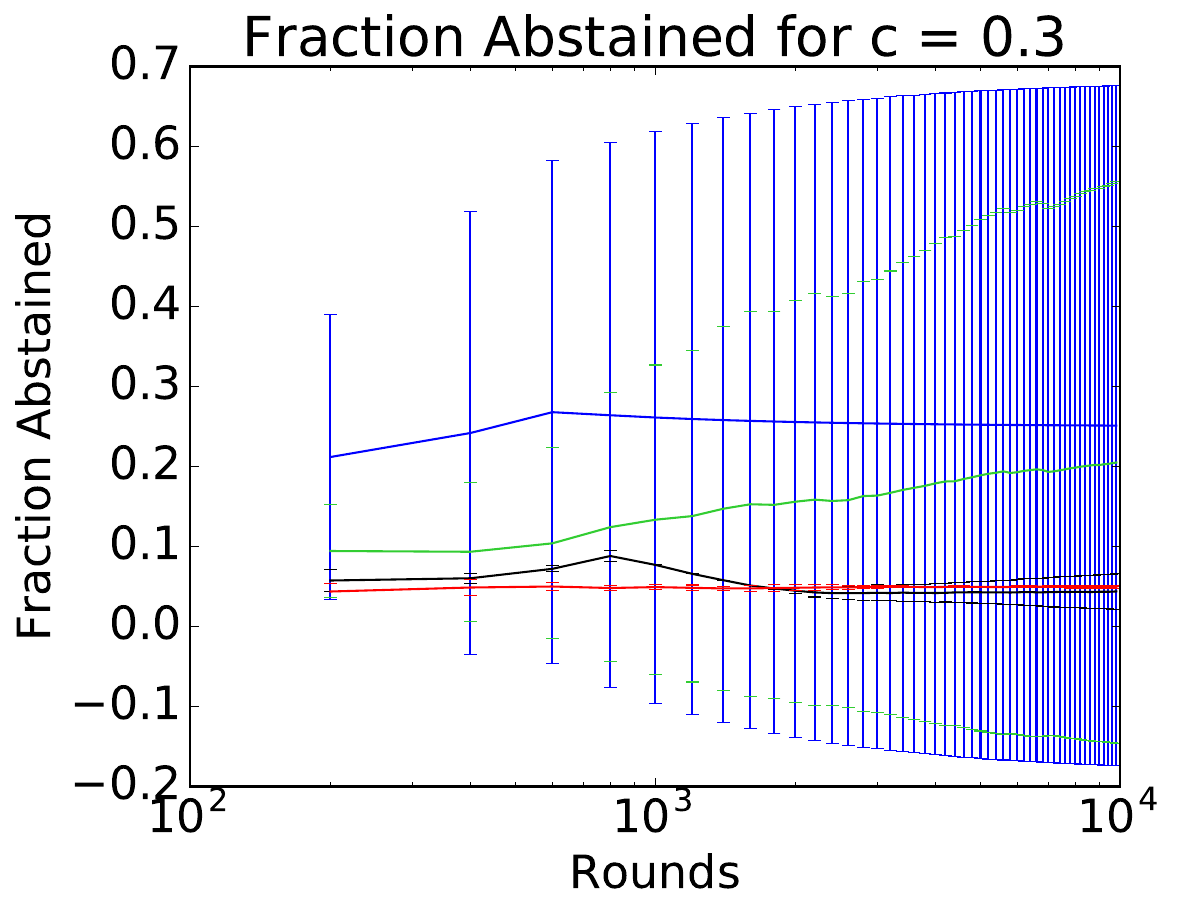} \\
\includegraphics[scale=0.25,trim= 5 10 10 5, clip=true]{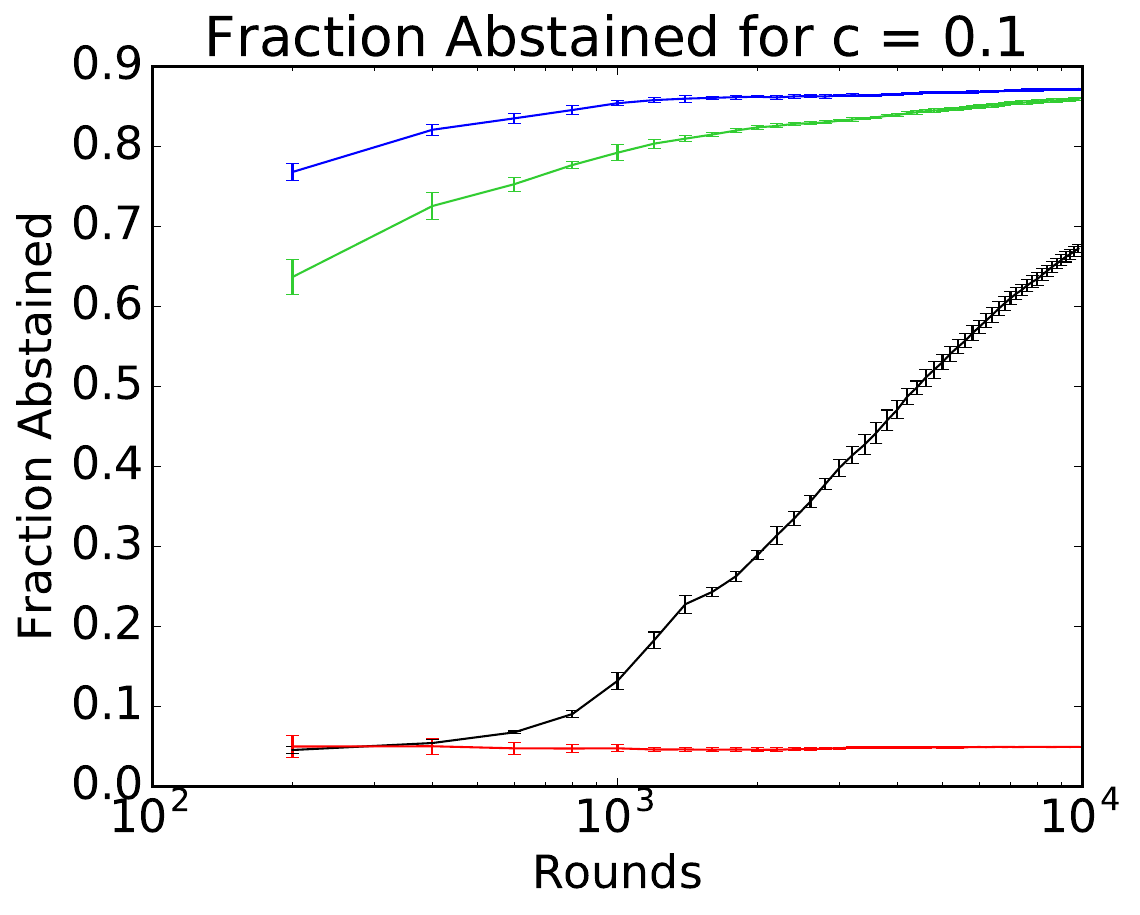} &
\hspace*{-5mm} \includegraphics[scale=0.25,trim= 5 10 10 5, clip=true]{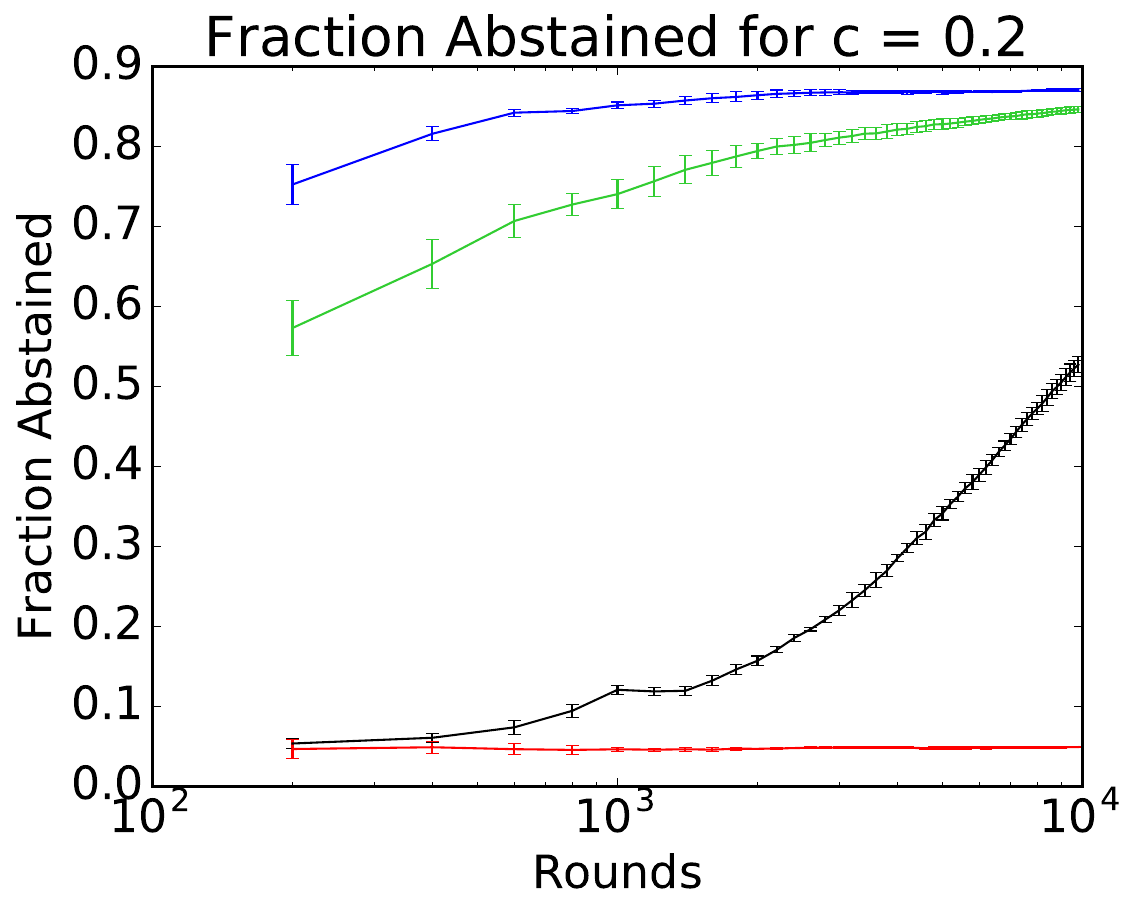}&
\hspace*{-5mm}\includegraphics[scale=0.25,trim= 5 10 10 5, clip=true]{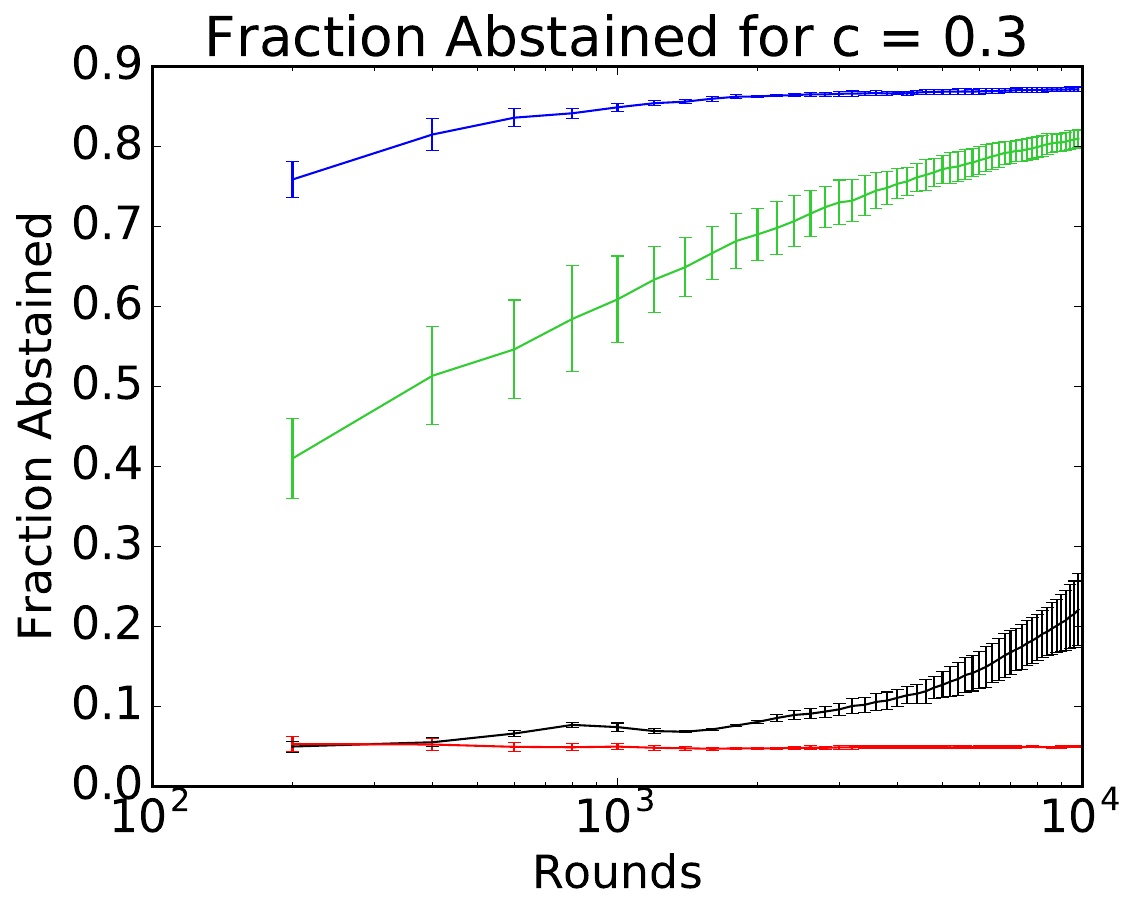} \\
\end{tabular}
\end{center}
\vskip -.15in
\caption{A graph of the averaged fraction of abstained points with
  standard deviations as a function of $t$ (log scale) for
  {\color[rgb]{0.16,0.67,0.16}\UCBGT }, \UCBNT , {\color{red} \UCB },
  and {\color{blue} \FTL } for different values of abstention costs.
  Each row is a dataset, starting from the top row we have: {\tt
    CIFAR}, {\tt ijcnn}, {\tt HIGGS}, {\tt phishing}, and {\tt
    covtype}. }
\label{fig:frac1}
\vskip -.1in
\end{figure*}

\begin{figure*}[!ht]
\begin{center}
\begin{tabular}{ c c c }
\includegraphics[scale=0.25,trim= 5 10 10 5, clip=true]{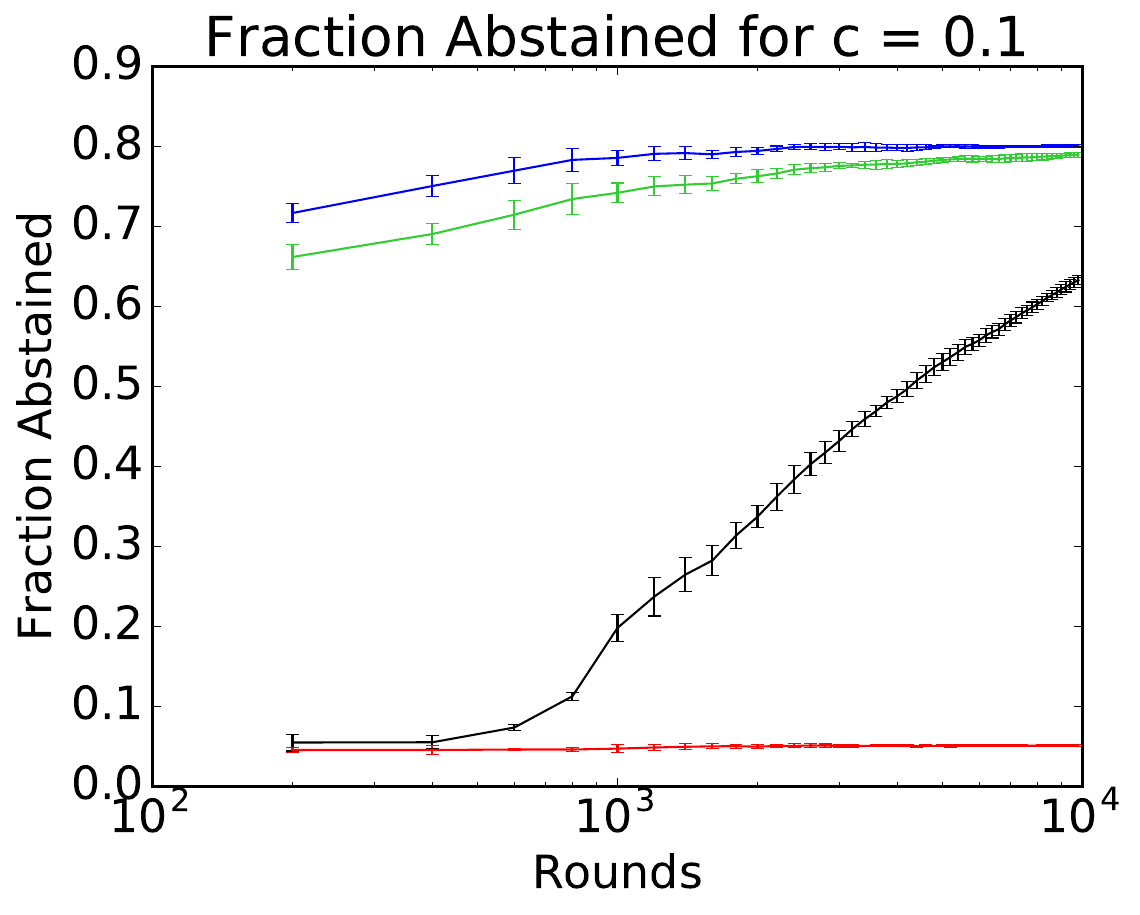} &
\hspace*{-5mm} \includegraphics[scale=0.25,trim= 5 10 10 5, clip=true]{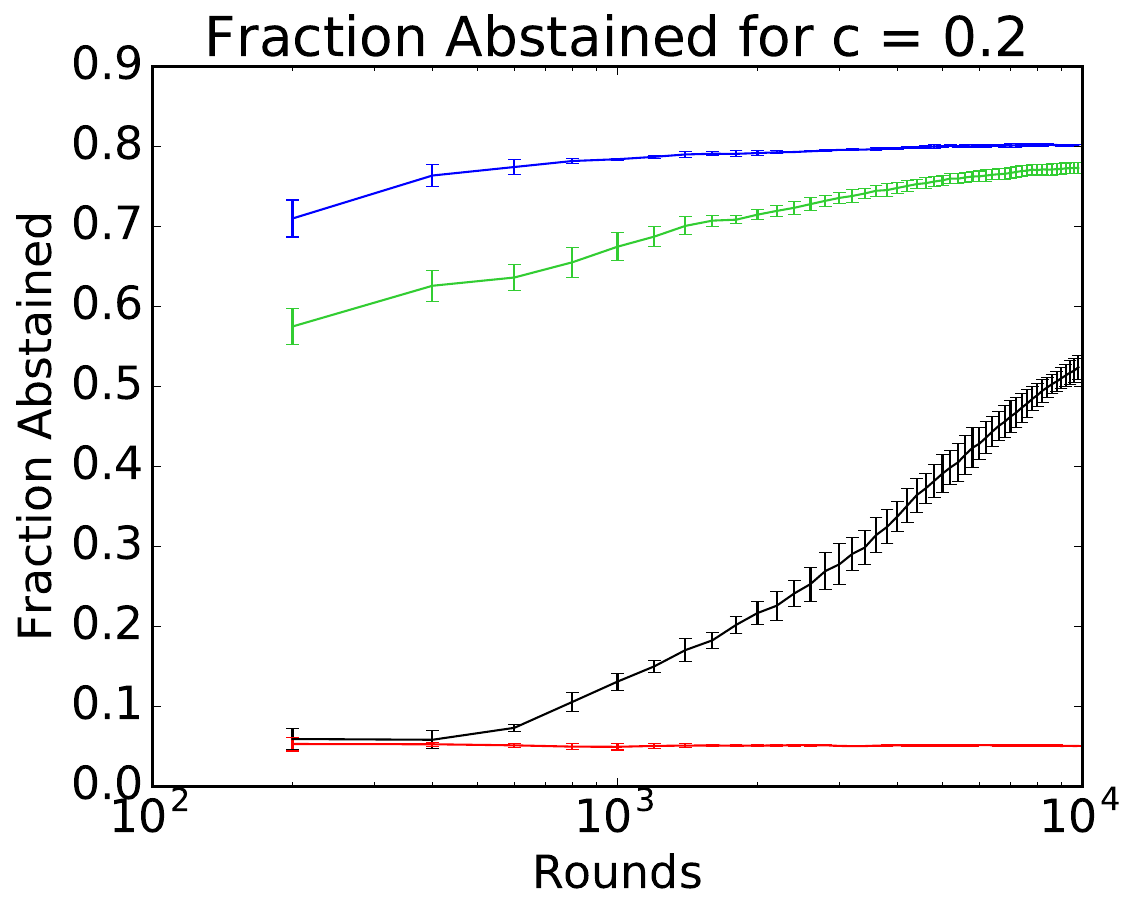} &
\hspace*{-5mm}\includegraphics[scale=0.25,trim= 5 10 10 5, clip=true]{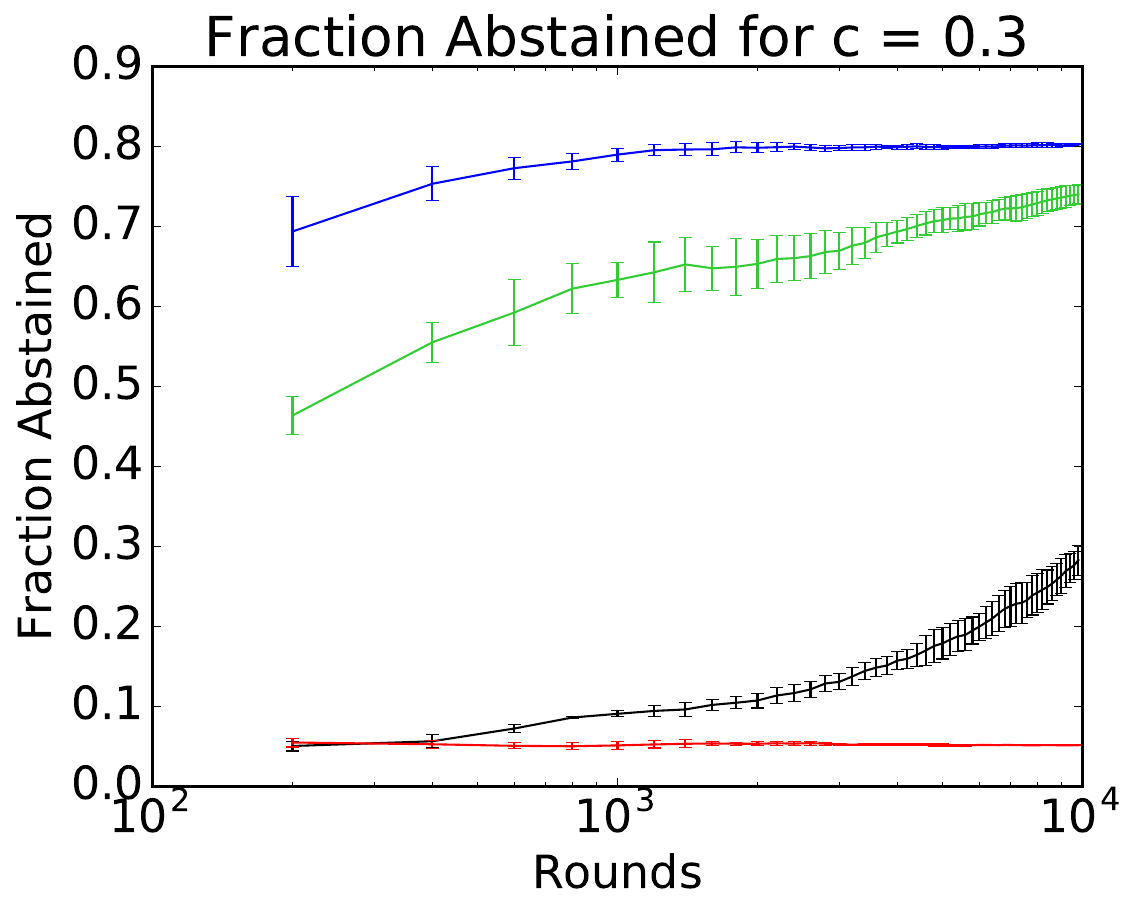} \\
\includegraphics[scale=0.25,trim= 5 10 10 5, clip=true]{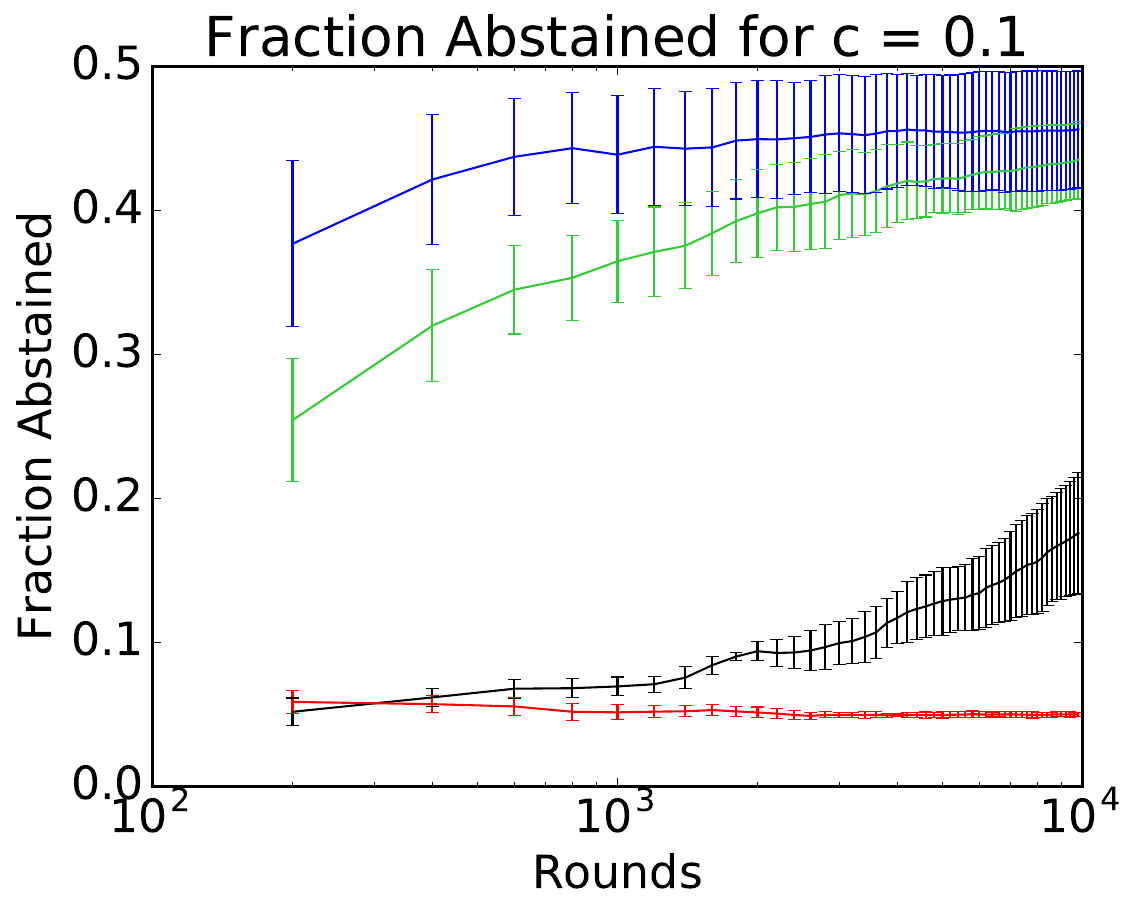} &
\hspace*{-5mm} \includegraphics[scale=0.25,trim= 5 10 10 5, clip=true]{exp_results3/counts_c02_d_8.pdf} &
\hspace*{-5mm}\includegraphics[scale=0.25,trim= 5 10 10 5, clip=true]{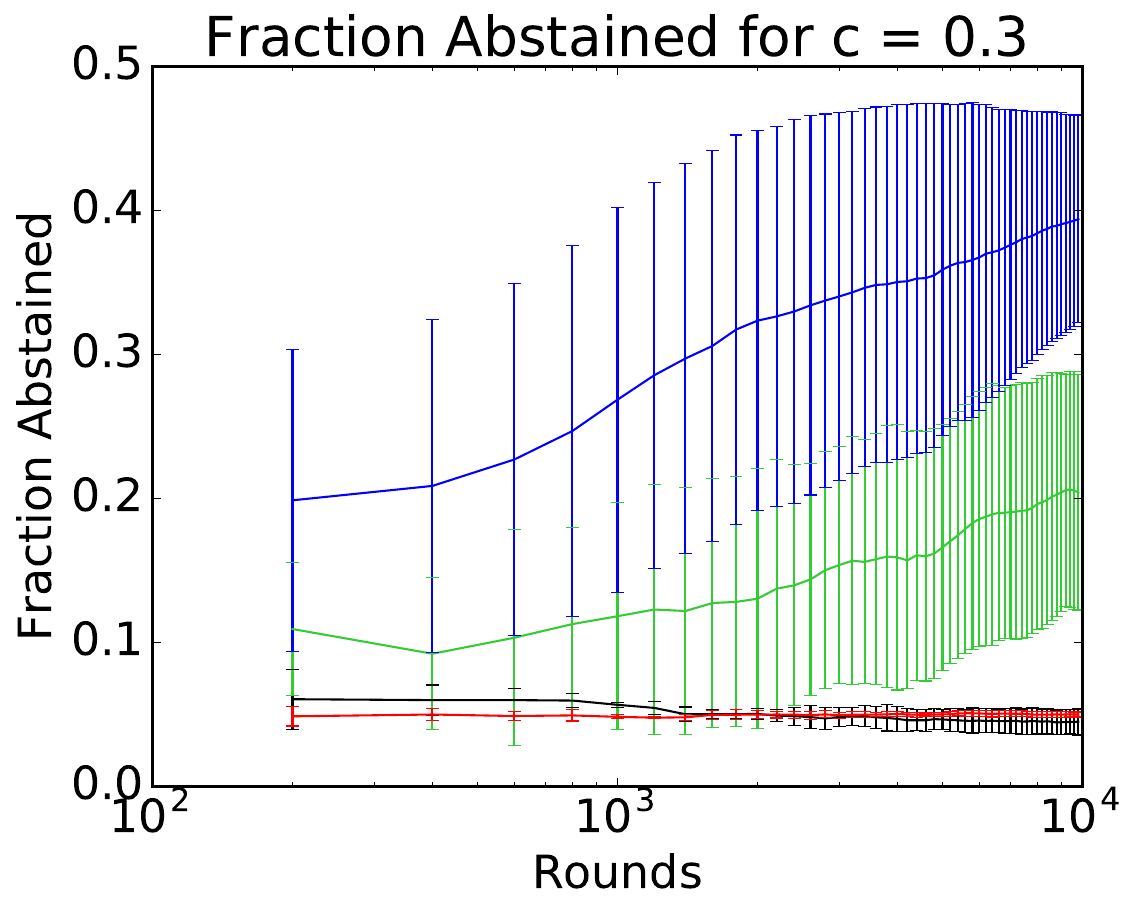} \\
\includegraphics[scale=0.25,trim= 5 10 10 5, clip=true]{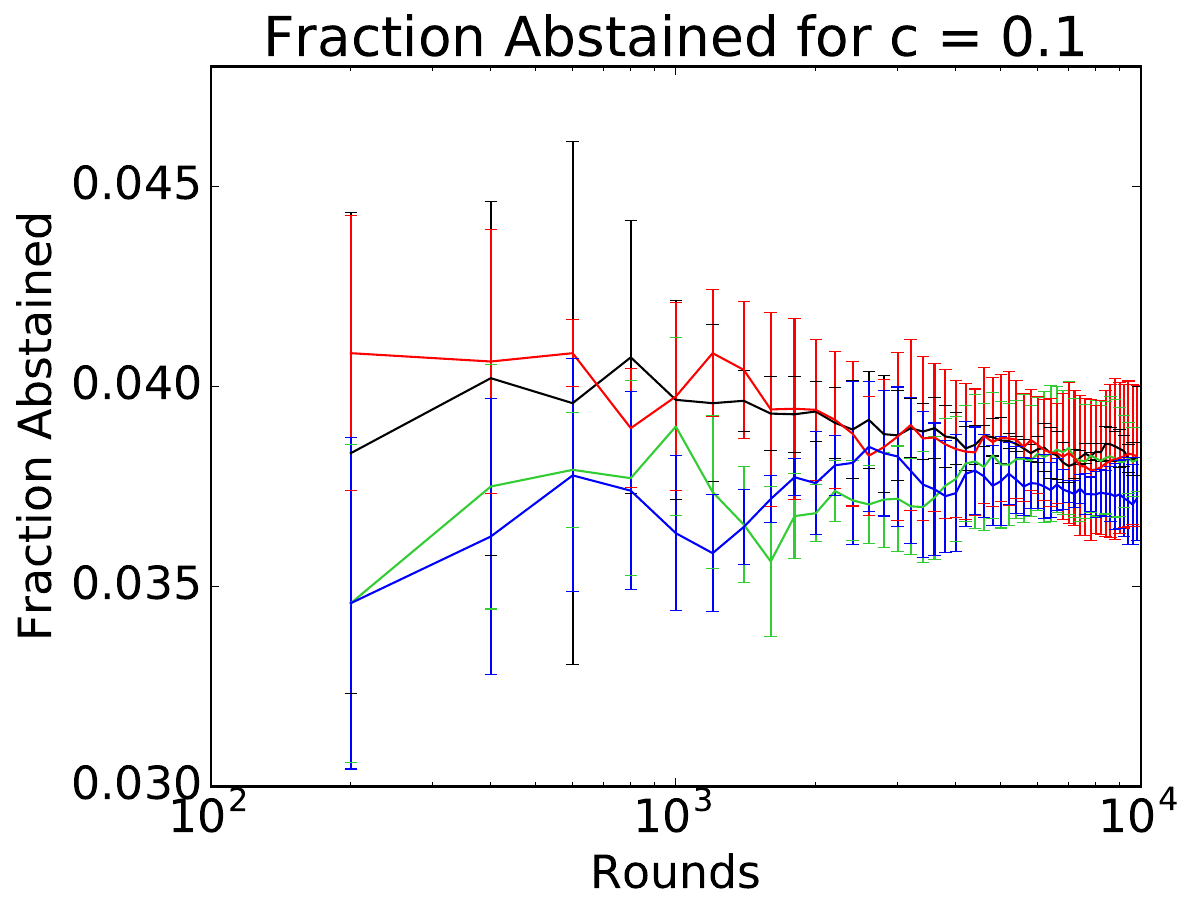} &
\hspace*{-5mm} \includegraphics[scale=0.25,trim= 5 10 10 5, clip=true]{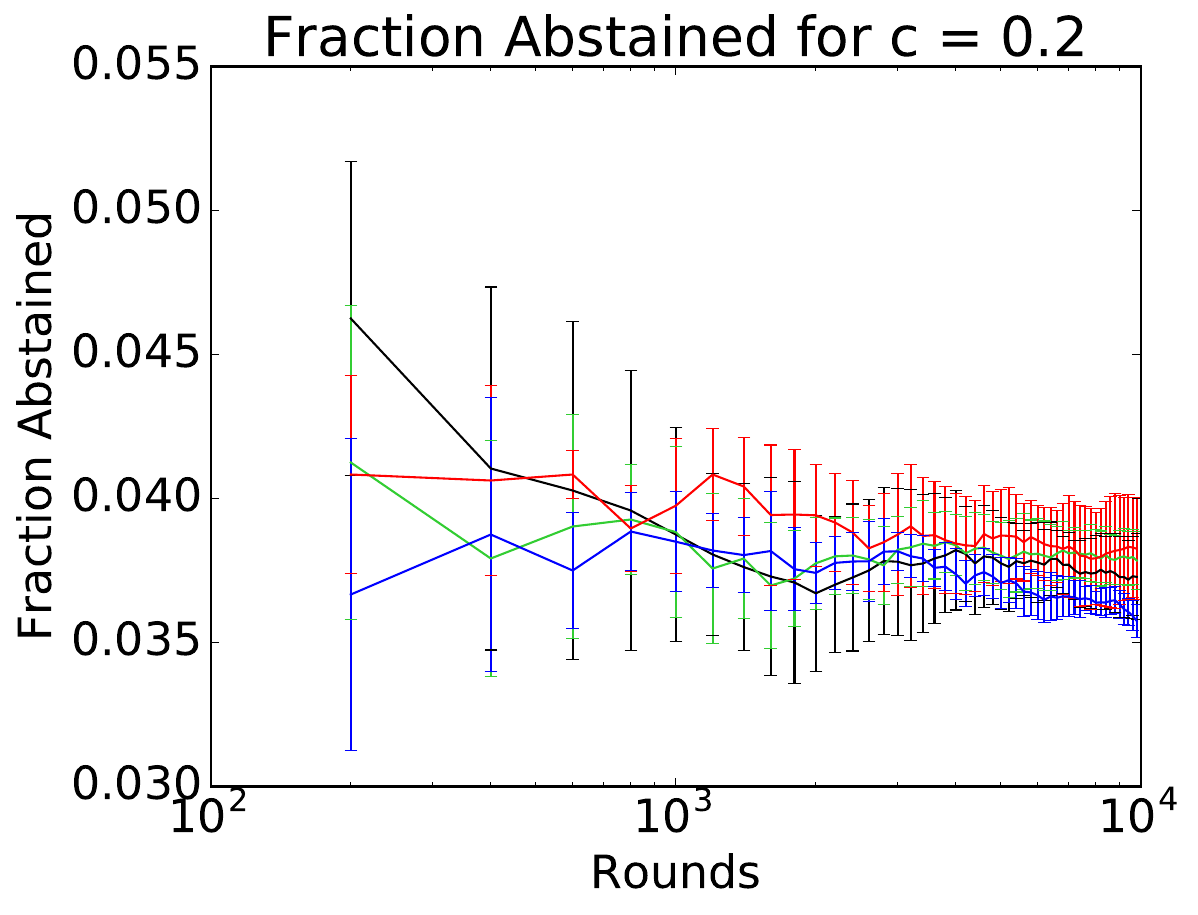} &
\hspace*{-5mm}\includegraphics[scale=0.25,trim= 5 10 10 5, clip=true]{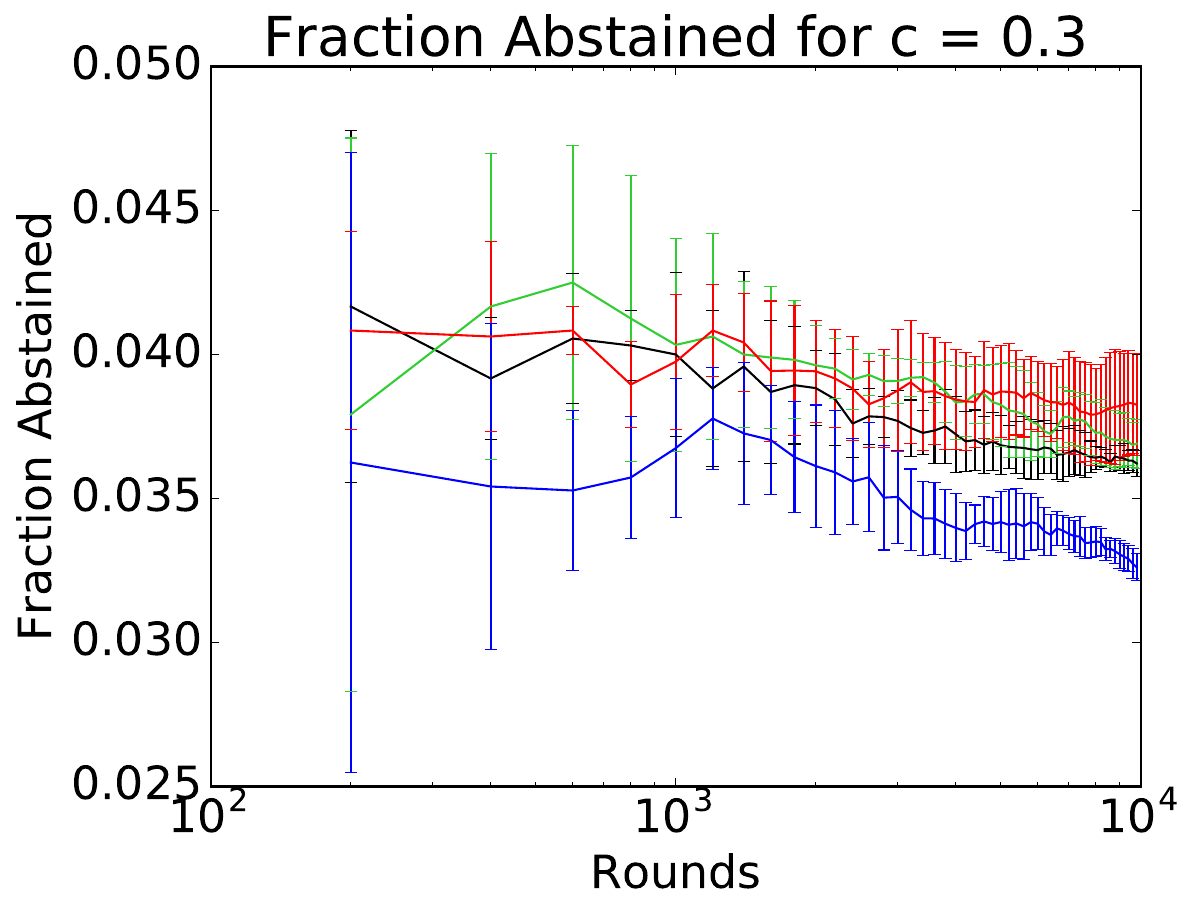} \\
\includegraphics[scale=0.25,trim= 5 10 10 5, clip=true]{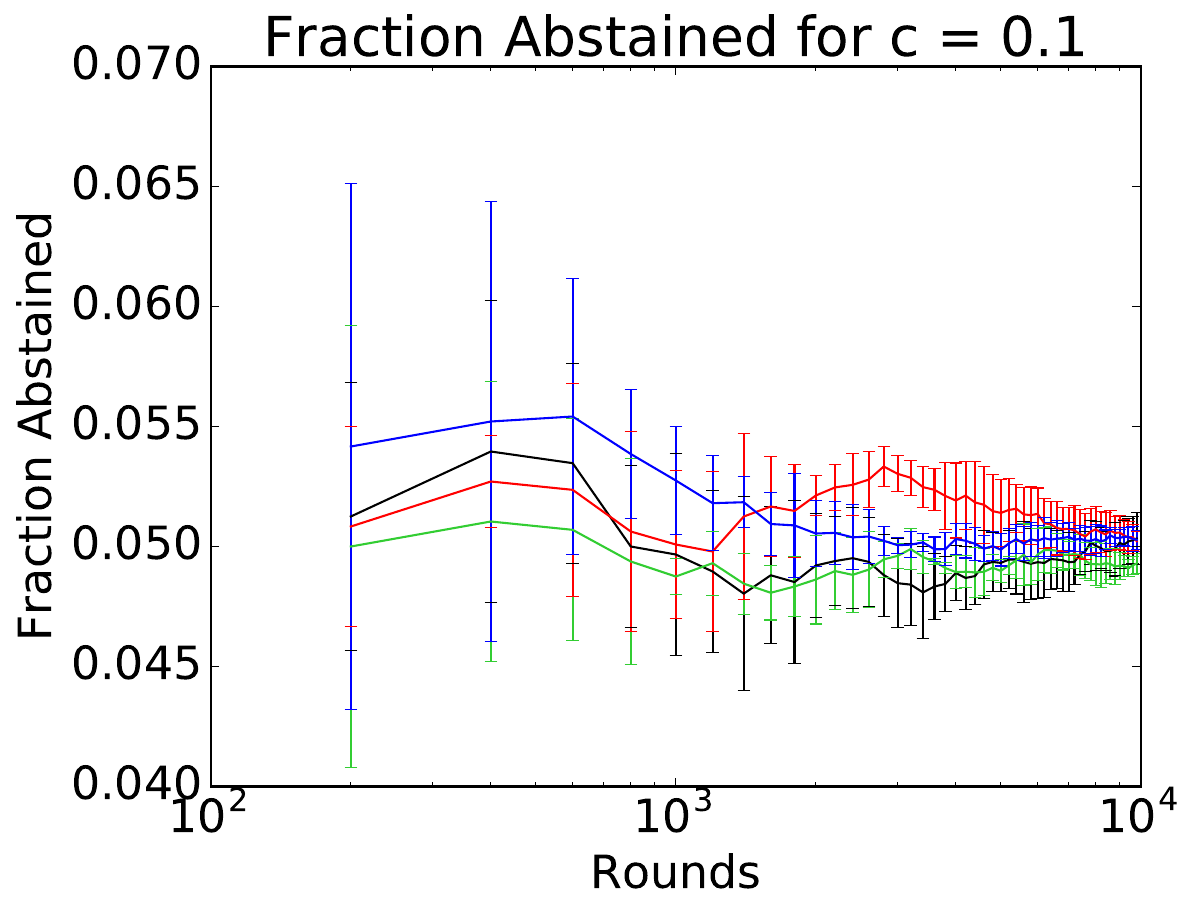} &
\hspace*{-5mm} \includegraphics[scale=0.25,trim= 5 10 10 5, clip=true]{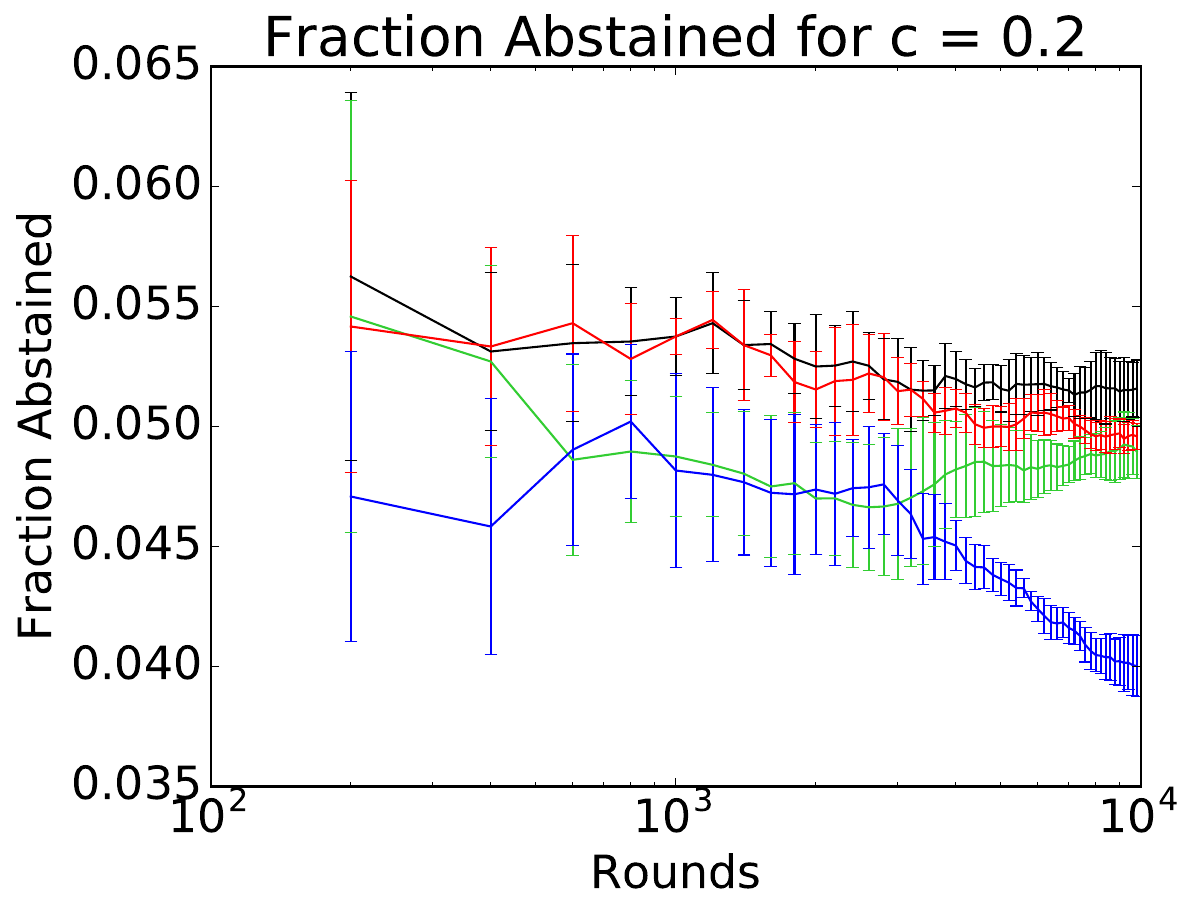} &
\hspace*{-5mm}\includegraphics[scale=0.25,trim= 5 10 10 5, clip=true]{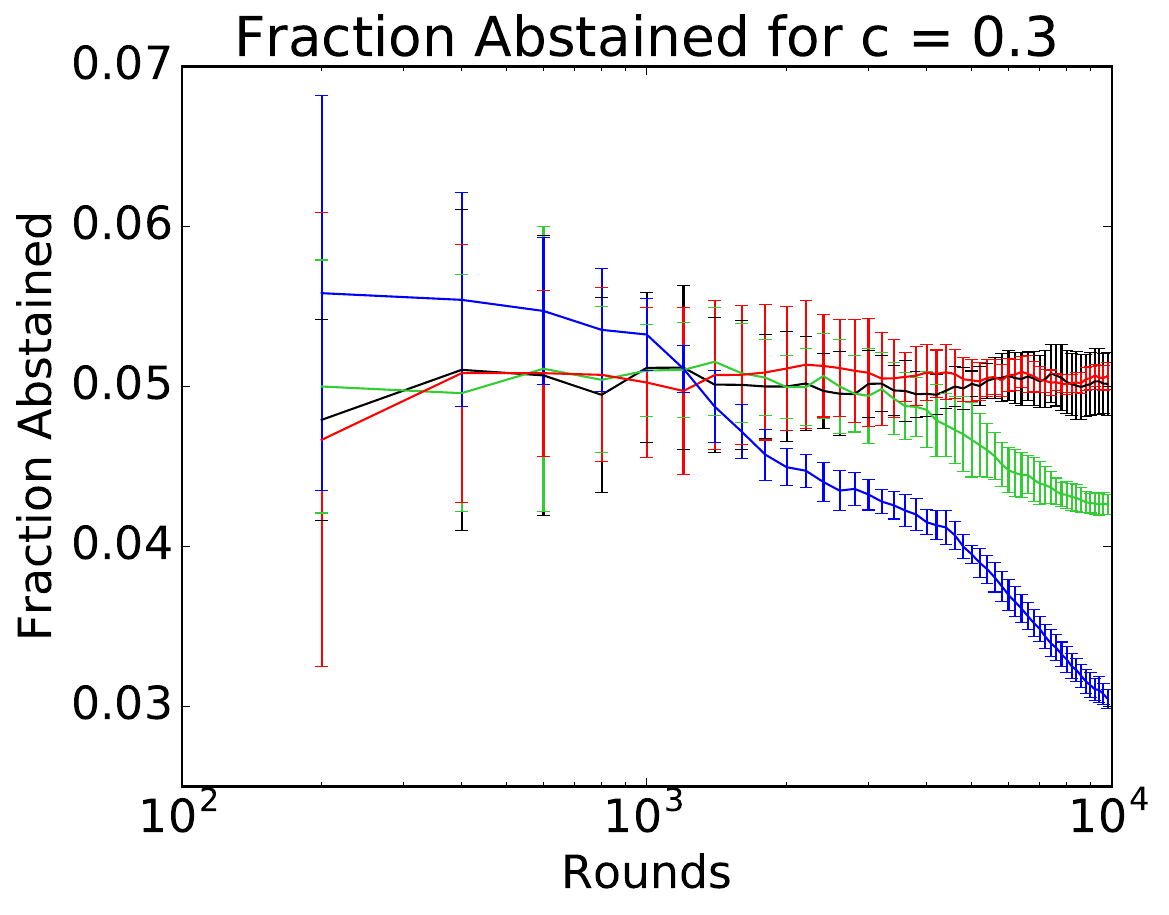} \\
\includegraphics[scale=0.25,trim= 5 10 10 5, clip=true]{exp_results3/counts_c01_d_4.pdf} &
\hspace*{-5mm} \includegraphics[scale=0.25,trim= 5 10 10 5, clip=true]{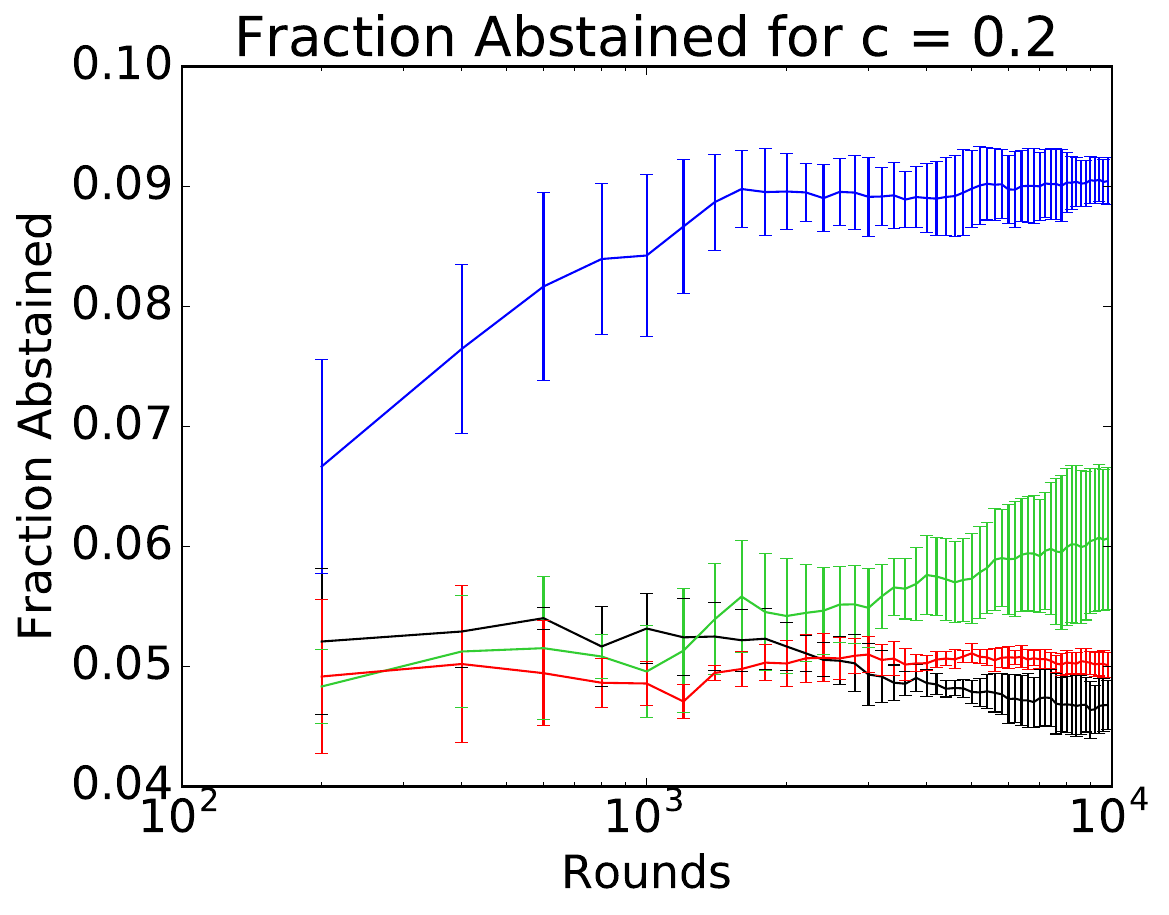}&
\hspace*{-5mm}\includegraphics[scale=0.25,trim= 5 10 10 5, clip=true]{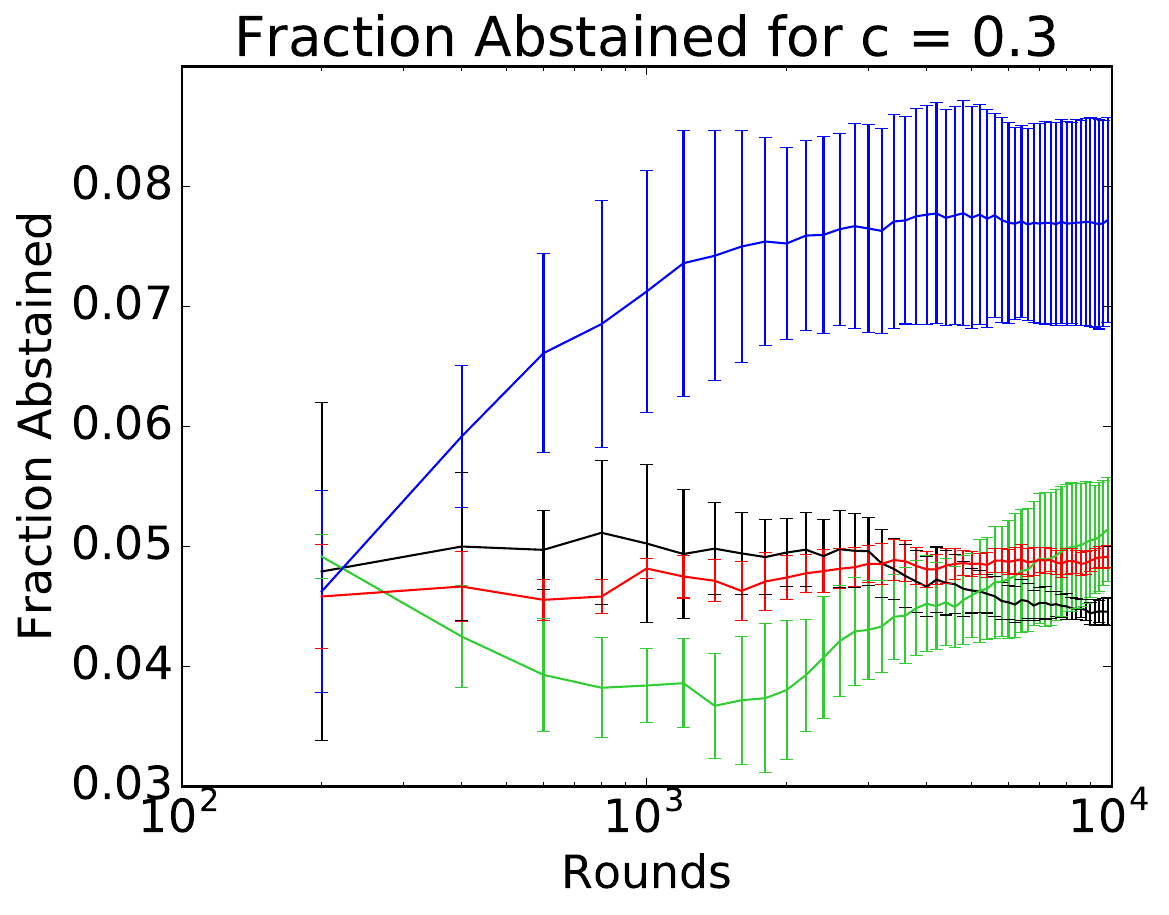} \\
\end{tabular}
\end{center}
\vskip -.15in
\caption{A graph of the averaged fraction of abstained points with
  standard deviations as a function of $t$ (log scale) for
  {\color[rgb]{0.16,0.67,0.16}\UCBGT }, \UCBNT , {\color{red} \UCB },
  and {\color{blue} \FTL } for different values of abstention costs.
  Each row is a dataset, starting from the top row we have: {\tt eye},
  {\tt cod-ran}, {\tt synthetic}, {\tt skin}, and {\tt guide}.  }
\label{fig:frac2}
\vskip -.1in
\end{figure*}

\begin{figure*}[!ht]
\begin{center}
\begin{tabular}{ c c c c }
\includegraphics[scale=0.2,trim= 5 10 10 5, clip=true]{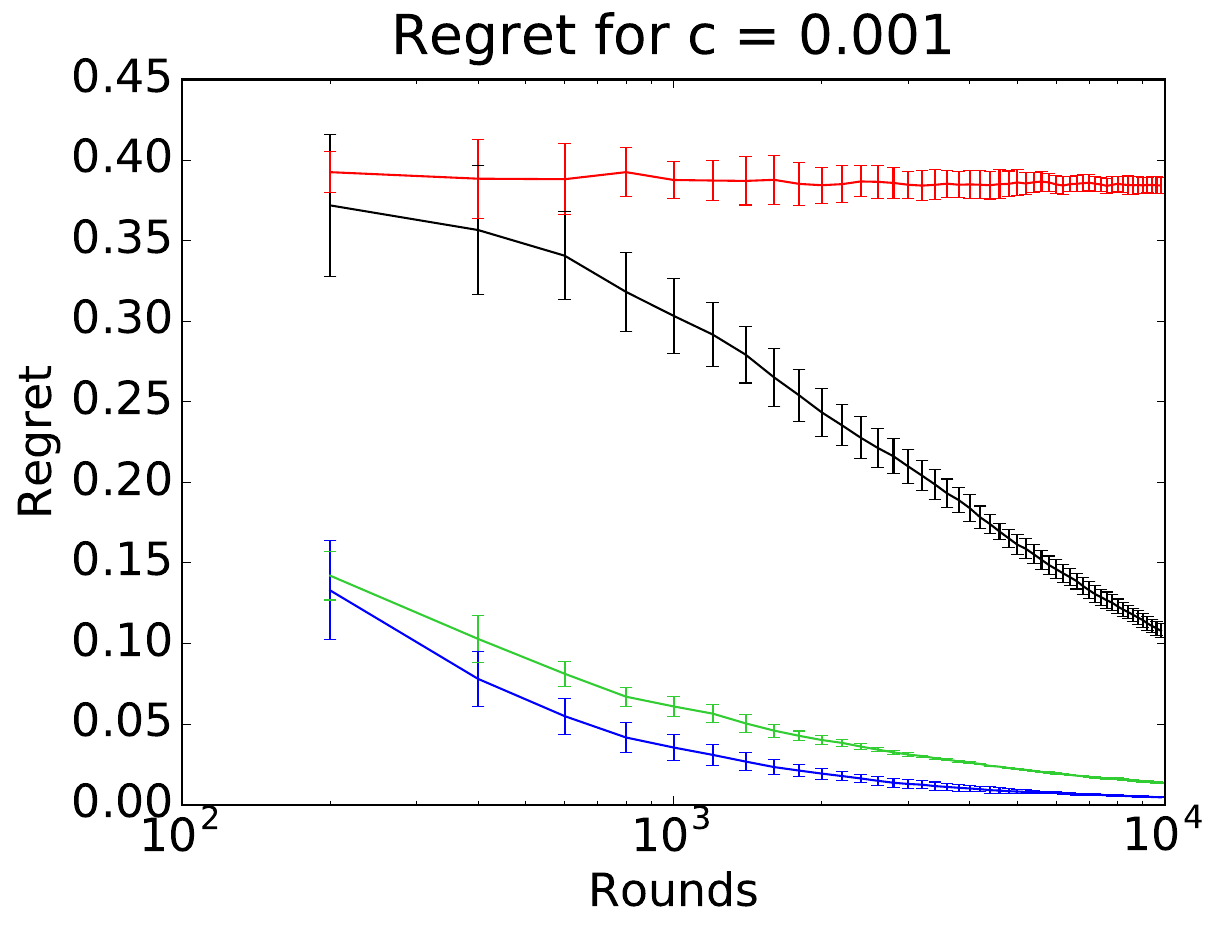} &
\hspace*{-5mm} \includegraphics[scale=0.2,trim= 5 10 10 5, clip=true]{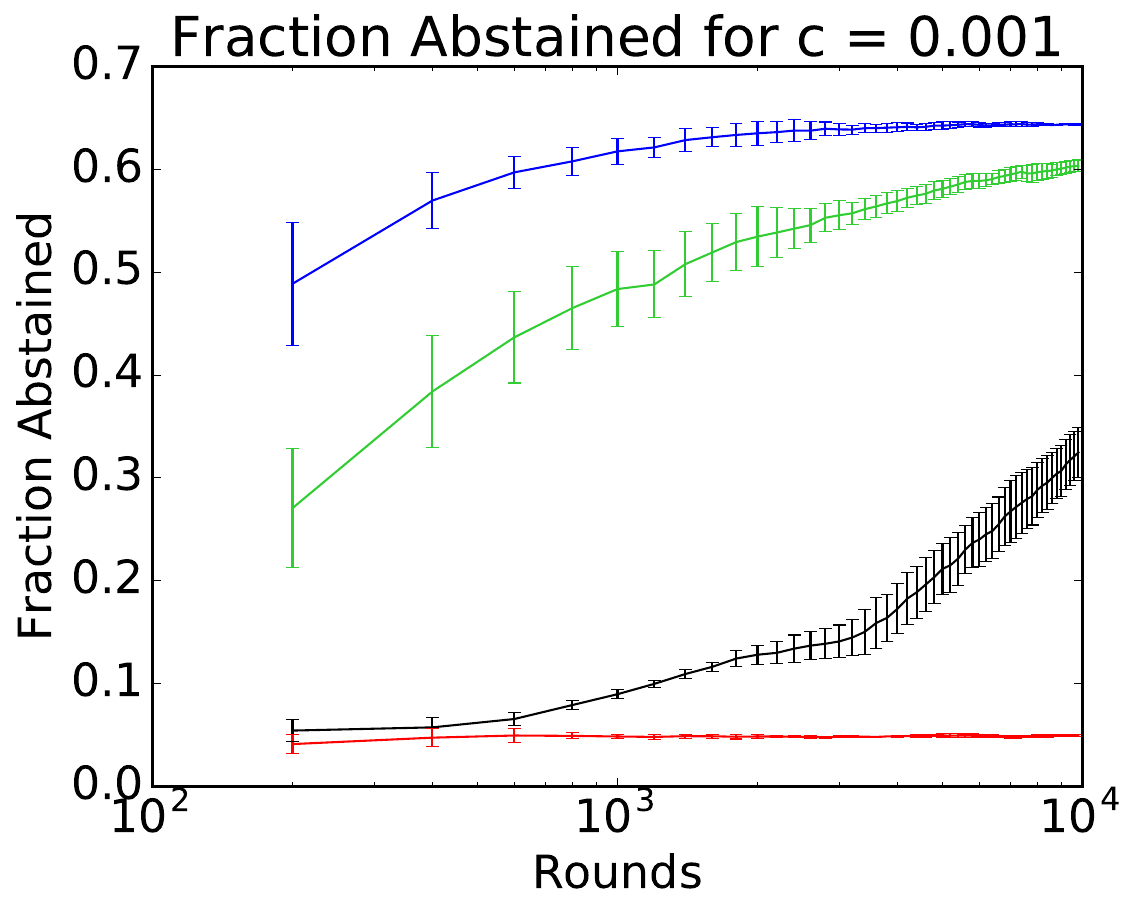} &
\hspace*{-5mm}\includegraphics[scale=0.2,trim= 5 10 10 5, clip=true]{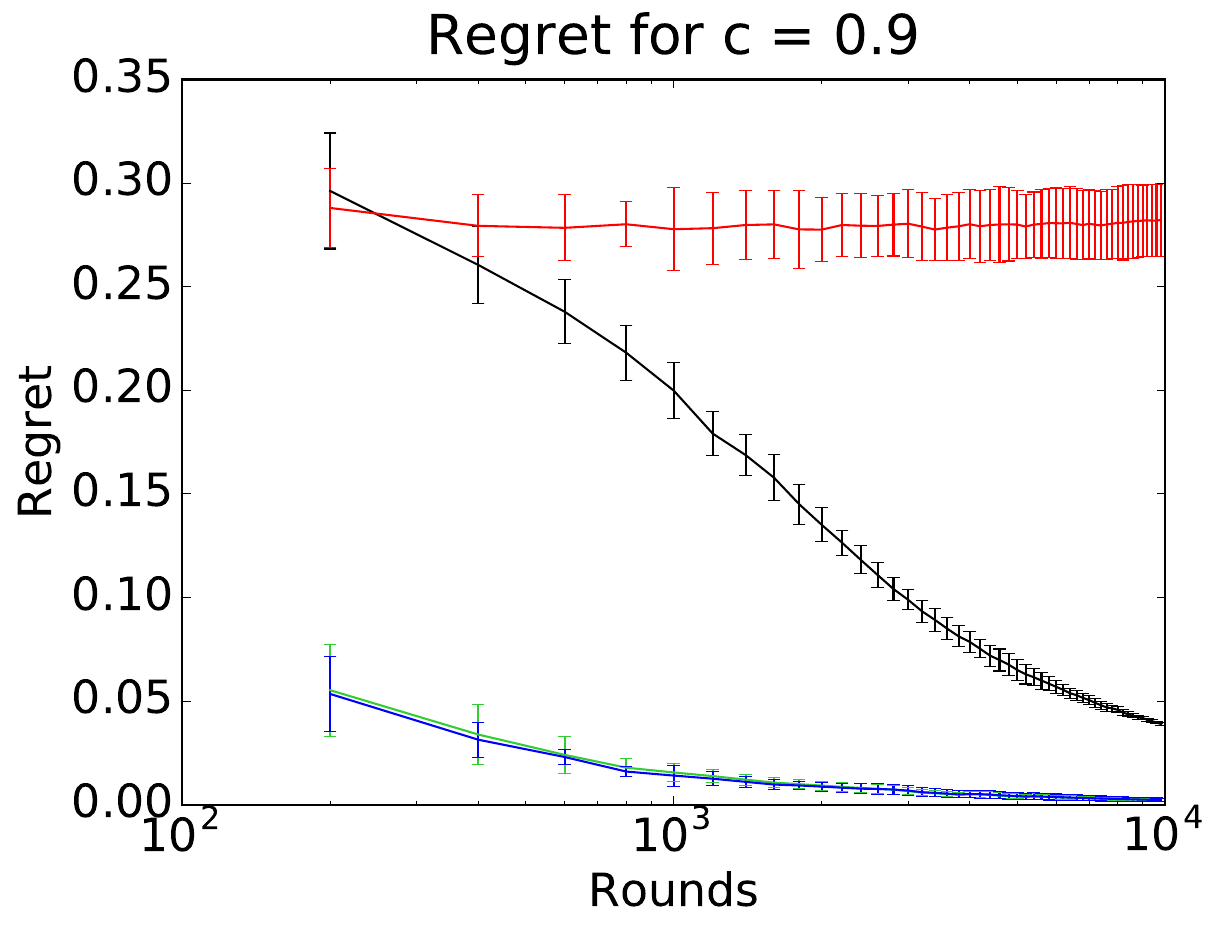} &
\hspace*{-5mm}\includegraphics[scale=0.2,trim= 5 10 10 5, clip=true]{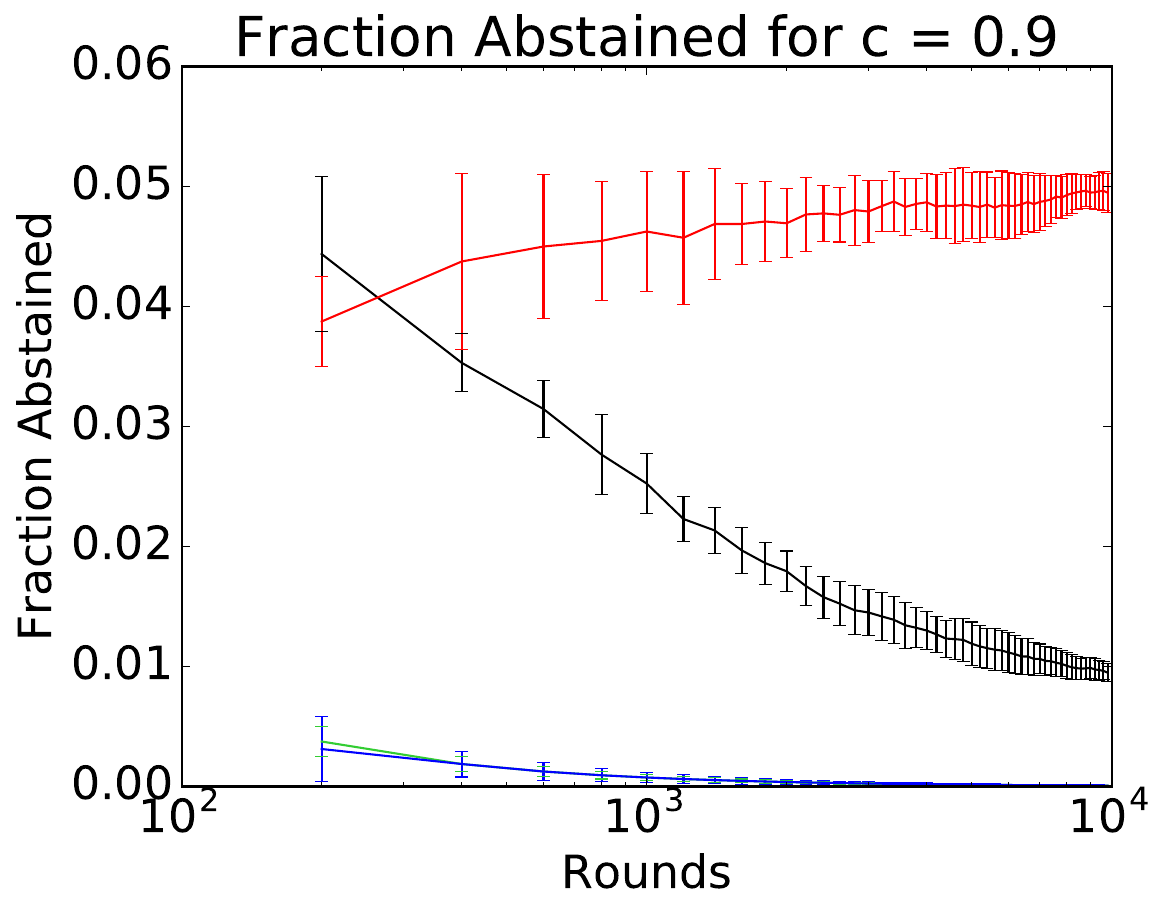} \\
\includegraphics[scale=0.2,trim= 5 10 10 5, clip=true]{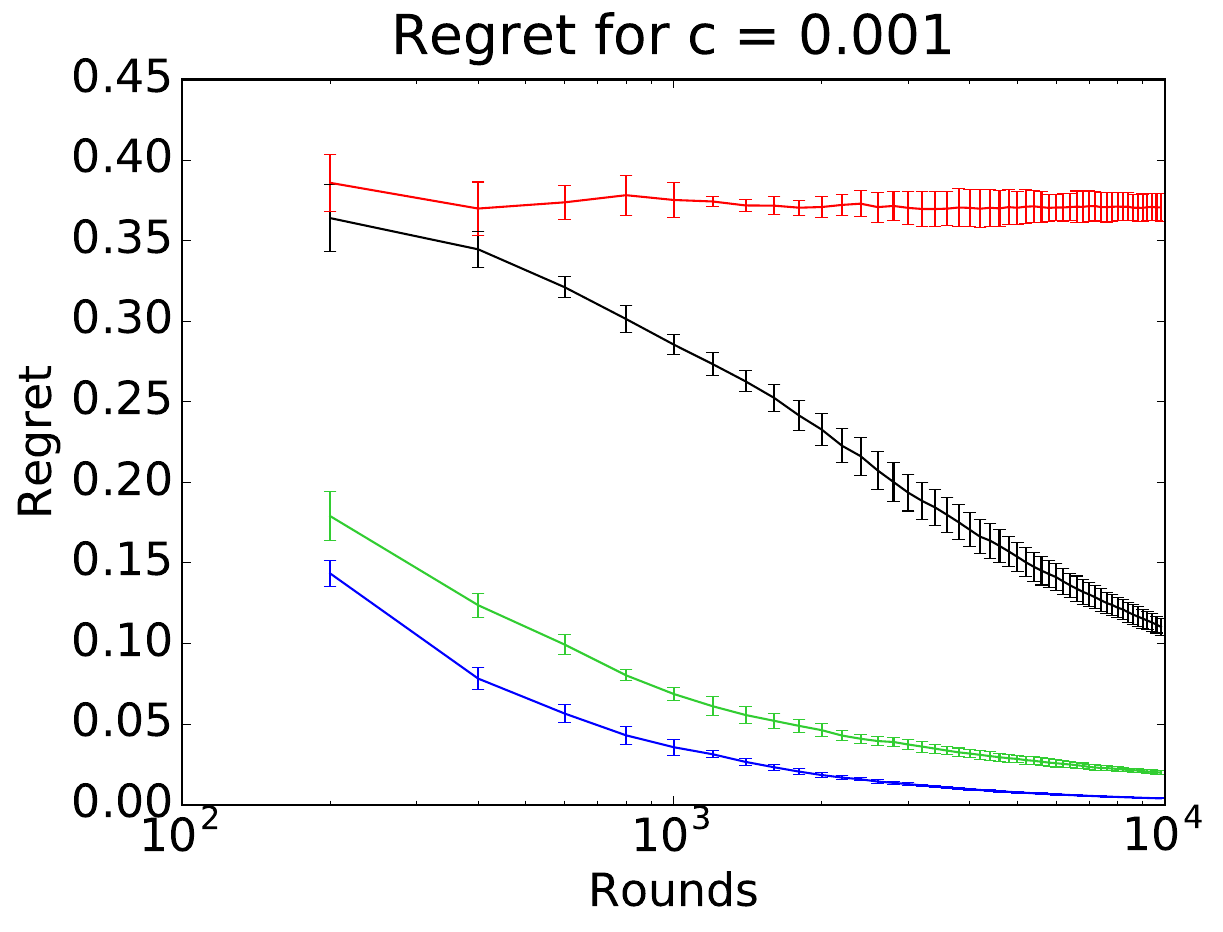} &
\hspace*{-5mm} \includegraphics[scale=0.2,trim= 5 10 10 5, clip=true]{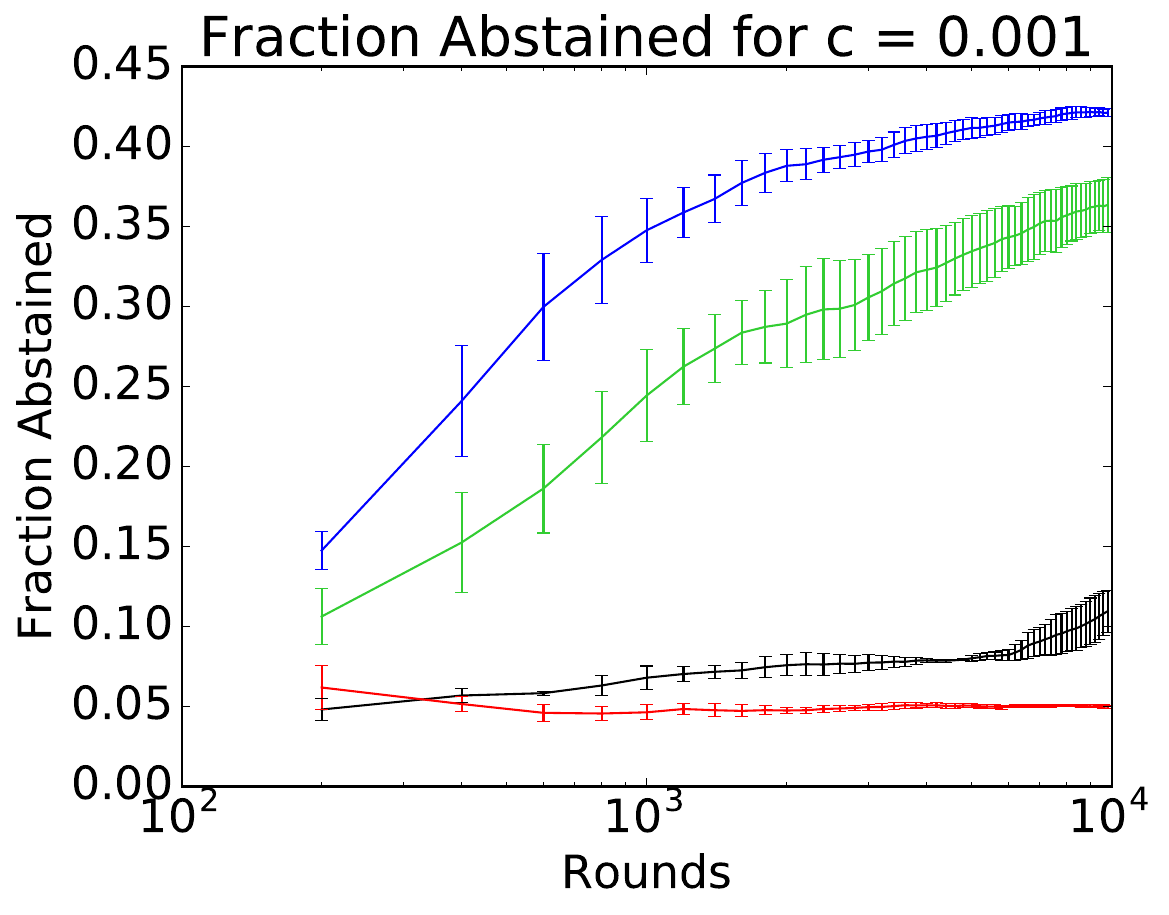} &
\hspace*{-5mm}\includegraphics[scale=0.2,trim= 5 10 10 5, clip=true]{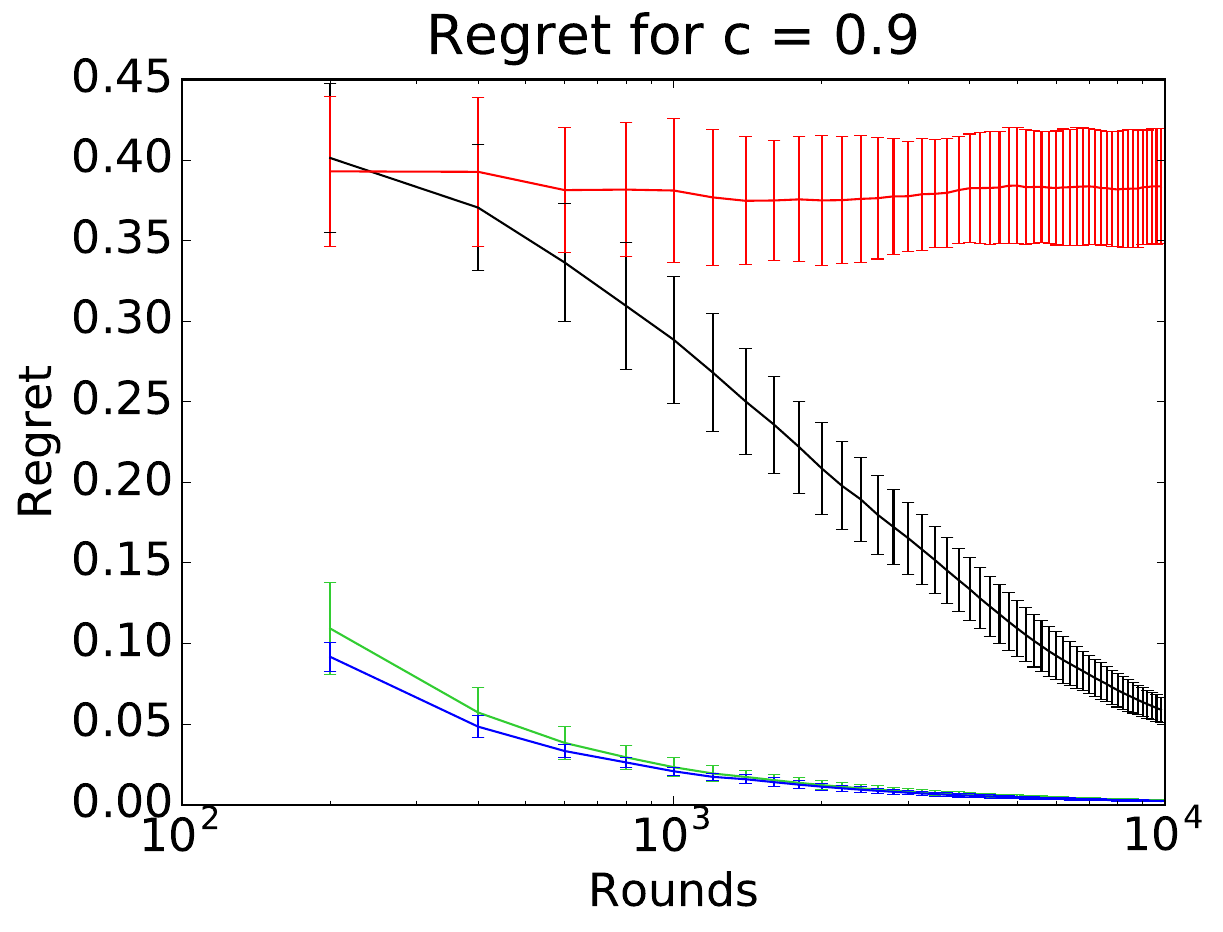} &
\hspace*{-5mm}\includegraphics[scale=0.2,trim= 5 10 10 5, clip=true]{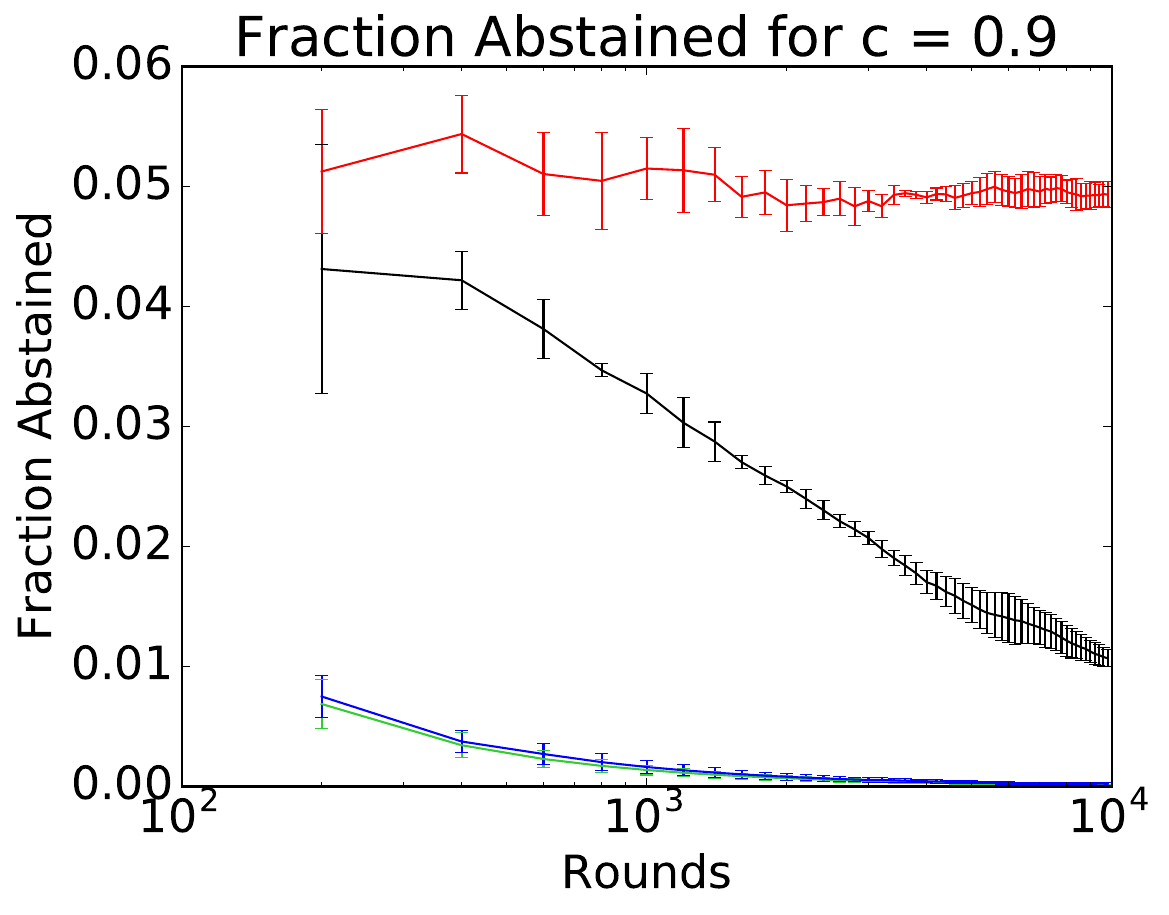} \\
\includegraphics[scale=0.2,trim= 5 10 10 5, clip=true]{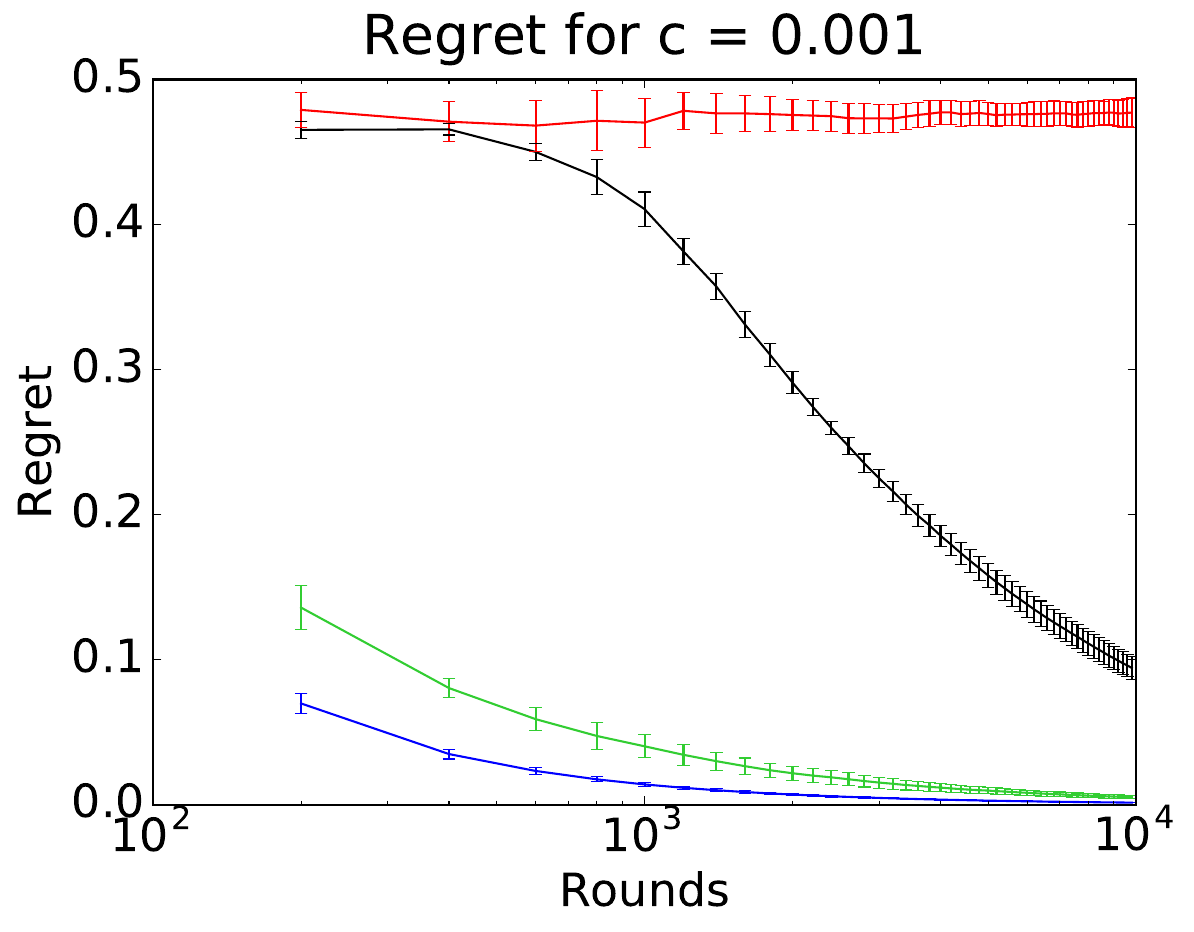} &
\hspace*{-5mm} \includegraphics[scale=0.2,trim= 5 10 10 5, clip=true]{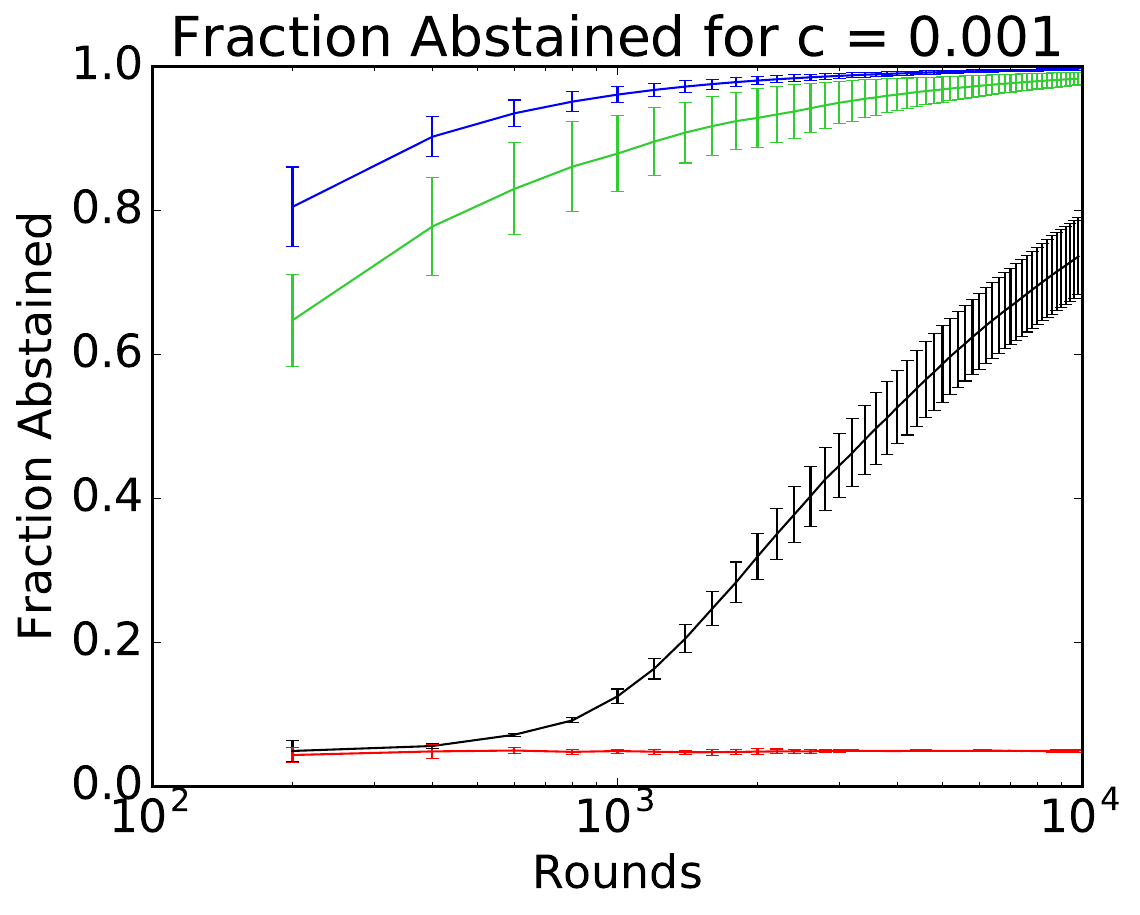} &
\hspace*{-5mm}\includegraphics[scale=0.2,trim= 5 10 10 5, clip=true]{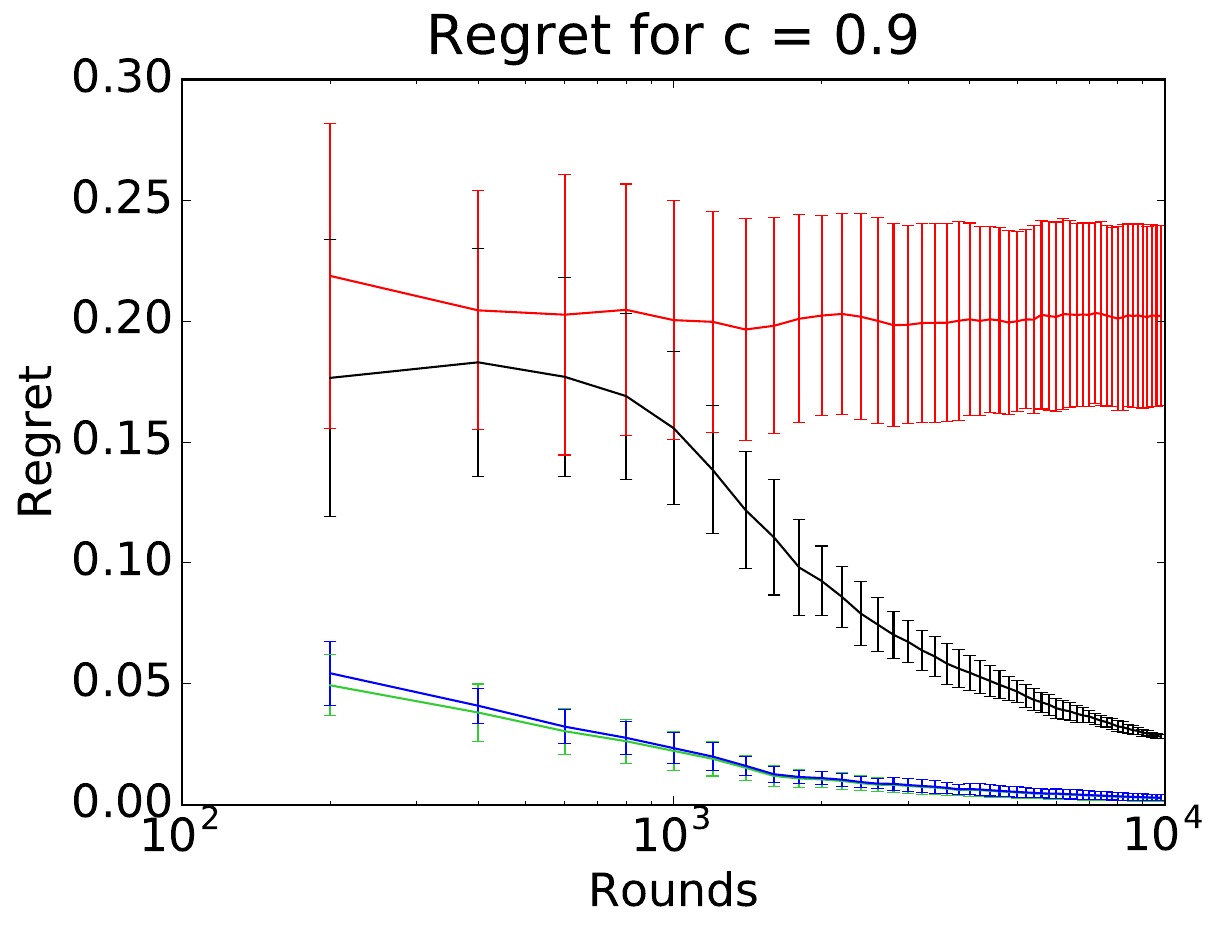} &
\hspace*{-5mm}\includegraphics[scale=0.2,trim= 5 10 10 5, clip=true]{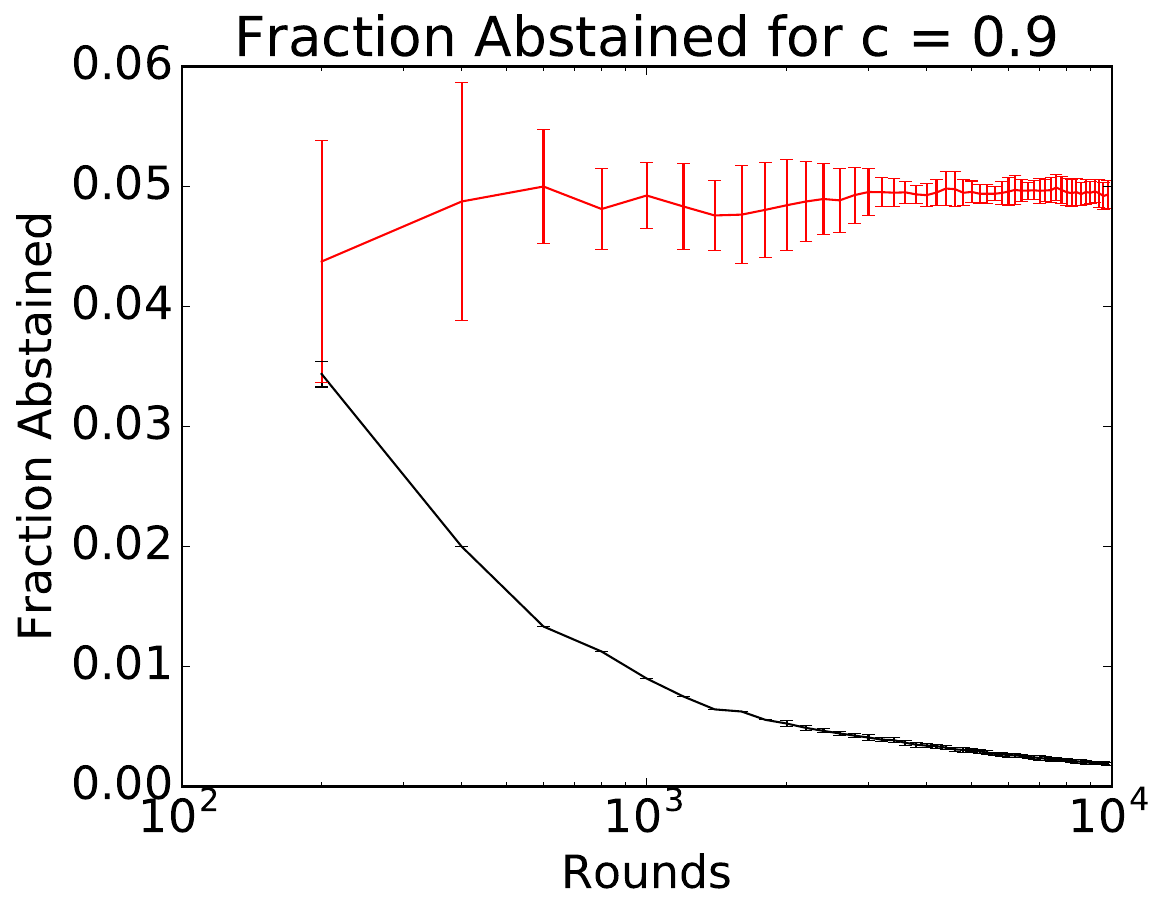} \\
\includegraphics[scale=0.2,trim= 5 10 10 5, clip=true]{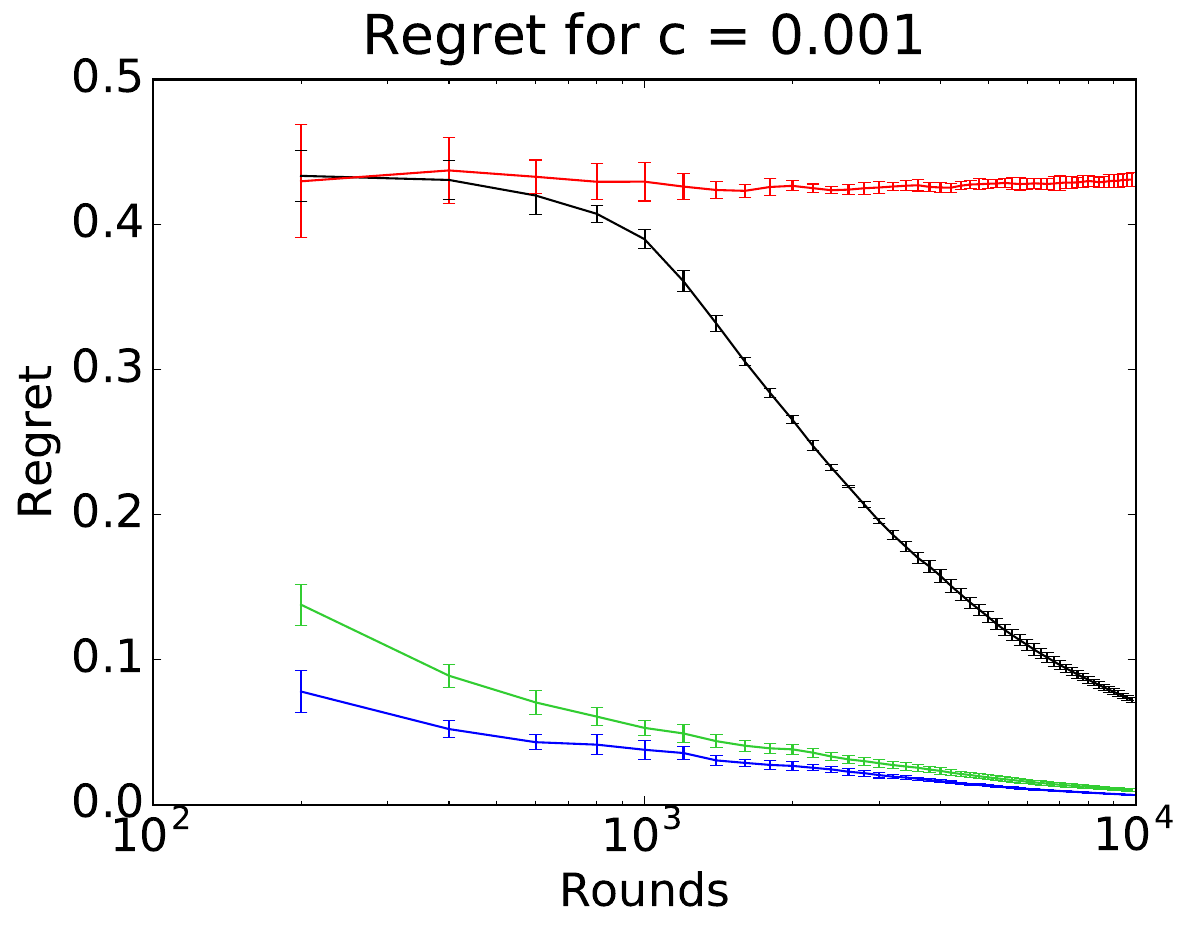} &
\hspace*{-5mm} \includegraphics[scale=0.2,trim= 5 10 10 5, clip=true]{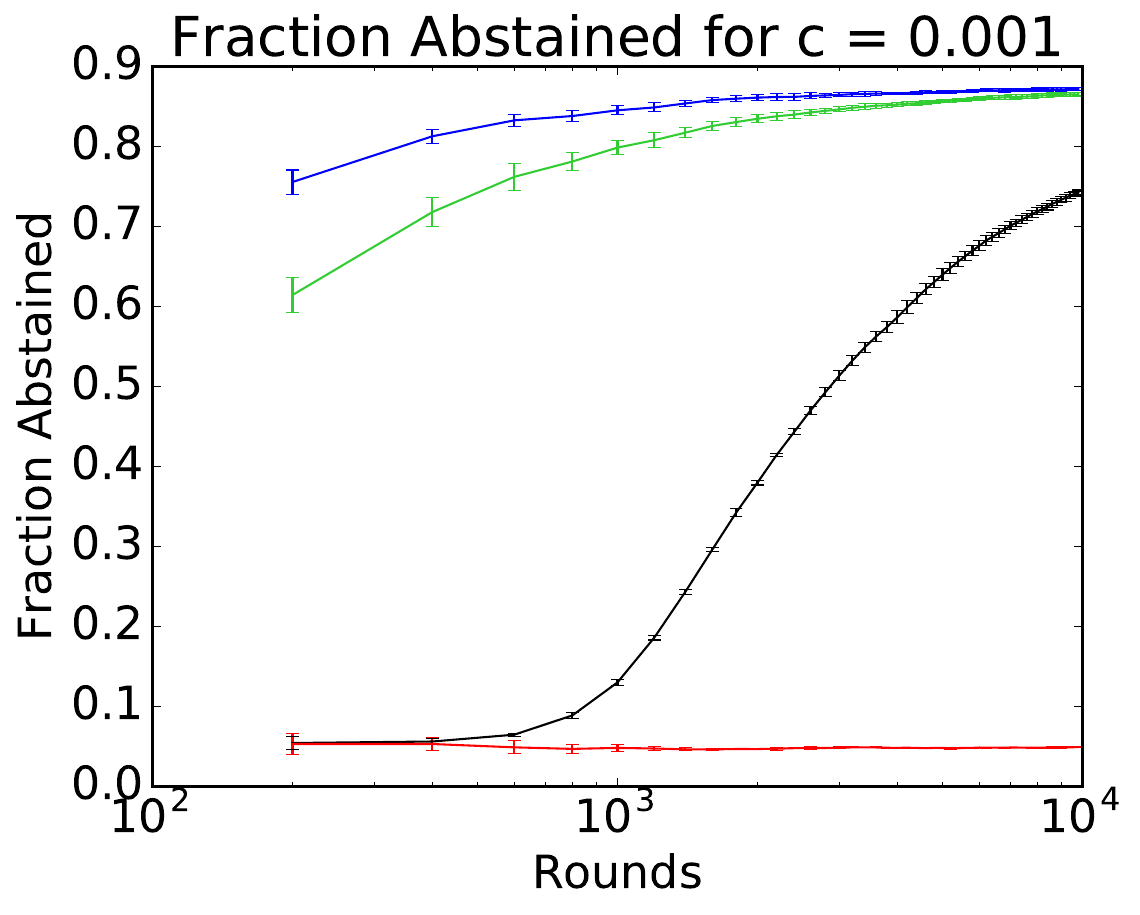} &
\hspace*{-5mm}\includegraphics[scale=0.2,trim= 5 10 10 5, clip=true]{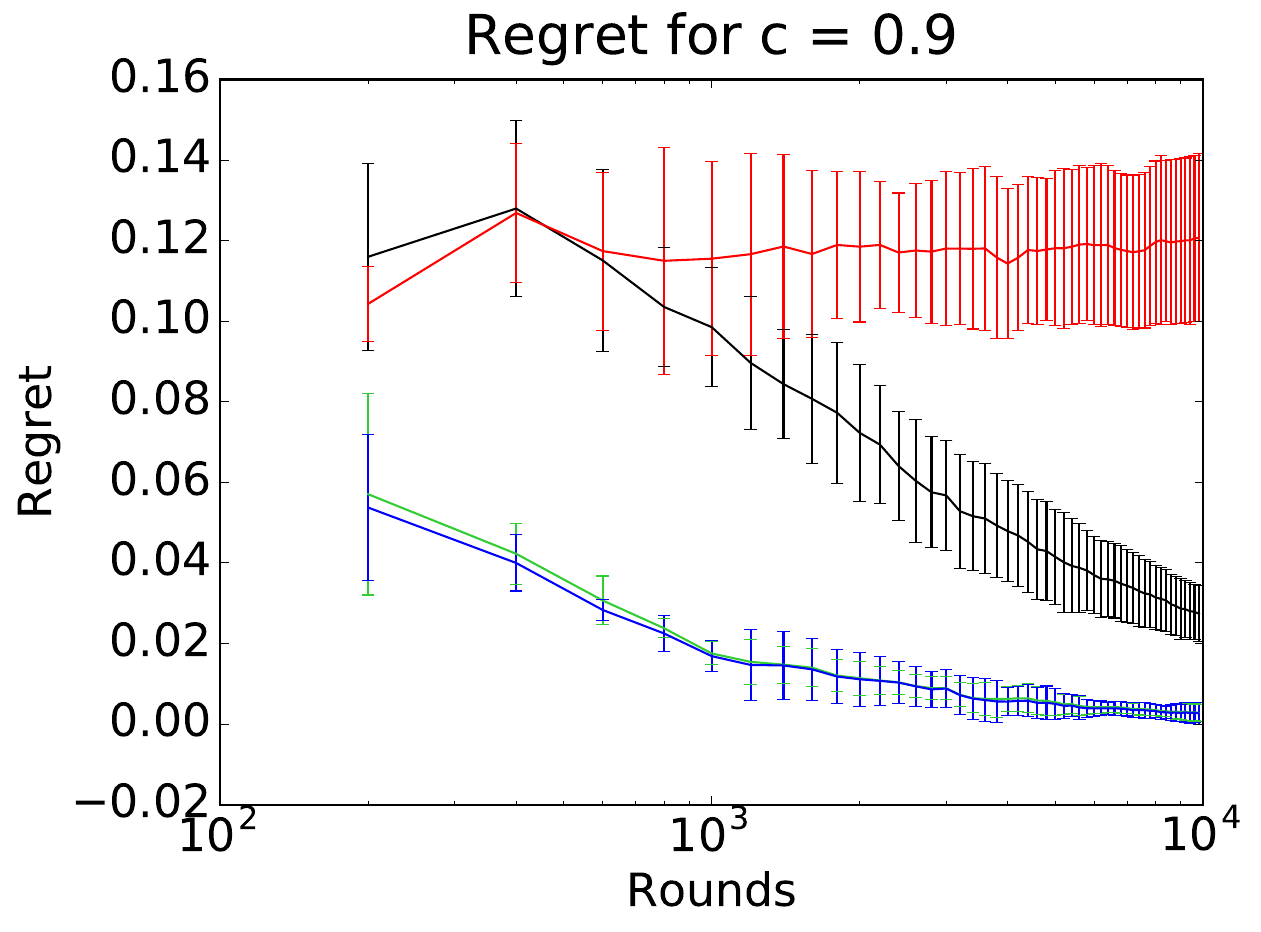} &
\hspace*{-5mm}\includegraphics[scale=0.2,trim= 5 10 10 5, clip=true]{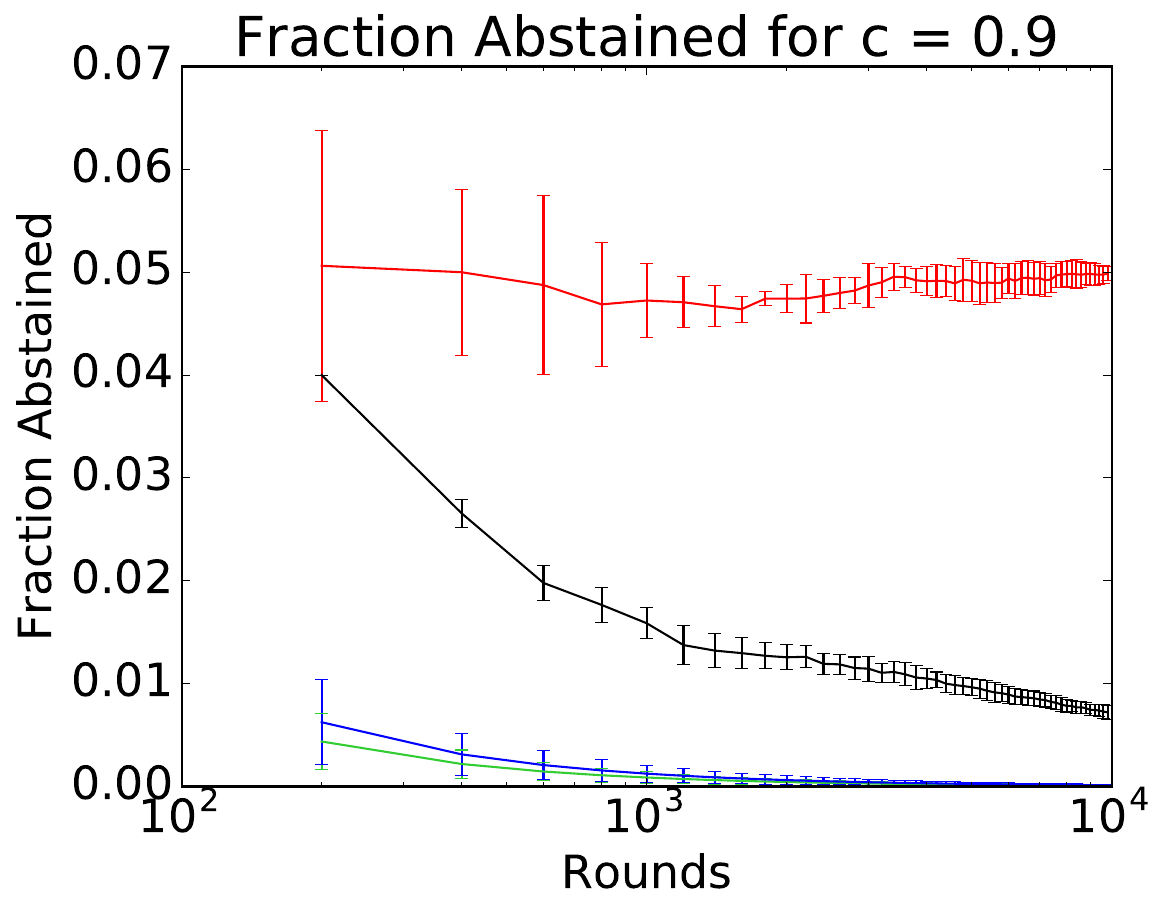} \\
\end{tabular}
\end{center}
\vskip -.15in
    \caption{
A graph of the averaged regret $R_t(\cdot)/t$ and fraction of points rejected with standard deviations as a function of $t$  (log scale) for
 {\color[rgb]{0.16,0.67,0.16}\UCBGT }, \UCBNT , {\color{red} \UCB }, and {\color{blue} \FTL }  for different values of abstention costs. The fraction of points decreases as the cost of abstention increases. The \UCBGT\ outperforms \UCBNT\  and \UCB\ while approaching the performance of \FTL\ even at these extreme values of $c$. Each row is a dataset, starting from the top row we have:    {\tt CIFAR}, {\tt ijcnn}, {\tt phishing}, and {\tt covtype}. }
     \label{fig:extremec}
\vskip -.1in
\end{figure*}

\begin{figure*}[!ht]
\begin{center}
\begin{tabular}{ c c c }
\includegraphics[scale=0.25,trim= 5 10 10 5, clip=true]{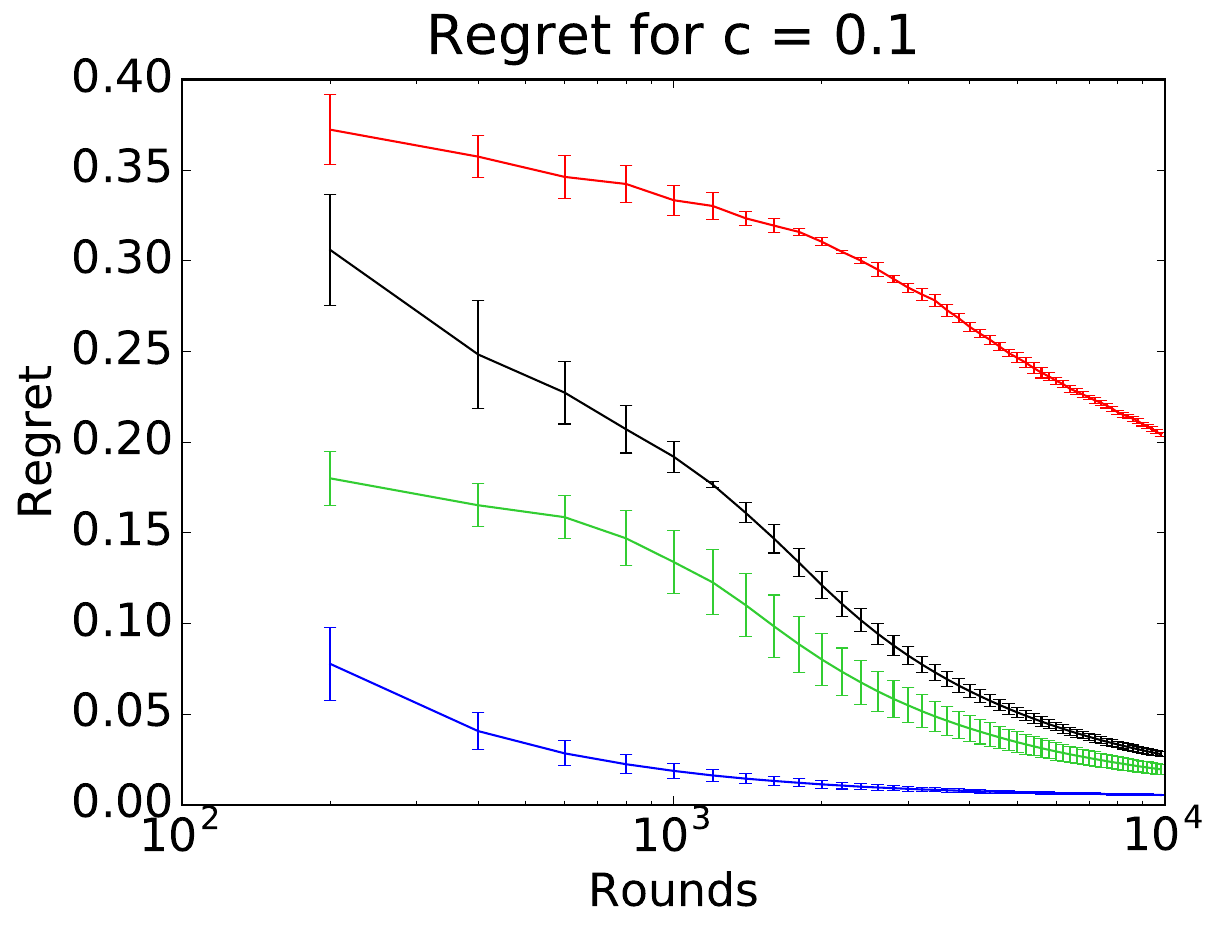} &
\hspace*{-5mm} \includegraphics[scale=0.25,trim= 5 10 10 5, clip=true]{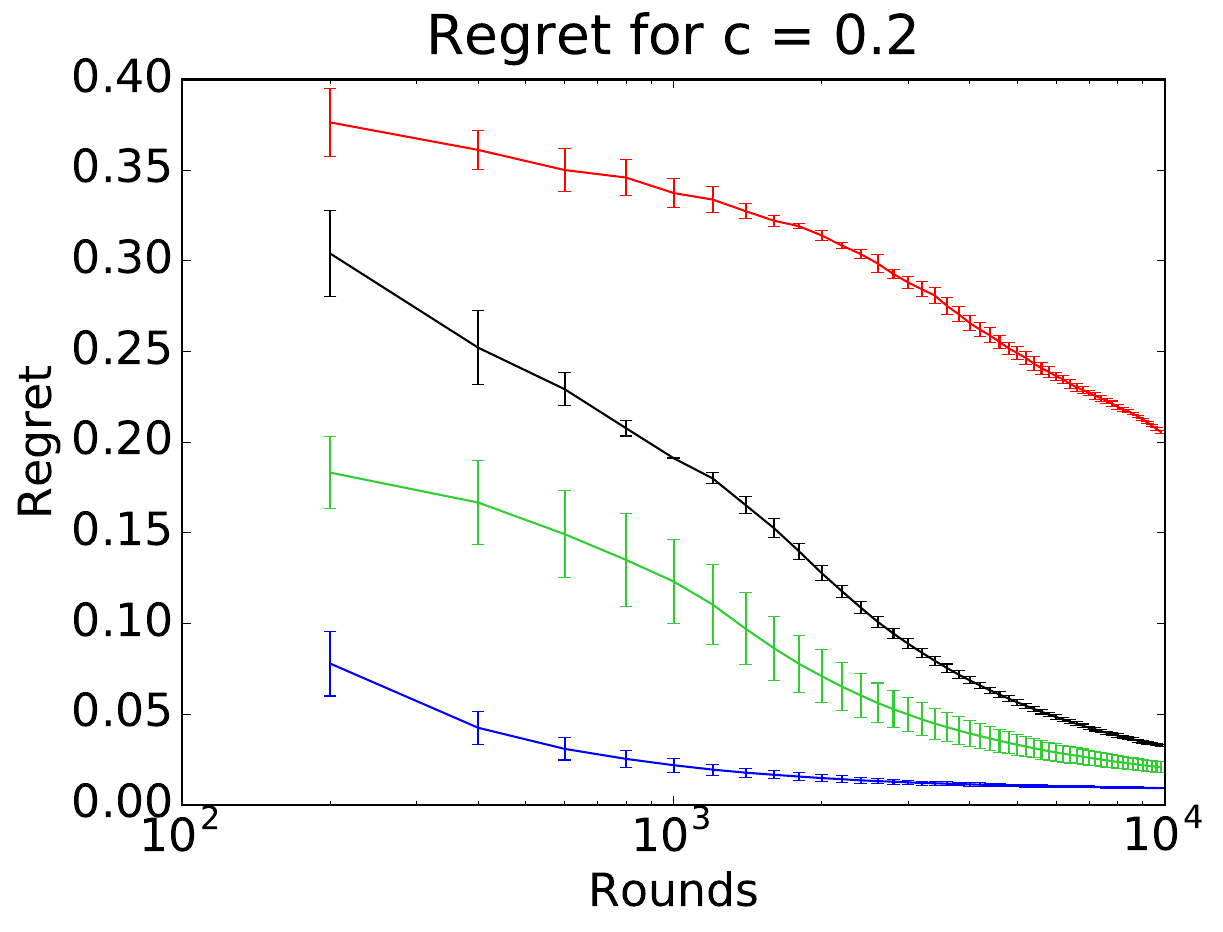} &
\hspace*{-5mm}\includegraphics[scale=0.25,trim= 5 10 10 5, clip=true]{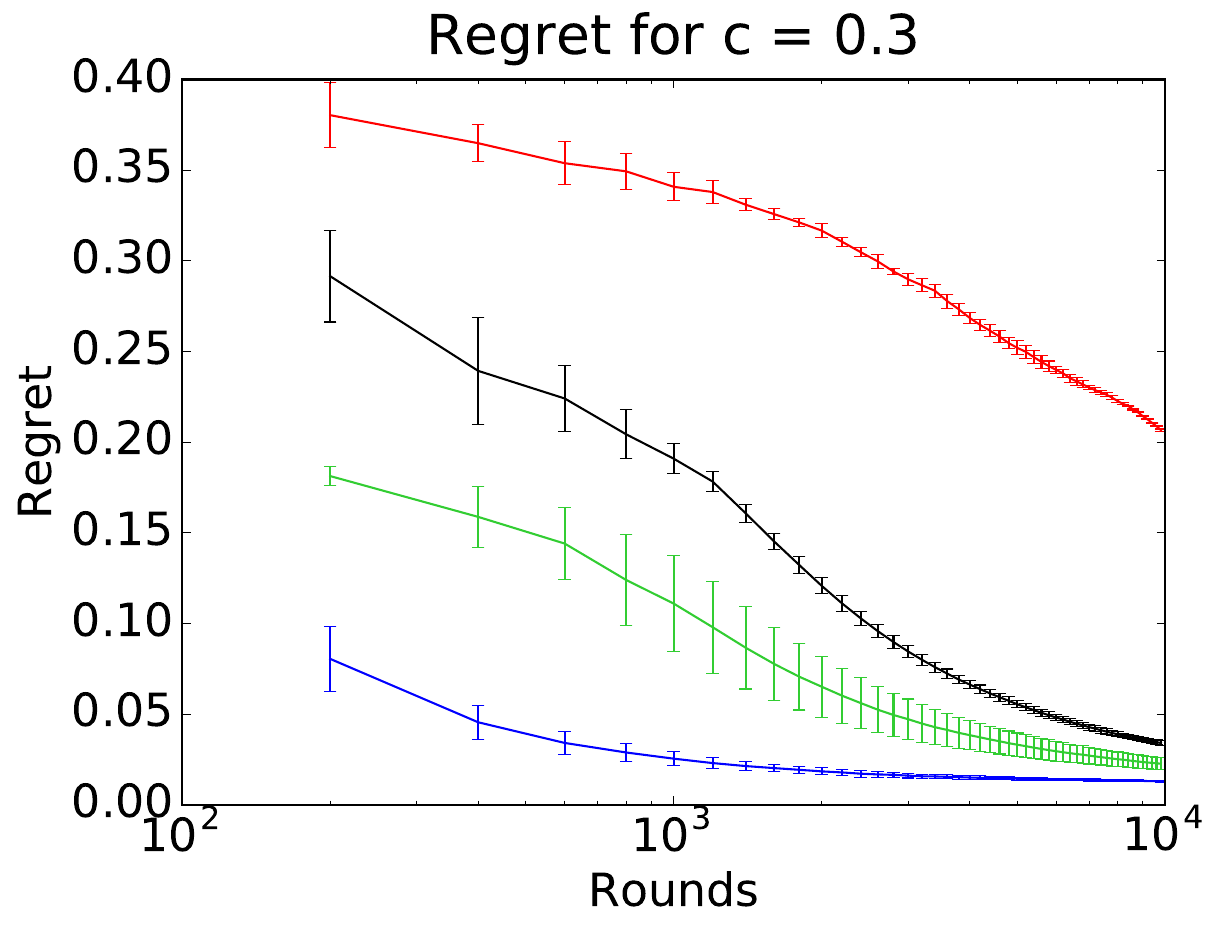} \\
\includegraphics[scale=0.25,trim= 5 10 10 5, clip=true]{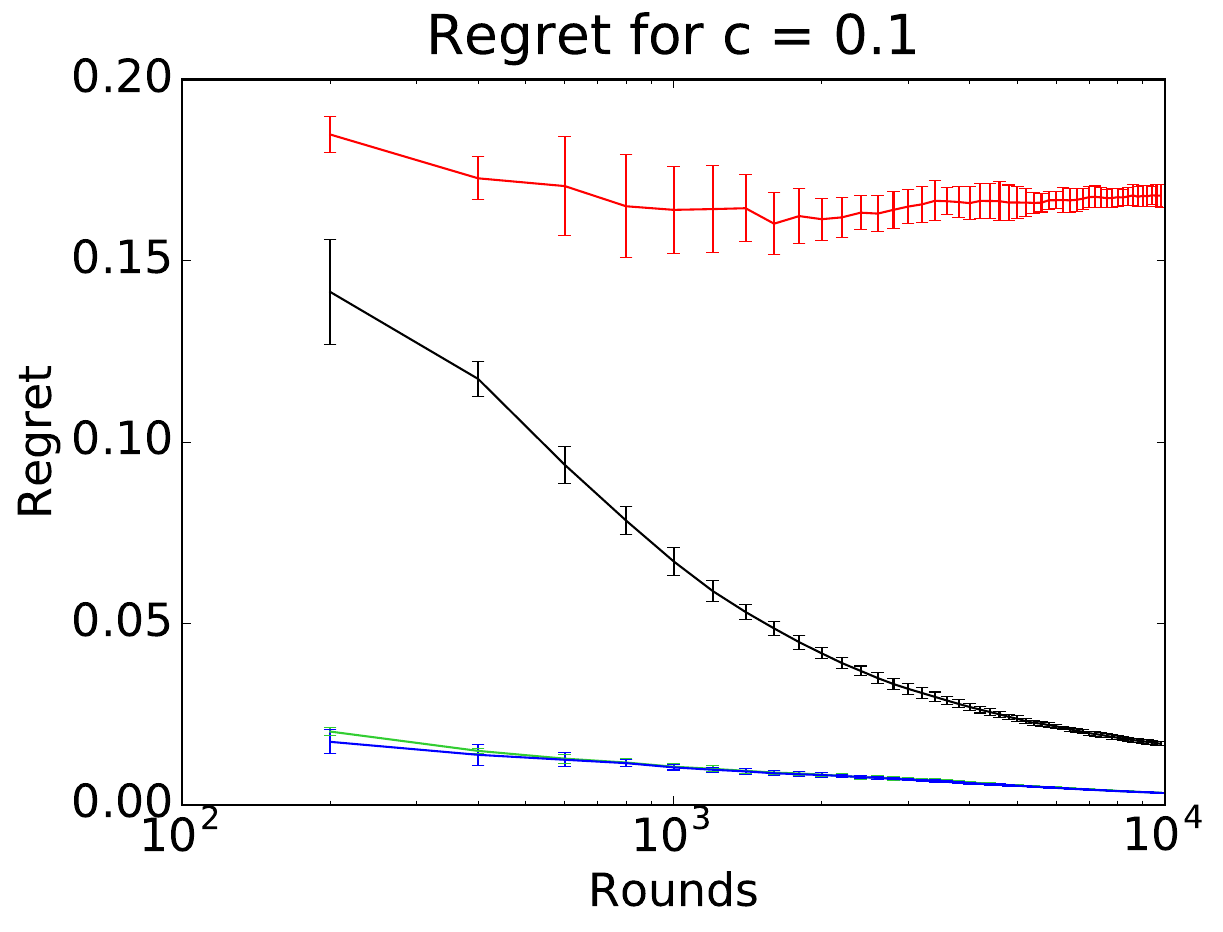} &
\hspace*{-5mm} \includegraphics[scale=0.25,trim= 5 10 10 5, clip=true]{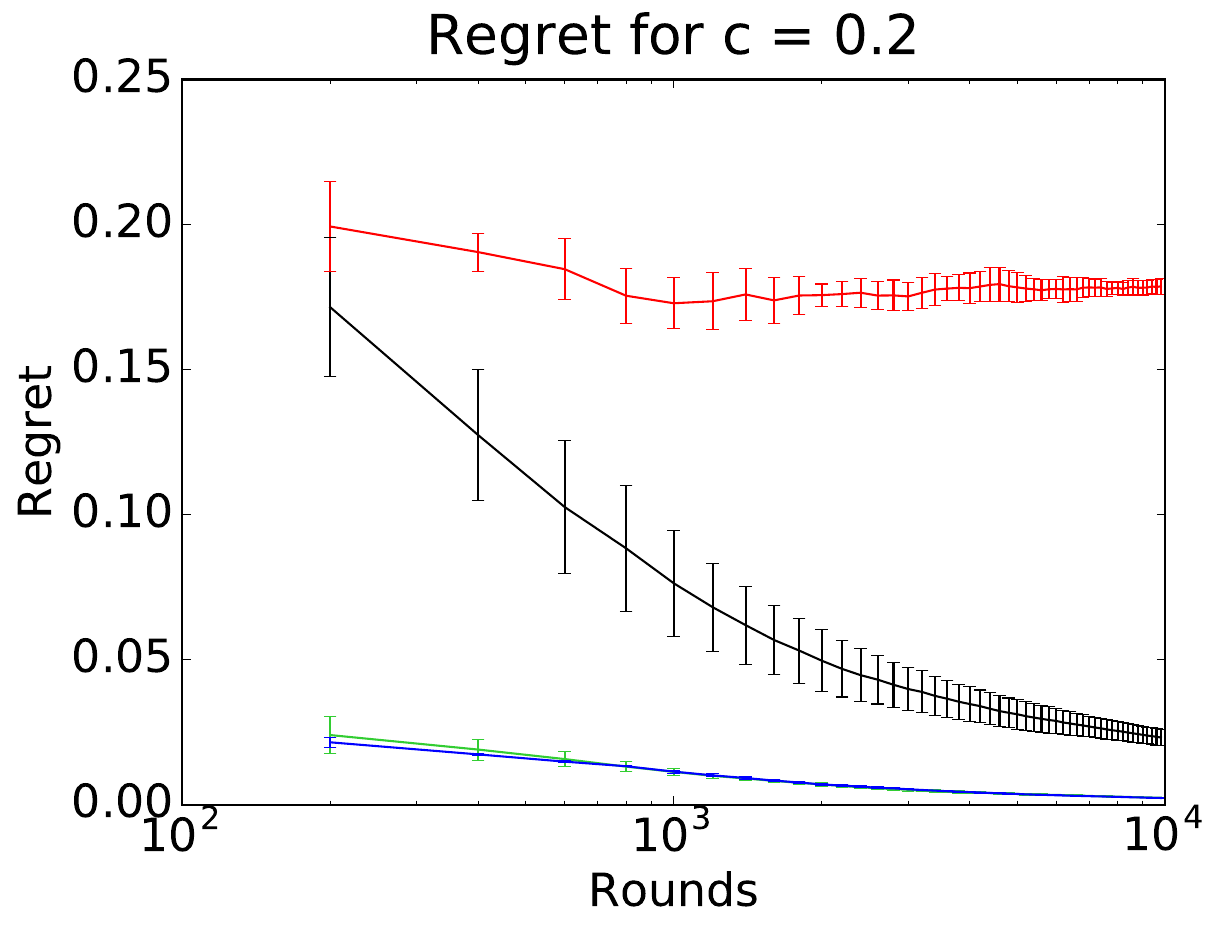} &
\hspace*{-5mm}\includegraphics[scale=0.25,trim= 5 10 10 5, clip=true]{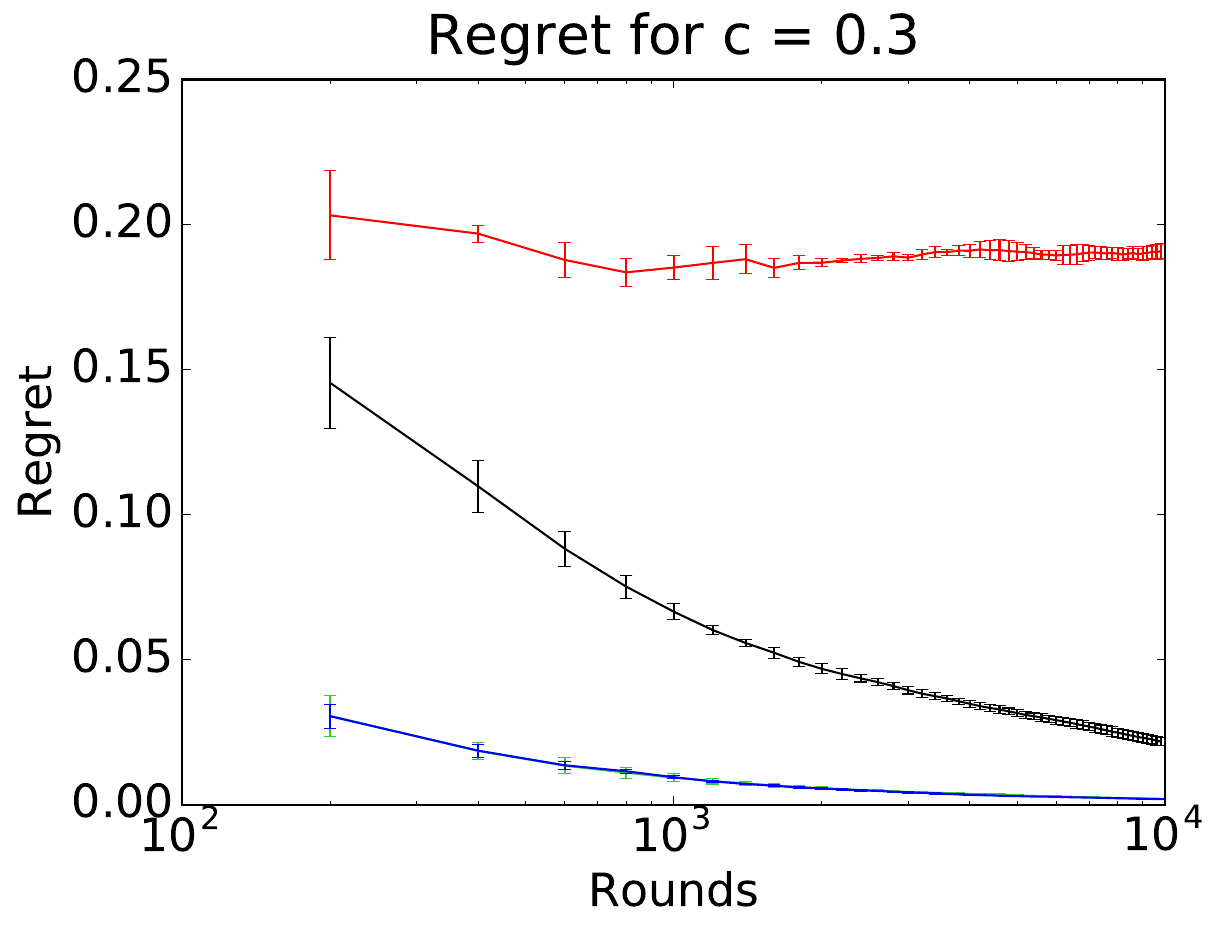} \\
\includegraphics[scale=0.25,trim= 5 10 10 5, clip=true]{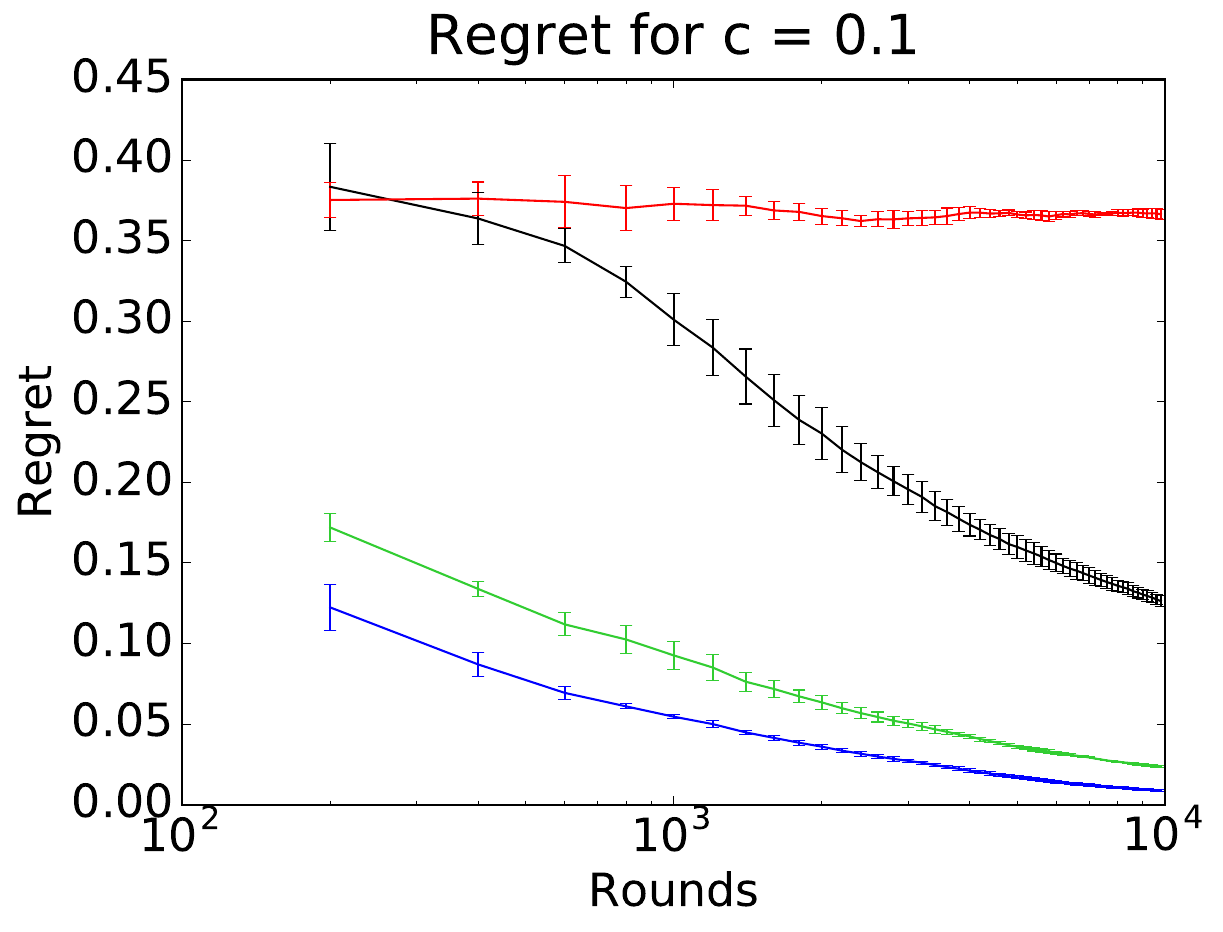} &
\hspace*{-5mm} \includegraphics[scale=0.25,trim= 5 10 10 5, clip=true]{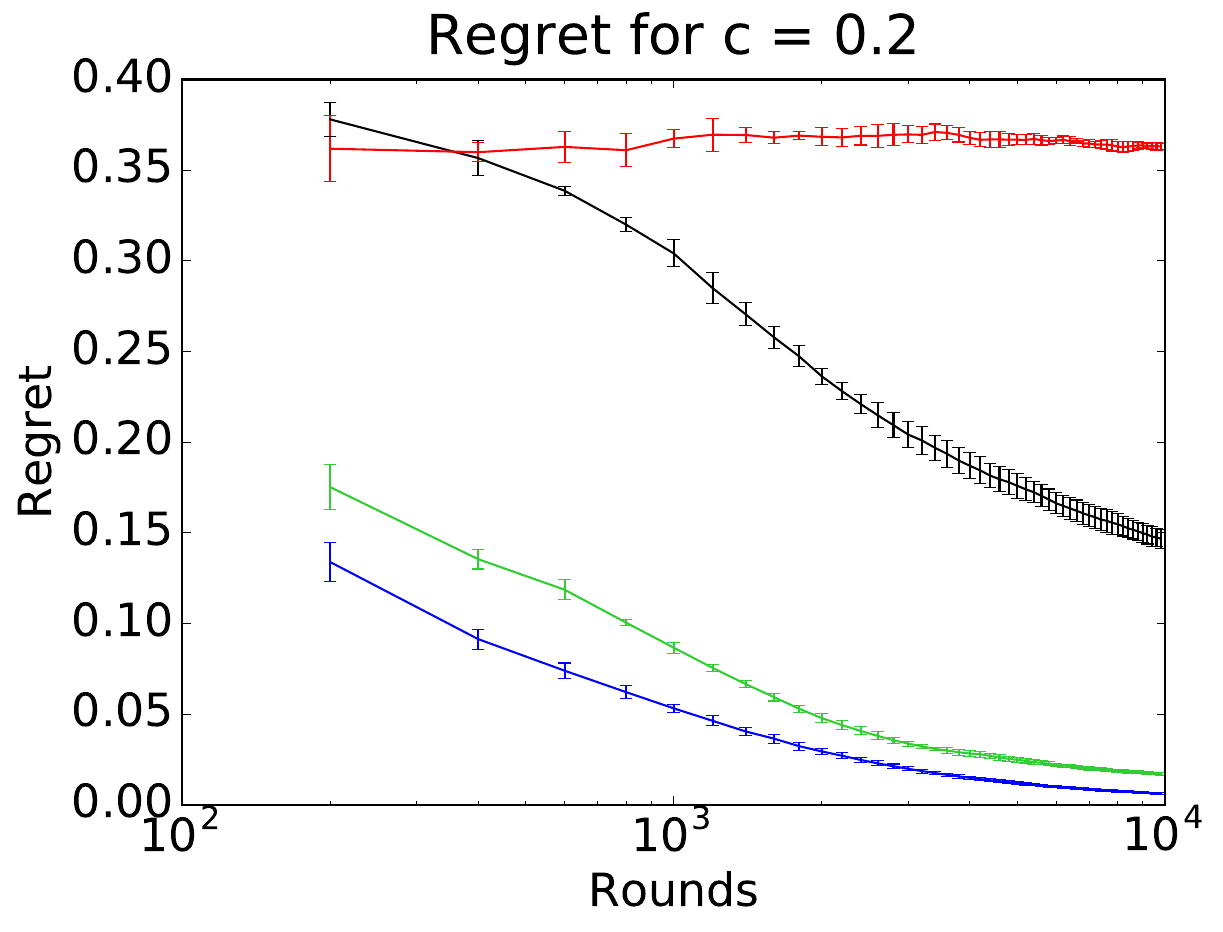} &
\hspace*{-5mm}\includegraphics[scale=0.25,trim= 5 10 10 5, clip=true]{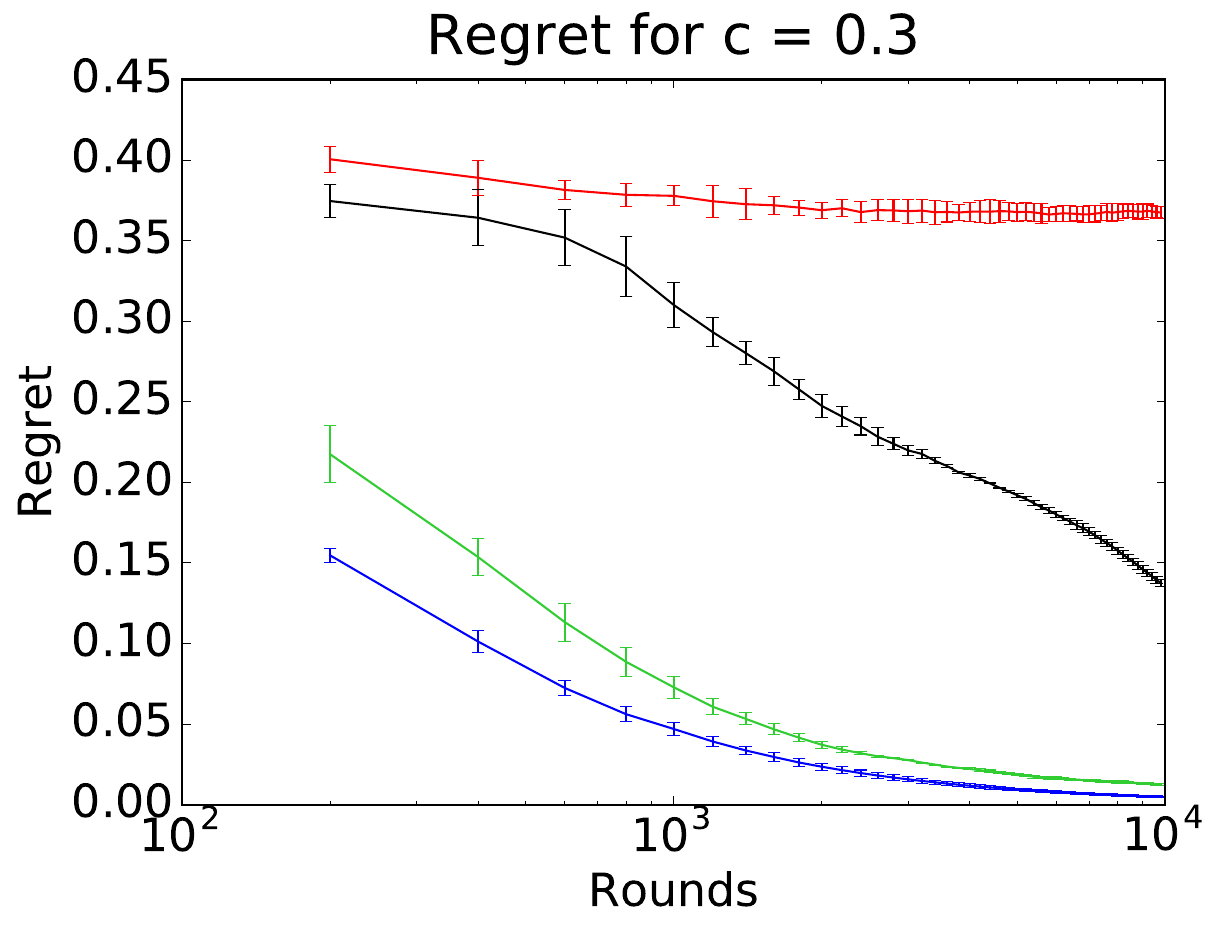} \\
\includegraphics[scale=0.25,trim= 5 10 10 5, clip=true]{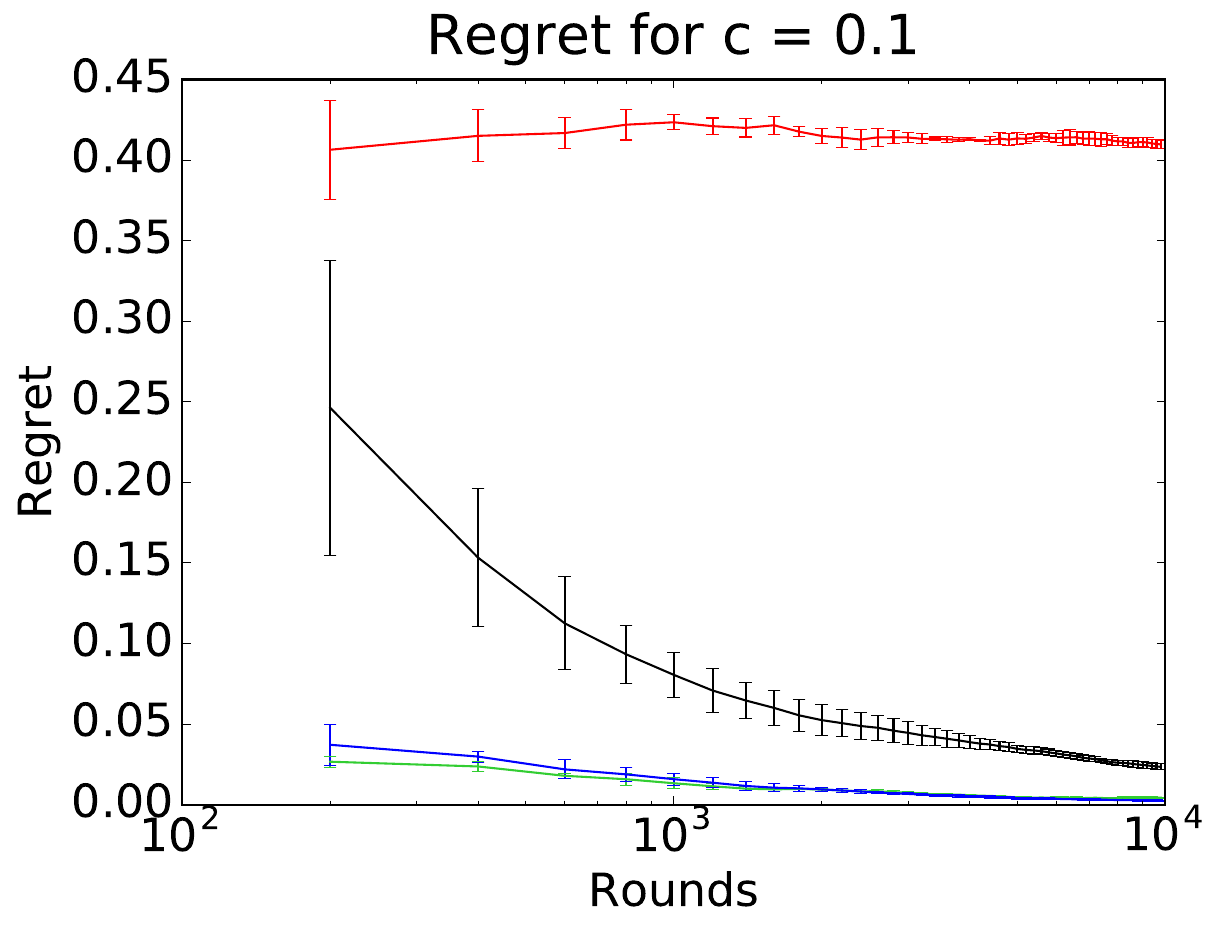} &
\hspace*{-5mm} \includegraphics[scale=0.25,trim= 5 10 10 5, clip=true]{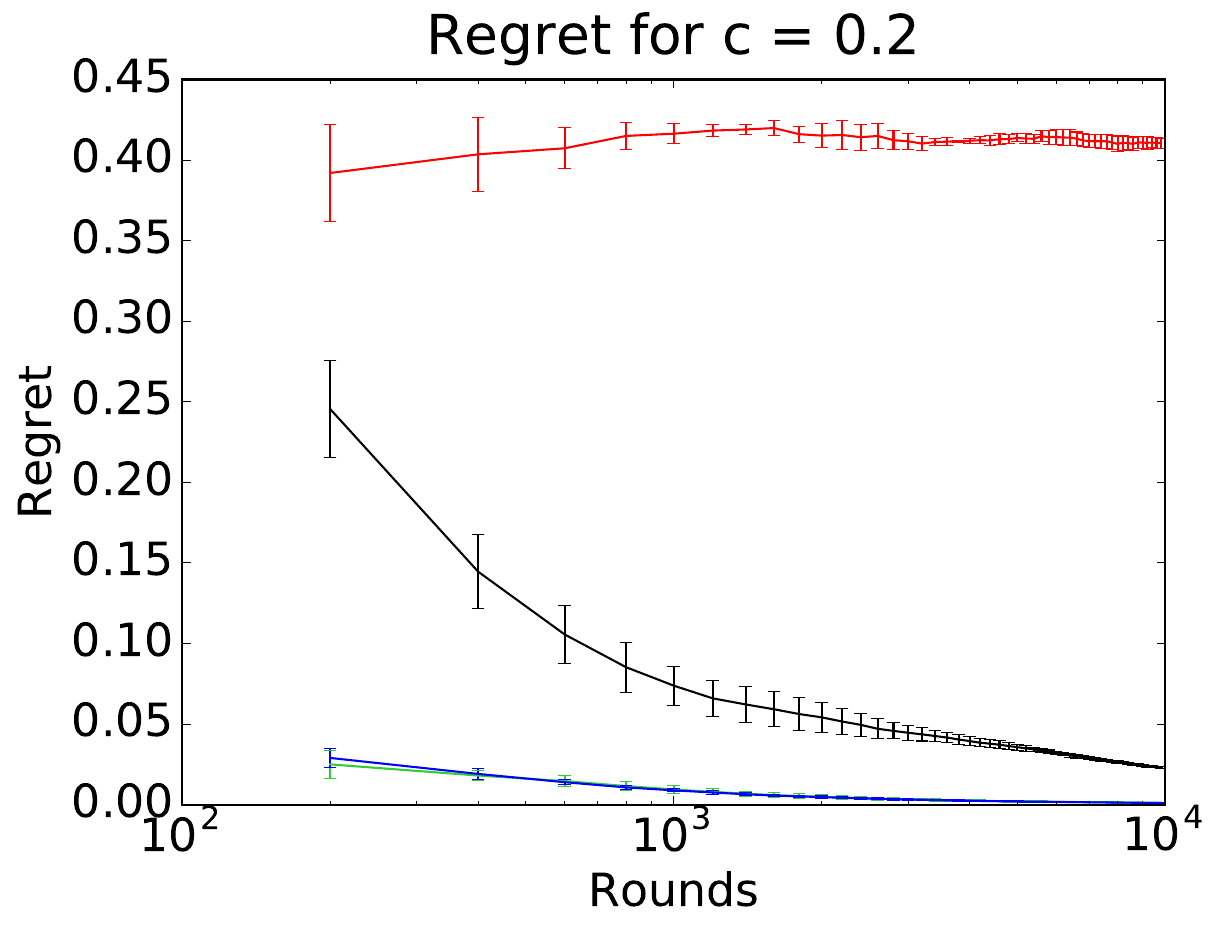} &
\hspace*{-5mm}\includegraphics[scale=0.25,trim= 5 10 10 5, clip=true]{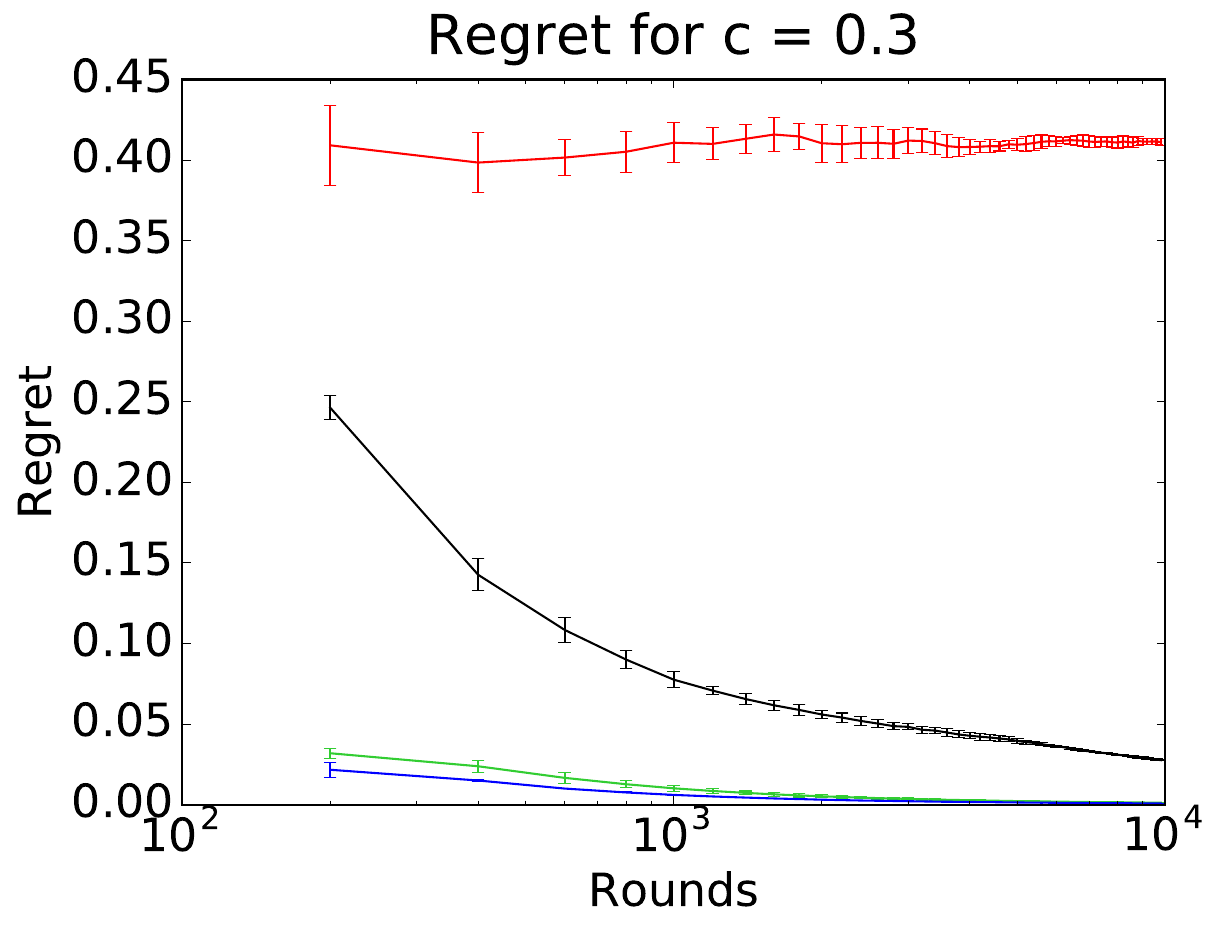} \\
\includegraphics[scale=0.25,trim= 5 10 10 5, clip=true]{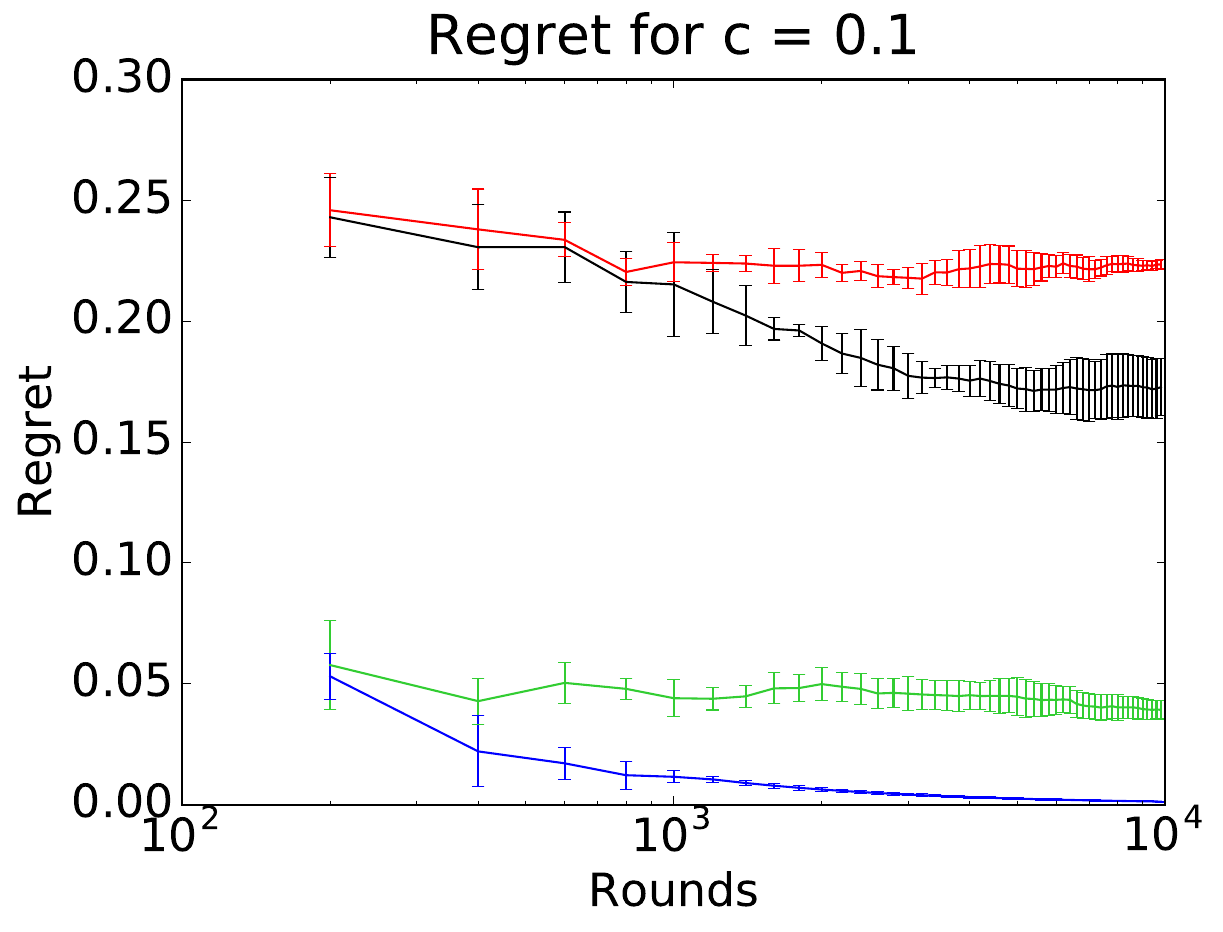} &
\hspace*{-5mm} \includegraphics[scale=0.25,trim= 5 10 10 5, clip=true]{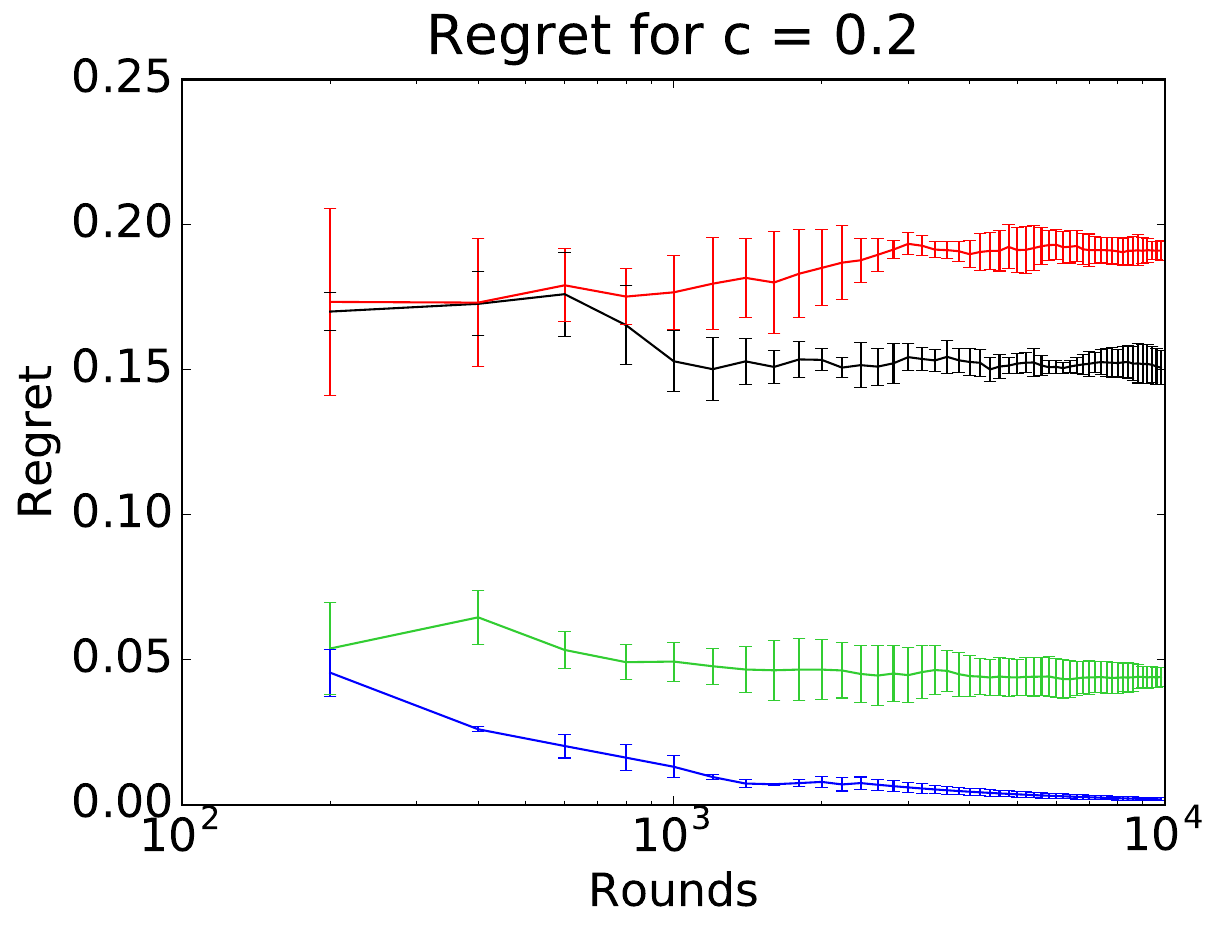} &
\hspace*{-5mm}\includegraphics[scale=0.25,trim= 5 10 10 5, clip=true]{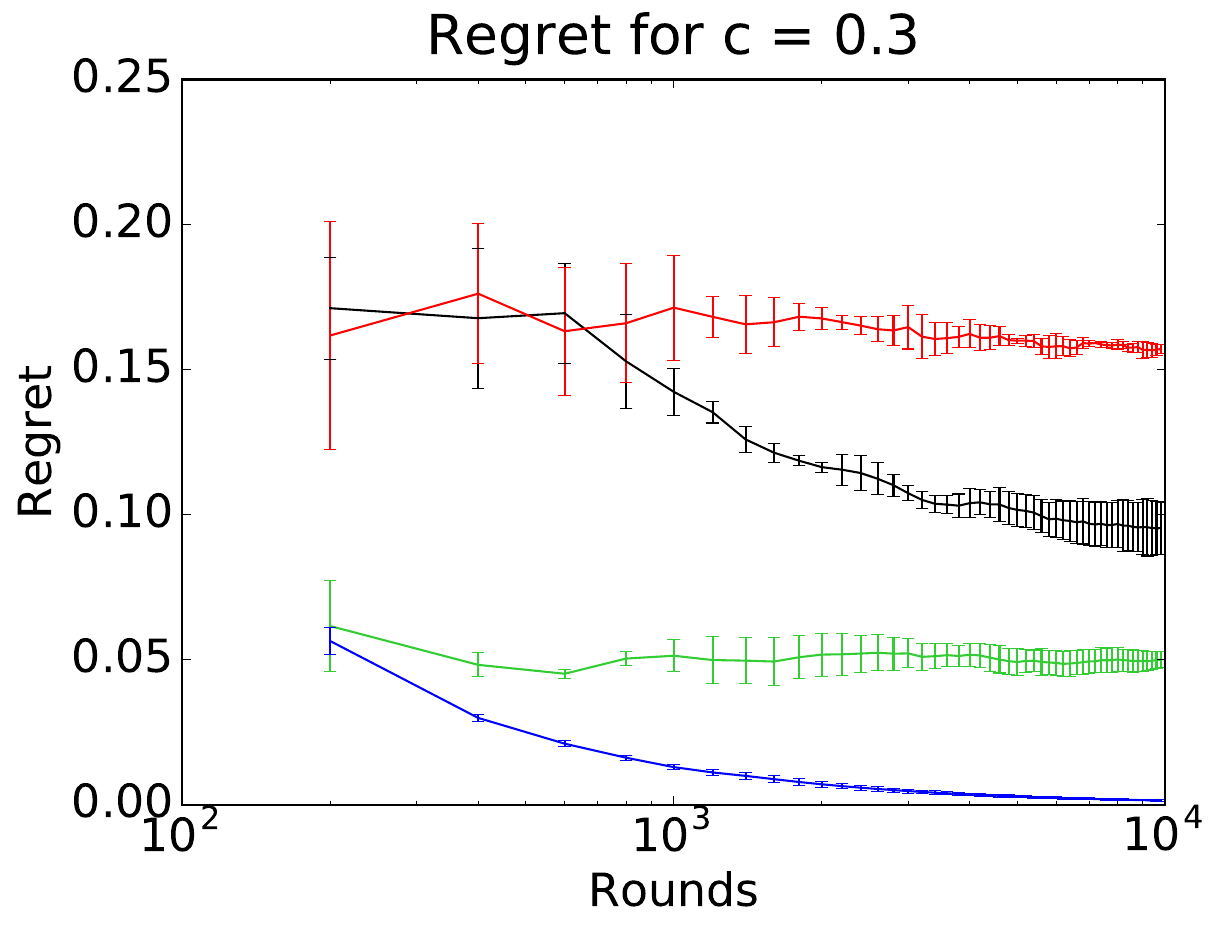} \\
\end{tabular}
\end{center}
\vskip -.15in
\caption{A graph of the averaged regret $R_t(\cdot)/t$ with standard
  deviations as a function of $t$ (log scale) when using the
  confidence based experts for {\color[rgb]{0.16,0.67,0.16}\UCBGT },
  \UCBNT , {\color{red} \UCB }, and {\color{blue} \FTL }. Each row is
  a dataset, starting from the top row we have: {\tt synthetic}, {\tt
    skin}, {\tt guide}, {\tt ijcnn} and { \tt CIFAR}.  }
\label{fig:confexpert}
\vskip -.1in
\end{figure*}

\begin{figure*}[!ht]
\begin{center}
\begin{tabular}{ c c c }
\includegraphics[scale=0.25,trim= 5 10 10 5, clip=true]{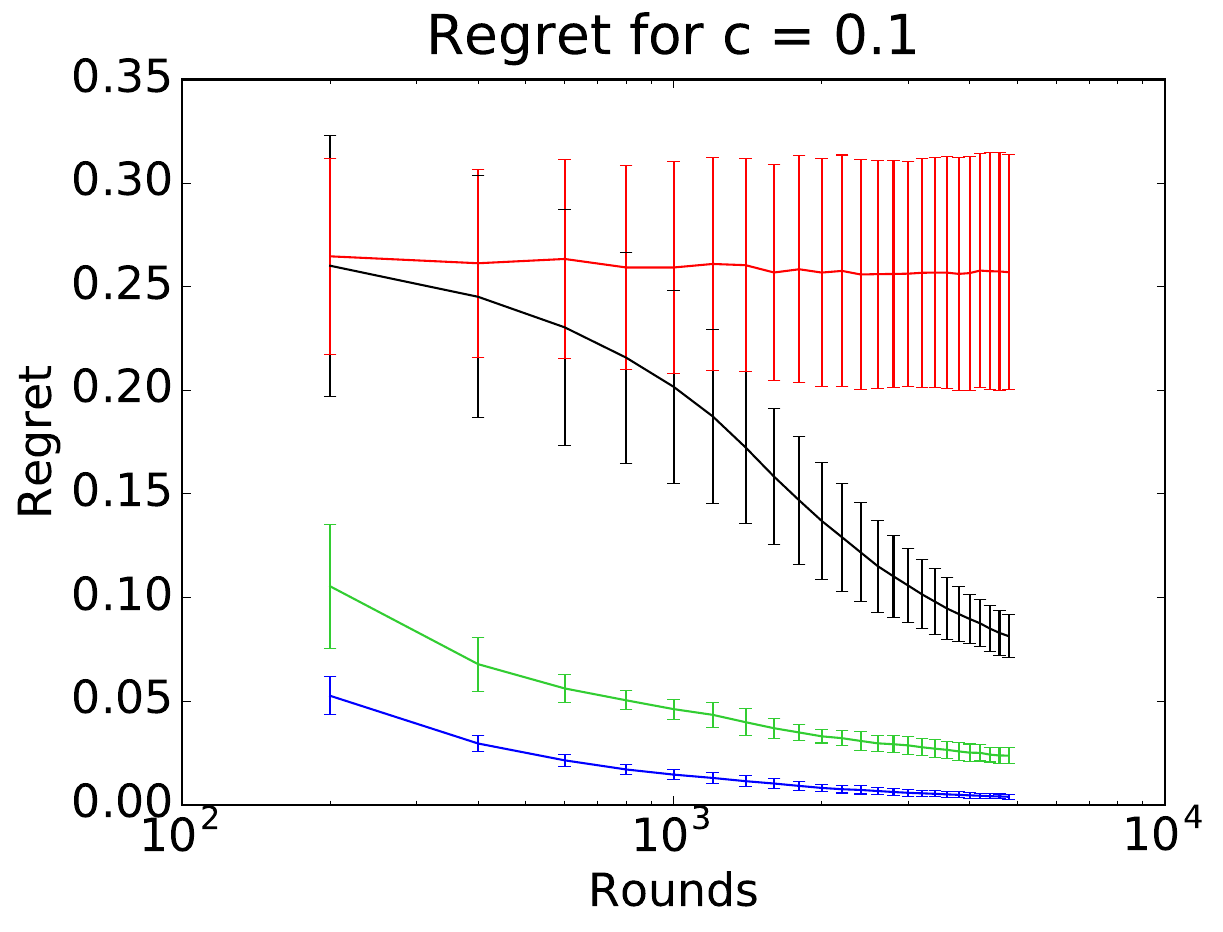} &
\hspace*{-5mm} \includegraphics[scale=0.25,trim= 5 10 10 5, clip=true]{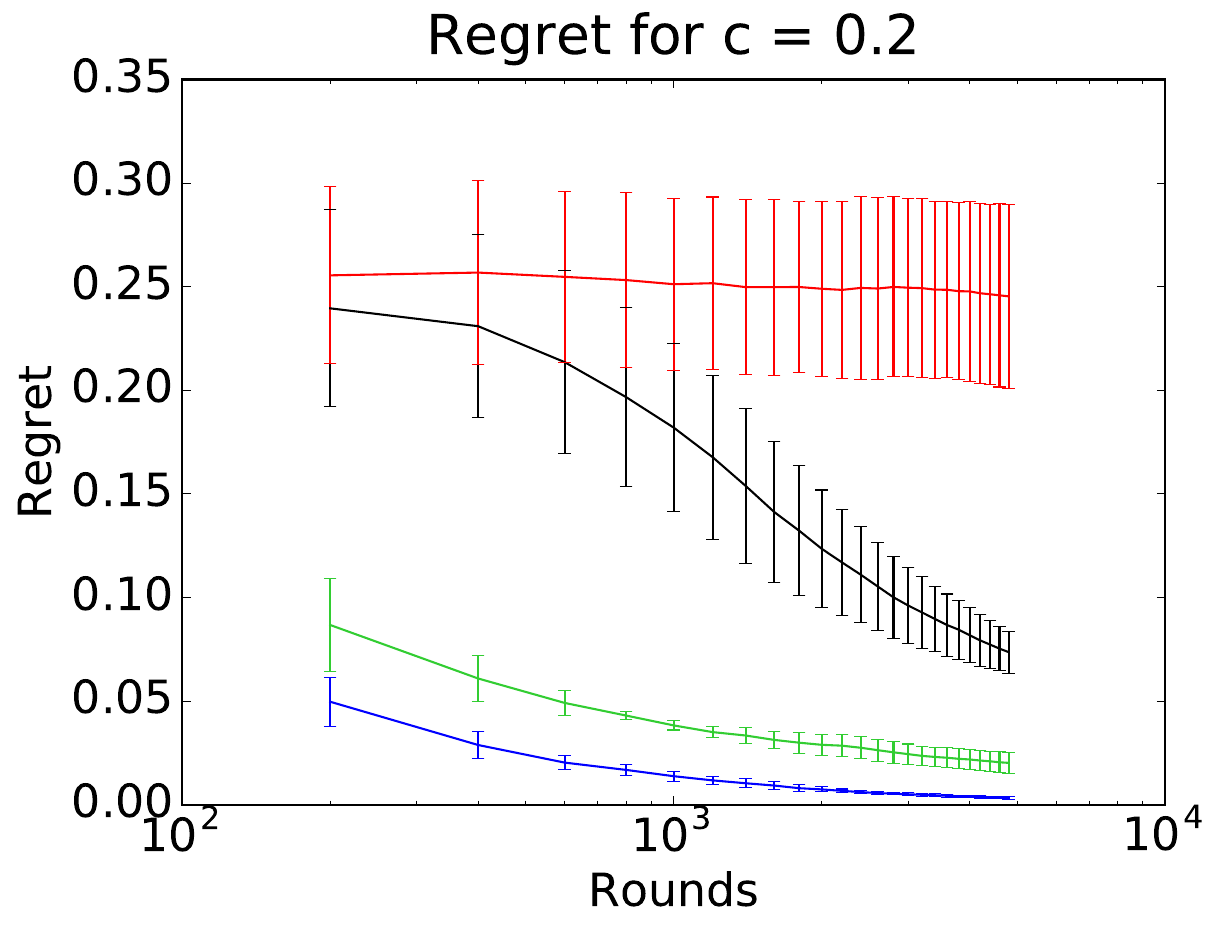}&
\hspace*{-5mm}\includegraphics[scale=0.25,trim= 5 10 10 5, clip=true]{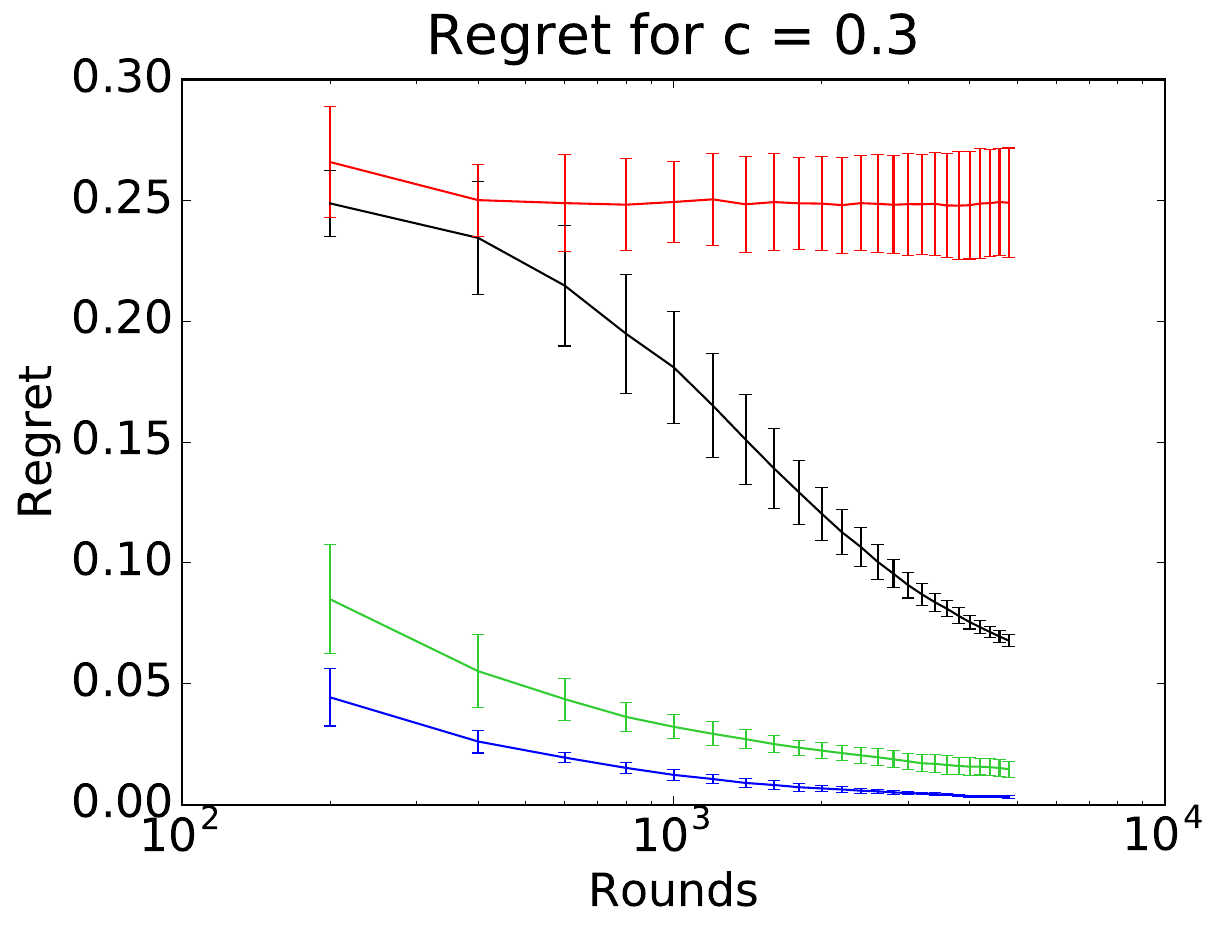} \\
\includegraphics[scale=0.25,trim= 5 10 10 5, clip=true]{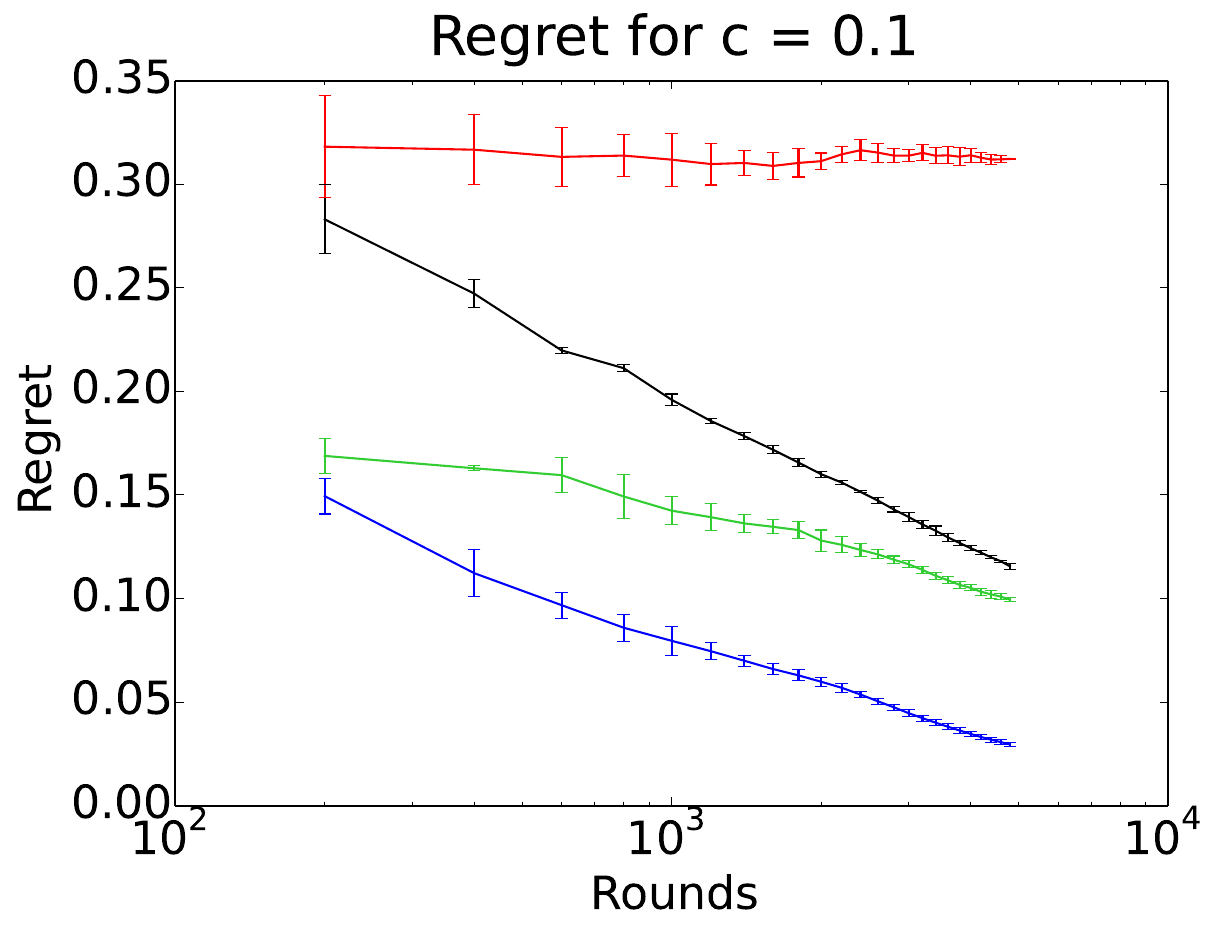} &
\hspace*{-5mm} \includegraphics[scale=0.25,trim= 5 10 10 5, clip=true]{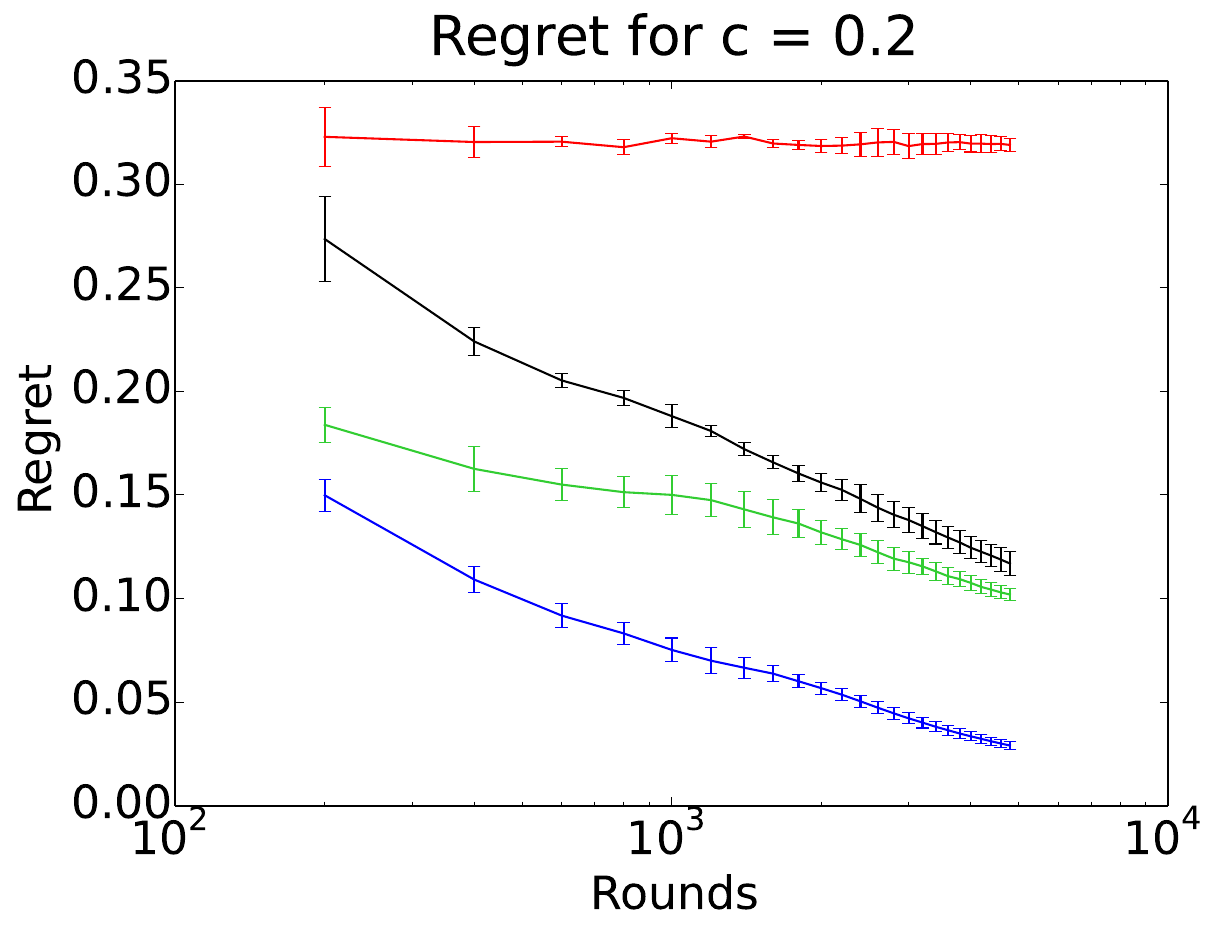}&
\hspace*{-5mm}\includegraphics[scale=0.25,trim= 5 10 10 5, clip=true]{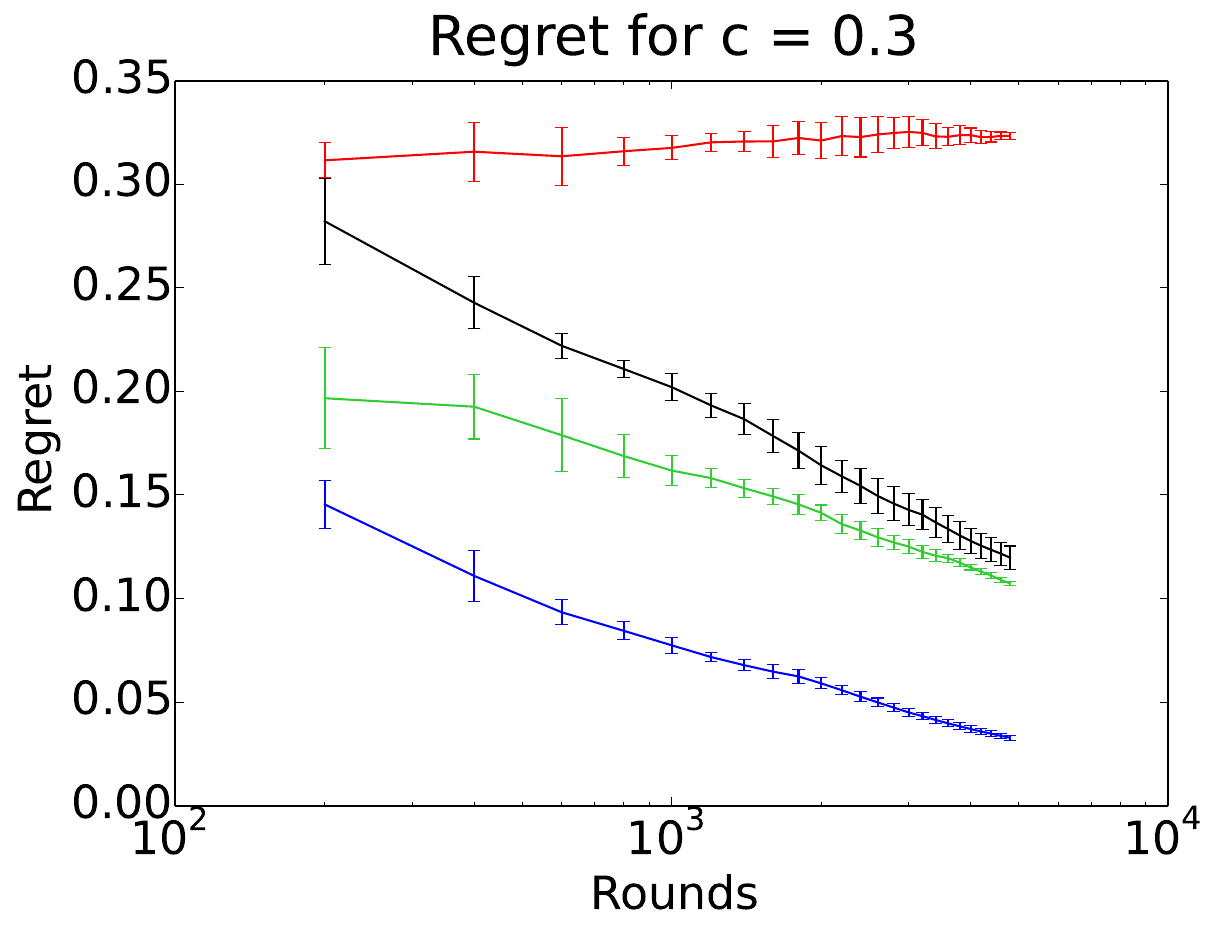} \\
\includegraphics[scale=0.25,trim= 5 10 10 5, clip=true]{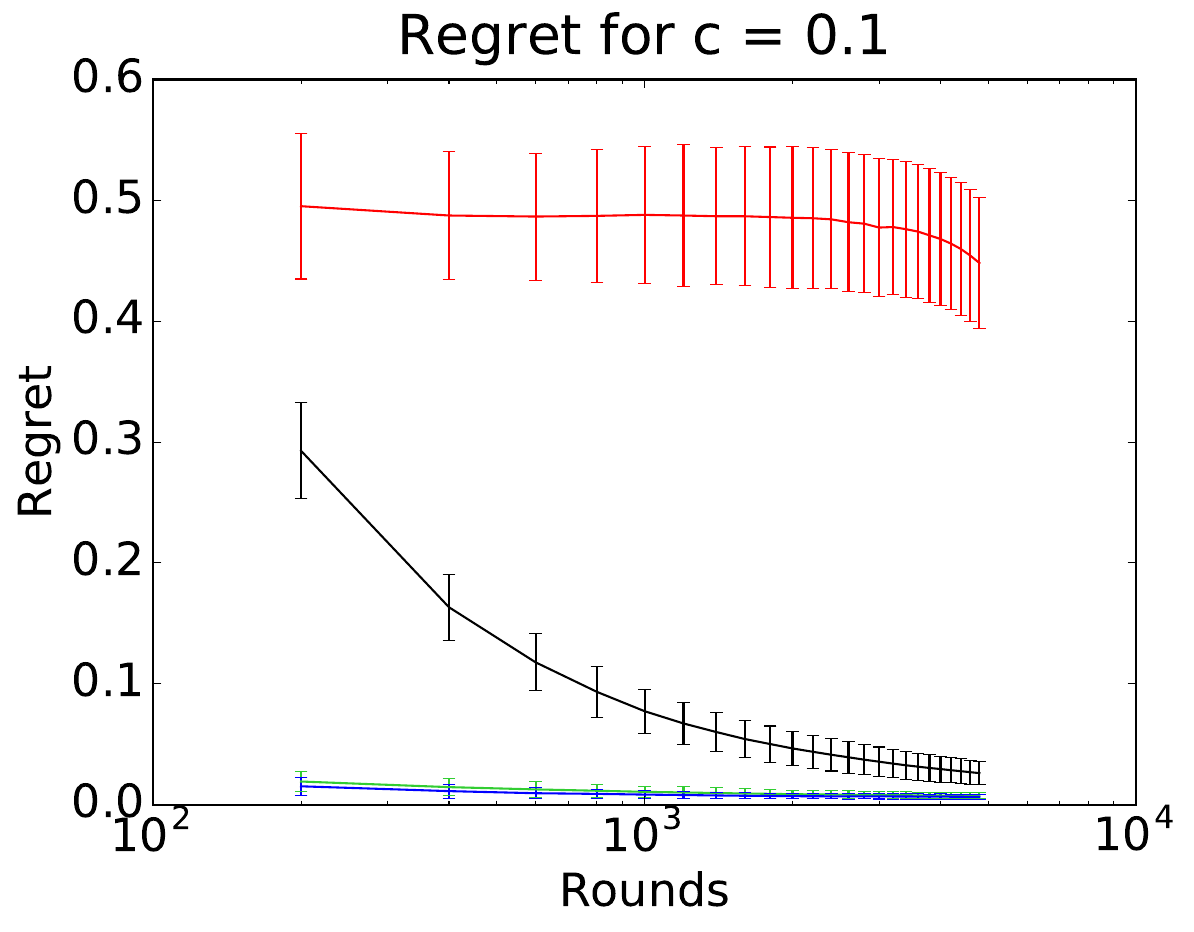} &
\hspace*{-5mm} \includegraphics[scale=0.25,trim= 5 10 10 5, clip=true]{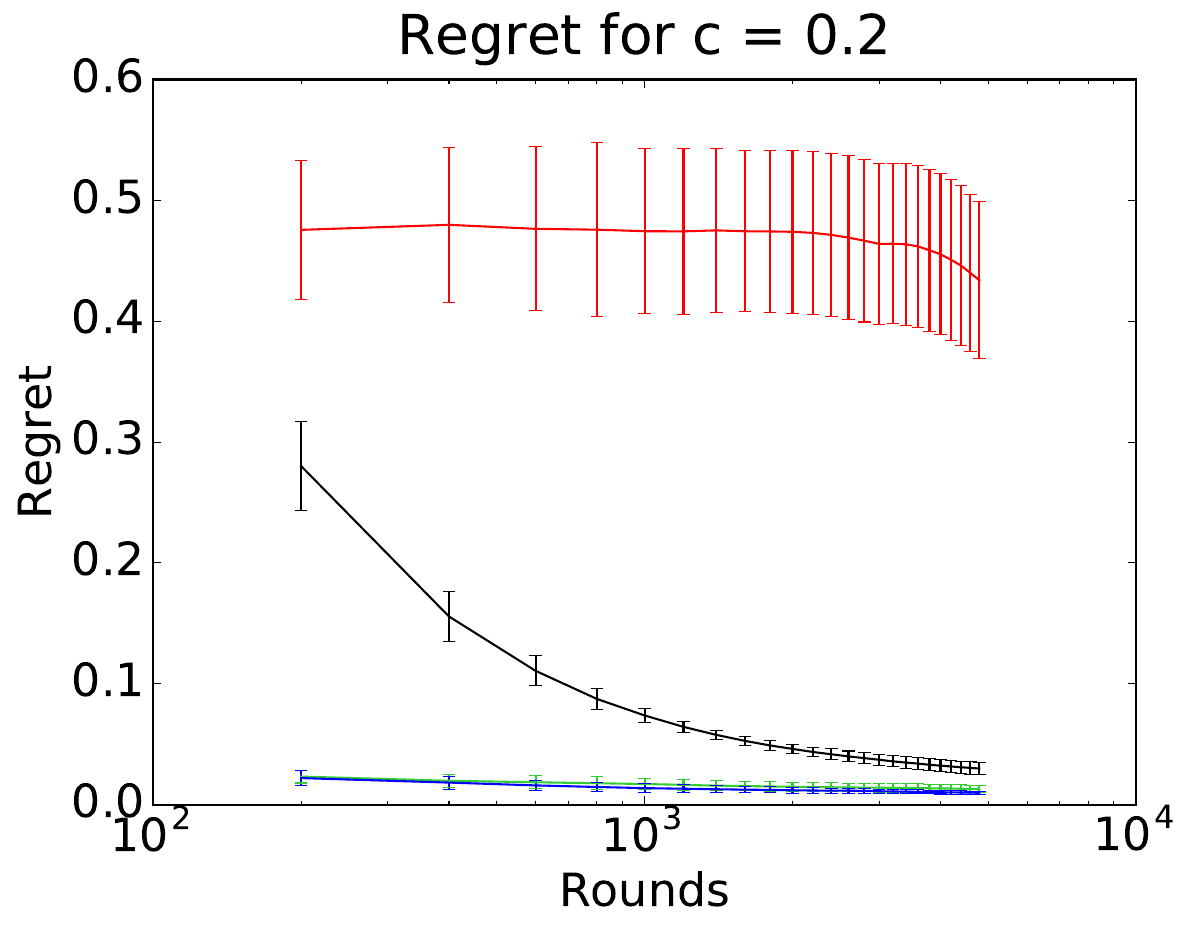}&
\hspace*{-5mm}\includegraphics[scale=0.25,trim= 5 10 10 5, clip=true]{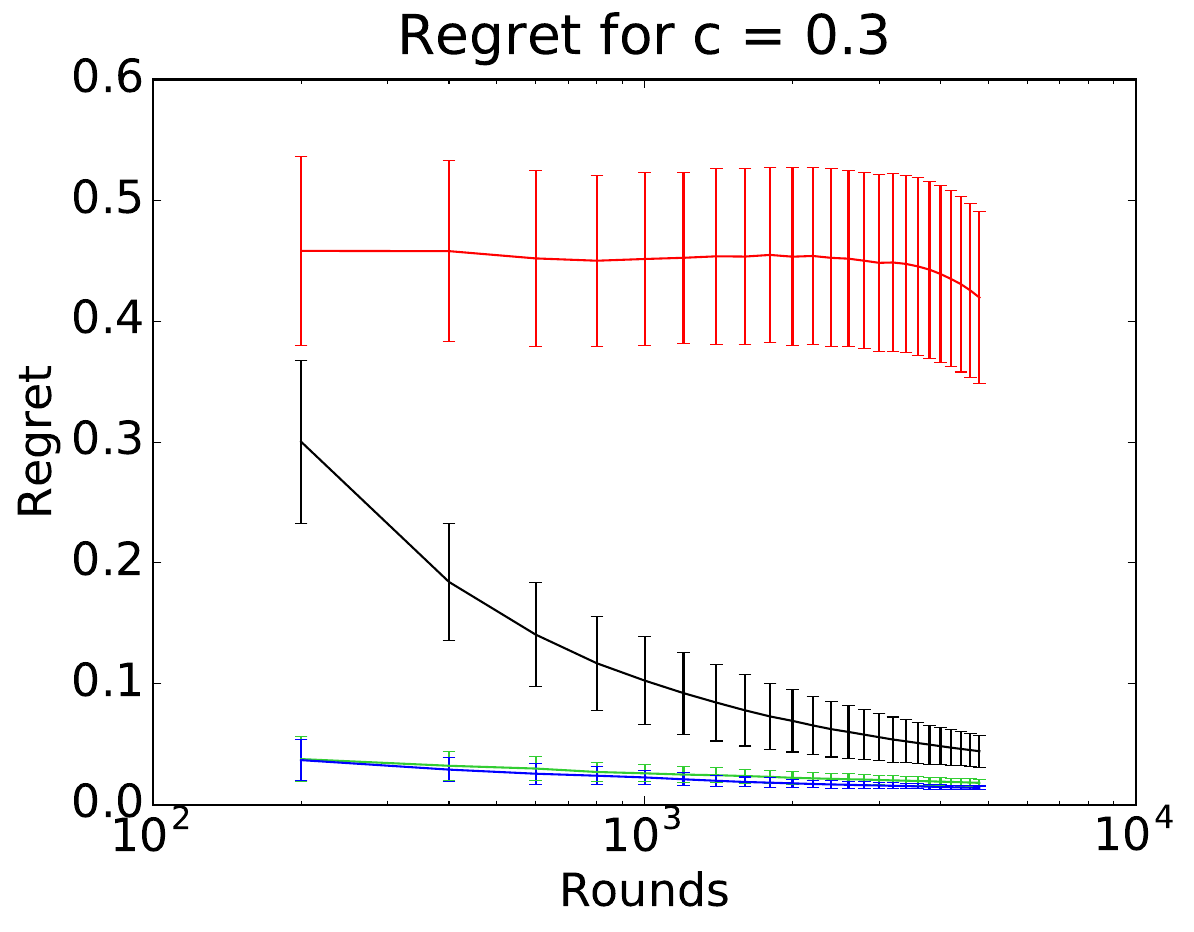} \\
\end{tabular}
\end{center}
\vskip -.15in
\caption{A graph of the averaged regret $R_t(\cdot)/t$ of abstained
  points with standard deviations as a function of $t$ (log scale) for
  {\color[rgb]{0.16,0.67,0.16}\UCBGT }, \UCBNT , {\color{red} \UCB },
  and {\color{blue} \FTL } for different values of abstention costs.
  Each row is a dataset, starting from the top row we have: {\tt
    guide}, {\tt synthetic}, and {\tt skin}. We used $K = 500$ experts
  and $T = 5\mathord,000$ rounds in order to see the effect when
  changing the number of experts used.  }
\label{fig:fullsmall}
\vskip -.1in
\end{figure*}

Figure~\ref{fig:frac1} and Figure~\ref{fig:frac2} show the fraction of
abstained points for all the datasets. Figure~\ref{fig:extremec} also
shows how the fraction of abstained points varies with abstention cost for two extreme values
$c\in\{0.001,0.9 \}$. Again \UCBGT\ admits a lower regret than \UCBNT\ and
\UCB\ and, as expected, the fraction of points decreases as the cost
of abstention increases.  Figure~\ref{fig:confexpert} shows the effect
of using confidence-based experts and suggests that the choice of experts
does not affect the relative performance of the
algorithms. We also tested the effect of varying the
number of experts: Figure~\ref{fig:fullsmall} shows the regret
of three datasets when the number of experts is $K = 500$ and
$T = 5\mathord,000$. For this set of experts, we find a similar pattern of
performance as above.  

Next, we describe in more detail the datasets and how they were
processed. In Table~\ref{tb:features}, we show the number of features
for each dataset. For all datasets, we normalized the features to be in
the interval $[-1, 1]$.  Note that the reason for choosing abstention
functions with radius range $(0, \sqrt{d})$ is to cover the entire
hypercube $[-1,1]^d$ with our concentric annuli.  For the {\tt CIFAR}
dataset, we extracted the first twenty-five principal components of
the {\tt horse} and {\tt boat} images, projected the images on these
components, and normalized the range of the projections to
$[-1,1]$. The features of the synthetic dataset are drawn from the
uniform distribution over $[-1,1]^2$ and the label is determined by
the sign of the projection of a point onto the normal of the diagonal
hyperplane $y=-x$.

The confidence-based abstention function has the form
$r(x) = |h(x)| - \theta$. In our experiments (Figure
~\ref{fig:confexpert}), we generated twenty abstention functions with
thresholds $\theta \in (0, \ldots, 0.25)$, which are paired with each
predictor. The predictors are axis-aligned planes along each feature
of the dataset. For each dataset, the number of predictors is
$\floor{100/d}$ where $d$ is the dimension of the dataset. We chose
twenty abstention functions and about 100 prediction functions in
order to match the experimental setup of the randomly drawn experts.
The total number of experts is then $\floor{100/d}\cdot 20\cdot d$.
Note that we only tested some of our datasets since for larger
dimensions $d$, the number of experts per feature was too small.

\clearpage
\subsection{Average regret for different abstention costs and datasets}
\label{app:exp_avgreg}


\clearpage
\subsection{Average fraction of abstention points for different
  abstention costs and datasets}
\label{app:exp_avgfracabs}

\clearpage
\subsection{Average regret and fraction of abstention points for extreme abstention costs}
\label{app:exp_extremecost}

\clearpage
\subsection{Average regret for confidence-based experts }
\label{app:exp_confidence}

\clearpage
\subsection{Average regret for a smaller set of experts }
\label{app:exp_fewerexperts}

\end{document}